%% file: main.tex
\documentclass[11pt,a4paper]{article}

\usepackage[T1]{fontenc}
\usepackage{lmodern}
\usepackage{amsmath, amssymb, amsthm, mathtools}
\usepackage{bm}
\usepackage{graphicx}
\graphicspath{{figures/}{./}}
\usepackage{multirow}
\usepackage{longtable,array}
\usepackage{booktabs}
\usepackage{placeins}
\usepackage[margin=1in]{geometry}
\usepackage{xcolor}
\usepackage[colorlinks=true, linkcolor=blue!50!black, citecolor=blue!50!black, urlcolor=blue!50!black]{hyperref}
\hypersetup{pdftitle={Stochastic Kernel Mode Decomposition: Information, Uncertainty, and Conditional Regime Inference}, pdfauthor={Boumediene Hamzi and Houman Owhadi}}
\usepackage{enumitem}
\usepackage{tikz}
\usetikzlibrary{arrows.meta, positioning, fit, backgrounds, calc, matrix}
\usepackage[numbers,sort&compress]{natbib}
\newtheorem{theorem}{Theorem}[section]
\newtheorem{definition}[theorem]{Definition}
\newtheorem{proposition}[theorem]{Proposition}
\newtheorem{corollary}[theorem]{Corollary}
\newtheorem{lemma}[theorem]{Lemma}

\newcommand{\R}{\mathbb{R}}
\newcommand{\E}{\mathbb{E}}
\newcommand{\Prob}{\mathbb{P}}
\newcommand{\N}{\mathcal{N}}

\DeclareMathOperator{\tr}{tr}
\DeclareMathOperator{\Var}{Var}

\usepackage{needspace}

\title{Stochastic Kernel Mode Decomposition:\\Information, Uncertainty, and Conditional Regime Inference}
\author{Boumediene Hamzi\thanks{Department of Computing and Mathematical
Sciences, California Institute of Technology, Pasadena, CA 91125, USA;
and The Alan Turing Institute, London NW1 2DB, UK.
Email: \href{mailto:boumediene.hamzi@gmail.com}{\texttt{boumediene.hamzi@gmail.com}}.}
\and Houman Owhadi\thanks{Department of Computing and Mathematical Sciences,
California Institute of Technology, Pasadena, CA 91125, USA.
Email: \href{mailto:owhadi@caltech.edu}{\texttt{owhadi@caltech.edu}}.}}
\date{12 November 2026}
\begin{document}
\maketitle
\begin{abstract}
When does the energy of each recovered mode retain enough information to
infer a regime, and how does the answer change with measurement quality?
Stochastic Kernel Mode Decomposition (sKMD) fits joint regime laws on
signed KMD mode coordinates and uses the reference mode posterior to
compute regime probabilities and conditional reconstructions. For
centered Gaussian regimes under independently varying known noise
variances, fixed coordinate energies are sufficient precisely when the
classes share every cross-covariance and class-dependent variances occur
only at isolated coordinates of that covariance graph. We extend the
criterion to nonzero means and show that, with freely varying correlated noise, even two
distinct centered regimes require the complete quadratic feature span for
a fixed linear log-odds decoder. At fixed quality a three-regime example
has no sufficient two-energy basis. Its smallest Bayes log-loss gap over
bases of condition number at most $\kappa$ has sharp order $\kappa^{-1}$.
Further bounds quantify the cost of limited quadratic features,
independently learned mode spans and sensor-noise error. Two controlled experiments find small log-loss gains
over strong signed classifiers, with sensitivity to calibration and
noise monitoring. Learned-mode and recorded-signal studies find benefits
mainly under specified synthetic degradations, with simpler controls
matching or outperforming integration in several native-quality and
transfer settings.
\end{abstract}

\clearpage
\setcounter{tocdepth}{1}
\tableofcontents

\section{Introduction}
\label{sec:introduction}
Kernel Mode Decomposition (KMD) separates signals into components described
by kernels and relates this recovery problem to Gaussian conditioning,
optimal recovery and adaptive estimation of amplitude, phase and waveform
\citep{owhadi_kernel_2020}. Many uses of a decomposition require a second
inference task. A vibration record may indicate a forcing regime; an
oscillatory waveform may need reconstruction despite missing samples; and
a change in observation quality may alter how much a small recovered mode
should count as evidence. These tasks depend on the joint law of the modes,
not only on their separate reconstruction energies.

Kernel representations also connect signal analysis with stochastic
transitions. Maximum mean discrepancy detects changes in distributions in
fast--slow systems \citep{mmd_kernels_bh} and has been studied for EEG
seizure detection \citep{hamzi2018seizure}. Transition-manifold embeddings
reduce metastable dynamics \citep{bittracher2021dimensionality}, and
kernel approximations of quasipotentials describe rare transitions in the
weak-noise setting \citep{lee2025quasipotential}. Microlocal,
time--frequency--phase kernels give a direct KMD precedent by reconstructing
signals around critical transitions in slow--fast models and EEG-related
seizure examples \citep{hamzipaillet}. This motivates retaining the
observation posterior as well as the reconstructed modes.
Related discrepancy constructions compare laws on countable spaces and
prefix trees \citep{hamzi2026discrepancies,hamzi2026algorithmictransport}
or drive artificial flows for matching prescribed measures
\citep{hamzi2026ksdmmd}. The inferential task here is to assign
probabilities to regimes of a measured process.

This paper develops \emph{stochastic KMD (sKMD)} as a regime-inference
extension of KMD. The extension fits joint class laws on signed mode
coordinates, retains the KMD reference posterior mean and covariance, and
uses a change of prior to obtain regime probabilities and
regime-conditioned reconstructions. A regime can describe a class, an
excitation law or a state in a specified temporal model. The same
interface handles a known observation mask and noise covariance. Every
method labeled sKMD in this paper is this construction: fitted regime laws
on KMD mode coordinates, combined with the reference posterior $(m,V)$
through Proposition~\ref{prop:posterior_transfer_v31}.

Building on KMD's signed modes and probabilistic interpretation, sKMD
introduces a fitted joint regime law. Comparing sKMD with separate mode energies tests
information discarded by a particular summary of KMD output. Comparing
its reconstructions with independent-mode or pooled full-covariance KMD
tests the value of the regime law. Comparing it with a quality-aware
signed classifier tests whether explicit integration adds predictive
value beyond retaining the relevant information. These are different
estimands and receive separate controls.

\subsection{The research question and contributions}
The central question is whether one representation of the modes remains
adequate as observation quality changes. A coordinate system that makes
energies sufficient at one noise level may fail at another. Conversely,
changing posterior coordinates can sometimes restore sufficiency. This
paper studies that distinction and uses it to specify the information
that an sKMD implementation should preserve.

\begin{enumerate}[leftmargin=*,itemsep=4pt]
\item \textbf{Representation characterizations under changing quality
(Section~\ref{sec:quality_theory_v32}).}
Theorem~\ref{thm:quality_diagonal_v32} characterizes, through the
covariance graph, when fixed coordinate energies remain sufficient for
every diagonal noise covariance in an open set.
Corollary~\ref{cor:noncentered_quality_extension} extends the
characterization to nonzero means and Corollary~\ref{cor:quality_full_v32}
to correlated noise. Theorem~\ref{thm:quality_quadratic_span_v32} shows
that correlated noise variation forces the complete quadratic span for any
fixed linear-in-features decoder, even for two fixed class covariances.
Proposition~\ref{prop:quality_feature_compression_v32} and the stable
approximation Proposition~\ref{prop:stable_quality_extension} bound the
error of a restricted quadratic feature budget under the actual observed
mixture, and Theorem~\ref{thm:quality_expansion_v32} gives the local size
of quality-induced energy loss.
\item \textbf{Fixed-quality geometry and a three-regime obstruction
(Section~\ref{sec:mode_information_v31}).}
Theorem~\ref{thm:energy_sufficiency_v31} gives the fixed-quality
energy-sufficiency criterion. Theorem~\ref{thm:three_class_obstruction_v31}
and Corollary~\ref{cor:two_quadratic_energies_v31} show that a specified
three-regime law has no sufficient nonredundant two-energy basis.
Theorems~\ref{thm:energy_uniform_gap_v31} and~\ref{thm:conditioned_basis_gap}
give positive information-loss bounds over orthogonal and
conditioning-controlled bases. Corollary~\ref{cor:sharp_conditioning_rate}
establishes the sharp order of the best gap under a conditioning limit.
Section~\ref{sec:energy_oracle_v31} estimates gap sizes using exact
Bayes energy posteriors.
\item \textbf{Learning bounds (Section~\ref{sec:adaptive_theory_v32}).}
Theorem~\ref{thm:weak_interactions_v32} gives a local information expansion
with a computable remainder. Lemma~\ref{lem:span_leakage_v32} and
Theorem~\ref{thm:learned_span_selection_v32} bound learned-span leakage for
an independently selected finite dictionary library, and
Theorem~\ref{thm:learned_benefit_v32} gives an explicit condition under
which a fitted interaction rule beats an energy oracle.
Proposition~\ref{prop:mode_perturbation_v31} bounds posterior log loss
under covariance perturbation.
\item \textbf{A mode-posterior interface (Section~\ref{sec:skmd_method}).}
Definition~\ref{def:skmd} and Proposition~\ref{prop:posterior_transfer_v31}
express regime inference and regime-conditioned reconstruction through the
KMD posterior mean and covariance, with an exact reduction to the reference
KMD model. The Gaussian transfer formula is standard Gaussian conditioning
expressed through the KMD posterior; matched latent-Gaussian
implementations give the same output and serve as implementation checks.
\item \textbf{Controlled tests and applications (Part~\ref{part:applications}).}
Two controlled experiments with 64 independent panels each separate
designed binary calibration from finite-sample recovery of partial energy
information in a three-regime family; a retrospective calibration analysis
compares binary inference with quality-dependent point calibration.
Studies with learned carriers, changing dynamics, ECG and bearings evaluate
the same mechanisms on learned and recorded data. In those studies
integration helped mainly under declared synthetic degradations; at native
quality or under operating-condition shift, simpler controls were as good
or better.
\end{enumerate}

\subsection{Relation to existing methods}
Joint Gaussian class laws, QDA, common-factor discriminant models and
heterogeneous-noise likelihoods are established tools
\citep{baek_commonfactor_2010,bovy_deconvolution_2011,bilson_uncertainty_2026}.
Using the same dictionary, class parameters and observation model in
extreme deconvolution or a latent-Gaussian classifier gives the same
probabilities as the Gaussian sKMD calculation. The posterior transfer is
therefore not a new probability calculation. The new content is the
characterization of which mode features retain the regime posterior as the
observation quality varies (Sections~\ref{sec:mode_information_v31}
and~\ref{sec:quality_theory_v32}) and the bounds on learning them from
finite data (Section~\ref{sec:adaptive_theory_v32}).

At fixed quality, the energy-sufficiency criterion follows from the
polynomial form of Gaussian log odds. Common principal components,
common-orientation discriminant models and simultaneous congruence
provide its coordinate geometry
\citep{flury_common_1984,bensmail_eigenvalue_1996,jiang_simultaneous_2016}.
Conditional mutual information and proper-score deficiencies measure
what a summary discards \citep{beraha_feature_2019,gyorfi_lossless_2023}.
Linear and quadratic sufficient features for Gaussian laws are classical,
including their use in heteroscedastic Gaussian-process models
\citep{le_heteroscedastic_2005}. Here one fixed class family is held constant
while the observation covariance varies over an open set. This uniform
constraint produces the covariance-graph characterization and the
quadratic-span result. The approximation result then uses the actual
quality-indexed class mixture to measure the cost of a limited feature
budget.

Kernel models for dynamical systems target several different quantities.
Balanced reduction uses controllability and observability to choose
reduced coordinates for nonlinear systems. Early conference and workshop
contributions preceded the journal treatment
\citep{BH4,bouvrie2013modelreduction,bh17}. Related
estimators recover these quantities and invariant measures from
stochastically forced systems \citep{bh12,hb17}.
Kernel approximations can estimate discrete-time update rules for linear
autonomous and controlled systems \citep{hamzi2019kernel,hamzi2019discrete_equations}.
Center-manifold approximation and learned Gaussian-process transformations
instead seek coordinates that simplify local dynamics
\citep{bhcm11,bhcm1,lee2025gpsimplify}. A modal representation in sKMD serves
another purpose. It must retain enough information from a noisy record to
distinguish its candidate regimes.

That inference target also differs from operator estimation. Kernel
spectral methods approximate the Koopman generator and the
Schr\"odinger operator \citep{bh2020b}. Approximating principal Koopman
eigenfunctions yields coordinates for the evolution of observables
\citep{lee2025koopmaneigenfunctions}. Transport and Feynman--Kac
representations yield kernels adapted to these eigenfunction problems
in deterministic and stochastic systems, respectively
\citep{hamzi2025kernel_lions,hamzi2026feynman_kac,hamzi2026feynman_ieee}.
Algorithmic kernels give another geometry for Koopman and transfer
operators \citep{hamzi2026algorithmicoperators}. Stability analysis
estimates Lyapunov functions directly from noisy observations \citep{lyap_bh}
or constructs them from approximate Koopman eigenfunctions
\citep{lee2024note,giesl2026lyapunov_koopman}. Kernel methods can also enforce
Hamiltonian structure in trajectory learning \citep{jalalian_deds_2023}
or approximate the value function in nonlinear optimal control
\citep{hamzi2026hjb}. A distinct use of dynamical models treats the training
process itself as a system, studying limits of trained weights or the
policies needed to steer training \citep{smirnov2022mean,hamzi2026learningentropy}.
Here the criterion is preservation of the posterior over regime labels.

\subsection{Organization}
The paper has two parts. Part~\ref{part:theory} (Theory) defines the
method in Section~\ref{sec:skmd_method}, which also collects the notation
(Table~\ref{tab:notation}). Section~\ref{sec:mode_information_v31} treats
the information retained by mode coordinates and energies at fixed
quality, Section~\ref{sec:quality_theory_v32} the effect of changing
observation quality, and Section~\ref{sec:adaptive_theory_v32} learning,
approximation and adaptive coordinates, ending with numerical checks of the
bounds.

Part~\ref{part:applications} (Applications) begins with the experimental
design, the evidence overview and the statement of which analyses are
confirmatory (Section~\ref{sec:experimental_design}).
Section~\ref{sec:controlled_studies} tests the mechanisms on known modal
observations. Section~\ref{sec:learned_studies} adds learned modes and
nonstationary signals, Section~\ref{sec:recorded_studies} recorded ECG and
bearing signals, and Section~\ref{sec:controlled_benchmarks} the exact
energy-oracle and finite-sample coupled-mode benchmarks.
Section~\ref{sec:discussion} discusses both parts and
Section~\ref{sec:conclusion} concludes. The appendices give the protocols of
the two controlled experiments (Appendices~\ref{sec:controlled64supp}
and~\ref{sec:three_regime_protocol}), the computational record of all
comparisons (Appendix~\ref{sec:focused_inventory}) and the computational
reproduction and verification record (Appendix~\ref{sec:reproduction}).

\Needspace{18\baselineskip}
\part{Theory}
\label{part:theory}

\section{The sKMD extension of KMD}
\label{sec:skmd_method}
\subsection{Reference decomposition and a joint regime law}
Let a finite-dimensional representation of a KMD dictionary be
$H=[H_1\ \cdots\ H_J]\in\R^{p\times r}$, with coefficient blocks
$A=(A_1^\top,\ldots,A_J^\top)^\top$. The synthesized signal is
$f=HA=\sum_j H_jA_j$. A reference model with independent blocks has
\[
 A\sim\N(0,\Gamma),\qquad
 \Gamma=\operatorname{diag}(\Gamma_1,\ldots,\Gamma_J)\succ0,
 \qquad Q_j=H_j\Gamma_jH_j^\top.
\]
Here $Q_j$ is the component kernel. If a known observation operator $B_o$
acts on the signal, then $L_o=B_oH$ and
$y=L_oA+\epsilon_o$, with independent Gaussian error of known positive-definite
covariance $R_o$. Standard KMD conditioning gives a reference coefficient
posterior $\N(m,V)$ and mode estimates $H_jm_j$. The corresponding
conditional mode covariances come from the blocks of $V$; the observation
can couple them even when the reference prior has independent blocks.
A pooled reference may also use a full covariance $\Gamma$.

\begin{definition}[Stochastic KMD regime extension]
\label{def:skmd}
Stochastic KMD (sKMD) is a supervised probabilistic extension of KMD.
Fix a KMD coordinate model and its shared observation law
\eqref{eq:reference_mode_observation_v31}, with reference coefficient
posterior $\N(m,V)$. Learn joint coefficient laws $A\mid Y=k$ from labeled
training data for regimes $Y\in\{0,\ldots,K-1\}$ with priors $\pi_k>0$.
The Gaussian implementation uses $A\mid Y=k\sim\N(\mu_k,C_k)$ with
$C_k\succ0$. Transferring the reference posterior to these laws gives the regime
posterior
$\Prob(Y=k\mid y,o)\propto\pi_k\,p_k(y\mid o)/p_{\rm ref}(y\mid o)$, with
the evidence ratio \eqref{eq:posterior_transfer_v31}, the conditional mode
laws \eqref{eq:conditional_mode_law_v31}, and their mixture over the regime
posterior (Proposition~\ref{prop:posterior_transfer_v31}).
\end{definition}

Off-diagonal entries of $C_k$ represent dependence between coordinates.
They may encode relative modal polarity when a trigger or phase
convention aligns records. The physical convention must be available to
both fitting and prediction. An arbitrary sign choice that changes from
record to record can erase this signal. Class-dependent means and
variances can also distinguish regimes; coupling is one source of
information, not a necessary feature of every useful regime model.

The fixed-model statements below condition on a supplied dictionary or
on independent fitting data used to estimate it. Adaptation on the scored
record needs a separate selection model; its limitations are addressed
in Section~\ref{sec:adaptive_theory_v32}. Finite-rank notation describes
the implemented examples and does not assert a finite-rank restriction
on KMD as a general framework.

\subsection{Coupled inference from a KMD posterior mean and covariance}
\label{sec:posterior_transfer_v31}

A useful extension retains the full reference mode posterior when observation
operators or noise vary. Let a prescribed KMD reference model have
\begin{equation}
 A\sim\N(0,\Gamma),\qquad y=L_oA+\epsilon_o,\qquad
 \epsilon_o\sim\N(0,R_o),\quad \Gamma\succ0,\ R_o\succ0.
 \label{eq:reference_mode_observation_v31}
\end{equation}
The metadata $o$ specifies the observed mask/operator and noise covariance;
$\Gamma$ may be diagonal. Its reference posterior is $\N(m,V)$, where
\[
 V^{-1}=\Gamma^{-1}+L_o^\top R_o^{-1}L_o,\qquad
 V^{-1}m=L_o^\top R_o^{-1}y.
\]
Regime $k$ has a joint mode prior $\N(\mu_k,C_k)$, $C_k\succ0$.
Assume that $o$ has the same distribution in every regime, or condition on a
fixed $o$. Otherwise its class likelihood must be included explicitly.

\begin{proposition}[Exact posterior transfer through the KMD mode law]
\label{prop:posterior_transfer_v31}
Define
\[
 M_k=V^{-1}+C_k^{-1}-\Gamma^{-1}
     =C_k^{-1}+L_o^\top R_o^{-1}L_o\succ0,\qquad
 b_k=V^{-1}m+C_k^{-1}\mu_k.
\]
The class evidence relative to the reference is
\begin{align}
 \frac{p_k(y\mid o)}{p_{\rm ref}(y\mid o)}
 &=\frac{|\Gamma|^{1/2}}
         {|C_k|^{1/2}|V|^{1/2}|M_k|^{1/2}}\nonumber\\
 &\quad\times\exp\!\left\{\tfrac12
 [b_k^\top M_k^{-1}b_k-m^\top V^{-1}m-\mu_k^\top C_k^{-1}\mu_k]
 \right\}.
 \label{eq:posterior_transfer_v31}
\end{align}
Normalizing these ratios times $\pi_k$ gives the regime posterior, and
\begin{equation}
 A\mid y,o,Y=k\sim\N(M_k^{-1}b_k,M_k^{-1}).
 \label{eq:conditional_mode_law_v31}
\end{equation}
Thus $(m,V)$ is an exact interface for both regime inference and
regime-conditioned reconstruction in this declared model family.
\end{proposition}
\begin{proof}
Bayes' formula gives the change-of-prior identity
\[
 \frac{p_k(y\mid o)}{p_{\rm ref}(y\mid o)}
 =\E_{A\sim\N(m,V)}\frac{p_k(A)}{p_{\rm ref}(A)}.
\]
Collecting the quadratic and linear terms of its Gaussian integral gives
$M_k,b_k$. Completing the square gives the determinant and exponential in
\eqref{eq:posterior_transfer_v31}, and the normalized integrand is the stated
conditional Gaussian.
\end{proof}

The reference mean alone is generally insufficient when $o$ varies. At the
same $m=0$, different posterior covariances $V$ give different determinant
evidence for latent variance classes; continuity makes the distinction persist
in neighborhoods. The posterior covariance records what a small coefficient
estimate means under its actual observation quality. Conversely, with one
fixed observation operator and fixed noise, $V$ is constant and $m$ determines
$V^{-1}m$; the reference mean already preserves the class information in this
model, even if the operator is rank deficient. The heterogeneous experiment
can test the benefit of retaining this observation uncertainty; comparisons
must also account for covariance estimation and regularization.

The final reconstructed mode law averages the conditional Gaussians over the
regime posterior. Its covariance includes both conditional uncertainty and
the covariance of the regime-specific means. The construction uses standard
Gaussian conditioning, change of prior, and EM for latent Gaussian parameters.
Common factor models and extreme deconvolution provide the corresponding
generative families and heterogeneous-observation likelihoods
\cite{baek_commonfactor_2010,bovy_deconvolution_2011}. The sKMD extension
expresses this inference through the KMD mode posterior, with matched dense
likelihoods and simpler mode features as controls.


\subsection{Reduction to KMD and the role of the controls}
\label{sec:reference_reduction}
If every regime law equals the reference law, $\mu_k=0$ and
$C_k=\Gamma$, then $M_k=V^{-1}$, $b_k=V^{-1}m$ and every evidence ratio
in \eqref{eq:posterior_transfer_v31} equals one. The regime posterior is
$\pi_k$, and the coefficient posterior is exactly $\N(m,V)$. Thus sKMD
contains the reference KMD calculation as an exact special case.

Removing covariance off-diagonals gives an independent-coordinate regime
ablation, but does not generally recover the reference model: the means
or diagonal variances can still differ by regime. A single pooled full
prior gives a stronger reconstruction control than separate independent
mode priors. A signed point classifier retains coefficient signs while
replacing the observation integration by a specified point-inference rule.
Each experiment states that rule, its calibration and its fitting budget.

\paragraph{Implementation checks.}
A matched latent-Gaussian calculation in observation space, with the same
regime laws, observation model and calibration, gives the same
probabilities and conditional means as the posterior transfer. In every
study of Part~\ref{part:applications}, the integrated predictions were
compared with such a direct dense Gaussian implementation (for the
finite-mixture, dictionary-mixture and hidden-Markov extensions, with the
matched dense mixture or Gaussian HMM). The largest recorded discrepancy is
$4.6\times10^{-7}$ in log odds (MIT--BIH ECG study); the values for every
study are listed in Appendix~\ref{sec:reproduction}. This agreement
verifies the implementation
of an algebraic identity; it is not evidence that the working model fits the
data, and it is not repeated in the individual studies.

\subsection{Fitting and computation}

The mode dictionary may encode known dynamics or be learned. For example,
an equilibrium covariance supplies a dynamics-informed Gaussian-process
prior for stochastic Navier--Stokes fields \citep{hamzi2025navierstokes}.
Parametric and nonparametric Kernel Flows adapt kernels by comparing fits
across subsamples \citep{BHPhysicaD,darcy2025nonparametric}. Sparse Kernel
Flows select among base kernels \citep{bh_sparse_kfs,hamzi2024aitkernels},
while kernel sum-of-squares optimization searches for kernel parameters
using a surrogate objective \citep{lengyel2025ksos}. Compression gives
another design principle for unsupervised kernels
\citep{hamzi2025clustering}. Work in this direction relates description
length to kernel spectra and learning rates
\citep{hamzi2025learningtheory,hamzi2026solomonofflearning}, or uses
program-length weights to construct Gaussian-process covariances
\citep{hamzi2026solomonoff}. A survey places these constructions among
other approximations to universal prediction \citep{hamzi2026solomonoffsurvey}.
These methods address representation design. The class-law fit below
conditions on a chosen dictionary.

Let $H\in\R^{p\times r}$ contain supplied carriers or a basis fitted
using only training observations. The class-law estimator depends on the
training observation process. For example, with known zero class means,
complete training windows, a common known noise covariance $N$, and
$H^\top H=I_r$, a moment estimator is
\[
 \widetilde C_k=\frac1{n_k}\sum_{i:Y_i=k}
 (H^\top v_i)(H^\top v_i)^\top-H^\top NH.
\]
With the same known centering and heterogeneous complete observations,
replace the final term by the average of $H^\top N_iH$ over the class.
Project eigenvalues onto $[\tau,\infty)$ to obtain a positive-definite
estimate, then optionally shrink off-diagonal entries using development
data. Unknown means require fitting them and accounting for their
estimation error and the effect of centering on the covariance estimator.
Each experiment specifies its fitting rule; the 64-panel study estimates
means and uses its declared finite-sample moment rule.

A learned truncated basis can omit signal outside its span; the theory's
loading-error term exposes that misspecification. Missing training entries
require an appropriate latent fit, such as established extreme-deconvolution
likelihood procedures \citep{bovy_deconvolution_2011}. The complete-window
moment example above is not a general missing-data estimator.

Kernel flow-map estimators accommodate irregular sample times
\citep{lee2023irregular}, and kernel graph completion uses partial state
observations to estimate dynamics \citep{hamzi2023partial}. Fitting drift
and diffusion from one trajectory \citep{darcy2023oneshot}, forecasting
with single learned kernels or ensembles \citep{hamzi2021geophysical,wei2026ensemble},
and assimilating longitudinal clinical measurements \citep{batlle2026alzheimer}
use kernel models to predict state evolution. Related trajectory tasks
include initializing reservoirs from short observation histories
\citep{grigoryeva2024cold} and inferring the past of a chaotic system
using algorithmic information \citep{dingle2026retrodicting}. Given a
chosen dictionary and observation law, sKMD uses its fitted class laws
to infer regime membership and latent mode coefficients.

\paragraph{Coupled sKMD procedure.}
\begin{enumerate}[leftmargin=*]
\item Fit $H$, the class priors, and joint coefficient laws $(\mu_k,C_k)$ on the fitting units. Select declared regularization parameters using development units. Choose a positive-definite reference covariance $\Gamma$.
\item For a new observation with operator $L_o$ and noise $R_o$, obtain the reference KMD posterior $(m,V)$. Retain the full covariance, including off-diagonal entries.
\item Form $M_k,b_k$ and the evidence ratios in \eqref{eq:posterior_transfer_v31}; normalize with the class priors to obtain $p_k$.
\item Return the regime probabilities and the reconstructed coefficient mean
$\bar a=\sum_kp_kM_k^{-1}b_k$. The coefficient covariance is
\[
 \sum_kp_k\left[M_k^{-1}+
 (M_k^{-1}b_k-\bar a)(M_k^{-1}b_k-\bar a)^\top\right].
\]
Apply the synthesis map to obtain the signal mean and covariance.
\end{enumerate}
The reference $\Gamma$ defines the intermediate KMD calculation, but cancels from the exact final class likelihood when the same observation model is used throughout. The implementation evaluates the small positive-definite systems and uses Cholesky log determinants. Once $L_o^\top R_o^{-1}L_o$ is formed, the class calculations cost $O(Kr^3)$ per distinct observation configuration and $O(Kr^2)$ for repeated observations with that configuration. Computing the Gram matrix costs $O(pr^2)$ for diagonal noise and a mask. This replaces an ambient $p\times p$ calculation by mode-dimensional systems
when $r\ll p$. Dictionary and class-covariance estimation contribute the
separate fitting cost.


The exact posterior identities refer to the untempered model. Experiments
that calibrate class scores with a development-selected temperature report
that predictive variant; such calibration is separate from the
change-of-prior identity.

\subsection{Extensions used in the experiments}
A finite Gaussian scale mixture replaces a class law by a weighted sum
of Gaussian laws. Applying the same evidence calculation to each
component and summing gives a robust integrated classifier and its
conditional reconstruction. A declared discrete dictionary distribution
can likewise be integrated with normalized evidence. A Markov regime
prior supports filtering through a specified transition model. These
are established mixture and state-inference operations; their value here
is evaluated through matched pooled, point and dictionary controls.
The exact Gaussian representation theorems apply to their stated
Gaussian families. They do not automatically cover an arbitrary fitted
mixture, non-Gaussian law or adaptive dictionary.

\subsection{Notation}
\label{sec:notation}
Table~\ref{tab:notation} collects the symbols used in more than one place.
A few letters are reused for local constants inside a single proof or
example; the table gives their scope. Natural logarithms are used
throughout, so information and NLL are in nats.

\begin{table}[htbp]
\centering\small
\caption{Notation. ``Scope'' gives the sections, or the statements, in which
a symbol has the stated meaning.}
\label{tab:notation}
\renewcommand{\arraystretch}{1.1}
\begin{tabular}{@{}>{\raggedright\arraybackslash}p{0.22\linewidth}>{\raggedright\arraybackslash}p{0.50\linewidth}>{\raggedright\arraybackslash}p{0.22\linewidth}@{}}
\toprule
Symbol & Meaning & Scope\\
\midrule
$Y$, $K$, $\pi_k$ & regime label in $\{0,\ldots,K-1\}$, number of regimes, regime priors & throughout\\
$A$; $C$, $c$ & latent mode coefficients; signed mode coordinates (random, observed value) & \S\ref{sec:skmd_method}; \S\ref{sec:mode_information_v31}--\ref{sec:adaptive_theory_v32}\\
$H$, $L_o$, $R_o$ & dictionary (synthesis) matrix, observation operator, observation-noise covariance & \S\ref{sec:skmd_method} and Part~\ref{part:applications}\\
$\Gamma$; $(m,V)$ & reference KMD prior covariance; reference posterior mean and covariance & throughout\\
$\mu_k$, $C_k$; $P_k$ & regime mean and covariance of the mode coefficients; regime precision & throughout\\
$M_k$, $b_k$ & regime posterior precision and information vector in the transfer formula & Proposition~\ref{prop:posterior_transfer_v31}\\
$\mathcal E(c)$, $E$ & coordinate energies $(c_1^2,\ldots,c_r^2)$ & \S\ref{sec:mode_information_v31}--\ref{sec:adaptive_theory_v32}\\
$\Delta_{\log}$, $\Delta_{\rm Br}$ & Bayes log-loss and Brier gaps of a summary & \S\ref{sec:mode_information_v31}\\
$\alpha$; $b_\alpha$ & interaction strength of the three-regime family \eqref{eq:three_class_precision_v31}; $\tfrac12\log(1-\alpha^2)$ & throughout; \S\ref{sec:mode_information_v31}\\
$\kappa$, $q$; $\varrho$ & condition-number bound and normalized condition number; radius of the integration ball & Theorems~\ref{thm:energy_uniform_gap_v31}--\ref{thm:conditioned_basis_gap}\\
$D$, $Z$ & diagonal (or general) observation-noise covariance; noisy coordinates & \S\ref{sec:quality_theory_v32}\\
$\lambda_-$, $\lambda_+$ & eigenvalue bounds on the observed covariances & Proposition~\ref{prop:stable_quality_extension}\\
$b$, $R$, $\tau$, $\beta_k$ & shared baseline off-diagonal precision, noise shape, noise scale, score coefficients & Theorem~\ref{thm:quality_expansion_v32}\\
$\mathsf D_U$ & divergence between the laws of a statistic $U$ under $p_\epsilon$ and $p_0$ & Lemma~\ref{lem:four_kl}\\
$D_k$, $B_k$, $\epsilon$ & diagonal baseline precision, interaction perturbation, its size & \S\ref{sec:adaptive_theory_v32}\\
$\bar b$, $\bar\beta$, $\lambda$ & $\max_k\|B_k\|_{\rm op}$, $\bar b/\lambda$, $\min_k\lambda_{\min}(D_k)$ & Theorem~\ref{thm:weak_interactions_v32}\\
$\Sigma_m$ & baseline covariance $\Gamma-V$ of the reference mean $m$ & \S\ref{sec:uncertainty_gain_v32}\\
$\Pi_j$, $L_j$, $L_*$, $G_D$ & candidate span projectors, signal leakage, its library optimum, expected leakage bound & \S\ref{sec:finite_kmd_selector_v32}\\
$W(q,n)$ & expected Wishart covariance-estimation loss & Theorem~\ref{thm:learned_benefit_v32}\\
$\rho$, $s$, $T(s)$ & binary correlation, noise variance, quality temperature & \S\ref{sec:quality_calibration}\\
panel & one independently generated and independently fitted simulation replicate & Part~\ref{part:applications}\\
\bottomrule
\end{tabular}
\end{table}

\section{Information retained by mode coordinates and energies}
\label{sec:mode_information_v31}
The sKMD construction retains a signed coordinate law. We now identify the
information lost if that law is replaced by separate energies. The results
concern the declared coordinate representation; KMD itself can retain the
signed reconstructions. We first give the fixed-quality criterion and its
multiclass obstruction, then quantify the cost of energy reduction.

\subsection{What the prescribed KMD energies retain}

For two orthonormal carriers collected in $H$, component kernels
$Q_i=\phi_i\phi_i^\top$, and isotropic noise kernel $\sigma^2I$, put
$S=\sum_iQ_i+\sigma^2I$. Prescribed KMD gives
\[
 w_i=Q_iS^{-1}v=\phi_i c_i,\qquad
 c_i=\frac{\phi_i^\top v}{1+\sigma^2},\qquad
 E_i=v^\top S^{-1}Q_iS^{-1}v=c_i^2.
\]
The signed vector $c$ contains more information than its separate energies.
For general linearly independent rank-one carriers the coefficient map is
still a nonsingular coordinate map on the informative mode subspace, after
whitening the common observation noise. The following characterization is
stated directly in those signed coordinates.

Let $Y\in\{0,\ldots,K-1\}$ have strictly positive priors, and suppose
\[
 C\mid Y=k\sim\N(\mu_k,\Sigma_k),\qquad \Sigma_k\succ0.
\]
In this section $P_k=\Sigma_k^{-1}$ denotes a precision matrix. Set
$A_k=P_k-P_0$ and $d_k=P_k\mu_k-P_0\mu_0$. The log odds are
\begin{equation}
 \ell_k(c)=\log\frac{\Prob(Y=k\mid c)}{\Prob(Y=0\mid c)}
 =\beta_k-\tfrac12c^\top A_kc+d_k^\top c.
 \label{eq:mode_logodds_v31}
\end{equation}
Write $\mathcal E(c)=(c_1^2,\ldots,c_r^2)$; positive fixed rescalings of
these coordinates do not affect sufficiency.

\begin{theorem}[When separate energies preserve the regime posterior]
\label{thm:energy_sufficiency_v31}
The posterior $\Prob(Y=\cdot\mid C)$ is a measurable function of
$\mathcal E(C)$ if and only if $d_k=0$ and $A_k$ is diagonal for every $k$.
For an invertible coordinate map $z=Tc$, replace these conditions by
$T^{-\top}d_k=0$ and diagonality of $T^{-\top}A_kT^{-1}$.
In the centered case, an orthogonal such map exists if and only if the
precision differences $A_k$ commute pairwise.
\end{theorem}

\begin{proof}
Outside the coordinate hyperplanes, the fibers of $\mathcal E$ are precisely
independent sign changes of the coordinates. All class densities are positive.
Posterior sufficiency is therefore equivalent to invariance of every log odd
under these sign changes, almost everywhere. Their polynomial form makes the
invariance an identity everywhere. Flipping coordinate $i$ forces its linear
coefficient and all cross-product coefficients involving $i$ to vanish.
This gives $d_k=0$ and $(A_k)_{ij}=0$ for $i\ne j$. Conversely the remaining
polynomial is a function of the squares. Substitution $c=T^{-1}z$ proves the
coordinate version. Finally commuting real symmetric matrices admit a common
orthonormal eigenbasis: each matrix preserves the eigenspaces of the others,
within which one diagonalizes successively. Diagonal matrices commute, giving
necessity as well.
\end{proof}

This is a decision-specific application of simultaneous diagonalization
\cite{jiang_simultaneous_2016}. It concerns \emph{precision differences}; a
shared off-diagonal part of all precisions cancels from the posterior.
Common covariance eigenvectors, as in common-principal-component and
common-orientation discriminant models
\cite{flury_common_1984,bensmail_eigenvalue_1996}, are sufficient but stronger
than this condition. For two centered classes a single symmetric difference
always diagonalizes. A binary example alone cannot establish that every
rotated energy basis loses the posterior.

\subsection{A three-regime obstruction in two modes}

Choose $0<\alpha<1$, equal priors, zero means and precisions
\begin{equation}
 P_0=I_2,\qquad
 P_1=I_2+\alpha\begin{pmatrix}1&0\\0&-1\end{pmatrix},\qquad
 P_2=I_2+\alpha\begin{pmatrix}0&1\\1&0\end{pmatrix}.
 \label{eq:three_class_precision_v31}
\end{equation}
All are positive definite. With $b_\alpha=\tfrac12\log(1-\alpha^2)<0$,
\begin{equation}
 \ell_1(c)=b_\alpha-\tfrac\alpha2(c_1^2-c_2^2),\qquad
 \ell_2(c)=b_\alpha-\alpha c_1c_2.
 \label{eq:three_class_logodds_v31}
\end{equation}

\begin{theorem}[No sufficient nonredundant real two-energy basis]
\label{thm:three_class_obstruction_v31}
For \eqref{eq:three_class_precision_v31} and every invertible real
$2\times2$ matrix $T$, the posterior is not measurable with respect to
$((TC)_1^2,(TC)_2^2)$.
\end{theorem}
\begin{proof}
If a congruence diagonalized both $A_1=P_1-P_0$ and $A_2=P_2-P_0$, the
determinant of their pencil would factor into two real linear forms.
However
\[
 \det(tA_1+sA_2)=-\alpha^2(t^2+s^2).
\]
A congruence multiplies this expression by a nonzero positive constant.
The displayed quadratic has no real linear factor. Thus no such congruence
exists, and Theorem~\ref{thm:energy_sufficiency_v31} applies.
\end{proof}

The pencil argument is an elementary use of classical simultaneous-congruence
theory. Its consequence here is that the binary rotation construction does not extend
to all three regimes through one nonredundant two-coordinate energy map.
The restriction also covers two more general nonnegative quadratic summaries
of these same coefficient coordinates.

\begin{corollary}[Two positive quadratic mode energies]
\label{cor:two_quadratic_energies_v31}
For the three-class example, let $Q_1,Q_2\succeq0$ be $2\times2$ matrices
with $Q_1+Q_2\succ0$. The pair $(C^\top Q_1C,C^\top Q_2C)$ does not
preserve the regime posterior.
\end{corollary}
\begin{proof}
Whiten $Q_1+Q_2$ to the identity and orthogonally diagonalize the first
transformed matrix. The second is its complement to the identity and is
diagonal as well. Both energies are therefore functions of squared
coordinates in an invertible basis, which already lose posterior information
by Theorem~\ref{thm:three_class_obstruction_v31}. A further function of those
coordinates cannot restore it.
\end{proof}

This corollary concerns quadratic forms on the two-dimensional mode
coordinates; it makes no claim about additional ambient residual features.
An additional total energy is another statistic when it is not determined
by the two reported energies; including it can change the conclusion.
For the original pair, optimal log loss and Brier risk are strictly worse than those of the
full coordinates, by \eqref{eq:energy_information_gap_v31}--\eqref{eq:energy_brier_gap_v31}
below. No positive constant uniform over all such quadratic forms is asserted.
The example embeds in a noisy mode model without singular priors. For
orthonormal $H\in\R^{p\times2}$, choose
$0<\sigma^2<1/(1+\alpha)$ and let
\begin{equation}
 v=HA+\eta,\qquad
 A\mid Y=k\sim\N(0,P_k^{-1}-\sigma^2I_2),\qquad
 \eta\sim\N(0,\sigma^2I_p).
 \label{eq:three_class_embedding_v31}
\end{equation}
Every latent covariance is positive definite; $H^\top v$ has covariance
$P_k^{-1}$. The orthogonal residual is independent class-invariant noise.
The shared scale of its KMD coefficients leaves the obstruction unchanged.

\subsection{An information gap, including a uniform quantitative bound}

For a statistic $E$ of $C$, write $\eta(C)=\Prob(Y=\cdot\mid C)$ and
$\eta_E(E)=\Prob(Y=\cdot\mid E)$. With natural logarithms, the optimal
log-loss difference is the standard conditional-information identity
\begin{equation}
 \Delta_{\log}(E)
 =H(Y\mid E)-H(Y\mid C)
 =I(Y;C\mid E)
 =\E\,D_{\rm KL}(\eta(C)\|\eta_E(E)).
 \label{eq:energy_information_gap_v31}
\end{equation}
It is strictly positive exactly when the posterior is not retained. For the
multiclass Brier loss $\sum_k(q_k-\bm1_{Y=k})^2$, the corresponding gap is
\begin{equation}
 \Delta_{\rm Br}(E)=\E\|\eta(C)-\eta_E(E)\|_2^2.
 \label{eq:energy_brier_gap_v31}
\end{equation}
These are proper-score statements. Posterior information loss alone does not
prove a particular excess classification-error rate.

\begin{theorem}[Uniform loss over orthogonal two-mode dictionaries]
\label{thm:energy_uniform_gap_v31}
For \eqref{eq:three_class_precision_v31}, let
$E_O(c)=((Oc)_1^2,(Oc)_2^2)$ with $O\in O(2)$. For any $\varrho>0$ put
\[
 J(\varrho)=1-e^{-\varrho^2/2}\left(1+\frac{\varrho^2}{2}+\frac{\varrho^4}{8}\right).
\]
Then every orthogonal dictionary satisfies
\begin{equation}
 \Delta_{\log}(E_O)
 \ \ge\ \frac{\alpha^2(1-\alpha^2)}{162}
              e^{-2\alpha \varrho^2}J(\varrho)>0.
 \label{eq:energy_uniform_bound_v31}
\end{equation}
The Brier gap is at least twice the displayed lower bound.
\end{theorem}

\begin{proof}
Set $z=Oc$. Under class zero $z\sim\N(0,I_2)$, and the four signs are
uniform given their squared coordinates. The sign-varying part of
$\ell_k$ is $-(OA_kO^\top)_{12}z_1z_2$. Direct rotation of the two matrices
in \eqref{eq:three_class_precision_v31} gives
\[
 \sum_{k=1}^2 (OA_kO^\top)_{12}^2=\alpha^2.
\]
Thus the sum of their conditional sign variances is
$\alpha^2z_1^2z_2^2$, independent of $O$.

On $\|z\|\le \varrho$, the range of the three log scores is at most
$|b_\alpha|+\alpha \varrho^2$. Every posterior component is therefore at least
$m_\varrho=e^{-|b_\alpha|-\alpha \varrho^2}/3$. The map
$q\mapsto(\log(q_1/q_0),\log(q_2/q_0))$ has Lipschitz constant at most
$\sqrt3/m_\varrho$ on this convex probability set: its derivative is
$B\operatorname{diag}(1/q)$, where $B$ has rows $e_1-e_0,e_2-e_0$ and
$\|B\|_{\rm op}=\sqrt3$. Applying the pairwise variance identity to the
four signs gives
\[
 \operatorname{Var}_{\rm sign}(\eta)
 \ge\frac{m_\varrho^2}{3}\operatorname{Var}_{\rm sign}(\ell),
\]
where a vector variance means the sum of coordinate variances.

The mixture density of $z$ is at least one third of the class-zero
density $\phi_0$, the standard bivariate normal density, and the ball is
energy-measurable.
Minimizing over all energy-based predictors therefore yields
\begin{align*}
 \Delta_{\rm Br}(E_O)
 &\ge\frac13\inf_f\E_0[
       \|\eta(O^\top z)-f(z_1^2,z_2^2)\|_2^2\bm1_{\|z\|\le \varrho}]\\
 &\ge\frac{\alpha^2m_\varrho^2}{9}
       \E_0[z_1^2z_2^2\bm1_{\|z\|\le \varrho}].
\end{align*}
Polar integration gives the last expectation as $J(\varrho)$: its angular factor
is $1/8$, and its squared radius has a chi-square law with two degrees of
freedom. Since $e^{-2|b_\alpha|}=1-\alpha^2$, this proves the Brier bound.
Pinsker's inequality, followed by $\|u\|_1\ge\|u\|_2$, gives the stated
log-loss bound.
\end{proof}

The constant is deliberately conservative. Section~\ref{sec:energy_oracle_v31}
evaluates the actual oracle proper-score gaps; it does not use this lower bound as an
estimate of their magnitude. The bound applies uniformly to orthogonal
bases. An important limitation arises for arbitrary ill-conditioned maps:
\begin{equation}
 T_\epsilon=\begin{pmatrix}1&\epsilon\\1&-\epsilon\end{pmatrix}.
 \label{eq:ill_conditioned_energy_v31}
\end{equation}
Its two energies reveal $c_1c_2$ and $c_1^2+\epsilon^2c_2^2$. Apart from a
common sign, the alternative orbit point is
$(\epsilon c_2,c_1/\epsilon)$. For almost every fixed $c$, its density under
every class tends to zero as $\epsilon\downarrow0$. The energy posterior
then approaches the full posterior. Bounded categorical entropy gives a
log-loss gap tending to zero. Thus Theorem~\ref{thm:three_class_obstruction_v31}
is an exact statement for each invertible map, and does not imply a uniform
positive gap over unbounded condition numbers.

\subsection{A uniform gap for invertible bases with bounded condition number}
\label{sec:conditioned_basis_extension}

The orthogonal bound extends to a specified range of conditioning. This
addresses the distinction between an exact obstruction for each invertible
basis and a quantitative guarantee over a family of bases. The result uses
the same three Gaussian regimes as \eqref{eq:three_class_precision_v31}, with
$0<\alpha<1$ and equal priors. It concerns a fixed representation of each
observation, independently of the test vector being classified.

For an invertible real $2\times2$ matrix $T$, write
\[
 \mathcal E_T(c)=((Tc)_1^2,(Tc)_2^2),\qquad
 \operatorname{cond}_2(T)=\|T\|_{\rm op}\|T^{-1}\|_{\rm op}.
\]
Let $\Delta_{\log}(\mathcal E_T)$ and $\Delta_{\rm Br}(\mathcal E_T)$ be the
Bayes log-loss and multiclass Brier gaps defined in
\eqref{eq:energy_information_gap_v31}--\eqref{eq:energy_brier_gap_v31}.

\begin{theorem}[Conditioning-controlled energy loss]
\label{thm:conditioned_basis_gap}
For every $\kappa\ge1$ and every invertible $T$ satisfying
$\operatorname{cond}_2(T)\le\kappa$,
\begin{equation}
 \Delta_{\log}(\mathcal E_T)
 \ge
 \frac{\alpha^2(1-\alpha^2)e^{-3}}{144(4\alpha+1)^3}
 \left(\frac1\kappa+\frac2{\kappa^3}+\frac1{\kappa^5}\right)>0.
 \label{eq:conditioned_basis_gap}
\end{equation}
The Brier gap is at least twice the right-hand side.
A sharper version uses $q$, the condition number after normalizing the two
rows of $T$ to equal Euclidean norm, in place of $\kappa$.
This normalization has $1\le q\le\operatorname{cond}_2(T)$.
\end{theorem}

\begin{proof}
We first establish the normalization and then bound the posterior variation
on sign orbits. Multiplying either row of $T$ by a positive constant multiplies
one reported energy by a known positive constant. The two energy vectors
therefore generate the same sigma-field, so both Bayes loss gaps are unchanged.
Normalize each row to unit norm, then multiply the whole matrix by a positive
constant to obtain a matrix $\widetilde T$ with $|\det\widetilde T|=1$.

To check the condition number, let the original row norms be $a,b>0$ and
let $s>0$ be the absolute sine of their angle. If the original condition
number is $\kappa_T$, its singular values give
\[
 \kappa_T+\kappa_T^{-1}
 =\frac{\|T\|_F^2}{|\det T|}
 =\frac{a^2+b^2}{a\,b\,s}
 \ge\frac2s=q+q^{-1}.
\]
The function $x+x^{-1}$ increases on $[1,\infty)$, hence $q\le\kappa_T$.
The equal row norms and unit absolute determinant imply
\begin{equation}
 \widetilde T\widetilde T^\top
 =\begin{pmatrix}g&h\\h&g\end{pmatrix},\qquad
 g^2-h^2=1,\qquad g=\frac{q+q^{-1}}2.
 \label{eq:balanced_basis_gram}
\end{equation}
The eigenvalues of this Gram matrix are $q$ and $q^{-1}$; thus
$\|\widetilde T^{-1}\|_{\rm op}^2=q$.

Put $z=\widetilde Tc$ and let $u,v$ be the columns of
$\widetilde T^{-1}$. For the two precision contrasts
$A_k=P_k-P_0$ in \eqref{eq:three_class_precision_v31}, direct multiplication
gives
\begin{align}
 \sum_{k=1}^2
 \bigl[(\widetilde T^{-\top}A_k\widetilde T^{-1})_{12}\bigr]^2
 &=\alpha^2\{(u_1v_1-u_2v_2)^2+(u_1v_2+u_2v_1)^2\}\nonumber\\
 &=\alpha^2\|u\|^2\|v\|^2=\alpha^2g^2.
 \label{eq:balanced_contrast_identity}
\end{align}
The last equality follows because the squared column norms of
$\widetilde T^{-1}$ are the diagonal entries of
$(\widetilde T\widetilde T^\top)^{-1}$, both equal to $g$.
Consequently the sum of the variances of the two log odds over the four
\emph{uniformly weighted} signs of $z$ is $\alpha^2g^2z_1^2z_2^2$.
These uniform signs will be used with an auxiliary isotropic Gaussian
measure; the transformed class-zero law itself can have unequal sign weights.

Let $\phi_I$ denote the standard bivariate Gaussian density. Since the
absolute determinant is one, the density $p_{0,z}$ under class zero satisfies
on the ball $\|z\|\le \varrho$
\begin{equation}
 p_{0,z}(z)
 =\frac1{2\pi}\exp\{-\|\widetilde T^{-1}z\|^2/2\}
 \ge e^{-(q-1)\varrho^2/2}\phi_I(z).
 \label{eq:balanced_density_minorization}
\end{equation}
Write $\eta(z)=\Prob(Y=\cdot\mid C=\widetilde T^{-1}z)$ and
$b_\alpha=\tfrac12\log(1-\alpha^2)$. On this ball,
$\|c\|^2\le q\varrho^2$. The range of the three log scores is therefore at most
$|b_\alpha|+\alpha q\varrho^2$, and every component of $\eta(z)$ is at least
\[
 m_{q,\varrho}=\frac{\sqrt{1-\alpha^2}}3e^{-\alpha q\varrho^2}.
\]
As in the orthogonal-basis proof, the map from posterior probabilities to
the two log odds has Lipschitz constant at most $\sqrt3/m_{q,\varrho}$ on this
convex probability set. The pairwise variance identity and
\eqref{eq:balanced_contrast_identity} imply
\[
 \operatorname{Var}_{\rm sign}(\eta)
 \ge\frac{m_{q,\varrho}^2}{3}\alpha^2g^2z_1^2z_2^2.
\]

The actual mixture density is at least $p_{0,z}/3$. Minimize the squared
posterior error over all predictors $f(z_1^2,z_2^2)$, restrict to the ball,
and then apply \eqref{eq:balanced_density_minorization}. Under $\phi_I$
the signs are uniform conditional on the squared coordinates, and the ball
is measurable in those coordinates. These facts give
\begin{align*}
 \Delta_{\rm Br}(\mathcal E_T)
 &\ge\frac{e^{-(q-1)\varrho^2/2}}3
       \inf_f\int_{\|z\|\le \varrho}\|\eta(z)-f(z_1^2,z_2^2)\|^2
                         \phi_I(z)\,dz\\
 &\ge\frac{\alpha^2(1-\alpha^2)g^2}{81}
       e^{-[2\alpha q+(q-1)/2]\varrho^2}J(\varrho),
\end{align*}
where the same radial integral as in Theorem~\ref{thm:energy_uniform_gap_v31}
is
\[
 J(\varrho)=\int_{\|z\|\le \varrho}z_1^2z_2^2\phi_I(z)\,dz
 =\frac18\int_0^\varrho r^5e^{-r^2/2}\,dr
 \ge e^{-\varrho^2/2}\frac{\varrho^6}{48}.
\]
Pinsker's inequality bounds the log-loss gap below by half this Brier bound.
Combining the exponents and setting
$\varrho^2=6/[q(4\alpha+1)]$ therefore yields
\[
 \Delta_{\log}(\mathcal E_T)
 \ge\frac{\alpha^2(1-\alpha^2)e^{-3}}{36(4\alpha+1)^3}
          \frac{g^2}{q^3}
 =\frac{\alpha^2(1-\alpha^2)e^{-3}}{144(4\alpha+1)^3}
          (q^{-1}+2q^{-3}+q^{-5}).
\]
The expression in parentheses decreases for $q\ge1$. Since $q\le\kappa$,
this proves the asserted bound. The Brier calculation gives twice the same
lower bound directly.
\end{proof}

\begin{corollary}[Sharp conditioning rate]
\label{cor:sharp_conditioning_rate}
For the family \eqref{eq:three_class_precision_v31} at fixed $0<\alpha<1$,
\[
 \inf_{\operatorname{cond}_2(T)\le\kappa}\Delta_{\log}(\mathcal E_T)
 =\Theta_{\alpha}(\kappa^{-1})\qquad(\kappa\to\infty).
\]
\end{corollary}
\begin{proof}
Theorem~\ref{thm:conditioned_basis_gap} gives the lower bound. For the upper
bound, use $T_\epsilon$ from \eqref{eq:ill_conditioned_energy_v31} with
$\epsilon=\kappa^{-1}$. Write $p$ for the mixture density, $\eta(c)$ for the
class posterior, and $J_\epsilon c=(\epsilon c_2,c_1/\epsilon)$ for the
alternative point in the energy fiber. Away from the zero-energy boundary,
which has probability zero, the energy map has a common absolute Jacobian
within each fiber.

Since the centered class laws are even, the energy posterior averages $\eta(c)$ and $\eta(J_\epsilon c)$ with weights
proportional to $p(c)$ and $p(J_\epsilon c)$, respectively. Each component of
this posterior is at least $\eta_k(c)p(c)/(p(c)+p(J_\epsilon c))$. Hence
\begin{align*}
 \Delta_{\log}(\mathcal E_{T_\epsilon})
 &\le\int p(c)\log\!\left(1+\frac{p(J_\epsilon c)}{p(c)}\right)\,dc\\
 &\le\int\sqrt{p(c)p(J_\epsilon c)}\,dc
 \le\frac{2\epsilon}{(1-\alpha)(1+\epsilon^2)}.
\end{align*}
The second inequality uses $\log(1+t)\le\sqrt t$ for $t\ge0$. The last
integrates the Gaussian envelope
$p(c)\le(2\pi)^{-1}\exp[-(1-\alpha)\|c\|^2/2]$ at the two orbit points.
The singular values of $T_\epsilon$ are $\sqrt2$ and $\sqrt2\epsilon$,
so this choice lies in the declared conditioning budget.
\end{proof}

The constant in the uniform lower bound is conservative and describes a population proper-score gap.
It does not estimate the numerical size of a fitted classifier's improvement
or imply a zero--one error gap. Row scaling can be removed before judging the geometry:
rescaling a reported energy changes numerical units but leaves its information
unchanged. The theorem supplies a quantitative conditioning-restricted extension
of the same sign-orbit argument. Redundant energies and observation-dependent
basis selection remain different representations.

\paragraph{Size of the explicit bounds.}
The explicit constants in Theorems~\ref{thm:energy_uniform_gap_v31}
and~\ref{thm:conditioned_basis_gap} are far below the gaps they bound.
Table~\ref{tab:bound_vs_gap} sets them beside the measured known-law gaps of
Section~\ref{sec:energy_oracle_v31}: at $\alpha=0.8$ and $\kappa=1$,
Theorem~\ref{thm:conditioned_basis_gap} gives $4.30\times10^{-6}$ nats,
more than three orders of magnitude below the measured gap of the
development-selected orthogonal rotation, $0.0262$ nats.
Theorem~\ref{thm:energy_uniform_gap_v31}, which covers the orthogonal case,
gives the sharper $5.23\times10^{-6}$ nats when maximized over the radius
($\varrho^\ast\approx1.237$); the numerically estimated minimum orthogonal population gap,
$0.0259$ nats, is larger than these two bounds by factors of about $6{,}000$
and $5{,}000$. At $\alpha=0.95$ and $\kappa=1/0.03$,
Theorem~\ref{thm:conditioned_basis_gap} gives $8.27\times10^{-9}$ nats,
against a quadrature estimate of $0.0152$ nats for the population $T_{0.03}$ gap (measured
$0.0156$), about $1.8\times10^{6}$ times larger. The two theorems
are therefore positivity statements with explicit dependence on $\alpha$
and on the condition number. The numerical estimates of gap size use the
exact Bayes posteriors of the energy representations.

\begin{table}[htbp]
\centering\small
\caption{Explicit lower bounds on the energy information gap (NLL, nats)
compared with the gaps measured in Table~\ref{tab:energy_oracle_paired_v31}
(eight independent panels of known-law test data) and with population gap
estimates from numerical quadrature. A dash means that the
theorem does not apply.}
\label{tab:bound_vs_gap}
\begin{tabular}{@{}lll@{}}
\toprule
Quantity & $\alpha=0.8$, orthogonal ($\kappa=1$) & $\alpha=0.95$, $T_{0.03}$ ($\kappa=1/0.03$)\\
\midrule
Bound of Theorem~\ref{thm:energy_uniform_gap_v31} & $5.23\times10^{-6}$ ($\varrho^\ast=1.237$) & ---\\
Bound of Theorem~\ref{thm:conditioned_basis_gap} & $4.30\times10^{-6}$ & $8.27\times10^{-9}$\\
Measured gap & $0.0262$ (selected rotation) & $0.0156$\\
Quadrature estimate & $0.0259$ (best rotation found) & $0.0152$\\
\bottomrule
\end{tabular}
\end{table}

\paragraph{An independently learned change of coordinates.}
Let $\mathcal D$ be fitting data independent of the new labeled vector
$(Y,C)$, and let $T(\mathcal D)$ be an invertible two-coordinate map with
$\operatorname{cond}_2(T(\mathcal D))\le\kappa$ almost surely. Conditional
on each fitting-data realization, the theorem applies to the unchanged
three-regime law of $(Y,C)$. The Bayes log-loss gap, with $\mathcal D$
available to both predictors, therefore obeys the same lower bound almost
surely and after averaging over $\mathcal D$. This covers learning a
conditioning-controlled change of the two specified coordinates. Estimated
spans with leakage and coordinates selected from the scored record require
their own analysis.


\subsection{Redundant energies and representation limits}

The features $c_1^2,c_2^2,c_1c_2$ recover both log odds in
\eqref{eq:three_class_logodds_v31}. Equivalently, one redundant energy suffices:
\begin{equation}
 c_1c_2=\tfrac12\{(c_1+c_2)^2-c_1^2-c_2^2\}.
 \label{eq:redundant_energy_repair_v31}
\end{equation}
Multinomial logistic regression on these quadratic features contains the
centered Gaussian posterior family and provides a compact
quadratic-discriminant comparator in the mode coordinates. Interaction
screening and sparse QDA have substantial existing theory
\cite{fan_interactions_2015,jiang_sparseqda_2018,cai_sparseqda_2021}.

Equation~\eqref{eq:redundant_energy_repair_v31} is postprocessing of fixed signed
coefficients. If a new component kernel is inserted in an actual KMD
reconstruction, its total kernel $S$ changes and must be recomputed. A genuine
redundant-kernel implementation can instead use the three unit directions at
angles $0,\pi/3,2\pi/3$. Their projectors sum to $3I_2/2$; with isotropic
noise the KMD total remains scalar on this subspace, and their three energies
span all symmetric quadratic forms. The obstruction therefore concerns a
nonredundant energy basis, not every possible KMD dictionary.

More generally, connect coordinates $i,j$ whenever a precision difference has
a nonzero $ij$ entry. Energies plus pair products on a spanning forest of this
interaction graph retain the centered Gaussian posterior almost surely.
Indeed those products determine relative signs within each connected
component, and the remaining common sign of a component changes no required
quadratic term. Conversely, for measurements consisting of energies and pair
products on a graph $F$, every discriminant interaction must lie within an
$F$-component: otherwise flipping that component changes a log odd while
preserving all measurements. If the discriminant graph has $n_{\rm c}$
connected components, $r-n_{\rm c}$ pair products suffice and are necessary
within this measurement family.
Linear terms can be handled by adjoining a fixed coordinate $c_0=1$ and its
required edges. This is an information statement; divisions near zero
coefficients are not a stable reconstruction algorithm. Relative-sign and
phase recovery on graphs, including the need for stable redundant
measurements, is established in polarization-based phase retrieval
\cite{alexeev_polarization_2014}.


\section{Observation quality and the required representation}
\label{sec:quality_theory_v32}

A mode interaction can become useful because observation quality changes,
even when separate mode energies were sufficient before that change.
Conversely, a quality-dependent posterior transformation can sometimes
restore energy sufficiency. This section distinguishes these cases exactly.
Measurement error in Gaussian graphical models is an established problem
\cite{byrd_measurement_2021}; shared precision entries also appear in joint
graphical estimation \cite{danaher_joint_2014}. Uncertainty-aware Gaussian
classification is established \cite{bovy_deconvolution_2011,bilson_uncertainty_2026}.
We ask which centered Gaussian regime families retain posterior sufficiency
of \emph{fixed coordinate squares} through an open family of observation
covariances. The proof uses Gaussian likelihood and matrix identities to
characterize sufficiency under a common representation across qualities.

\subsection{A complete characterization for independently varying noise levels}

Let $Y\in\{0,\ldots,K-1\}$ have positive fixed priors, and let
\begin{equation}
 A\mid Y=k\sim\N(0,C_k),\quad C_k\succ0,\qquad
 Z=A+\xi_D,\quad \xi_D\sim\N(0,D),\quad
 D=\operatorname{diag}(d_1,\ldots,d_r)\succ0.
 \label{eq:quality_model_v32}
\end{equation}
Noise is independent of $(A,Y)$, its covariance is known, and coordinates
are fixed. Sufficiency below means
$\Prob(Y=\cdot\mid Z,D)=\Prob(Y=\cdot\mid Z_1^2,\ldots,Z_r^2,D)$.
For deterministic $D$, this is the usual posterior sufficiency condition;
if quality varies between observations, its value remains part of both
summaries. No claim here concerns an unobserved mixture over noise levels.

\begin{theorem}[Uniform sufficiency under diagonal observation noise]
\label{thm:quality_diagonal_v32}
The following statements are equivalent.
\begin{enumerate}
\item Coordinate energies are sufficient for every $D$ in some nonempty
open subset of $(0,\infty)^r$.
\item Coordinate energies are sufficient for every positive diagonal $D$.
\item All $C_k$ have the same off-diagonal entries, and each nonzero common
entry $(C_k)_{ij}$ forces both $(C_k)_{ii}$ and $(C_k)_{jj}$ to be identical
across classes.
\item After a common coordinate permutation,
\begin{equation}
 C_k=\operatorname{diag}(v_{ki}:i\in I)\oplus C_*,
 \label{eq:quality_structure_v32}
\end{equation}
where $C_*$ is shared by all classes and $I$ consists of the isolated
vertices of the common covariance graph. Thus every component containing
an edge has the same class covariance, and class-dependent variances occur
only at coordinates independent of all the others.
\end{enumerate}
The graph in this statement is the \emph{covariance} graph, not the graph
of conditional dependencies defined by the precision matrix.
\end{theorem}

\begin{proof}
For fixed $D$, Theorem~\ref{thm:energy_sufficiency_v31} says that sufficiency
is equivalent to every contrast
$(C_k+D)^{-1}-(C_0+D)^{-1}$ being diagonal. Each off-diagonal entry of this
contrast is a rational function of $d_1,\ldots,d_r$, with nonzero denominator
on the positive orthant. If it vanishes on a nonempty open set, its
polynomial numerator vanishes identically. This proves (1)$\Rightarrow$(2).

Fix classes $k,\ell$ and coordinates $i\ne j$. Send all other diagonal
noise entries to infinity while keeping $d_i,d_j$ fixed. The Schur
complement formula shows that the corresponding $2\times2$ principal
block of the inverse converges to the inverse of the principal covariance
block plus $\operatorname{diag}(d_i,d_j)$. Hence, writing
$a_k=(C_k)_{ii}$, $c_k=(C_k)_{jj}$ and $b_k=(C_k)_{ij}$,
\[
 \frac{b_k}{(a_k+d_i)(c_k+d_j)-b_k^2}
 =\frac{b_\ell}{(a_\ell+d_i)(c_\ell+d_j)-b_\ell^2}
 \qquad(d_i,d_j>0).
\]
Cross-multiplication and comparison of the $d_id_j$ coefficients give
$b_k=b_\ell=b$. If $b\ne0$, comparison of the $d_i$ and $d_j$ coefficients
also gives $c_k=c_\ell$ and $a_k=a_\ell$. This proves (2)$\Rightarrow$(3).
The graph formulation gives (3)$\Leftrightarrow$(4). Finally, the inverse
of \eqref{eq:quality_structure_v32} plus diagonal $D$ has a common block
on the complement of $I$ and diagonal class contrasts on $I$, proving
(4)$\Rightarrow$(2); (2)$\Rightarrow$(1) is immediate. For $r=1$ the
shared block may be empty and every centered variance family qualifies.
\end{proof}

The theorem excludes an apparent escape through identical, nonzero
precision interactions: such interactions can cancel from the class odds
at one quality, but generally cease to cancel when different mode noise
levels change independently. The next results quantify the local information change.

\begin{corollary}[Arbitrary correlated observation noise]
\label{cor:quality_full_v32}
For $r\ge2$, fixed coordinate energies are sufficient for every known
noise covariance in a nonempty open subset of the full positive-definite
cone if and only if all $C_k$ are identical. The same assertion holds in
any fixed invertible coordinate basis.
\end{corollary}
\begin{proof}
The rational-identity argument extends the equality to the entire
positive-definite cone. Theorem~\ref{thm:quality_diagonal_v32}, applied
to its diagonal subfamily, gives common off-diagonal entries. For a pair
$i\ne j$, choose $N=\eta I+s(e_ie_j^\top+e_je_i^\top)\succ0$ with
$|s|<\eta$ and $s\ne-(C_k)_{ij}$. The energies remain sufficient under
noise $N+D$ for every positive diagonal $D$. Applying the theorem to the
covariances $C_k+N$ forces their diagonal entries at $i,j$ to agree, since
their common off-diagonal entry is now nonzero. All diagonal entries agree
by varying the pair, proving $C_k=C_0$. The converse is immediate.
A fixed invertible basis maps the full noise cone onto itself by congruence.
\end{proof}

\subsection{A quality-dependent posterior energy can be sufficient}
A quality-dependent basis is outside
Theorem~\ref{thm:quality_diagonal_v32} and
Corollary~\ref{cor:quality_full_v32}. For two centered Gaussian classes,
the precision difference at each known quality can be orthogonally
diagonalized; squared coordinates in that changing basis are sufficient.
A reference posterior mean $m=A(D)Z$ can likewise mix coordinates and
must not be substituted silently for the fixed $Z$. If both the reference
prior and $D$ are diagonal, $A(D)$ is an invertible diagonal rescaling,
so $(m_i^2,D)$ and $(Z_i^2,D)$ are equivalent. For correlated quality
this equivalence generally fails.

The following exact counterexample sharpens this limitation. With
reference covariance $\Gamma=I_r$, define
\begin{equation}
 K_D=(I+D)^{-1},\qquad m=K_DZ,\qquad V=I-K_D.
 \label{eq:quality_reference_v32}
\end{equation}
Here $V$ is the actual reference posterior covariance. Suppose the
baseline class covariance is $C_0=I_r$, and each other class has a
single-coordinate update $C_k=I_r+t_ke_{i_k}e_{i_k}^\top$, $t_k>-1$.
For \emph{every} known $D\succ0$, including correlated noise,
\begin{equation}
 \frac{p_k(Z\mid D)}{p_0(Z\mid D)}
 =(1+t_k(K_D)_{i_ki_k})^{-1/2}
 \exp\!\left\{\frac{t_km_{i_k}^2}
 {2[1+t_k(K_D)_{i_ki_k}]}\right\}.
 \label{eq:quality_reference_exception_v32}
\end{equation}
The determinant lemma and Sherman--Morrison inverse formula prove this
identity directly. Consequently $(m_1^2,\ldots,m_r^2,V)$ is sufficient
for this nontrivial multiclass family under every quality. Confusing
$K_D$ with $V$ would give the wrong denominator. Retaining signed posterior
means and full covariance supports general regime families, but this
example shows that it is not necessary for every family.

\subsection{Nonzero means and a fixed energy center}
\label{sec:noncentered_quality_extension}
The same argument gives a useful extension beyond centered class laws.
The center of the reported energies must be fixed and declared in advance.

\begin{corollary}[Noncentered Gaussian classes]
\label{cor:noncentered_quality_extension}
Let $A\mid Y=k\sim\N(\mu_k,C_k)$, $C_k\succ0$, with positive priors,
and use energies $(Z_i-b_i)^2$ around one fixed $b\in\R^r$.
These energies preserve the posterior for every diagonal positive noise
covariance in a nonempty open set if and only if:
\begin{enumerate}
\item the covariances have the block structure in
Theorem~\ref{thm:quality_diagonal_v32};
\item all class means equal a common vector $\mu$;
\item $\mu_i=b_i$ at every isolated coordinate whose variance differs
between at least two classes.
\end{enumerate}
For $r\ge2$, sufficiency throughout an open subset of the full noise cone
holds if and only if all class means and covariances coincide.
\end{corollary}
\begin{proof}
Translate by $b$ and write $\nu_k=\mu_k-b$. The fixed-quality sign criterion
requires diagonal precision contrasts and
$(C_k+D)^{-1}\nu_k=(C_0+D)^{-1}\nu_0$. As in
Theorem~\ref{thm:quality_diagonal_v32}, rational identities extend both
conditions from a nonempty open set to all positive diagonal $D$.
The quadratic condition gives the covariance block structure. Set $D=tI$
and multiply the linear identity by $t$. Its limit as $t\to\infty$ gives
$\nu_k=\nu_0=\nu$. On the shared covariance block the linear identities
then hold automatically. On an isolated coordinate they reduce to
\[
 \frac{\nu_i}{v_{ki}+d_i}=\frac{\nu_i}{v_{0i}+d_i}\qquad(d_i>0),
\]
which holds for all classes exactly when $\nu_i=0$ or all their variances
at that coordinate agree. This also proves sufficiency. In the full noise
cone, Corollary~\ref{cor:quality_full_v32} forces common covariances, and
any one invertible common precision then forces common means.
\end{proof}

A common nonzero mean is allowed on a class-invariant correlated block.
Only an informative variance coordinate must be centered at that mean.
Thus arbitrary recentering can change the information in an energy summary.
This corollary extends the Gaussian characterization, not its distributional
scope: non-Gaussian class laws need a separate analysis.

\subsection{The quadratic feature budget for a quality-aware linear decoder}

The preceding result concerns separate energies. A stronger representation
cost follows if a classifier uses any fixed quadratic features and makes
its log odds linear in those features, while allowing the coefficients to
depend on the known observation quality.

\begin{theorem}[Noise variation spans every quadratic interaction]
\label{thm:quality_quadratic_span_v32}
Let $C_0,C_1\succ0$ be distinct $r\times r$ covariance matrices, and let
$\mathcal U$ be any nonempty open subset of the full positive-definite
noise cone. Then
\begin{equation}
 \operatorname{span}\{(C_1+D)^{-1}-(C_0+D)^{-1}:D\in\mathcal U\}
 =\operatorname{Sym}_r,
 \label{eq:quality_quadratic_span_v32}
\end{equation}
where $\operatorname{Sym}_r$ is the space of real symmetric matrices.
Consequently, suppose the exact binary Gaussian log odds can be written
for every $D\in\mathcal U$ as
\begin{equation}
 \ell_D(z)=a(D)+\sum_{h=1}^m w_h(D)\,z^\top B_hz,
 \label{eq:quality_linear_decoder_v32}
\end{equation}
with fixed symmetric feature matrices $B_h$. Then these matrices span
$\operatorname{Sym}_r$. In particular at least $r(r+1)/2$ linearly
independent quadratic features are necessary, and the full set of
coordinate squares and pair products is sufficient.
\end{theorem}

\begin{proof}
If the span were proper, it would admit a nonzero symmetric annihilator
$T$ under the trace inner product. The rational function
\[
 \operatorname{tr}\{T[(C_1+D)^{-1}-(C_0+D)^{-1}]\}
\]
would vanish on $\mathcal U$, hence on the entire positive-definite cone.
Fix $A\succ0$, set $D=tA^{-1}$, and let $t\to\infty$. The inverse
expansion gives, with $\Delta=C_1-C_0\ne0$,
\[
 t^2[(C_1+tA^{-1})^{-1}-(C_0+tA^{-1})^{-1}]
 \longrightarrow -A\Delta A.
\]
Thus $\operatorname{tr}(TA\Delta A)=0$ for every $A\succ0$.
This is a polynomial identity in the symmetric entries of $A$, so it
holds for every symmetric $A$. In particular, $A=uu^\top$ gives
\[
 (u^\top Tu)(u^\top\Delta u)=0\qquad\hbox{for every }u\in\mathbb R^r.
\]
The second polynomial is not identically zero. Since real polynomials
have no zero divisors, the first vanishes identically, which forces $T=0$,
a contradiction. This proves \eqref{eq:quality_quadratic_span_v32}.
In \eqref{eq:quality_linear_decoder_v32}, positivity of Gaussian densities
makes an almost-everywhere equality a polynomial identity. Comparing its
quadratic coefficients puts every precision contrast in
$\operatorname{span}\{B_h\}$, proving the feature lower bound. The standard
quadratic monomials give the converse.
\end{proof}

The lower bound applies even to two fixed, distinct class covariances:
varying observation quality alone generates the full linear span of
quadratic decision forms. It is a \emph{linear-decoder} statement, not a
lower bound on all sufficient statistics. Nonlinear recovery from suitable
redundant intensity measurements, or a quality-dependent feature basis,
can evade it. For example, binary quality-specific eigenvectors diagonalize
each contrast separately, and the posterior-energy exception in
\eqref{eq:quality_reference_exception_v32} does not
use fixed $z^\top B_hz$ features. The theorem requires no uniform separation
between $C_0$ and $C_1$ and supplies no uniform positive risk gap or stable
numerical rank bound; when the class covariances nearly coincide, the
quadratic forms may span the space while being difficult to distinguish.
The underlying fact that unrestricted QDA uses quadratic monomials is
classical. The result specifies that a full observation-quality family can
force this feature span even when only one fixed class contrast is present
before observation noise is varied.

\begin{proposition}[A reference metric for a declared quality ensemble]
\label{prop:quality_feature_compression_v32}
Fix finitely many known qualities $D_l$ with weights $w_l>0$ summing to
one, and their precision contrasts $B_l=(C_1+D_l)^{-1}-(C_0+D_l)^{-1}$.
Let $\mathcal S\subseteq\operatorname{Sym}_r$ be the linear span of the
fixed quadratic feature matrices. Under the common reference
$Z\sim\N(0,I_r)$, allowing a free intercept for each quality, the minimum
weighted mean squared error in the exact log odds is
\begin{equation}
 \frac12\sum_l w_l\|B_l-\operatorname{Proj}_{\mathcal S}B_l\|_F^2.
 \label{eq:quality_compression_metric_v32}
\end{equation}
Use symmetric vectorization $\operatorname{svec}$ that leaves diagonal
entries unchanged and multiplies upper off-diagonal entries by $\sqrt2$,
so Euclidean and Frobenius norms agree. If
\[
 G=\sum_lw_l\operatorname{svec}(B_l)\operatorname{svec}(B_l)^\top
\]
has eigenvalues $\lambda_1\ge\cdots\ge\lambda_{r(r+1)/2}\ge0$, the best
$m$-dimensional feature space has error
\begin{equation}
 \frac12\sum_{j>m}\lambda_j(G),
 \label{eq:quality_compression_tail_v32}
\end{equation}
and is given by the leading eigenvectors reshaped as symmetric matrices.
\end{proposition}
\begin{proof}
For an approximation $\widetilde B_l\in\mathcal S$, set
$F_l=B_l-\widetilde B_l$. Optimizing the intercept centers the quadratic
residual, giving $-\tfrac12(Z^\top F_lZ-\operatorname{tr}F_l)$.
Its squared expectation is $\tfrac12\|F_l\|_F^2$ by the Gaussian fourth
moment. Frobenius projection proves
\eqref{eq:quality_compression_metric_v32}. Optimizing the feature subspace
is the standard weighted principal-subspace approximation
\cite{eckart_approximation_1936}, whose residual is the eigenvalue tail.
\end{proof}

This operational design diagnostic selects fixed interaction features under
a \emph{specified common reference distribution}. It uses the classical
Gaussian fourth moment and principal-subspace optimization to minimize
mean squared log-odds approximation error for a fixed quadratic feature
space with a quality-dependent linear decoder. The following proposition
provides predictive log-loss control under the actual class mixture.
A change in the quality ensemble or reference law changes the design
criterion. The eigenvalue tail quantifies the interaction budget for a
chosen approximation accuracy.

\subsection{Conditioning bounds under the observed Gaussian mixture}
\label{sec:stable_quality_extension}

The reference-distribution diagnostic has a quantitative counterpart under the
actual centered Gaussian class mixture. It requires bounds on the observed
covariances and a declared distribution of observation qualities.

\begin{proposition}[Stable quadratic approximation and log-loss control]
\label{prop:stable_quality_extension}
Consider two centered Gaussian classes with positive priors and finitely many
known qualities $D_l$, drawn independently of the class with probabilities
$w_l>0$ summing to one. Let
$\Sigma_{kl}=C_k+D_l$ satisfy $\lambda_-I\preceq\Sigma_{kl}\preceq\lambda_+I$ for
$0<\lambda_-\le\lambda_+<\infty$, and define $B_l$ and $G$ as in
Proposition~\ref{prop:quality_feature_compression_v32}.
Write $d=r(r+1)/2$ and $T_m=\sum_{j>m}\lambda_j(G)$.
Among fixed $m$-dimensional quadratic feature spaces with freely chosen
quality-dependent coefficients and intercepts, let $\mathcal R_m$ be the
minimum mean squared error in the exact log odds, evaluated under the actual
class mixture at each quality. Then
\begin{equation}
 \frac{\lambda_-^2}{2}T_m\ \le\ \mathcal R_m
 \ \le\ \frac{(r+2)\lambda_+^2}{4}T_m.
 \label{eq:stable_quality_sandwich}
\end{equation}
Moreover, the feature space given by the leading eigenvectors of $G$ admits
a logistic decoder whose excess population NLL is at most
\begin{equation}
 \frac{(r+2)\lambda_+^2}{32}T_m.
 \label{eq:stable_quality_logloss}
\end{equation}
These are population approximation bounds; the coefficients use the stated
class laws rather than finite-sample estimates.
\end{proposition}
\begin{proof}
Fix quality $l$ and a quadratic residual matrix $F=B_l-\widetilde B_l$.
The best intercept subtracts its mixture mean, leaving log-odds residual
$-\tfrac12\{Z^\top FZ-\E_l(Z^\top FZ)\}$. By conditional variance and the
Gaussian fourth-moment identity,
\[
 \Var_l(Z^\top FZ)
 \ge \sum_k\pi_k\,2\|\Sigma_{kl}^{1/2}F\Sigma_{kl}^{1/2}\|_F^2
 \ge 2\lambda_-^2\|F\|_F^2.
\]
Also $\Var_l(Z^\top FZ)\le\E_l[(Z^\top FZ)^2]$, and each Gaussian
second moment of this quadratic form is
\[
 2\|\Sigma_{kl}^{1/2}F\Sigma_{kl}^{1/2}\|_F^2
       +(\tr(F\Sigma_{kl}))^2
 \le (r+2)\lambda_+^2\|F\|_F^2.
\]
Multiplication by $1/4$, quality averaging and minimization over coefficients
and feature spaces give the lower bound in \eqref{eq:stable_quality_sandwich}.
Frobenius projection onto the leading eigenspace of $G$ gives the upper bound,
because its weighted squared matrix residual is $T_m$.

For binary log odds $\ell$ and approximation $\widetilde\ell$, conditional
excess NLL is the Bregman remainder of $A(t)=\log(1+e^t)$:
\[
 A(\widetilde\ell)-A(\ell)-A'(\ell)(\widetilde\ell-\ell)
 \le\tfrac18(\widetilde\ell-\ell)^2,
\]
since $0<A''(t)\le1/4$. Apply this inequality to the projection decoder
with the optimal intercept and average over qualities and the actual
mixtures. This proves \eqref{eq:stable_quality_logloss}.
\end{proof}

If $G$ is positive definite, the first bound is at least
$\tfrac12\lambda_-^2(d-m)\lambda_d(G)$ for $m<d$. Thus stability depends on
both observation conditioning and coverage of the precision contrasts by the
quality ensemble. Bounded covariance condition numbers alone do not bound
$\lambda_d(G)$ away from zero: nearly identical classes or a narrow quality
range can make it arbitrarily small. The lower bound concerns squared
log-odds error; a predictive log-loss lower bound requires further control of
posterior saturation. The upper bound provides an actual-mixture predictive
approximation guarantee, with estimation and adaptive selection still to be
analyzed separately.

\subsection{A numerical interaction budget}
\begin{figure}[tbp]
\centering\includegraphics[width=\linewidth]{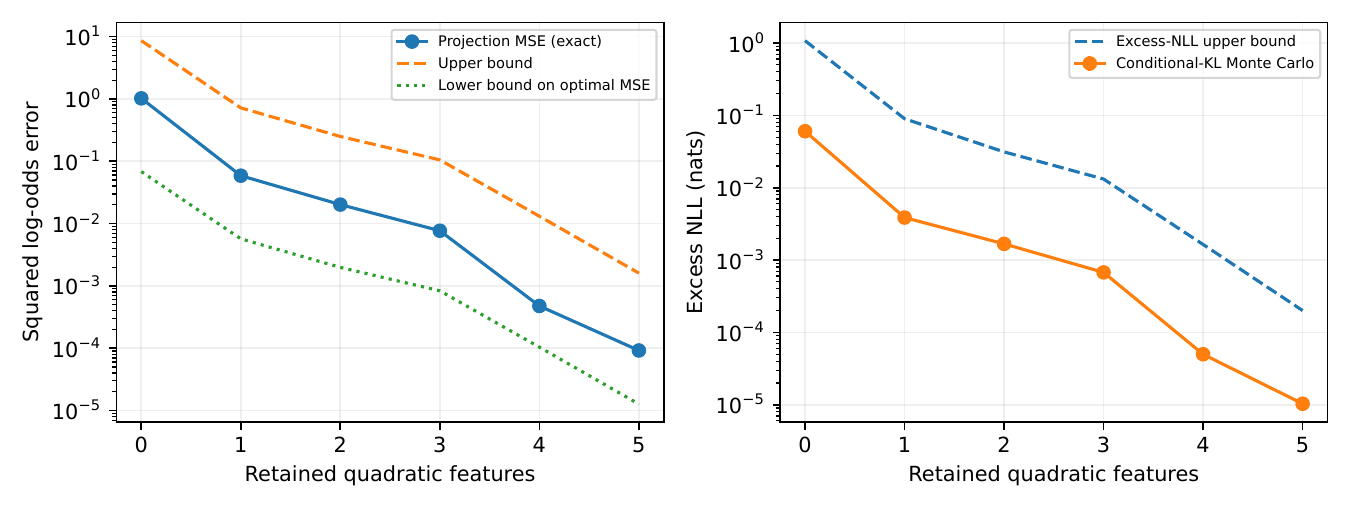}
\caption{The precision-contrast spectrum yields a quantitative feature budget.
The projection decoder has an exactly evaluated MSE between the theorem's
bounds. Conditional-KL Monte Carlo approximates its population excess NLL;
error bars show numerical integration uncertainty. The lower bound concerns
the optimal MSE, not predictive NLL. With all six features both errors vanish
up to rounding; that endpoint is omitted from logarithmic axes.}
\label{fig:focused_budget}
\end{figure}

To separate exact span from approximation accuracy, consider three modes,
$C_0=I_3$, $C_1=\operatorname{diag}(4,1,0.5)$, and 12 fixed positive-definite
qualities $D_l$ with equal weights, generated by a deterministic script
(Appendix~\ref{sec:reproduction}). Table~\ref{tab:feature_budget_qualities}
lists the 12 qualities and the resulting constants $\lambda_-$ and
$\lambda_+$ of Proposition~\ref{prop:stable_quality_extension}, so that the
bounds below can be recomputed from the paper. The contrast Gram
eigenvalues are approximately
\[
 (0.21193,0.012581,0.003885,0.002473,0.000310,0.0000430).
\]
All six are positive, yet two leading quadratic features give population
excess NLL about $0.001673$ nats for the projection decoder; its numerical
standard error is $5.4\times10^{-6}$. The conservative bound
\eqref{eq:stable_quality_logloss} is $0.03126$ nats. With four features, the
estimated population excess NLL is $0.00004984$ nats and the bound is
$0.001642$ nats. The actual-mixture MSE is computed exactly from Gaussian
fourth moments. NLL is a Monte Carlo average of conditional KL over 100,000
observations at each quality, with known class laws and no fitting.
Figure~\ref{fig:focused_budget} shows both quantities for every feature
budget.

\input{tables/feature_budget_qualities}

\subsection{Noise can expose discarded information while reducing total information}

A scalar noise sweep does not test the open-set requirement. For example,
let
\begin{equation}
 P_0=\begin{pmatrix}a&b\\b&c\end{pmatrix},\qquad
 P_1=\begin{pmatrix}c&b\\b&a\end{pmatrix},\qquad
 C_k=P_k^{-1},\qquad ac>b^2,\ a,c>0.
 \label{eq:quality_swap_v32}
\end{equation}
The baseline precision interaction $b$ is shared, so separate energies
are sufficient. After noise $\tau\operatorname{diag}(r_1,r_2)$ with
$r_1,r_2>0$,
\begin{equation}
 [(C_k+\tau\operatorname{diag}(r_1,r_2))^{-1}]_{12}
 =\frac{b}{1+\tau(a_kr_1+c_kr_2)
               +\tau^2r_1r_2(ac-b^2)},
 \label{eq:quality_swap_inverse_v32}
\end{equation}
where $(a_0,c_0)=(a,c)$ and $(a_1,c_1)=(c,a)$.
For isotropic noise, $r_1=r_2$, the entries agree at every noise level.
For $a\ne c$, $b\ne0$, $r_1\ne r_2$ and every finite $\tau>0$, they
are unequal. Thus the optimal log-loss deficiency
\[
 G(\tau)=I(Y;Z_\tau\mid Z_{\tau,1}^2,Z_{\tau,2}^2)
\]
is zero at $\tau=0$ and strictly positive thereafter. At the same time,
$I(Y;Z_\tau)$ is nonincreasing in $\tau$ by Gaussian-channel data processing.
The noise exposes information lost by the fixed feature map; it does not
create new information about the class.

For any positive-definite $R$ and $\tau>0$, the Gaussian KL to the
noise-only reference $\N(0,\tau R)$ gives the useful limiting bound
\begin{equation}
 0\le G(\tau)\le I(Y;Z_\tau)
 \le\frac{1}{4\tau^2}\sum_k\pi_k
 \|R^{-1/2}C_kR^{-1/2}\|_F^2.
 \label{eq:quality_large_noise_v32}
\end{equation}
Indeed the mixture information is at most the class-averaged KL to any
reference law, and $x-\log(1+x)\le x^2/2$ for $x\ge0$ bounds the Gaussian
KL eigenvalue terms. Hence the positive intermediate-noise deficiency
returns to zero as noise tends to infinity.

The local size of the deficiency, here and in
Theorem~\ref{thm:weak_interactions_v32}, rests on the following exact
decomposition of a conditional mutual information into four divergences.
For a family of joint laws $p_\epsilon$ of $(Y,X)$ and a statistic $U$ of
$(Y,X)$, write $\mathsf D_U(\epsilon)=D_{\rm KL}(p_{\epsilon,U}\|p_{0,U})$
for the divergence between the laws of $U$ under $p_\epsilon$ and $p_0$.

\begin{lemma}[Four-KL identity]
\label{lem:four_kl}
Let $(p_\epsilon)$ be a family of joint laws of $(Y,X)$, with $Y$ discrete,
such that $p_\epsilon\ll p_0$, and let $S=S(X)$ be a statistic of $X$.
Suppose that under the baseline law
\begin{equation}
 p_0(Y\mid X)=p_0(Y\mid S)\qquad\text{almost surely.}
 \label{eq:four_kl_hypothesis}
\end{equation}
Then, whenever the four divergences below are finite,
\begin{equation}
 I_\epsilon(Y;X\mid S)
 =\mathsf D_{(Y,X)}(\epsilon)-\mathsf D_X(\epsilon)
  -\mathsf D_{(Y,S)}(\epsilon)+\mathsf D_S(\epsilon).
 \label{eq:four_kl_v32}
\end{equation}
\end{lemma}
\begin{proof}
Since $S$ is a function of $X$,
$I_\epsilon(Y;X\mid S)=\E_{p_\epsilon}\log[p_\epsilon(Y\mid X)/p_\epsilon(Y\mid S)]$.
The chain rule for divergences gives
$\mathsf D_{(Y,X)}-\mathsf D_X=\E_{p_\epsilon}\log[p_\epsilon(Y\mid X)/p_0(Y\mid X)]$
and
$\mathsf D_{(Y,S)}-\mathsf D_S=\E_{p_\epsilon}\log[p_\epsilon(Y\mid S)/p_0(Y\mid S)]$.
Subtracting, the right side of \eqref{eq:four_kl_v32} equals
\[
 I_\epsilon(Y;X\mid S)
 -\E_{p_\epsilon}\log\frac{p_0(Y\mid X)}{p_0(Y\mid S)}.
\]
The cross term vanishes because its integrand is zero
$p_0$-almost surely by \eqref{eq:four_kl_hypothesis}, hence
$p_\epsilon$-almost surely.
\end{proof}

In both applications below, $X$ is a coordinate vector, $S$ is its vector
of coordinate energies, and \eqref{eq:four_kl_hypothesis} holds because,
under the baseline law, $Y$ is conditionally independent of the coordinate
signs given the energies. The signs need not be uniform.

\begin{theorem}[Local size of quality-induced energy loss]
\label{thm:quality_expansion_v32}
In two coordinates, suppose all baseline class precisions have the same
off-diagonal entry $b$, with possibly different positive-definite diagonal
parts. Let $\Sigma_k(\tau)=P_k(0)^{-1}+\tau R$ for a fixed symmetric $R$,
and restrict $\tau$ to an interval where all covariances are positive
definite. Set
\[
 \beta_k=-(P_k(0)RP_k(0))_{12},\qquad
 \eta_k^0(e)=\Prob_0(Y=k\mid Z_1^2=e_1,Z_2^2=e_2).
\]
Then, as $\tau\to0$,
\begin{equation}
 G(\tau)=\frac{\tau^2}{2}\mathcal A_R+O(|\tau|^3),\qquad
 \mathcal A_R=\mathbb E_0\!\left[
 E_1E_2\operatorname{sech}^2(b\sqrt{E_1E_2})
 \operatorname{Var}_{k\sim\eta^0(E)}(\beta_k)\right].
 \label{eq:quality_expansion_v32}
\end{equation}
If the $\beta_k$ are not all equal, then $\mathcal A_R>0$.
For diagonal $R=\operatorname{diag}(r_1,r_2)$,
$\beta_k=-b(r_1(P_k)_{11}+r_2(P_k)_{22})$.
An explicit uniform remainder is given below.
\end{theorem}

\begin{proof}
Under the baseline law, $Y$ and the signs of $Z$ are conditionally
independent given $E=(Z_1^2,Z_2^2)$; these signs need not be uniform. Since
$Z$ is determined by $E$ and its signs, $p_0(Y\mid Z)=p_0(Y\mid E)$, and
Lemma~\ref{lem:four_kl} with $X=Z$, $S=E$ and $\epsilon=\tau$ gives
$G(\tau)=\mathsf D_{(Y,Z)}(\tau)-\mathsf D_Z(\tau)-\mathsf D_{(Y,E)}(\tau)
+\mathsf D_E(\tau)$.
For any of these four statistics $U$, the second derivative at zero is the
squared $L^2(p_0)$ norm of the conditional score
$\mathbb E_0[s_Y(Z)\mid U]$. Write $Q=Z_1Z_2$.
The Gaussian score has the form
$s_k(z)=u_k(E)-\beta_k Q$, where $u_k$ depends only on the energies.
Conditioning on $E=e$, the sign-product probabilities are proportional
to $\exp(-bQ)$, independently of $k$. Therefore
\[
 \mathbb E_0[Q\mid E=e]=-\sqrt{e_1e_2}\tanh(b\sqrt{e_1e_2}),\qquad
 \operatorname{Var}_0(Q\mid E=e)
 =e_1e_2\operatorname{sech}^2(b\sqrt{e_1e_2}).
\]
Subtracting the four squared conditional-score norms cancels every
energy-only contribution and leaves
$\mathbb E_0[\operatorname{Var}_0(Q\mid E)
\operatorname{Var}_{\eta^0(E)}(\beta_k)]$. This is $G''(0)$.
Positive Gaussian densities and positive class priors make every
$\eta_k^0(e)$ positive; the variance is therefore positive for almost
every $e$ when two $\beta_k$ differ. The bounds below justify the
Taylor expansion and differentiation under integration.
\end{proof}

\paragraph{An explicit remainder.}
Put $\nu=\|R\|_{\rm op}$,
$\Lambda=\max_k\lambda_{\max}(P_k(0)^{-1})$, and choose $t>0$ such that
$\lambda_t=\min_k\lambda_{\min}(P_k(0)^{-1})-t\nu>0$ and
$t\nu\Lambda/\lambda_t^2<1$. For $u\ge0$, set (the dimension is two)
\[
 L_1(u)=\frac{\nu}{\lambda_t}+\frac{\nu u}{2\lambda_t^2},\quad
 L_2(u)=\frac{\nu^2}{\lambda_t^2}+\frac{\nu^2u}{\lambda_t^3},\quad
 L_3(u)=\frac{2\nu^3}{\lambda_t^3}+\frac{3\nu^3u}{\lambda_t^4},\quad h(u)=tL_1(u).
\]
For $W\sim\chi_2^2$, define the finite, explicit constant
\begin{equation}
 M_t=\mathbb E\!\left[e^{h(\Lambda W)}
 \{4L_1^3+3L_1L_2+h(L_1^3+3L_1L_2+L_3)\}(\Lambda W)\right].
 \label{eq:quality_remainder_v32}
\end{equation}
Then
\begin{equation}
 |G(\tau)-\tfrac12\tau^2\mathcal A_R|
 \le\tfrac23 M_t|\tau|^3\qquad(|\tau|\le t).
 \label{eq:quality_cubic_bound_v32}
\end{equation}
To check the bound, if $P=\Sigma_k(\tau)^{-1}$, the first three derivatives
of the log Gaussian density are
\[
 -\tfrac12\operatorname{tr}(PR)+\tfrac12z^\top PRPz,\quad
 \tfrac12\operatorname{tr}((PR)^2)-z^\top PRPRPz,\quad
 -\operatorname{tr}((PR)^3)+3z^\top PRPRPRPz.
\]
Their absolute values are bounded by $L_1(u),L_2(u),L_3(u)$ respectively.
For a density ratio $\rho_\tau$ to baseline,
$|\log\rho_\tau|\le h(u)$ and its first three derivatives are bounded by
$\rho_\tau L_1$, $\rho_\tau(L_1^2+L_2)$, and
$\rho_\tau(L_1^3+3L_1L_2+L_3)$.
The same bounds hold after each of the four coarsenings, since each
retains $u=\|z\|^2$ and its ratio is a baseline conditional expectation.
Differentiating $x\log x-x+1$ at $x=\rho_\tau$ three times gives the envelope in
\eqref{eq:quality_remainder_v32}. Under every baseline class,
$\|Z\|^2$ is stochastically bounded by $\Lambda W$; the displayed
integrability condition permits dominated differentiation. Taylor's
remainder for four KL terms is $4M_t|\tau|^3/6$.
The constant can be evaluated using the chi-square exponential moments
stated in Section~\ref{sec:adaptive_theory_v32}. It is conservative;
ordinary quadrature of $\mathcal A_R$ is a numerical check, not a formal
certificate of a positive lower bound. The local KL expansion uses
established information-geometric machinery \cite{prelov_verdu_2004};
the coefficient here allows correlated baseline signs and specifies which
quality changes expose lost information.

\subsection{A correlated-noise example with a sufficient residual}

Independent latent modes do not prevent quality-induced interaction
information. Take equal priors,
\begin{equation}
 C_0=I_2,\qquad C_1=\operatorname{diag}(4,1),\qquad
R=\begin{pmatrix}1&0.8\\0.8&1\end{pmatrix}.
 \label{eq:quality_independent_example_v32}
\end{equation}
Here $Z=A+\xi_R$ with $\xi_R\sim\N(0,R)$ independent of $(A,Y)$;
the noise multiplier is one.
The latent energies are sufficient, but the noisy physical energies are
not. The residual $U=Z_1-0.4Z_2$ is independent of $Z_2$ in each class,
and has variances $v_0=1.68$ and $v_1=4.68$. Thus $U^2$ is sufficient,
whereas $(Z_1^2,Z_2^2)$ omits the cross product needed to compute it.
With equal priors the full Bayes accuracy is exactly
\begin{equation}
 \tfrac12+\Phi(c/\sqrt{v_0})-\Phi(c/\sqrt{v_1}),\qquad
 c^2=\frac{\log(v_1/v_0)}{1/v_0-1/v_1}.
 \label{eq:quality_accuracy_v32}
\end{equation}
Here physical energies also incur strictly greater zero--one loss:
take $\sqrt{e_1}=c$ and $0<0.4\sqrt{e_2}<c$. At these energies and on
an open neighborhood, the two sign products give
$|Z_1-0.4Z_2|$ on opposite sides of $c$. Both signs have positive density,
so the full optimal decisions disagree within the energy fiber. This
specific argument is needed; strict log-loss deficiency alone would not
imply strict classification error in every example.

\begin{figure}[htbp]
 \centering
 \includegraphics[width=0.95\linewidth]{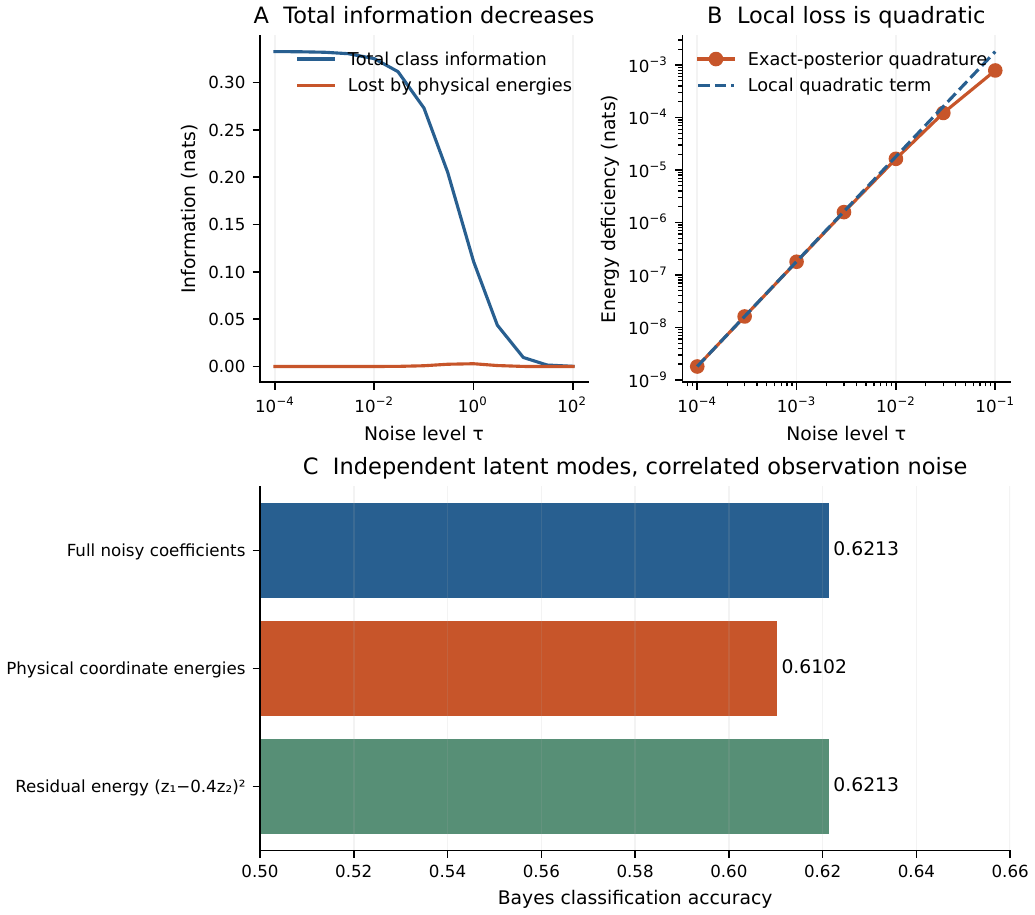}
 \caption{Known observation quality changes the information retained by
 fixed physical energies. The swapped-precision example
 \eqref{eq:quality_swap_v32} uses $(a,c,b)=(0.4,2,0.7)$ and
 $R=\operatorname{diag}(1,4)$. Upper panels show total class information
 and energy deficiency, and the local quadratic approximation; ordinary
 Gaussian quadrature is used. The lower panel compares sufficient and
 lossy summaries in the independent-latent-mode example
 \eqref{eq:quality_independent_example_v32}. The mixing transformation is
 quality dependent, so its equality with the full posterior is consistent
 with the fixed-coordinate theorem. Numerical uncertainty and convergence
 checks are recorded with the results; none is a fitted-model confidence
 interval.}
 \label{fig:quality_theory_v32}
\end{figure}

\paragraph{Numerical verification and scope.}
The accompanying observation-quality verification script checks randomized
covariance structures under diagonal and correlated noise, the two exact
sufficient-transform controls, and the local expansion. It uses specified
Gaussian class laws, fixed coordinates, and an analytic four-sign energy
posterior to verify the structural theorem and quantify its numerical size.
Table~\ref{tab:quality_theory_v32} and Figure~\ref{fig:quality_theory_v32}
report the results: the local coefficient $\mathcal A_R$ of
Theorem~\ref{thm:quality_expansion_v32} for the swapped-precision example,
and the accuracy and log-loss deficiency of physical energies in the
correlated-noise example.
\begin{table}[htbp]
\centering
\caption{Known-law quality checks. Accuracy uncertainty is Monte Carlo integration error, not training variability. The full accuracy is analytic; the energy gap is in nats.}
\label{tab:quality_theory_v32}
\begin{tabular}{lr}
\toprule
Quantity & Value\\
\midrule
Local coefficient $\mathcal A_R$ (600-point quadrature per axis) & 0.360836\\
Coefficient difference, 400 vs. 600 points per axis & 3.8e-09\\
Independent-latent example: full Bayes accuracy & 0.621314\\
Independent-latent example: physical-energy accuracy & 0.610228\\
Conservative Monte Carlo SE of energy accuracy & 6.8e-05\\
Physical-energy log-loss deficiency & 0.007520\\
Conservative Monte Carlo SE of deficiency & 1.8e-05\\
Sufficient residual-energy posterior error (maximum) & 1.4e-14\\
\bottomrule
\end{tabular}
\end{table}

\FloatBarrier

\section{Learning, approximation and adaptive coordinates}
\label{sec:adaptive_theory_v32}

This section quantifies the sign information discarded by KMD energies and
identifies conditions under which it can be estimated from finite data.
The analysis combines classical local KL expansions, Gaussian quadratic
concentration, and Wishart identities. It builds on Gaussian
mutual-information feature selection and its prediction interpretation
\cite{lefakis_jointly_2016,beraha_feature_2019}, information expansions
\cite{prelov_verdu_2004}, general lossless-transform and excess-risk theory
\cite{gyorfi_lossless_2023}, and Gaussian divergence-oriented projections
\cite{chen_projection_2024}. We specialize these tools and the established
compression--risk identity to KMD sign information, its attenuation by an
observation posterior, and an explicit error budget for selecting a mode
span before fitting regime covariances.

\subsection{A local formula with a uniform, computable remainder}

Fix positive class priors $\pi_k$. In signed mode coordinates suppose
\begin{equation}
 C\mid Y=k\sim\N(0,P_k(\epsilon)^{-1}),\qquad
 P_k(\epsilon)=D_k+\epsilon B_k,
 \label{eq:weak_precision_v32}
\end{equation}
where $D_k$ is positive diagonal and $B_k$ is symmetric with zero diagonal.
Write $\mathcal E=(C_1^2,\ldots,C_r^2)$ and let
$\eta_k^0(e)=\Prob_0(Y=k\mid\mathcal E=e)$. Define
\begin{equation}
 \mathcal A=\mathbb E_0\left[
 \sum_{i<j}\mathcal E_i\mathcal E_j
 \operatorname{Var}_{k\sim\eta^0(\mathcal E)}((B_k)_{ij})\right].
 \label{eq:interaction_coefficient_v32}
\end{equation}
The expectation includes the baseline Gaussian mixture. A large precision
interaction shared by every class contributes zero; interactions count only
where the energy posterior still leaves class uncertainty.

The following constants give an explicit remainder for the expansion. Let $\lambda=\min_k\lambda_{\min}(D_k)>0$,
$\bar b=\max_k\|B_k\|_{\rm op}$, and choose $t>0$ with $t\bar b<\lambda$.
Set $\bar\beta=\bar b/\lambda$ and
\[
 a_2=\frac{r\bar\beta^2}{2(1-t\bar\beta)^2},\qquad
 a_3=\frac{r\bar\beta^3}{(1-t\bar\beta)^3},\qquad
 h(u)=\frac{t\bar b}{2}u+\frac{t^2a_2}{2},\qquad
 L(u)=\frac{\bar b}2u+ta_2.
\]
For $Z\sim\chi_r^2$, put
\begin{equation}
 \mathcal M_t=\mathbb E\left[e^{h(Z/\lambda)}
 \left\{4L(Z/\lambda)^3+3a_2L(Z/\lambda)
 +h(Z/\lambda)\big[L(Z/\lambda)^3+3a_2L(Z/\lambda)+a_3\big]\right\}\right].
 \label{eq:weak_remainder_constant_v32}
\end{equation}
This finite expectation is a polynomial times a chi-square exponential
moment. In particular
$\mathbb E[Z^j e^{sZ}]=2^j(r/2)_j(1-2s)^{-r/2-j}$ for $2s<1$,
so no unknown distributional constant is hidden in it.

\begin{theorem}[Size of the energy information loss]
\label{thm:weak_interactions_v32}
For every $|\epsilon|\le t$,
\begin{equation}
 \left|I_\epsilon(Y;C\mid\mathcal E)
       -\frac{\epsilon^2}{2}\mathcal A\right|
 \le \frac23\mathcal M_t|\epsilon|^3.
 \label{eq:weak_information_v32}
\end{equation}
Consequently the right side subtracted from
$\epsilon^2\mathcal A/2$ is a certified lower bound, truncated at zero.
This is a fixed-coordinate result; optimizing a basis changes both $D_k$
and $B_k$ and is a separate problem.
\end{theorem}

\begin{proof}
Let $p_\epsilon$ be the joint law of $(Y,C)$ and $p_0$ its baseline.
Under $p_0$ every class law has independent coordinates, so conditional on
$\mathcal E$ the signs of $C$ are uniform and independent of $Y$. Hence
$p_0(Y\mid C)=p_0(Y\mid\mathcal E)$, and Lemma~\ref{lem:four_kl} with
$X=C$ and $S=\mathcal E$ gives
\[
 I_\epsilon(Y;C\mid\mathcal E)
 =\mathsf D_{(Y,C)}(\epsilon)-\mathsf D_C(\epsilon)
  -\mathsf D_{(Y,\mathcal E)}(\epsilon)+\mathsf D_{\mathcal E}(\epsilon);
\]
the cross term of the chain rule vanishes exactly because of this equality
of baseline posteriors. The joint score at zero is
$s_k(c)=-c^\top B_kc/2=-\sum_{i<j}(B_k)_{ij}c_ic_j$:
the determinant derivative vanishes because $D_k$ is diagonal and $B_k$
has zero diagonal. For any of the four pushforwards $U$,
$\mathsf D_U(0)=\mathsf D_U'(0)=0$ and
$\mathsf D_U''(0)=\mathbb E_0[(\mathbb E_0[s_Y(C)\mid U])^2]$.
The scores conditional on $(Y,\mathcal E)$ and on $\mathcal E$ vanish.
Given $\mathcal E$, distinct sign products $\operatorname{sign}(C_iC_j)$
are orthogonal. Substituting the remaining two scores in
\eqref{eq:four_kl_v32} yields $I''_0=\mathcal A$.

The Taylor remainder can be bounded before integration.
The joint density ratio is
$r_\epsilon(k,c)=\exp\{\theta_k(\epsilon)-\epsilon c^\top B_kc/2\}$,
where $\theta_k=\tfrac12\log\det(I+\epsilon D_k^{-1/2}B_kD_k^{-1/2})$.
On $[-t,t]$, $|\theta_k''|\le a_2$, $|\theta_k'''|\le a_3$,
$|\theta_k'|\le ta_2$, and $|\theta_k|\le t^2a_2/2$.
Each of the four statistics retains $u=\|c\|^2$; taking conditional
expectations under $p_0$ therefore preserves the following ratio bounds:
\[
 |\log r_{\epsilon,U}|\le h(u),\quad
 |r_{\epsilon,U}'|\le r_{\epsilon,U}L(u),\quad
 |r_{\epsilon,U}''|\le r_{\epsilon,U}(L(u)^2+a_2),
\]
\[
 |r_{\epsilon,U}'''|\le
 r_{\epsilon,U}(L(u)^3+3L(u)a_2+a_3).
\]
For $\varphi(x)=x\log x-x+1$, differentiation of
$\varphi(r_{\epsilon,U})$ three times bounds its absolute value by the
integrand in \eqref{eq:weak_remainder_constant_v32}, with $u$ replacing
$Z/\lambda$. Under every baseline class, $\|C\|^2$ is stochastically
bounded by $Z/\lambda$. The envelope is increasing and integrable since
$t\bar b<\lambda$. Thus $|\mathsf D_U'''|\le\mathcal M_t$; dominated differentiation
is justified by the same envelope. There are four terms in
\eqref{eq:four_kl_v32}, giving remainder
$4\mathcal M_t|\epsilon|^3/6$.
\end{proof}

The remainder is conservative. Its positivity certificate may require a
much smaller coupling than the range where the leading term predicts the
numerical gain accurately. The verification reports both facts; it does not
replace the bound by an empirically fitted remainder. Nor does a positive
log-loss gain imply an accuracy gain or justify selecting every nonzero
interaction from finite data.
In the numerical checks, $\mathcal A$ is evaluated by ordinary Gaussian
quadrature. A positive floating-point evaluation of the lower-bound formula
is not itself a rigorous numerical certificate without controlling that
quadrature and rounding error.

\paragraph{An exact approximate-sufficiency upper bound.}
For any positive-definite $\widetilde P_k=R+D_k'$ with common symmetric
$R$ and class-dependent diagonal $D_k'$, the corresponding Gaussian
posterior is energy-measurable. With true $P_k\succ0$, therefore,
\begin{equation}
 I(Y;C\mid\mathcal E)
 \le\frac12\sum_k\pi_k\left[
 \operatorname{tr}(\widetilde P_kP_k^{-1})-r
 -\log\det(\widetilde P_kP_k^{-1})\right].
 \label{eq:approximate_energy_upper_v32}
\end{equation}
Indeed the best energy rule is no worse than this particular rule, whose
posterior KL is bounded by the joint Gaussian KL. This elementary bound
allows a common off-diagonal precision component; setting every class
precision diagonal would unnecessarily penalize information that cancels
from the decision.

\subsection{How observation uncertainty attenuates useful interactions}
\label{sec:uncertainty_gain_v32}

Let $A\mid k\sim\N(0,(D+\epsilon B_k)^{-1})$ with common positive diagonal
$D$ and $B_k$ symmetric with zero diagonal as in \eqref{eq:weak_precision_v32},
and observe $y=LA+\xi$, $\xi\sim\N(0,R)$, $R\succ0$.
The observation map is fixed independently of $y$. Put $\Gamma=D^{-1}$,
and let $(m,V)$ be the posterior under the common reference prior
$\N(0,\Gamma)$. Then $m$ has baseline covariance $\Sigma_m=\Gamma-V$.
The class score at $\epsilon=0$ is
\begin{equation}
 s_k(y)=-\tfrac12\operatorname{tr}
       [B_k(mm^\top+V-\Gamma)]
       =-\tfrac12\operatorname{tr}[B_k(mm^\top-\Sigma_m)].
 \label{eq:uncertain_interaction_score_v32}
\end{equation}
This follows by differentiating the latent density under its Gaussian
observation integral. If $\overline B=\sum_k\pi_kB_k$, a Gaussian fourth
moment gives
\begin{equation}
 I_\epsilon(Y;y)
 =\frac{\epsilon^2}{4}\sum_k\pi_k
   \operatorname{tr}[\Sigma_m(B_k-\overline B)\Sigma_m(B_k-\overline B)]
   +O(|\epsilon|^3).
 \label{eq:observed_total_information_v32}
\end{equation}
When $\Sigma_m$ is diagonal, deleting its zero-variance coordinates if necessary,
the energy loss has the simpler expansion
\begin{equation}
 I_\epsilon(Y;y\mid(m_1^2,\ldots,m_r^2))
 =\frac{\epsilon^2}{2}\sum_{i<j}(\Sigma_m)_{ii}(\Sigma_m)_{jj}
       \operatorname{Var}_{k\sim\pi}((B_k)_{ij})
       +O(|\epsilon|^3).
 \label{eq:uncertainty_attenuation_v32}
\end{equation}
To verify the second expression, $m$ is sufficient for this fixed observation
family and its baseline signs are uniform conditional on its squares;
because $\Sigma_m$ is diagonal and $B_k$ has zero diagonal,
$\operatorname{tr}(B_k\Sigma_m)=0$, so $s_k=-\sum_{i<j}(B_k)_{ij}m_im_j$ and
$\mathbb E[s_k\mid m_1^2,\ldots,m_r^2]=0$. The variance of
$\sum_{i<j}(B_k)_{ij}m_im_j$ is the displayed sum. Thus poorly observed
modes, for which $(\Sigma_m)_{ii}$ is small, contribute little even when their latent
precision interaction is large. With general correlated $V$, the trace
formula remains valid, but the separate-edge formula does not.

The $O(|\epsilon|^3)$ statements here are ordinary local expansions for
positive-definite Gaussian covariance paths on a fixed finite observation
space, not the numerical constant in
\eqref{eq:weak_remainder_constant_v32}. Smooth matrix inverses and a compact
positive-definite neighborhood give an integrable polynomial--Gaussian
envelope for three derivatives, justifying Taylor's theorem. The explicit
uniform remainder in Theorem~\ref{thm:weak_interactions_v32} applies to its
linear precision path. Neither statement treats a carrier selected from the
same test signal as a fixed observation operator.

\subsection{Information lost by an imperfect learned mode span}

Whiten a known common noise shape and write
\begin{equation}
 X=HA+\xi,\qquad \xi\sim\N(0,\tau I_p),\qquad
 S=H\operatorname{Cov}(A)H^\top.
 \label{eq:learned_span_model_v32}
\end{equation}
The noise is independent of $(Y,A)$, and $Y\to A\to X$ is a Markov chain.
The latent law need not be Gaussian in the following lemma. Let $\Pi$ be
an orthogonal projector fixed independently of the new observation, and
let $Z=U^\top X$, with orthonormal columns of $U$ spanning $\Pi$.

\begin{lemma}[Prediction cost of signal leakage]
\label{lem:span_leakage_v32}
If $\operatorname{Cov}(A)$ is finite, then
\begin{equation}
 I(Y;X\mid Z)
 \le\frac12\log\det\left[I_p+\tau^{-1}(I_p-\Pi)S(I_p-\Pi)\right]
 \le\frac{\operatorname{tr}((I_p-\Pi)S)}{2\tau}.
 \label{eq:span_leakage_v32}
\end{equation}
\end{lemma}
\begin{proof}
Complete $U$ to an orthogonal basis and denote the residual coordinates by
$X_\perp$. Conditional data processing gives
$I(Y;X_\perp\mid Z)\le I(A;X_\perp\mid Z)$.
The two orthogonal noise components are independent. Conditional on $Z=z$,
the entropy bound by a Gaussian with the same covariance gives
\[
 I(A;X_\perp\mid Z=z)
 \le\tfrac12\log\det
 [I+\tau^{-1}U_\perp^\top H\operatorname{Cov}(A\mid Z=z)H^\top U_\perp].
\]
Concavity of log determinant, the total covariance identity, and
$\mathbb E\operatorname{Cov}(A\mid Z)\preceq\operatorname{Cov}(A)$ prove
the first inequality. The second uses $\log(1+x)\le x$.
\end{proof}

This measures prediction error through leaked signal variance, rather than
requiring each named carrier to be recovered. A sign, permutation, or basis
rotation within the same span has zero leakage. Small physical frequency
errors and an unresolved weak mode have different consequences through $S$.

\subsection{A sampled KMD candidate selector with an off-grid guarantee}
\label{sec:finite_kmd_selector_v32}

The selection objective should match the inference problem. For Koopman
approximation, kernel EDMD learns kernels for a dictionary of observables
\citep{bolager2026dict_kernel_edmd}. For slow-coordinate recovery, adaptive
kernelized diffusion maps use a time-lagged objective to distinguish slow
dynamics from a misleading static spectrum \citep{aboussaad2026diffusion}.
The selector below estimates the signal span from independent records.
The analysis separates the predictive cost of span leakage from the error
of fitting the regime laws.

Consider a finite library of $J$ rank-$q$ mode spans with projectors
$\Pi_1,\ldots,\Pi_J$. The library is either specified before data are read,
or generated by an upstream KMD routine on a separate data split. On $n_D$
fresh independent centered Gaussian records, select
\begin{equation}
 \widehat j\in\arg\max_j\frac1{n_D}\sum_{i=1}^{n_D}X_i^\top\Pi_jX_i.
 \label{eq:projection_energy_selection_v32}
\end{equation}
The class composition of these records is fixed at the target priors
(equal counts suffice for equal priors); their covariances may differ by
class. The selection rule uses no class labels. Assume
$\|\operatorname{Cov}(X_i)\|_{\rm op}\le M$,
$\operatorname{rank}(S)=q$, and the smallest nonzero eigenvalue of $S$ is
at least $\lambda_s>0$. Define the signal leakage and its library optimum by
\[
 L_j=\operatorname{tr}((I-\Pi_j)S),\qquad L_* =\min_jL_j.
\]

The selector uses controlled or stratified records with class composition
fixed at the target priors and labels withheld during selection.
Collection can use regime labels or controlled generation to achieve this
composition; the $n_D$ records count toward the sampling budget. An analysis
of ordinary IID draws from an unlabeled Gaussian mixture would additionally
account for random class-count fluctuations.

\begin{theorem}[Independent energy selection controls mode leakage]
\label{thm:learned_span_selection_v32}
Condition on an independently generated library. With probability at least
$1-\delta$, for $x=\log(J/\delta)$,
\begin{equation}
 \sqrt{L_{\widehat j}}\le\sqrt{L_*}
 +2M\sqrt{\frac2{\lambda_s}}
       \left(\sqrt{\frac{x}{n_D}}+\frac{x}{n_D}\right).
 \label{eq:span_oracle_v32}
\end{equation}
Writing $\ell=\log J$, define
\begin{equation}
 G_D=\frac{16M^2}{\lambda_s}
 \left[\frac{\ell+1}{n_D}
       +\frac{\ell^2+2\ell+2}{n_D^2}\right].
 \label{eq:expected_span_bound_v32}
\end{equation}
Then $\mathbb E L_{\widehat j}\le(\sqrt{L_*}+\sqrt{G_D})^2$.
\end{theorem}

\begin{proof}
Let $\Pi_0$ project onto the rank-$q$ range of $S$. Since
$S\succeq\lambda_s\Pi_0$,
$\|\Pi_j-\Pi_0\|_F^2\le2L_j/\lambda_s$.
Fix a population-optimal library index $j_*$. For
$\Delta\Pi_j=\Pi_j-\Pi_{j_*}$, the triangle inequality gives
\[
 \|\Delta\Pi_j\|_F\le\sqrt{2/\lambda_s}(\sqrt{L_j}+\sqrt{L_*}),\qquad
 \|\Delta\Pi_j\|_{\rm op}\le\|\Delta\Pi_j\|_F.
\]
The population score difference is $L_*-L_j$: the common noise contributes
the same $q\tau$ to both projectors. Gaussian quadratic concentration gives,
simultaneously for every $j$, upper deviation at most
\[
 2M\|\Delta\Pi_j\|_F\sqrt{x/n_D}
 +2M\|\Delta\Pi_j\|_{\rm op}x/n_D.
\]
This inequality follows directly by diagonalizing each centered Gaussian
quadratic form: its log moment generating function is at most
$t^2\|\Sigma_i^{1/2}\Delta\Pi_j\Sigma_i^{1/2}\|_F^2/
(1-2|t|\|\Sigma_i^{1/2}\Delta\Pi_j\Sigma_i^{1/2}\|_{\rm op})$.
Chernoff optimization and a union bound prove the display; these are the
standard Gaussian quadratic ingredients of \cite{laurent_massart_2000}.
For the selected index the empirical difference is nonnegative, hence
$L_{\widehat j}-L_*$ is no larger than the deviation bound. Factor
$L_{\widehat j}-L_*=(\sqrt{L_{\widehat j}}-\sqrt{L_*})
(\sqrt{L_{\widehat j}}+\sqrt{L_*})$ and cancel the nonnegative second
factor; the zero case is immediate. This proves
\eqref{eq:span_oracle_v32}.

For the expectation, set $x=\log J+t$ and integrate the resulting
$e^{-t}$ tail for $(\sqrt{L_{\widehat j}}-\sqrt{L_*})_+$.
Using $(\sqrt a+a)^2\le2(a+a^2)$, and
$\mathbb E T=1$, $\mathbb E T^2=2$ for $T\sim\operatorname{Exp}(1)$,
its second moment is at most $G_D$. Cauchy--Schwarz completes the result.
\end{proof}

\paragraph{A sharper constant for a specified observation design.}
The universal $M$ bound can be replaced by the actual quadratic variances.
For $\Delta\Pi_j=\Pi_j-\Pi_{j_*}$ and class observation covariances $\Sigma_k$, put
\[
 F_j^2=\sum_k\pi_k\operatorname{tr}(\Sigma_k\Delta\Pi_j\Sigma_k\Delta\Pi_j),\qquad
 O_j=\max_k\|\Sigma_k^{1/2}\Delta\Pi_j\Sigma_k^{1/2}\|_{\rm op},
\]
\[
 a_F=\max_{j:\Delta\Pi_j\ne0}\frac{2F_j}{\sqrt{L_j}+\sqrt{L_*}},\qquad
 a_O=\max_{j:\Delta\Pi_j\ne0}\frac{2O_j}{\sqrt{L_j}+\sqrt{L_*}}.
\]
The maximum over an empty set is zero. The same proof gives
$\sqrt{L_{\widehat j}}\le\sqrt{L_*}+a_F\sqrt{x/n_D}+a_Ox/n_D$.
Integrating its tail and applying Cauchy--Schwarz yields the alternative
\begin{equation}
 G_D^{\rm design}=\left[
 a_F\sqrt{\frac{\ell+1}{n_D}}+
 \frac{a_O\sqrt{\ell^2+2\ell+2}}{n_D}\right]^2.
 \label{eq:design_span_bound_v32}
\end{equation}
Either bound, or their minimum, can be used below. These population design
constants can be evaluated exactly for a specified simulation. Estimated
versions require simultaneous upper confidence bounds to provide a
certificate. The variance expression is particularly useful when nearby
candidate spans exchange a small signal direction with a noisy residual
direction.

The fixed-rank restriction is substantive: otherwise the noise energy
$q\tau$ favors larger dictionaries. Independently generated libraries can
include candidate ridges from several KMD bandwidths or refinements; the
same records cannot both generate arbitrary candidates and justify this
finite-library union bound. The selector above is a global projection-energy
reranking of candidate mode spans. It is not a theorem for every upstream
local ridge-tracking rule.

\paragraph{A concrete frequency-grid approximation term.}
For sampled times $|t_i|\le T$, use real chirp columns
$p^{-1/2}\cos(2\pi(ft_i+a t_i^2/2))$ or the corresponding sine.
A frequency error $\Delta f$ and chirp-rate error $\Delta a$ change each
column by at most $d_\phi=2\pi(T|\Delta f|+T^2|\Delta a|/2)$ in Euclidean
norm. If the library includes the corresponding $q$-column approximation
$H_*$, then $(I-\Pi_*)H_*=0$ and
\begin{equation}
 L_*\le\|\operatorname{Cov}(A)\|_{\rm op}\|H-H_*\|_F^2
 \le q\|\operatorname{Cov}(A)\|_{\rm op}d_\phi^2.
 \label{eq:chirp_grid_leakage_v32}
\end{equation}
Columns here retain the displayed $p^{-1/2}$ scaling; normalizing them
requires including the normalization constants. Taking both sine and cosine
columns makes a constant phase change a within-pair rotation. Close or weak
modes can make $\lambda_s$ small and correctly weaken the selection bound.

\subsection{A finite-sample benefit condition for the complete procedure}

Now assume the latent law is centered Gaussian within each class, with known
priors and known zero means. After selecting the span, use a third independent
split with $n_k>q+1$ observations of each class to fit the uncentered
covariance of $Z=U_{\widehat j}^\top X$. Let $\widehat\eta$ be the resulting
Gaussian regime posterior. Define
\begin{equation}
 W(q,n)=\frac12\left[
 \frac{nq}{n-q-1}-q+
 \sum_{i=1}^q\psi\!\left(\frac{n+1-i}{2}\right)
 +q\log(2/n)\right],
 \label{eq:wishart_risk_v32}
\end{equation}
where $\psi$ is the digamma function.

\begin{theorem}[A learned interaction rule can beat an energy oracle]
\label{thm:learned_benefit_v32}
Condition on a fixed or independently generated candidate library.
Under the preceding independent-split assumptions, the expected population
log loss, averaged over dictionary-selection and covariance-fitting data,
satisfies
\begin{equation}
 \mathbb E\mathcal R(\widehat\eta)-H(Y\mid X)
 \le \frac{(\sqrt{L_*}+\sqrt{G_D})^2}{2\tau}
       +\sum_k\pi_k W(q,n_k).
 \label{eq:learned_risk_v32}
\end{equation}
If the right side is smaller than the true energy-oracle deficiency
$\Delta_E=I(Y;X\mid\mathcal E_0(X))$ for a specified energy map
$\mathcal E_0$, the learned rule has strictly smaller expected population
log loss than even the Bayes rule using those energies. A lower bound from
Theorem~\ref{thm:weak_interactions_v32} may replace $\Delta_E$ when its
assumptions hold. Alternatively $\Delta_E$ can be computed for a specified
simulation law; numerical quadrature then supplies numerical evidence,
not a universal analytic lower bound.
\end{theorem}

\begin{proof}
Condition on the selected span. The Bayes loss in its coordinates exceeds
$H(Y\mid X)$ by $I(Y;X\mid Z)$, bounded by
Lemma~\ref{lem:span_leakage_v32}. Each projected class is Gaussian with some
positive-definite covariance $B_k$, regardless of whether the span is exact.
The KL chain rule bounds fitted posterior loss over the projected Bayes loss
by $\sum_k\pi_k D_{\rm KL}(\N(0,B_k)\|\N(0,\widehat B_k))$.
For $n\widehat B\sim\operatorname{Wishart}_q(B,n)$,
\[
 \mathbb E\operatorname{tr}(\widehat B^{-1}B)=\frac{nq}{n-q-1},\quad
 \mathbb E\log\det\widehat B-\log\det B
 =\sum_{i=1}^q\psi((n+1-i)/2)+q\log(2/n).
\]
These standard Wishart expectations \cite{muirhead_multivariate_1982} give exactly
\eqref{eq:wishart_risk_v32}. Averaging the leakage bound using
Theorem~\ref{thm:learned_span_selection_v32} proves
\eqref{eq:learned_risk_v32}. Finally the energy Bayes risk is
$H(Y\mid X)+\Delta_E$.
\end{proof}

Here $W(q,n)=q(q+1)/(4n)+O(n^{-2})$ for fixed $q$. The bound includes
dictionary-selection observations explicitly; it is not a claim that
$n_k$ labels are the entire development budget. It is an expectation over
training, not a promise for every fitted model. It does not require unique
physical labels for the modes, but requires independent training stages,
Gaussian class laws for the covariance step, and common observation noise.
Small $L_*$ and large $n_D$ cannot compensate for $n_k\le q+1$, where the
unregularized expected inverse-covariance loss is infinite.

\paragraph{Estimated noise scale and changed observation quality.}
With unchanged full observations, fitting their projected covariance directly
already includes the noise. Subtracting and re-adding the same estimated
noise variance cancels exactly if no eigenvalue clipping intervenes.
For a change from training variance $\tau_0$ to test variance $\tau_1$, use
$\widehat B_{k,1}=\widehat B_{k,0}+(\widehat\tau_1-\widehat\tau_0)I$.
If independent calibration supplies $N_a$ whitened noise-only scalar
observations at variance $\tau_a$, then, simultaneously for $a=0,1$ with
probability at least $1-\delta_N$,
\[
 |\widehat\tau_a-\tau_a|\le
 e_a=2\tau_a\left(\sqrt{x_N/N_a}+x_N/N_a\right),\qquad
 x_N=\log(4/\delta_N).
\]
The bound is a direct chi-square concentration consequence. Conditional on
the learned span, let $\varepsilon_S=2u+u^2$ with
$u=(\sqrt q+\sqrt{2\log(2K/\delta_C)})/\sqrt{\min_kn_k}$.
If $\|B_{k,0}\|_{\rm op}\le b_0$ and
$\lambda_{\min}(B_{k,1})\ge b_1>0$, then simultaneous relative covariance
error is at most
\begin{equation}
 \gamma=\frac{b_0\varepsilon_S+e_0+e_1+e_{\rm reg}}{b_1}.
 \label{eq:noise_covariance_budget_v32}
\end{equation}
Here $e_{\rm reg}$ bounds any additional operator perturbation introduced
by clipping or regularization; it is zero for the stated unclipped fit.
For $\gamma<1$, the projected fitted posterior excess is at most
$q\gamma^2/[4(1-\gamma)^2]$ on that event. Combine this with the
high-probability leakage bound, using test noise $\tau_1$, to obtain a
noise-calibrated version of the benefit condition. Its simultaneous event
has probability at least $1-\delta_D-\delta_C-\delta_N$ when
$\delta_D$ is used for dictionary selection. Unknown unrestricted
noise shape, dependent calibration residuals, and heterogeneous masks that
alter the learned span require further analysis; they are not covered by
this scalar-scale result.

\subsection{A conditional robustness bound in the mode dimension}
\label{sec:mode_perturbation_v31}

The learned-span bounds above control the leakage of a selected span. The
following proposition complements them with a conditional bound on the
posterior loss caused by errors in the fitted noise, loadings and latent
covariances. Suppose centered true and fitted class models are
$\Sigma_k=N+HC_kH^\top$ and
$\widehat\Sigma_k=\widehat N+\widehat H\widehat C_k\widehat H^\top$,
with common positive-definite noise within each model, positive-semidefinite
latent covariances, the same known class priors in both models, and loading
ranks at most $r$.
Let $U$ be an orthonormal basis for
\[
 \operatorname{col}(N^{-1}H)+
 \operatorname{col}(\widehat N^{-1}\widehat H),
 \qquad d=\dim U\le2r.
\]
Both posteriors depend only on $U^\top v$, by the Gaussian likelihood ratio
and the Woodbury identity. Put $B_k=U^\top\Sigma_kU$ and
$\widehat B_k=U^\top\widehat\Sigma_kU$.

\begin{proposition}[Posterior log-loss under covariance perturbation]
\label{prop:mode_perturbation_v31}
If
$\|B_k^{-1/2}(\widehat B_k-B_k)B_k^{-1/2}\|_{\rm op}
\le\varepsilon_k<1$, the fitted posterior has excess population log loss
for a fresh observation at most
\begin{equation}
 \sum_k\pi_k\frac{d\varepsilon_k^2}{4(1-\varepsilon_k)^2}.
 \label{eq:mode_perturbation_bound_v31}
\end{equation}
If $\lambda_{\min}(N)\ge\nu$, $\|H\|_{\rm op}\le M$,
$\|C_k\|_{\rm op}\le M_C$, and operator errors in noise, loadings and latent
covariance are $e_N,e_H,e_{C,k}$, respectively, a sufficient choice is
\begin{equation}
 \varepsilon_k=
 \frac{e_N+M_C(2Me_H+e_H^2)+(M+e_H)^2e_{C,k}}{\nu}<1.
 \label{eq:mode_error_components_v31}
\end{equation}
\end{proposition}
\begin{proof}
The chain rule for joint KL bounds expected posterior KL by the average
class-conditional KL in the sufficient union coordinates. If a relative
covariance eigenvalue is $1+\lambda$, its Gaussian KL contribution is
$\{(1+\lambda)^{-1}-1+\log(1+\lambda)\}/2$.
The derivative of the braced expression is $\lambda/(1+\lambda)^2$.
Integration bounds the contribution by
$\lambda^2/[4(1-\varepsilon_k)^2]$. Summing at most $d$ terms proves
\eqref{eq:mode_perturbation_bound_v31}. Expanding the two low-rank covariance
factors, applying the operator norm triangle inequality, and compressing
with $U$ gives \eqref{eq:mode_error_components_v31}.
\end{proof}

\begin{corollary}[Relative sensor-noise error]
\label{cor:relative_sensor_noise}
In Proposition~\ref{prop:mode_perturbation_v31}, suppose the centered class
laws and loadings agree. Set $E_N=\widehat N-N$ and assume
$\delta_{\rm obs}=\|N^{-1/2}E_NN^{-1/2}\|_{\rm op}<1$. With
$F_k=B_k^{-1/2}U^\top E_NUB_k^{-1/2}$, the assumed Gaussian posterior has
excess population log loss at most
\begin{equation}
 \frac{\sum_k\pi_k\|F_k\|_F^2}{4(1-\delta_{\rm obs})^2}
 \le \frac{d\delta_{\rm obs}^2}{4(1-\delta_{\rm obs})^2},\qquad d\le2r.
 \label{eq:relative_sensor_noise}
\end{equation}
\end{corollary}
\begin{proof}
The relative error assumption gives
$-\delta_{\rm obs}N\preceq E_N\preceq\delta_{\rm obs}N$. Compress this
inequality by $U$ and use $B_k\succeq U^\top NU$. Thus the eigenvalues of
$F_k$ lie in $[-\delta_{\rm obs},\delta_{\rm obs}]$. Summing their squares
in the scalar Gaussian KL bound of the preceding proof gives
\eqref{eq:relative_sensor_noise}.
\end{proof}

These bounds condition on the stated estimation errors. In the clean case of
known sufficient coordinates, known zero means, and $n$ independent Gaussian
training vectors per class, the usual Gaussian singular-value bounds
\cite{vershynin_highdim_2018} give
relative covariance error at most $\varepsilon=2u+u^2$ simultaneously, where
\[
 u=\frac{\sqrt r+\sqrt{2\log(2K/\delta)}}{\sqrt n}.
\]
With probability at least $1-\delta$, provided $\varepsilon<1$, fitting the
uncentered empirical covariances has excess log loss at most
$r\varepsilon^2/[4(1-\varepsilon)^2]$. If this is below
\eqref{eq:energy_uniform_bound_v31}, the fitted interaction-preserving model
beats every orthogonal energy-only oracle in population log loss on that
training event. The guarantee depends on the mode dimension, while its
conservative constants can require impractically large $n$. Correlated
windows, learned coordinates and heterogeneous-observation EM are outside
this IID covariance argument; they are evaluated separately.

\subsection{Selection from the scored record}
Substituting a record-dependent dictionary $H(x)$ into a fixed-dictionary
Gaussian density generally fails to account for its selection event.
If a discrete selection rule $J(x)$ is used, the correct density conditional
on $J=j$ has a class-dependent normalizer:
\[
 p_k(x\mid J=j)=\frac{p_k(x)\,\bm1_{\{J(x)=j\}}}
 {\Prob_k(J(X)=j)}.
\]
The corresponding class prior is
\[
 \pi_k(j)=\frac{\pi_k\Prob_k(J(X)=j)}
 {\sum_h\pi_h\Prob_h(J(X)=j)}.
\]
For deterministic $J(X)$ these factors cancel the truncation denominators,
so $\Prob(Y\mid X,J)=\Prob(Y\mid X)$. Here $p_k$ must be a coherent
full-record density. A selected-dictionary working score is generally
neither that density nor its correctly normalized conditional density;
treating it as either requires a coherent generative model. Retaining the
original priors with truncated likelihoods instead specifies a different
class-sampling convention. Integrating a declared latent dictionary index
is also different from conditioning on deterministic selection. The finite-library bound analyzes independent selection data. Extending
this analysis to same-record carrier adaptation is a further research
direction.

The conditioning-controlled representation bound and the fitted-risk
comparison quantify different sources of error. At $\alpha=0.8$ and
$\kappa=1$, Theorem~\ref{thm:conditioned_basis_gap} gives the uniform energy
lower bound $4.30\times10^{-6}$ nats; the sharper orthogonal-basis bound of
Theorem~\ref{thm:energy_uniform_gap_v31}, optimized over the radius
$\varrho$ ($\varrho^\ast\approx1.237$), gives $5.23\times10^{-6}$ nats
(Table~\ref{tab:bound_vs_gap}). Both are far below the Wishart estimation
term $W(2,300)=0.005089$, even before span leakage. The finite-sample benefit
illustrated in Section~\ref{sec:adaptive_verification_v32} therefore uses one
specified energy oracle, rather than a uniform fitted advantage over every
allowed energy basis.

\subsection{Numerical verification across conditions}
\label{sec:adaptive_verification_v32}

The accompanying adaptive-theory verification script
checks the expansion against Gaussian quadrature, the exact expected
Wishart loss against independent covariance draws, and a finite chirp-library
selector with separate selection, covariance-fitting, and test samples.
The companion result file records all sample budgets, off-grid errors and
noise-scale perturbations. Candidate selection is learned from observations;
latent truth is used only for evaluation and for evaluating the theorem's
known simulation constants.

\begin{table}[htbp]
\centering\small
\caption{Independent learned chirp-library check. Each condition uses 300 new fitting records per class and eight independent panels with 6,000 test records each. The bound uses population design constants; the energy-oracle gap is numerically 0.041575 nats. Empirical gains are paired mean NLL reductions against that energy oracle, with standard errors across panels.}
\label{tab:adaptive_theory_v32}
\begin{tabular}{lrrrr}
\toprule
Library condition & $n_D$ & $L_*$ & Excess bound & Observed gain \\
\midrule
On grid & 3,000 & 0.0000 & 0.0199 & $0.0372 \pm 0.0007$ \\
Off grid & 12,000 & 0.0097 & 0.0286 & $0.0401 \pm 0.0011$ \\
Outside library & 3,000 & 3.1653 & 3.2457 & $-0.1873 \pm 0.0025$ \\
\bottomrule
\end{tabular}
\end{table}

\begin{figure}[htbp]
 \centering
 \includegraphics[width=\linewidth]{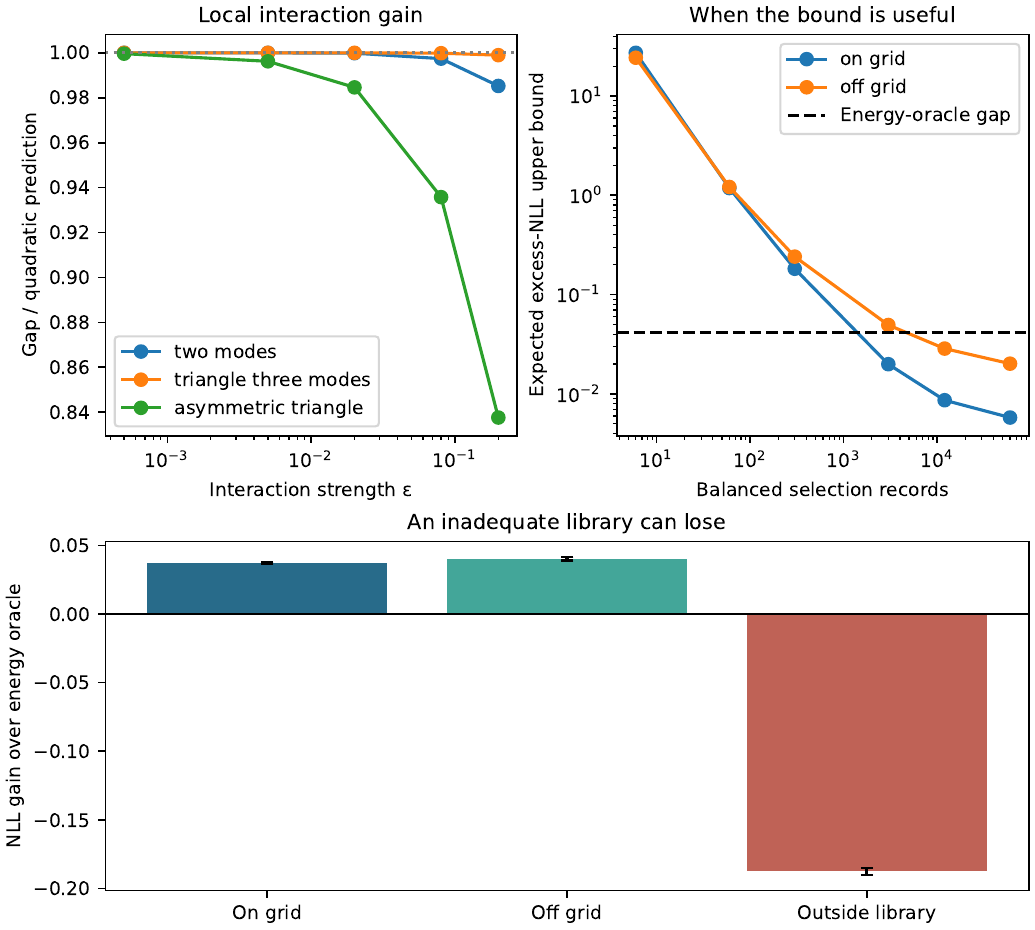}
 \caption{Checks of the quantitative theory. Upper left: the actual energy gap
 divided by the weak-coupling prediction, including a three-mode example
 with a nonzero cubic correction. Upper right: the design-specific expected
 error budget compared with the fixed energy-oracle gap. Lower: observed
 paired log-loss gains for the three conditions in
 Table~\ref{tab:adaptive_theory_v32}; bars show one standard error across
 eight independent panels. The lower panels include the outside-library
 setting.}
 \label{fig:adaptive_theory_v32}
\end{figure}

The 48-sample chirp experiment uses nine predetermined two-dimensional
candidate spans, common noise variance $0.5$, and the three precisions in
\eqref{eq:three_class_precision_v31} with $\alpha=0.8$. Each setting has
eight independent panels. Rank two and zero class means are specified
in advance; they are not estimated. Selection records are generated with
exactly equal class counts, with their labels withheld from the selector.
In the on-grid case the design-specific
expected excess bound is $0.01992$ nats with $3{,}000$ such selection
records and $300$ covariance-fitting records per class. For frequencies
offset by $0.015$ cycles per observation interval from the grid, the bound
is $0.02861$ with $12{,}000$ selection records. Both are below the specified
energy oracle's numerical deficiency of $0.041575$ nats. Quadrature orders
$100$ and $140$ differ by $8.9\times10^{-8}$; this is a numerical convergence
check, not an interval-arithmetic proof of that deficiency. The much looser
universal bound only becomes informative on grid at $60{,}000$ selection
records. When the true frequencies fall outside the candidate library,
the bound makes no positive prediction and the fitted rule loses to the
energy oracle (Table~\ref{tab:adaptive_theory_v32} and
Figure~\ref{fig:adaptive_theory_v32}).

The observed gains are ordered against the leakage: the off-grid gain
($0.0401$) exceeds the on-grid gain ($0.0372$), although positive leakage
$L_*=0.0097$ would predict the opposite. The two condition means differ by
$0.0029$ nats (standard error $0.0013$ over the two independent sets of
eight panels), with eight panels per condition; the off-grid condition also
uses four times as many selection records. The difference comes from test-sample variation in the energy-oracle deficiency
against which the gains are measured: its population value $0.041575$ is
the same in both conditions, but its eight-panel means are $0.0394$ on grid
and $0.0432$ off grid. The directly measured excess of the fitted rule over
the true-span Bayes rule is ordered as leakage predicts: $0.0022$ on grid
and $0.0030$ off grid.

With a fixed span and complete observations, the classifier is zero-mean
QDA on signed coordinates. The experiment checks the error budget for the
specified projection-energy selector and compares the fit with its energy oracle.

\FloatBarrier

\Needspace{12\baselineskip}
\part{Applications}
\label{part:applications}

\section{Experimental design and evidence overview}
\label{sec:experimental_design}
\label{sec:evaluation-design-v32}

We use complementary levels of validation to study representation,
estimation, inference, and transfer. An exactly specified Gaussian
experiment checks the information-loss and estimation calculations. A
nonlinear oscillator experiment tests extracted modes and a fitted working
regime law. Triggered ringdowns, chirped and fading responses, and explicitly
time-inhomogeneous switching test distinct mechanisms beyond critical
transitions. These examples complement microlocal KMD for slow--fast
critical transitions \citep{hamzipaillet}. An external bearing study assesses transfer to independent
physical units and a change in operating conditions. The
prescribed-mode experiment isolates the coupling mechanism under its stated
assumptions. Each study addresses its own target and sampling protocol.

\paragraph{Training-only and per-record adaptation.}
When a dictionary is learned from independent fitting records and frozen,
the Gaussian posterior transfer is an exact calculation conditional on the
fitted dictionary and the assumed observation model. Its error relative to
the true data law includes dictionary and parameter errors. By contrast,
if a dictionary $H(y)$ is selected from the same record $y$ being scored,
substituting $H(y)$ into a Gaussian density does not generally give a
normalized density for $y$. Its conditional covariance also excludes
frequency, phase and model-selection uncertainty. We call such outputs
\emph{working scores} and evaluate their development-calibrated probabilities
on independent records. Exact Gaussian integration conditional on a selected
kernel does not establish exact Bayesian inference for the adaptive pipeline.

\paragraph{Coordinates and physical phases.}
Mode ordering and the sine--cosine convention must be fixed by an extraction
rule available at prediction time. A common coordinate sign change is a
reparameterization if the corresponding fitted class laws change with it.
Randomizing a physical relative phase changes the data law and can destroy
the very interaction used for classification. Simulation ground truth is
reserved for evaluating reconstruction; it cannot align test modes or
construct the classifier inputs. Tests of time shifts and phase variation
therefore have a scientific role beyond checking a numerical implementation.

\paragraph{Comparators and information budgets.}
Separate energies, signed modes, and interaction features are distinct
representations. Each statistical comparison specifies the same observation
footprint and available noise metadata. Strong controls include shrinkage
QDA, quadratic logistic regression, a calibrated direct sparse-QDA control,
and dynamical or spectral summaries appropriate to the task. The sparse-QDA
control reimplements the precision-contrast and linear-index objectives
\citep{jiang_sparseqda_2018}; reference-class contrasts and development-only
temperature/intercept calibration supply its declared multiclass wrapper.
The unpenalized implementation is checked against ordinary Gaussian QDA,
and diagonal penalized cases against closed forms. The dense-Gaussian
implementation checks of the integrated predictions are described once, in
Section~\ref{sec:reference_reduction}.

\paragraph{Replication units and proper scores.}
The fitting and development labels both count toward the available sample
budget. Throughout Part~\ref{part:applications}, a \emph{panel} is one
independently generated and independently fitted simulation replicate,
including its own fitting, development and test data; panels are the
replication units of every simulation study. In the recorded-data studies
the replication units are patients (ECG) and physical bearings. Individual
events, trajectories and windows remain the prediction units within a panel
or a physical unit. Several windows or corrupted views of a
unit do not create additional independent units. Preprocessing, feature
selection, regularization, and probability calibration use fitting or
development data only. Test predictions are retained for every specified
unit, including difficult decompositions. Negative log likelihood is the
primary probability score, accompanied by Brier score and classification
metrics. Reconstruction against a simulated latent signal measures a
different target from restoration toward an originally recorded, noisy
waveform. Uncertainty intervals use the stated independent unit, and their
interpretation accounts for the small number of physical bearings.

\paragraph{Comparison targets.}
An energy-classifier contrast measures the effect of its specified
representation and estimator. Signed-mode and spectral controls assess
richer representations. Probability loss, reconstruction error, and
estimation or computational cost provide separate comparison targets.
The finite-library theory uses its stated sampling and identifiability
conditions; the nonlinear and external studies evaluate fitted working
models.

\paragraph{Effect size and decision tolerance.}
A mean NLL difference $\Delta$ corresponds to a ratio $\exp(-\Delta)$
of geometric mean true-class probabilities, providing a score-scale
interpretation. A symmetric $\varepsilon$-nat tolerance permits score ratios
between $e^{-\varepsilon}$ and $e^{\varepsilon}$. Application-specific
decision costs provide a basis for choosing such a tolerance. The binary
protocol's $0.02$-nat TOST calculations are reported as sensitivity analyses;
the three-regime primary comparison estimates the NLL difference and its
uncertainty.

\paragraph{Confirmatory and exploratory analyses.}
Only three analyses in this paper are confirmatory, in the sense that the
endpoint, comparator family and multiplicity adjustment were fixed before
the corresponding test data were evaluated and a claim is made from the
result: the two-contrast primary family of the binary controlled
experiment (Table~\ref{tab:controlled64primary}), the single primary
endpoint of the three-regime experiment
(Section~\ref{sec:three_regime}), and the two-comparison primary
classification family of the INCART ECG study
(Section~\ref{sec:ecg_repair_v32}). Every other interval and test in
Part~\ref{part:applications}, including all entries of
Table~\ref{tab:focused_scorecard} other than these three, is exploratory or
descriptive, whether or not its comparison was declared in a protocol. The
MIT--BIH and absolute-frequency bearing studies declared their test
families and success criteria in advance; both failed, and their cohorts
then served as development data, so we list them as exploratory.
Retrospective analyses, such as the quality-aware calibration of
Section~\ref{sec:calibration_audit} and the centered three-regime control,
are exploratory by construction.

\paragraph{Interval reporting with few units.}
For studies with at most $12$ independent units (the ringdown, oscillator,
nonstationary, dictionary-mixture, switching, energy-oracle and coupled-mode
studies), the paired $t$ interval across units is the reporting interval.
Percentile bootstrap intervals over so few units are known to undercover
and can take only a small number of distinct values
(Section~\ref{sec:focused_evidence}); they are given only as a sensitivity
analysis. Where the
saved summaries of a study contained only bootstrap intervals (the
oscillator, diagonal-quality ringdown, energy-oracle and coupled-mode
studies), the paired $t$ intervals were computed for this revision from the
saved per-panel results (Appendix~\ref{sec:reproduction}).

\subsection{Outcomes and their statistical interpretation}
\label{sec:focused_evidence}
Table~\ref{tab:focused_scorecard} summarizes the experiments of
Sections~\ref{sec:controlled_studies}--\ref{sec:recorded_studies}.
NLL comparisons use the strongest relevant signed control or clearly name a
representation-restricted comparator. Improvements in reconstruction are
reported separately. The last two columns state whether the analysis is
confirmatory in the sense of Section~\ref{sec:experimental_design} and
whether a prespecified practical criterion (typically a $0.02$-nat NLL
improvement or a $5\%$ MSE reduction) was met.

\begin{table}[tbp]
\centering\footnotesize
\caption{Outcomes against informative controls. NLL is in nats, lower is
better. Values summarize outcomes in the stated evaluation conditions;
Sections~\ref{sec:controlled_studies}--\ref{sec:recorded_studies} give the
conditions and comparisons. Status: Conf.\ = confirmatory, Expl.\ =
exploratory or descriptive (Section~\ref{sec:experimental_design}). The
MIT--BIH and absolute-frequency bearing families were declared in advance
but failed and became development data; they are listed as exploratory.
Criterion: whether the prespecified practical criterion was met (``none''
if the paper states no such criterion for that study).}
\label{tab:focused_scorecard}
\setlength{\tabcolsep}{2.5pt}\renewcommand{\arraystretch}{1.12}
\begin{tabular}{@{}>{\raggedright\arraybackslash}p{0.15\linewidth}>{\raggedright\arraybackslash}p{0.28\linewidth}>{\raggedright\arraybackslash}p{0.32\linewidth}>{\raggedright\arraybackslash}p{0.095\linewidth}>{\raggedright\arraybackslash}p{0.105\linewidth}@{}}
\toprule Study / units & Result against stated control & Evaluation conditions & Status & Criterion met\\\midrule
Binary signed modes / 64 panels (\S\ref{sec:controlled64})
& Integrated NLL $0.5826$, selected signed $0.5846$, same-law point $0.6511$.
& The point contrast reflects the point rule's temperature grid; quality-aware calibration removes $97.7\%$ of it (\S\ref{sec:calibration_audit}).
& Conf.\ (primary family)
& Illustrative margin only\\
Three-regime continuous quality / 64 panels (\S\ref{sec:three_regime})
& Integrated NLL $1.03371$, selected signed $1.03617$, centered signed sensitivity $1.03487$.
& NLL improves by $0.00246$ over the selected signed control and by $0.00116$ over the retrospective centered control; biased monitoring favors the centered control.
& Conf.\ (primary endpoint); centered: Expl.
& None declared\\
Adaptive oscillators / 8 panels (\S\ref{sec:adaptive_experiments_v32})
& Energy-interaction logistic means $0.615/0.724$, versus sKMD $0.636/0.749$.
& Source contrast inconclusive: $+0.0210$, $t$ $[-0.0334,0.0755]$; shift contrast unresolved: $+0.0242$, $t$ $[-0.0079,0.0563]$ (bootstrap $[0.00010,0.0489]$).
& Expl.
& None stated\\
Triggered ringdown / 6 panels (\S\ref{sec:ringdown_v32})
& Source sKMD $1.021$ versus signed OAS $0.831$. Noise-shift NLL contrast inconclusive.
& One $179$-nat event; trigger/dynamics shift also fails. Noisy MSE improves $5.4\%$ over pooled KMD.
& Expl.
& No\\
Chirps and fading / 12 panels each (\S\ref{sec:nonstationary_v32})
& Fading noise-shift gain over OAS is $0.031$; its interval does not certify $0.02$ nats. Chirp contrast inconclusive.
& Fading MSE improves $26\%$ over independent-mode KMD, $2.2\%$ over pooled KMD.
& Expl.
& No (fading NLL point estimate only)\\
Switching / 12 panels (\S\ref{sec:switching_v32})
& Noise-shift NLL gain over point OAS is $0.264$; changed hazards reverse the comparison.
& MSE gain is $9.5\%$ over pooled KMD, $2.0\%$ over point-weighted conditional means.
& Expl.
& Yes for noise-shift NLL and MSE vs pooled; no elsewhere\\
Carrier averaging / 12 panels per family (\S\ref{sec:dictionary_mixture_v32})
& In-library MSE improves $24.0/21.3\%$ over coupled MAP selection.
& Same-library pooled mixtures recover most of it: extra coupling gain $1.52\%$ (chirp, resolved) / $1.08\%$ (fading, unresolved); independent-mode mixture better for fading.
& Expl.
& No\\
Global MIT--BIH / 44 patients (\S\ref{sec:ecg_v32})
& Gaussian-shift sKMD $0.338$, point QDA $0.298$, waveform PCA/RBF $0.280$.
& No coupled gain survives the declared Holm family; PCA reconstruction is better.
& Expl.
& No\\
Robust INCART / 31 patients (\S\ref{sec:ecg_repair_v32})
& Added-noise NLL $0.0394$ versus point $0.0965$ and PCA43 $0.0631$. One-sided zero-margin Holm $p=0.0423$: suggestive.
& Neither $0.02$-nat margin passes. Point two-sided $t$ interval and both fold-level intervals cross zero; Global7 wins three conditions; reconstruction worse than pooled prior and PCA43.
& Conf.\ (primary family)
& No\\
Absolute-frequency bearings / 15 bearings (\S\ref{sec:external-bearing-v32})
& At unseen speed all 12 learned heads exceed uniform NLL $1.099$; sKMD gives $10.346$.
& Unseen-speed transfer with absolute-frequency coordinates.
& Expl.
& No\\
Order-coordinate bearings / 14 bearings (\S\ref{sec:paderborn_repair_v32})
& Severe simulated telemetry: integrated $0.831$, point $1.026$, quality forest $1.065$.
& Milder telemetry favors point inference; native integration selects zero; one healthy unit.
& Expl.
& No\\
\bottomrule
\end{tabular}
\end{table}

\paragraph{Observation quality in recorded-data tests.}
The main favorable ECG comparison adds Gaussian noise and missing samples to
recordings; its noise monitor is supplied by that experiment. Bearing
telemetry likewise changes a declared observation operator. These tests
assess classification and reconstruction under supplied corruption laws.
Reconstruction is evaluated against recorded reference waveforms, which
include physiological and sensor variation.

\paragraph{Small panels, cross-fitting and multiplicity.}
The six-panel bootstrap has only $\binom{11}{6}=462$ possible count vectors
(and potentially fewer distinct statistic values). Increasing the number
of Monte Carlo bootstrap draws does not create more independent panels.
The corresponding intervals are descriptive. The controlled studies
use independently fitted panels. ECG patient intervals also condition
on overlapping training cross-fits; 31 patients are not 31 independently
trained models.

The comparison inventory enumerates every focal-method comparison in a
declared set of result summaries, with its source, condition, metric and
statistical unit. Appendix~\ref{sec:focused_inventory} reports
the counts by study. Holm adjustments control error only within their
declared local families and give no study-wide guarantee. Claims rest only
on the three confirmatory families listed in
Section~\ref{sec:experimental_design}; all other intervals in
Table~\ref{tab:focused_scorecard} are exploratory.

\section{Controlled mechanism and finite-sample tests}
\label{sec:controlled_studies}
These controlled studies connect population information with finite-sample
inference under specified observation laws. The binary construction gives
an exact energy-information calculation and a scalar quality correction.
A separate three-regime experiment retains partial energy information and
introduces a noise-dependent quadratic term outside the affine span of the
noiseless log-odds contrasts.
\subsection{When a quality-dependent calibration replaces integration}
\label{sec:quality_calibration}
An integration comparison depends on the permitted calibration of its point
control. Two Gaussian examples make that dependence explicit. The identities
below concern known population laws; separately estimated means and
covariances need not preserve their exact symmetry.

\begin{proposition}[Scalar calibration in the binary correlation family]
\label{prop:binary_quality_temperature}
Let $C_\pm=\left(\begin{smallmatrix}1&\pm\rho\\\pm\rho&1\end{smallmatrix}\right)$,
$0<\rho<1$. Assume $A\mid Y=\pm\sim N(0,C_\pm)$ with equal priors
and $Z=A+\varepsilon$, where $\varepsilon\sim N(0,sI_2)$ is independent
of $(A,Y)$ and $s\ge0$ is known. Write $\ell_s(z)=\log[p_+(z\mid s)/p_-(z\mid s)]$. Then
\[
 \ell_s(z)=\frac{2\rho z_1z_2}{(1+s)^2-\rho^2}
 =\frac{\ell_0(z)}{T(s)},\qquad
 T(s)=\frac{(1+s)^2-\rho^2}{1-\rho^2}.
\]
Thus quality-dependent temperature scaling of the noiseless likelihood
log odds exactly reproduces Gaussian integration in this family. For unequal
priors the likelihood log odds are scaled and the prior log odds are added
unchanged.
\end{proposition}
\begin{proof}
The observed covariances have common determinant $(1+s)^2-\rho^2$.
Their inverses have diagonal entries $(1+s)/[(1+s)^2-\rho^2]$ and
opposite off-diagonal entries $\mp\rho/[(1+s)^2-\rho^2]$.
Subtracting the Gaussian log densities cancels determinant and diagonal
terms, giving the displayed cross-product coefficient. Dividing by its
value at $s=0$ proves the temperature identity.
\end{proof}

For $\rho=0.65$ and $s=\sigma^2$, the source qualities
$\sigma=0.05,0.3,0.8,1.6$ require temperatures approximately
$1.009,1.326,3.926,21.214$; at $\sigma=3$ the temperature is $172.429$.
A single global temperature cannot implement this dependence. In
particular, the binary protocol's grid, which ends at $4$, also restricts
its range. The comparison with that point rule measures both restrictions.
A fitted quality-aware control is needed to determine their empirical cost.

\begin{proposition}[A radial statistic beyond affine logit calibration]
\label{prop:three_regime_calibration}
Let $P_0=I_2$, $P_k=I_2+\alpha Q_k$ for $k=1,2$, where
$0<\alpha<1$, $Q_1=\operatorname{diag}(1,-1)$ and
$Q_2=\left(\begin{smallmatrix}0&1\\1&0\end{smallmatrix}\right)$.
Assume $A\mid Y=k\sim N(0,P_k^{-1})$ with equal priors and
$Z=A+\varepsilon$, where $\varepsilon\sim N(0,sI_2)$ is independent of
$(A,Y)$ and $s\ge0$ is known. Define
\[
 D_s=(1+s)^2-\alpha^2s^2,\quad
 a_s=-\frac{s\alpha^2}{(1+s)D_s},\quad b_s=\frac{\alpha}{D_s},\quad
 \beta_s=\frac12\log\frac{(1-\alpha^2)(1+s)^2}{D_s}.
\]
The class-versus-zero log odds are
\[
 \ell_k(z;s)=\beta_s-\tfrac12\{a_s\|z\|^2+b_s z^\top Q_kz\},\qquad k=1,2.
\]
For every $s>0$, these log odds cannot be affine functions of the two
noiseless log odds with coefficients depending only on $s$.
\end{proposition}
\begin{proof}
Since $Q_k^2=I_2$, direct inversion gives
\[
 (P_k^{-1}+sI_2)^{-1}
 =\frac{1+s(1-\alpha^2)}{D_s}I_2+\frac{\alpha}{D_s}Q_k.
\]
Subtracting $(1+s)^{-1}I_2$ yields $a_sI_2+b_sQ_k$.
The observed determinants are $D_s/(1-\alpha^2)$ for $k=1,2$ and
$(1+s)^2$ for class zero, yielding $\beta_s$. At $s=0$, the quadratic
parts of both logits are traceless. Every affine combination retains
that property, whereas the noisy contrasts have trace $2a_s\ne0$.
Equality of these Gaussian posterior functions would imply equality of
the quadratic polynomials, which is impossible.
\end{proof}

This obstruction includes scalar temperature and separate class temperatures
with quality-only offsets. It is specifically an affine-decoder statement:
if $u=z^\top Q_1z$ and $v=z^\top Q_2z$, then
$\|z\|^2=\sqrt{u^2+v^2}$. A nonlinear decoder of the two noiseless logits
can therefore recover the radial statistic. A full signed quadratic
feature head also retains it directly. Unlike the binary reflection
example, separate coordinate energies carry partial class information
through the different marginal variances. The finite-sample experiment
below consequently compares estimated procedures against useful energy
references and strong signed controls.

\subsection{Binary mechanism example: signed information and calibration}
\label{sec:controlled64}
A fixed-carrier experiment separates two questions: what is lost by replacing signed coefficients with energies, and what is gained by integrating measurement uncertainty relative to a strong signed classifier. The study follows a specified protocol (Appendix~\ref{sec:controlled64supp}). Each of $64$ independent panels contains $1200$ fitting events, $800$ development events and $2400$ events per test condition. The statistical unit is the panel.

The two equal-prior coefficient laws are
\[
 C\mid K=k\sim\mathcal N\!\left(0,
 \begin{bmatrix}1&(-1)^k\rho\\(-1)^k\rho&1\end{bmatrix}\right),
 \qquad \rho=0.65,\qquad Z=C+\varepsilon,
 \quad \varepsilon\mid\sigma\sim\mathcal N(0,\sigma^2 I_2).
\]
Appendix~\ref{sec:controlled64supp} specifies all fitting rules and stress conditions.
The dictionary consists of two prescribed orthonormalized damped cosine atoms. Quality $\sigma$ is known to every method and independently takes values $0.05,0.3,0.8,1.6$ in fitting, development and the primary test. These synthetic modal observations use the supplied noise levels. The signed and energy interaction classifiers have equal feature dimensions and tuning budgets. Development NLL selects a signed comparator from quality-aware interaction logistic, OAS-QDA and same-law point inference; the first is selected in all $64$ panels.

\begin{table}[htbp]
\centering\small
\caption{Primary NLL differences: integrated Gaussian minus control. The
first two rows are the prespecified primary family, with two-sided Holm
adjustment; intervals use $64$ paired panel means. The same-law point rule
calibrates one global temperature on a grid ending at $4$, whereas the exact
correction at $\sigma=1.6$ is $21.214$
(Proposition~\ref{prop:binary_quality_temperature}), so its contrast
largely reflects that grid restriction. The last row, from the retrospective
analysis in Section~\ref{sec:calibration_audit}
(Table~\ref{tab:quality_calibration_audit}), is not part of the Holm family.}
\label{tab:controlled64primary}
\begin{tabular}{@{}lrrr@{}}\toprule
Control & Difference & Paired $t$ 95\% interval & Holm $p$\\\midrule
Selected signed & $-0.00200$ & $[-0.00301,-0.00099]$ & $0.000184$\\
Same-law point, bounded grid & $-0.06847$ & $[-0.07159,-0.06535]$ & $1.29\,10^{-48}$\\\midrule
\multicolumn{4}{@{}l}{\emph{Retrospective, exploratory:}}\\
Smooth quality-temperature point & $-0.00155$ & $[-0.00214,-0.00095]$ & ---\\\bottomrule
\end{tabular}
\end{table}

\paragraph{Integration and a strong signed head.}
Primary mean NLLs are $0.58260$ for Gaussian integration, $0.58460$ for the selected signed head and $0.65107$ for same-law point inference. The integrated-minus-selected difference is statistically resolved, with a $90\%$ interval of $[-0.00284,-0.00116]$ nats. A $0.00200$-nat reduction corresponds
to a factor $e^{0.00200}=1.00200$ in geometric mean true-class probability.
The protocol's illustrative $\pm0.02$-nat margin gives TOST Holm
$p=1.46\,10^{-43}$, but no decision cost justifies that margin here.
It therefore supplies a sensitivity calculation, not evidence of practical
interchangeability. The integration-only contrast exceeds $0.02$ nats
(one-sided practical-margin Holm $p=6.94\,10^{-40}$), against the specific
globally calibrated point rule described in the protocol. Point inference shares the selected Gaussian law and receives its own development calibration.
This contrast was prespecified and is reported as such, but it largely
measures the point rule's temperature grid rather than the value of
integration: the grid ends at $4$, Proposition~\ref{prop:binary_quality_temperature}
requires $21.214$ at $\sigma=1.6$, and the quality-aware calibration of
Section~\ref{sec:calibration_audit} removes $97.7\%$ of the contrast
(last row of Table~\ref{tab:controlled64primary}). Accuracy is almost unchanged: $65.23\%$ versus $65.25\%$. The gain primarily concerns probability calibration across quality levels.

\paragraph{Signed information and the randomized-sign control.}
The direct matched-budget signed-minus-energy logistic difference is $-0.10906$ nats, with a secondary pointwise $t$ interval $[-0.11107,-0.10706]$. Independently randomizing the second mode sign in every fitting, development and test event makes the two observation laws identical; the corresponding difference is $-0.00006$, $[-0.00037,0.00025]$. In the positive arm, reflection of coordinate two exchanges the classes
while preserving both energies, so the population energy posterior is exactly
uniform. The signed-versus-energy result illustrates the reflection symmetry in finite samples.
The null arm checks the generator and the fitted procedure's behavior when
the signed signal is also absent. It concerns the prescribed energy representation; rotating the dictionary can preserve the interaction for these two classes.

\begin{figure}[t]
\centering\includegraphics[width=0.93\linewidth]{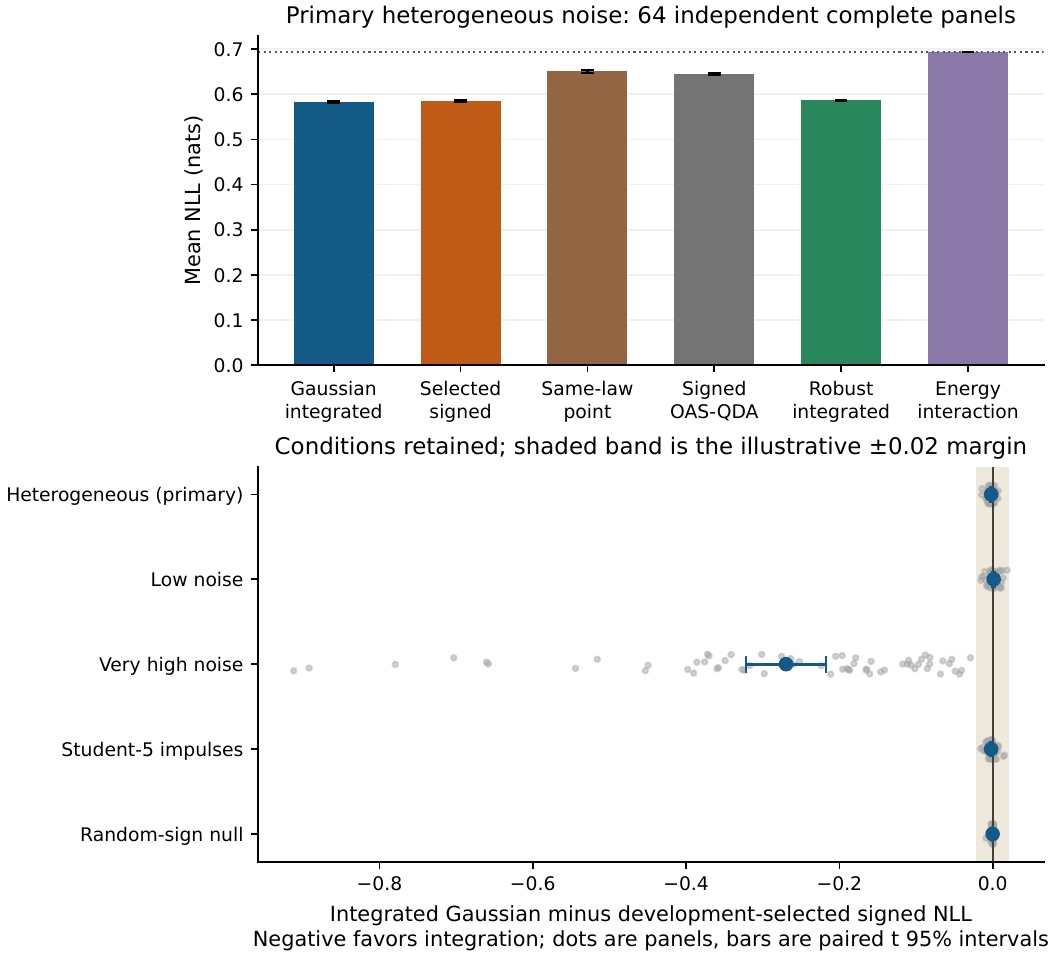}
\caption{Controlled modal observations. Top: primary NLL means and panel $t$ 95\% intervals; the dotted line is uniform prediction. Bottom: Gaussian integration minus the development-selected signed control across every condition. Dots are panel differences; bars are paired $t$ intervals. Only the heterogeneous-noise comparison belongs to the primary family. The shaded band is the illustrative $\pm0.02$-nat margin.}
\label{fig:controlled64}
\end{figure}

Figure~\ref{fig:controlled64} compares the stress conditions. At out-of-range $\sigma=3$, integrated NLL $0.69122$ is close to uniform $\log2$, and accuracy is only $51.98\%$: conservative probabilities do not recover useful discrimination. Under covariance-preserving Student-$5$ \emph{latent impulses}, the robust integrated scale mixture improves NLL from $0.60603$ to $0.59735$; observation errors remain Gaussian. Primary coefficient MSE improves $1.08\%$ over the pooled full prior and $0.11\%$ over selected-signed-weighted conditional means. The experiment shows how calibration and representation choices affect
fitted risk. Section~\ref{sec:quality_calibration} and the following analysis
separate the effects of global and quality-dependent calibration.

Two additional controlled studies separate population information from
fitting error. Section~\ref{sec:controlled_benchmarks} gives exact
energy-oracle comparisons for the three-regime obstruction and the complete
supplied/learned-carrier fitting grid. They use different class laws and
replication units from the 64-panel experiment, and are not pooled with it.

\subsection{Quality-aware temperature calibration}
\label{sec:calibration_audit}
The population identity motivates a retrospective calibration analysis of
the binary study. The fitted Gaussian laws and test observations are fixed,
and calibration parameters are estimated using development observations
only. This analysis assesses sensitivity to the permitted calibration
family.

\Needspace{9\baselineskip}
Let $\delta_i$ be the untempered point log odds and $v_i=\sigma_i^2$.
We minimize development binary log loss
\[
 \frac1{n_{\rm dev}}\sum_i
 \{\log(1+e^{\beta(v_i)\delta_i})-y_i\beta(v_i)\delta_i\}
\]
with stable numerical evaluation. Three declared families are fitted:
a constant $\beta\ge0$; four separate nonnegative inverse temperatures,
one per observed source quality; and
\[
 \beta(v)=\frac{a}{1+bv+cv^2},\qquad a,b,c\ge0.
\]
The smooth family is informed by the isotropic observation model and
contains the exact population correction, but its parameters are learned
from development labels without using the generating correlation. It uses
eight declared optimization starts. The constant and four-quality fits
are convex one-dimensional problems. The added parameter counts are
$1,4,3$ beyond the shared fitted Gaussian law; these are matched-data and
matched-law comparisons, not equal-complexity models. All three are
reported without selecting a winner on test data. Four-quality calibration
has no defined output at unseen $\sigma=3$.

\begin{table}[t]
\centering\small
\caption{Retrospective quality-aware calibration on the 64 source-quality panels. All added calibrators use development observations only and use the same fitted Gaussian law. Negative differences favor integration; intervals use 64 paired panel means. The known-law Bayes NLL of the design ($\rho=0.65$, equiprobable $\sigma\in\{0.05,0.3,0.8,1.6\}$) calibrates the size of the remaining differences.}
\label{tab:quality_calibration_audit}
\begin{tabular}{lrr}
\toprule
Method & Mean NLL & Integrated minus method (95\% CI)\\
\midrule
Point, bounded grid & 0.65107 & -0.06847 [-0.07159, -0.06535]\\
Global temperature & 0.64579 & -0.06319 [-0.06496, -0.06142]\\
Four quality temperatures & 0.58486 & -0.00227 [-0.00295, -0.00158]\\
Smooth quality temperature & 0.58414 & -0.00155 [-0.00214, -0.00095]\\
Selected signed control & 0.58460 & -0.00200 [-0.00301, -0.00099]\\
Integrated Gaussian & 0.58260 & ---\\
\midrule
Known-law Bayes oracle (population) & 0.57842 & ---\\
\bottomrule
\end{tabular}
\end{table}

\begin{figure}[tbp]
\centering\includegraphics[width=0.94\linewidth]{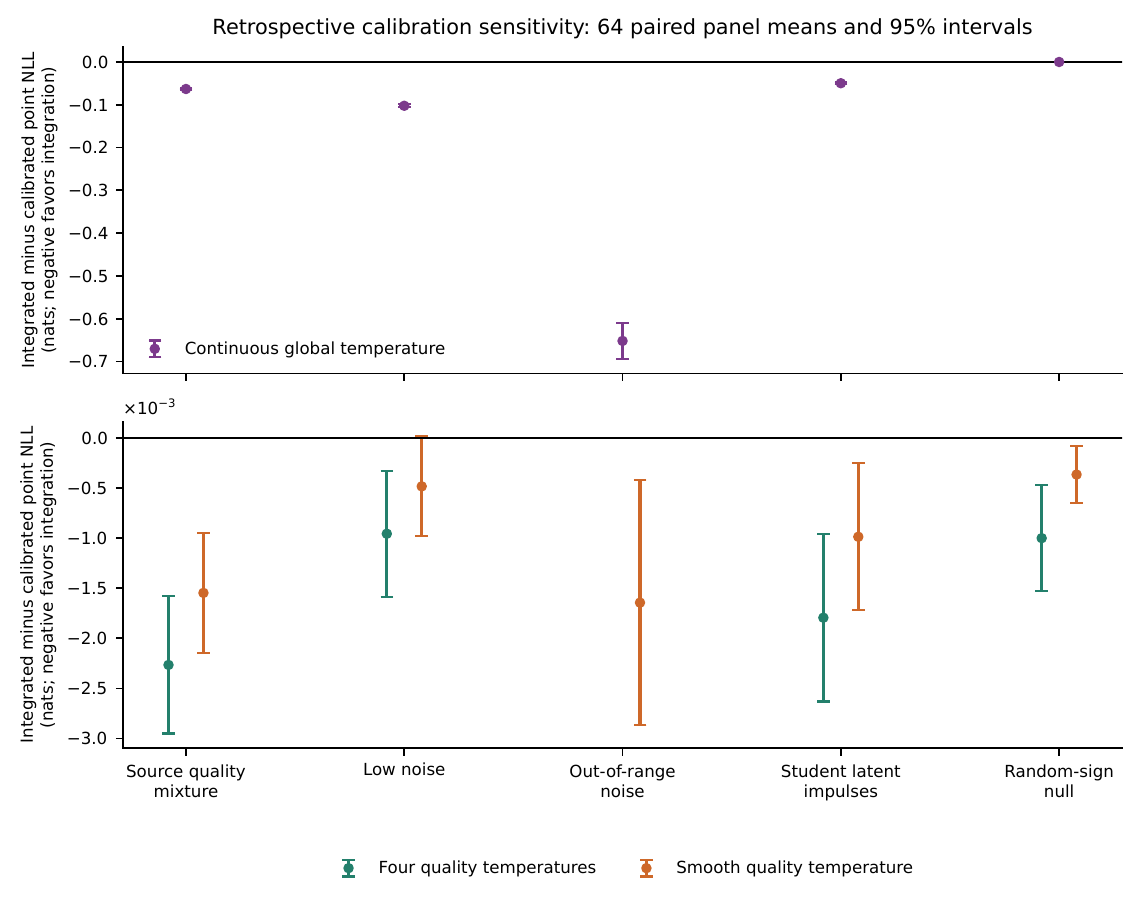}
\caption{All retrospective calibration contrasts on the same binary panels, with
separate vertical scales for global and quality-aware calibration.
Error bars are pointwise paired $t$ intervals. Smooth model-informed calibration accounts for most of the NLL difference between integration and globally calibrated point inference. Four-quality calibration is omitted at the
unseen quality because its rule has no output there.}
\label{fig:quality_calibration_audit}
\end{figure}

On the source mixture, continuous global calibration gives NLL $0.64579$,
compared with $0.65107$ for the bounded global grid. Most of the difference
remains. Four quality temperatures give $0.58486$, and smooth calibration
gives $0.58414$. The integrated-minus-smooth difference is $-0.001547$ nats,
with paired pointwise $95\%$ interval $[-0.002144,-0.000950]$. Thus allowing
quality calibration removes about $97.7\%$ of the $0.06847$-nat contrast
against the bounded-grid point rule (Table~\ref{tab:quality_calibration_audit}
and Figure~\ref{fig:quality_calibration_audit}).

The remaining integrated-minus-smooth residual is statistically resolved.
We conjecture that it reflects finite-sample fitting error and departures
of the estimated laws from exact reflection symmetry, but no analysis here
tests that attribution. For scale, the known-law Bayes NLL of the design
($\sigma$ observed by every method) is $0.5784$ nats, so the fitted
integrated rule ($0.5826$) is $0.0042$ nats above it; on the saved
$64\times2400$ primary test events the realized oracle NLL is $0.5789$. In
population, the integrated method's distance from the oracle bounds what any
calibration or representation change could still gain.

At $\sigma=3$, smooth calibration gives NLL $0.69287$ versus integrated
$0.69122$, a difference of $-0.001645$ with interval
$[-0.002870,-0.000419]$. It therefore also captures nearly all the conservative
probability behavior that the quality-interaction logistic head misses
outside its fitting support. The near-chance discrimination in this
condition persists. Figure~\ref{fig:quality_calibration_audit} also shows
the calibration contrasts in the random-sign null and latent-tail stress
conditions.

\subsection{Three regimes with partial energy information}
\label{sec:three_regime}
The binary example makes the energy posterior uniform by symmetry. We next
use the three-regime precision family in
\eqref{eq:three_class_precision_v31}, with $\alpha=0.65$, equal priors and
known zero means. Observations are $Z=A+\epsilon$, with
$\epsilon\mid\sigma\sim\N(0,\sigma^2I_2)$ and independent, continuously
varying $\sigma\sim\operatorname{Unif}[0.05,1.6]$. Coordinate energies now
retain class information. The noise-dependent radial term in
Section~\ref{sec:quality_calibration} makes scalar calibration of noiseless
logits insufficient, while a full signed quadratic control can retain it.

\paragraph{Independent finite-sample test.}
The protocol specifies $64$ independent panels, fitting budgets
of $600$ and $2400$ events, $1200$ development events, and $3000$ test events
per condition. Every split has balanced classes. Budgets use separate draws;
conditions within a budget share a fitted model. The fixed two-carrier
experiment supplies modal measurements directly and performs no carrier
extraction. The primary endpoint is integrated Gaussian minus
development-selected signed NLL at $2400$ fitting events and continuous
source quality. All other contrasts and stress conditions are secondary.
No practical or equivalence margin is asserted.

\begin{figure}[tbp]
\centering\includegraphics[width=\linewidth]{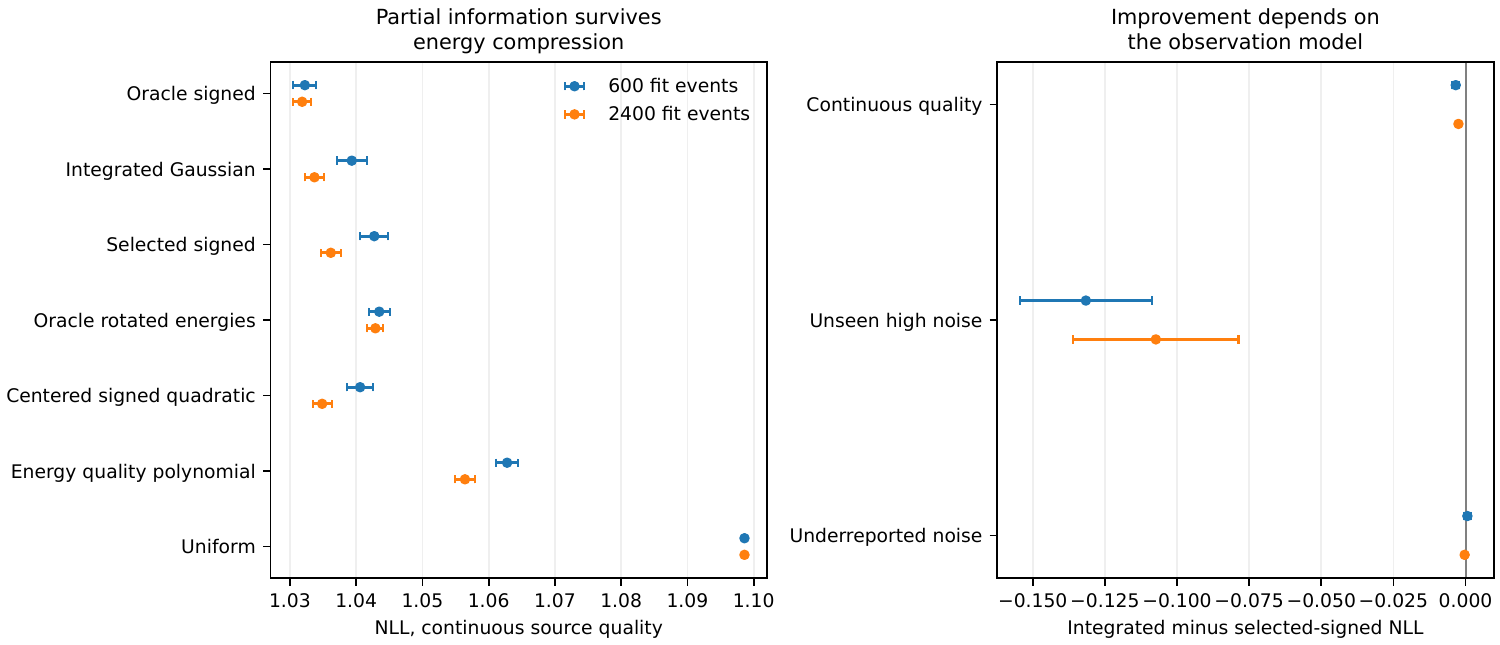}
\caption{Three-regime experiment with continuous observation quality.
Left: source NLL and panel $t$ intervals at two fitting budgets;
energy references retain useful information. Right: integrated minus
selected-signed NLL for every condition. Only the larger-budget source
comparison is primary. The high-noise condition combines lower NLL with near-chance accuracy; underreported observation noise gives an unresolved integrated-versus-selected contrast.}
\label{fig:three_regime}
\end{figure}

The integrated estimator fits noise-corrected second moments, selects
covariance shrinkage on development observations, and receives a continuous
nonnegative inverse temperature. The point control receives the same
continuous calibration range. The signed selector chooses between a
17-feature quality-aware quadratic classifier, an 8-feature calibration of
point logits and quality, and signed OAS-QDA. The fitted energy polynomial
control has the same 17-feature size and tuning grid as the signed
quadratic control. Separate known-law signed and energy posteriors measure
representation loss; a rotated-energy oracle chooses among $37$ orthogonal
bases using development NLL. Its true parameters are privileged information,
so it is an information reference, not a matched fitted competitor.
Appendix~\ref{sec:three_regime_protocol} gives the complete fitting rules.
Table~\ref{tab:three_regime_primary} reports the source-condition NLL of every
method at both budgets, and Figure~\ref{fig:three_regime} shows the paired
differences in every condition.

\begin{table}[htbp]
\centering\small
\caption{Continuous-quality three-regime NLL, averaged over 64 independently
fitted panels. Energies carry partial information: both energy oracles improve
on uniform prediction, $\log3=1.09861$. Only the $2400$-event integrated-versus-selected-signed
contrast is primary.}
\label{tab:three_regime_primary}
\begin{tabular}{@{}lrr@{}}\toprule
Method & 600 fit events & 2400 fit events\\\midrule
Integrated Gaussian & 1.03934 & 1.03371\\
Development-selected signed & 1.04273 & 1.03617\\
Centered signed quadratic (sensitivity) & 1.04059 & 1.03487\\
Point with continuous global calibration & 1.05992 & 1.05494\\
Quality calibration of point logits & 1.04399 & 1.03612\\
Fitted energy polynomial & 1.06278 & 1.05642\\
Known-law signed oracle & 1.03226 & 1.03185\\
Known-law fixed-energy oracle & 1.05206 & 1.05196\\
Known-law development-rotated energy oracle & 1.04348 & 1.04290\\\bottomrule
\end{tabular}
\end{table}

\paragraph{Recoverable information and fitted performance.}
The primary paired difference is $-0.002463$ nats, with $95\%$ interval
$[-0.002853,-0.002074]$ and two-sided $p=6.44\times10^{-19}$. This corresponds
to a geometric mean true-class probability ratio of $1.00247$, about a
$0.247\%$ increase. The signed selector chooses quality-logit calibration
in $41$ panels and the full quadratic head in $23$. A retrospective
symmetry-matched check enforces central symmetry by omitting odd features
from the signed head, matching the Gaussian fit's known-zero-mean assumption.
This 11-feature
centered control gives NLL $1.03487$; the integrated-minus-centered
difference is $-0.001162$, with pointwise interval
$[-0.001569,-0.000754]$. This retrospective sensitivity analysis is separate
from the primary comparison; it halves the estimated benefit of integration,
and it reverses under the biased monitor below.

At that budget, the known-law signed posterior improves NLL by $0.020115$
over fixed energies and $0.011057$ over the development-rotated energy
oracle. The fitted integrated estimator has a $0.009198$ advantage over
that selected energy oracle, with secondary pointwise interval
$[0.008296,0.010100]$ for the improvement. This demonstrates that the
specified information gap can survive estimation in this experiment.

With $600$ fitting events the selected-signed improvement is $0.003397$
with pointwise interval $[0.002226,0.004567]$; its larger uncertainty reflects
less stable estimation.

\paragraph{Stress conditions and reconstruction.}
At unseen $\sigma=3$, integrated NLL is $1.09578$ versus $1.20310$ for the
selected signed head, but integrated accuracy is only $35.74\%$ against
$33.33\%$ chance. This is mainly a calibration advantage under weak
remaining information. If the noise monitor reports half the true variance,
the integrated-minus-selected difference is $-0.000316$ with pointwise
interval $[-0.000986,0.000355]$; the source advantage is unresolved under
that misspecification. The centered signed sensitivity control is better
than integration there: its NLL is $1.04442$, and the integrated-minus-centered
difference is $+0.000942$, with pointwise interval $[0.000252,0.001632]$.

\Needspace{9\baselineskip}
As a separate population sensitivity check, hold the generating class laws
fixed and use the untempered Gaussian posterior. When the supplied sensor
variance is half its true value, the mean excess NLL is approximately
$0.00943$ nats relative to the same posterior supplied with the true variance.
This calculation uses $24$-point Gauss--Legendre quadrature over source
noise SD and a $64$-point Gauss--Hermite rule in each observation coordinate
within each class. Corollary~\ref{cor:relative_sensor_noise} gives the
corresponding noise-only bound. The fitted contrast above also depends on
the estimated class laws and the calibration of the two heads.

On the primary source condition, coefficient MSE is $0.444668$ for regime
integration, $0.444852$ for the same class-conditional means weighted by the
selected signed head, $0.447044$ for pooled full-prior KMD and $0.453705$ for
independent-prior KMD. The relative reductions are approximately $0.041\%$,
$0.531\%$ and $1.99\%$, respectively. Under the biased monitor, pooled KMD
has lower MSE ($0.488874$ versus $0.491243$). The reconstruction gain of
regime integration over pooled full-prior KMD is therefore about half a
percent at source quality and reverses when the noise monitor is biased.

\section{Learned modes and nonstationary signals}
\label{sec:learned_studies}
The controlled coefficient experiments fix the representation. This section
adds extraction error, time-varying carriers and regime dynamics. The sequence
isolates a progression: energy-sufficient random-phase signals, triggered
signed impulses, changing waveforms, uncertainty about the dictionary, and
changing regime persistence. Each study holds its own observation and fitting
budget fixed across the named controls.
\subsection{Random-phase oscillators: when energy paths suffice}
\label{sec:adaptive_experiments_v32}

Eight independent panels of nonlinear oscillators compare modal interaction
representations with amplitude- and energy-based classifiers.

\paragraph{Adaptive KMD front end.}
We use the released KMD implementation~\cite{owhadi_kernel_2020} and apply
its \texttt{max\_squeeze\_refine} Gaussian energy calculation directly. Two
frequency-ordered, separated peaks initialize local frequency searches; smoothed
ridges define phases with zero at the observed time origin. Joint local
cosine/sine recovery uses the Gaussian weighted least-squares algebra of
\texttt{micloc\_KMD}. This bounded automatic subset handles noncrossing sinusoidal modes; the
complete implementation additionally supports interactive global peeling.
We also run the released complete semimanual algorithm on two prespecified
two-mode chirps. Both runs recover two modes automatically. Clean-signal RMSE is
$1.03\times10^{-5}$ without noise and $0.0507$ with sensor SD $0.15$; the subset
gives $0.00750$ and $0.0351$, respectively.

\paragraph{Nonlinear independent trajectories.}
Writing $u_j=\log r_j$, the simulated amplitude-coupled oscillators obey
\begin{align*}
du_j&=\{1-e^{2u_j}+\kappa\tanh(u_{3-j})\}\,dt+0.60\,dB_j,\\
d\theta_j&=2\pi f_j(t)\,dt+0.03\,dW_j,
\qquad x(t)=\sum_{j=1}^2r_j(t)\cos\theta_j(t).
\end{align*}
The log-coordinate equation defines the stochastic dynamics; its nonlinear
radial drift is the Stuart--Landau radial drift with bounded cross-coupling.
The three labels are $\kappa\in\{-0.9,0,0.9\}$. All initial phases are independent
and uniform. After five seconds of burn-in, each independent trajectory supplies
eight seconds at 128 Hz. Source carrier frequencies are near 6 and 14 Hz with
slow sinusoidal drift, and independent sensor SD is uniform on $[0.15,0.40]$.
The prespecified shift changes carriers to approximately 6.7 and 12.8 Hz, doubles
frequency-drift and phase-noise amplitudes, and increases sensor SD to
$[0.40,0.70]$. The fitting, development, source-test and shifted-test sets contain
30, 20, 30 and 30 independent trajectories per class in each of eight panels.
Fitting, development and test partitions contain disjoint trajectories.
The integration step is $1/512$ s. A separate 12-trajectory shared-Brownian
refinement to $1/1024$ s changes clean signals by RMS $4.15\times10^{-4}$ and
mean recovered amplitudes by $1.13\times10^{-4}$. This comparison measures
sensitivity to the integration step.

\paragraph{Working uncertainty and comparison methods.}
The noise scale is estimated from each observed record's high-frequency power,
without simulator truth. For local synthesis matrix $B$, Gaussian weights $W$
and estimated noise variance $\widehat\sigma^2$, the coefficient sampling
covariance is the sandwich
\[
\widehat\sigma^2(B^\top W^2B)^{-1}B^\top W^4B
                 (B^\top W^2B)^{-1}.
\]
This sampling covariance conditions on the estimated carriers and a locally
constant coefficient model, with carrier selection treated as fixed. A delta approximation transfers it to positive amplitudes. Class
Gaussian amplitude models subtract this estimated measurement variance during
fitting and add the test variance in their working likelihood. Covariance
shrinkage and a single score temperature are chosen by independent-trajectory
development NLL. All controls have explicit, bounded development grids.
The direct sparse-QDA wrapper instead uses its declared continuous inverse
temperature and class-intercept calibration; this gives that comparator a
more flexible calibration model.

The comparison methods include amplitude point QDA; amplitude quadratic logistic;
logistic regression on squared-amplitude means, second moments and cross-products;
direct sparse QDA~\cite{jiang_sparseqda_2018} on the same five amplitude moments; and forests on pooled
energies, marginal energy summaries, or all 40 local energies. A further control
uses envelope correlation and demodulated residual-phase consistency.
The signed quadrature QDA control retains a fixed observed time origin;
random phases make its Gaussian summary ineffective. The complete local
energy trajectory determines the positive amplitudes through square roots,
so both representations retain the same amplitude information in this experiment.

\begin{table}[t]
\centering\small
\caption{Adaptive nonlinear oscillator classification, averaged over eight
independent panels. All methods use the same extracted modes; smaller NLL and
Brier are better. Full accuracy, macro-AP, panel variation and paired intervals
are in the numerical package.}
\label{tab:adaptive_oscillator_v32}
\resizebox{\linewidth}{!}{
\begin{tabular}{lrrrr}
\toprule
Method & Source NLL & Shift NLL & Source Brier & Shift Brier \\
\midrule
Coupled sKMD & 0.636 & 0.749 & 0.394 & 0.461 \\
Amplitude point QDA & 0.636 & 0.763 & 0.394 & 0.465 \\
Signed quadrature QDA & 1.108 & 1.111 & 0.672 & 0.674 \\
Amplitude quadratic logistic & 0.645 & 0.749 & 0.393 & 0.456 \\
Energy interaction logistic & 0.615 & 0.724 & 0.376 & 0.445 \\
Pooled KMD energies + RF & 0.948 & 1.008 & 0.566 & 0.605 \\
Marginal energy summaries + RF & 0.968 & 1.021 & 0.574 & 0.608 \\
Full energy sequence + RF & 1.000 & 1.006 & 0.595 & 0.603 \\
Envelope/phase summaries + logistic & 0.769 & 0.915 & 0.460 & 0.525 \\
Calibrated direct sparse QDA & 0.677 & 0.820 & 0.416 & 0.482 \\
\bottomrule
\end{tabular}
}
\end{table}

\begin{figure}[tp]
\centering\includegraphics[width=\linewidth]{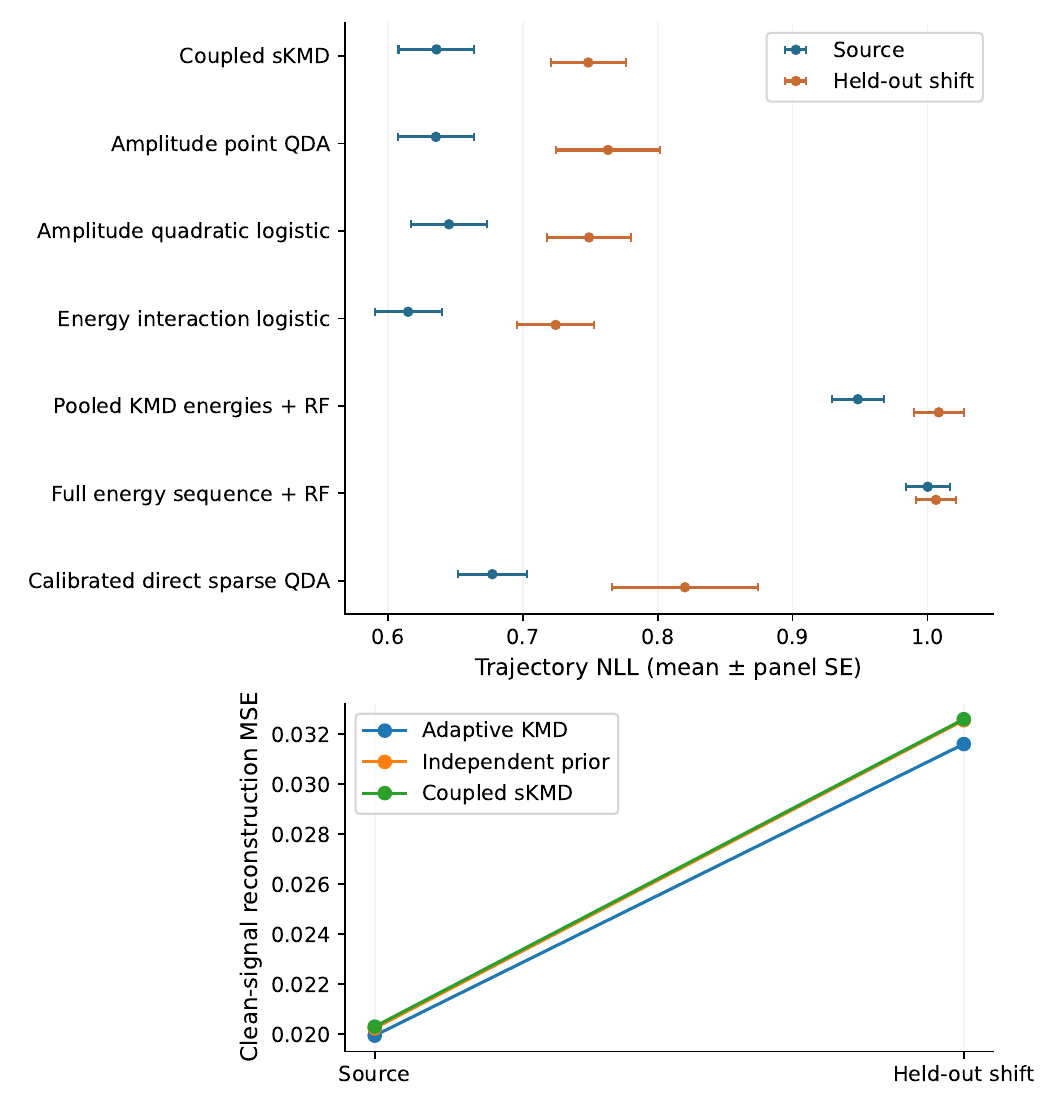}
\caption{Adaptive nonlinear oscillator classification and reconstruction.
Classification error bars are panel standard errors; paired confidence
intervals are reported in the text.}
\label{fig:adaptive_oscillator_v32}
\end{figure}

\paragraph{Classification and reconstruction outcomes.}
No test trajectory has a missing detected peak; mean carrier errors are
$0.038$ Hz in source tests and $0.066$ Hz under the shift. Table~\ref{tab:adaptive_oscillator_v32}
and Figure~\ref{fig:adaptive_oscillator_v32} show that coupled sKMD improves
on the energy forests, while the more explicit energy-interaction logistic
control has lower mean NLL in both conditions; with eight panels neither
paired contrast is resolved. In the source condition, the paired NLL
difference (sKMD minus energy-interaction logistic) is $+0.0210$ with paired
$t$ interval $[-0.0334,0.0755]$ (7 df; bootstrap sensitivity interval
$[-0.0228,0.0599]$), so the source contrast is inconclusive. Under the
shift, it is $+0.0242$; its paired $t$ interval $[-0.0079,0.0563]$
includes zero, so the shift contrast is also unresolved, although the
saved pointwise 95\% panel-bootstrap sensitivity interval
$[0.00010,0.0489]$ just excludes zero.
For point amplitude QDA and amplitude quadratic logistic, the paired NLL
intervals against sKMD include zero in both conditions. Against calibrated
direct sparse QDA the source NLL difference is $-0.0414$ with paired $t$
interval $[-0.0799,-0.0029]$ (bootstrap $[-0.0735,-0.0137]$), which still
favors sKMD; its shift interval includes zero. These comparisons
use pointwise, unadjusted intervals, and all are exploratory.

Reconstruction MSE provides a complementary comparison. Adaptive micro-local recovery
gives MSE $0.01995$ on source tests and $0.03160$ under the shift. The independent
prior gives $0.02024$ and $0.03255$; the coupled class mixture gives $0.02031$
and $0.03260$. The coupled prior gives slightly higher reconstruction MSE here. Its
covariance approximation conditions on estimated carriers and locally
constant coefficients, and the class prior can introduce bias.

All frequency searches, failures and outcomes are retained, and
hyperparameters use development data only.

\FloatBarrier

\subsection{Triggered ringdowns: a signed modal mechanism}
\label{sec:ringdown_v32}

In the random-phase experiment in Section~\ref{sec:adaptive_experiments_v32},
positive envelopes are recoverable from their full energy trajectories.
Here an external trigger fixes a phase reference, so the relative signs of
modal impulses encode the direction of the excitation. We evaluate this
mechanism using a prespecified protocol and independent panels.

\paragraph{Damped weakly nonlinear normal modes.}
Two coordinates satisfy
\[
\ddot q_j+2\gamma_j\dot q_j+
\omega_j^2(q_j+0.0002q_j^3)=0,
\qquad \omega=2\pi(6,14),\quad\gamma=(0.22,0.35).
\]
A single sensor observes $q_1+q_2$. The modal impulses are
$a=F_k g$, $g\sim N(0,I)$, with $F_kF_k^\top=P_k^{-1}$ and
\[
P_0=I,\qquad P_1=I+0.9\operatorname{diag}(1,-1),\qquad
P_2=I+0.9\begin{pmatrix}0&1\\1&0\end{pmatrix}.
\]
Thus the labels describe isotropic, axis-oriented and oblique excitation
ensembles of the same instrument. For observed-trigger offset $\delta$,
$q_j(0)=a_j\sin(\omega_j\delta)$ and
$\dot q_j(0)=a_j\omega_j\cos(\omega_j\delta)$.
Source offsets are uniform on $[-0.5,0.5]$ ms. The physical mechanism uses
stable modes and an externally aligned trigger. RK4 uses step $1/1024$ s for four seconds, observed at
128 Hz. A separate 12-event halving check changes the clean signal by RMS
$5.92\times10^{-5}$.

\paragraph{Learn the dictionary before classification.}
Each of six panels uses 30 separate calibration events to learn
frequency trajectories and damping through the adaptive
KMD front end described above. Median frequency trajectories and median log-amplitude decay
fits define a damped sine/cosine pair for each mode. This bank is complete,
has sensor SD $0.25$, and contains exactly ten events from each excitation
ensemble with their labels withheld from the learner. The calibration
bank therefore uses controlled balanced collection. These 30 events are part of the total
training budget. Across panels, median
frequencies range over $6.004$--$6.007$ and $14.003$--$14.011$ Hz; learned damping
ranges over $0.215$--$0.223$ and $0.342$--$0.350$ s$^{-1}$. No test signal, label
or true modal trajectory enters this fit. All 180 calibration events return
two detected peaks. The resulting learned kernels are fixed for independent
fitting, development and test events. Dictionary calibration is thus
independent of the signal whose likelihood is evaluated.

Each panel then uses 40 fitting and 25 development events per class, and
60 test events per class in each of three conditions. Source sensor SD is
uniform on $[0.3,1.3]$, and the event's observation-retention probability is
uniform on $[0.25,1]$. A complete 64-sample pretrigger baseline estimates sensor
noise; the generating variance is unavailable to every fitted method. The
first test condition repeats this distribution. The second increases noise to
$[0.8,2.0]$ and reduces retention to $[0.15,0.40]$. The third additionally changes
damping to $(0.32,0.50)$, frequencies to $(6.04,13.96)$ Hz, and trigger offsets
to $[-10,10]$ ms. All development events use source conditions.

\paragraph{Conditional integration and its exact comparator.}
For learned synthesis matrix $H$ and observed samples $y_O$, the four signed
quadratures and their working noise covariance are
\[
z=(H_O^\top H_O)^{-1}H_O^\top y_O,\qquad
V=\widehat\sigma^2(H_O^\top H_O)^{-1}.
\]
Symmetric impulse ensembles justify zero class means. The class covariance
estimate subtracts fitting measurement variance, projects eigenvalues to at
least $0.001$, and selects a diagonal-shrinkage weight on development data.
Coupled scores integrate the observation variance:
$\log N(z;0,C_k+V)$. A common development-selected temperature calibrates each
classifier, except that the direct sparse-QDA wrapper uses its declared
continuous inverse temperature and class-intercept calibration.
Weak nonlinear dynamics and estimated noise make the fitted model a working
Gaussian approximation.

The independent KMD reference uses the diagonal of the pooled fitted covariance.
A stronger reconstruction control uses the full pooled covariance, while the
coupled reconstruction averages class-conditional posterior means using the
development-calibrated class probabilities, giving a calibrated mixture estimator.
The diagonal class-covariance ablation removes all quadrature off-diagonals,
including both within-mode and cross-mode terms.
The classification controls include
empirical-mean OAS-QDA, zero-mean point QDA, signed quadratic logistic, the
calibrated direct sparse-QDA implementation~\cite{jiang_sparseqda_2018}, and energy methods. Both the energy
forest and energy-interaction logistic receive noise and missingness quality
features. A stronger forest receives the full 16-point squared reconstruction
of each mode. All controls use the same learned dictionary and observed data;
none receives true impulses. Complete per-mode energies retain within-mode
structure but discard a whole mode's sign in the ideal separated model.
Finite noisy recovery can also mix modes; empirical error gaps therefore
combine sign information with recovery effects.

\begin{table}[t]
\centering\small
\caption{Triggered ringdown NLL in six independent panels. All test
conditions are prespecified.}
\label{tab:ringdown_v32}
\resizebox{\linewidth}{!}{
\begin{tabular}{lrrr}
\toprule
Method & Source & Noise/missingness shift & Trigger/dynamics shift \\
\midrule
Coupled sKMD / matched Gaussian & 0.812 & 0.785 & 0.935 \\
Diagonal class covariance + noise & 0.869 & 0.865 & 0.889 \\
Signed OAS-QDA & 0.797 & 0.809 & 0.962 \\
Zero-mean point QDA & 0.882 & 1.334 & 3.166 \\
Signed quadratic logistic & 0.832 & 1.262 & 2.923 \\
Calibrated direct sparse QDA & 0.854 & 1.356 & 3.146 \\
KMD energies + RF & 0.924 & 1.020 & 1.000 \\
Full energy paths + RF & 0.838 & 0.873 & 0.876 \\
Energy interaction logistic & 0.865 & 1.848 & 1.898 \\
\bottomrule
\end{tabular}
}
\end{table}

\begin{table}[t]
\centering\small
\caption{Clean-signal reconstruction MSE for the same held-out ringdowns.
The changed trigger and dynamics reverse the advantage over independent KMD.}
\label{tab:ringdown_reconstruction_v32}
\resizebox{\linewidth}{!}{
\begin{tabular}{lrrr}
\toprule
Reconstruction & Source & Noise/missingness shift & Trigger/dynamics shift \\
\midrule
Independent KMD & 0.01721 & 0.05748 & 0.15253 \\
Pooled full covariance & 0.01594 & 0.04931 & 0.20717 \\
Coupled class mixture & 0.01524 & 0.04601 & 0.19851 \\
Unregularized least squares & 0.01904 & 0.07572 & 0.13713 \\
\bottomrule
\end{tabular}
}
\end{table}

\begin{figure}[tp]
\centering\includegraphics[width=\linewidth]{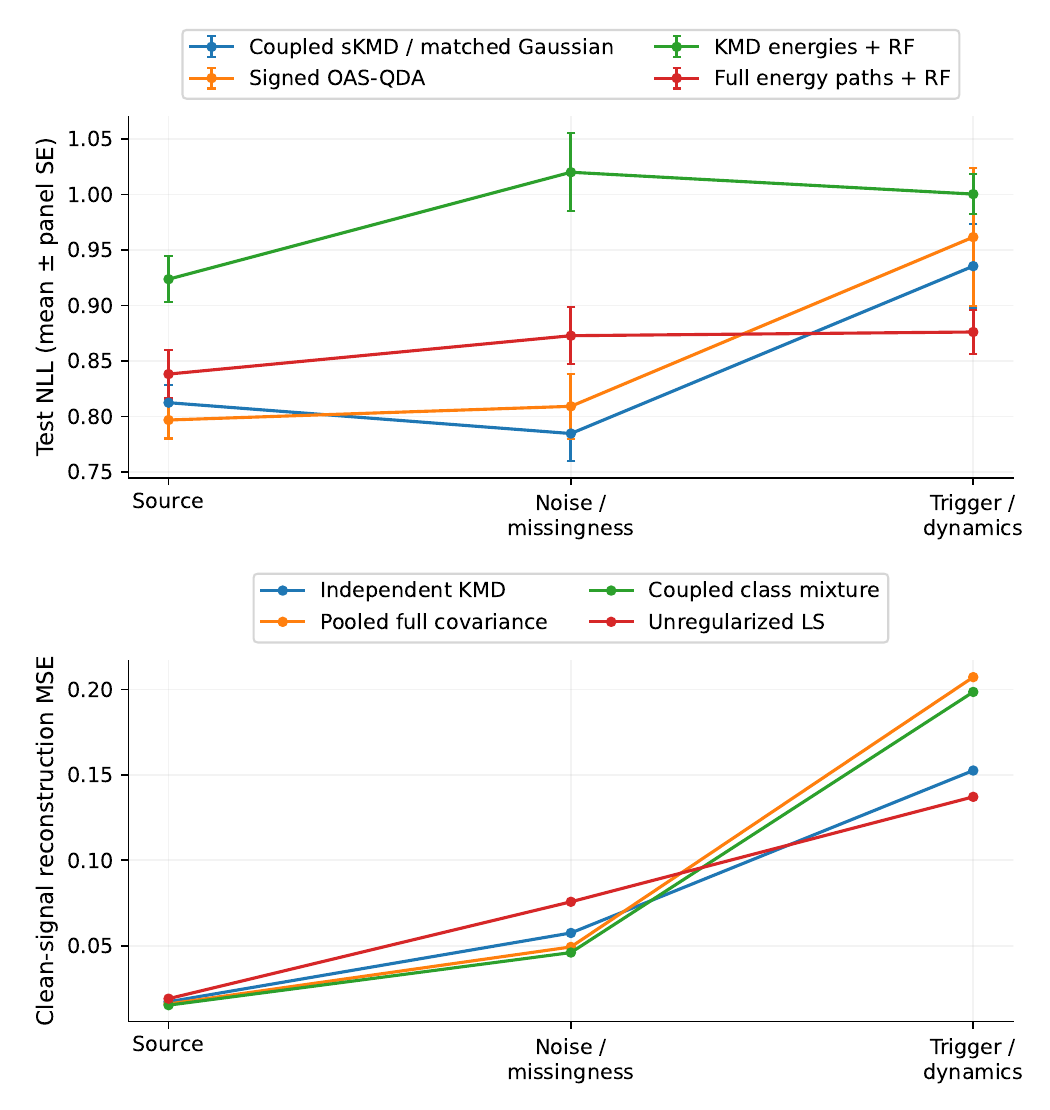}
\caption{A learned, externally triggered mode reference permits a useful
coupled prior under changing observation noise. Changing the trigger and
dynamics increases reconstruction error. Classification bars are panel standard errors.}
\label{fig:ringdown_v32}
\end{figure}

\paragraph{Where coupling helps.}
Tables~\ref{tab:ringdown_v32} and~\ref{tab:ringdown_reconstruction_v32} and
Figure~\ref{fig:ringdown_v32} report all methods and conditions.
In source tests, coupled sKMD has NLL $0.812$, versus $0.924$ for pooled KMD
energies, $0.838$ for full energy paths, and $0.797$ for signed OAS-QDA.
The sKMD--OAS paired interval includes zero, leaving their source-condition
NLL difference unresolved at this precision.
Under increased noise and missingness, the corresponding NLLs are
$0.785$, $1.020$, $0.873$ and $0.809$. The paired difference from OAS-QDA is
$-0.0245$ and the difference from the full-energy-path forest is $-0.0881$.
Their paired $t$ intervals over the six panels (5 df) are
$[-0.0424,-0.0066]$ and $[-0.1450,-0.0312]$, both excluding zero; the saved
summary gives pointwise 95\% panel-bootstrap sensitivity intervals
$[-0.0363,-0.0114]$ and $[-0.1258,-0.0463]$. In this condition retaining
each event's observation covariance improves NLL, but the independent
validation below does not resolve the OAS-QDA comparison.

Reconstruction MSE decreases from $0.05748$ with independent KMD to $0.04601$
with the coupled mixture in that condition, a $20.0\%$ reduction. Against the
stronger pooled full covariance, the reduction from $0.04931$ is $6.7\%$;
the paired MSE difference is $-0.00330$ (paired $t$ interval
$[-0.00432,-0.00228]$; bootstrap sensitivity interval
$[-0.00402,-0.00259]$). Source improvements are smaller but have the same sign.

\paragraph{Trigger and dynamics shift.}
The third condition changes the phase reference and dynamics simultaneously.
Coupled reconstruction MSE rises to $0.1985$, above independent KMD's $0.1525$
and unregularized least squares' $0.1371$. The full-energy-path forest also has
lower mean NLL, $0.876$ versus $0.935$, although that paired interval includes
zero.

All intervals in this study are pointwise and unadjusted for multiplicity;
six panels limit their resolution.

\subsubsection{Full observation-covariance controls}
\label{sec:ringdown_fullcov_v32}

The generic heads in the diagonal-quality experiment
receive only the four diagonal entries of $V$, sensor noise and retained
fraction, while the integrated model uses all ten unique covariance entries.
Two analyses examine this information difference. An exploratory
sensitivity analysis, specified after those outcomes were available, refits
generic heads on the same six panels with the complete upper triangle of
$V$. A six-panel validation, specified before its outcomes, uses
independent calibration, fitting, development and test seeds. Both analyses
use the same physical model, sample budgets and three test conditions.

The full-quality signed quadratic logistic head receives all signed coefficients,
squares, pairwise products and full observation-covariance information.
A signed RBF-SVM receives the coefficients and the same quality variables;
its nine fixed hyperparameter settings give it a larger search than the
three-setting logistic or forest controls. Full-energy-path forest and
energy-interaction logistic controls also receive all ten covariance entries.
Development data alone select parameters and temperatures. The integrated
model and OAS-QDA use the fitting procedures specified above. The sensitivity and validation analyses use separate panels and
intervals.

\begin{table}[t]
\centering\small
\caption{Six-panel validation with full observation covariance supplied to
generic controls. Test events are independent of those in
Table~\ref{tab:ringdown_v32}, with the same prespecified observation conditions.}
\label{tab:ringdown_fullcov_v32}
\resizebox{\linewidth}{!}{
\begin{tabular}{lrrr}
\toprule
Method & Source & Noise/missingness shift & Trigger/dynamics shift \\
\midrule
Coupled sKMD / matched Gaussian & 1.021 & 0.808 & 1.138 \\
Signed OAS-QDA & 0.831 & 0.826 & 0.962 \\
Full-V signed quadratic logistic & 0.926 & 2.110 & 3.492 \\
Full-V signed RBF-SVM & 1.051 & 1.178 & 1.192 \\
Full-V energy interaction logistic & 0.946 & 1.940 & 1.917 \\
Full-V energy paths + RF & 0.879 & 0.858 & 0.896 \\
\bottomrule
\end{tabular}
}
\end{table}

\begin{figure}[tp]
\centering\includegraphics[width=\linewidth]{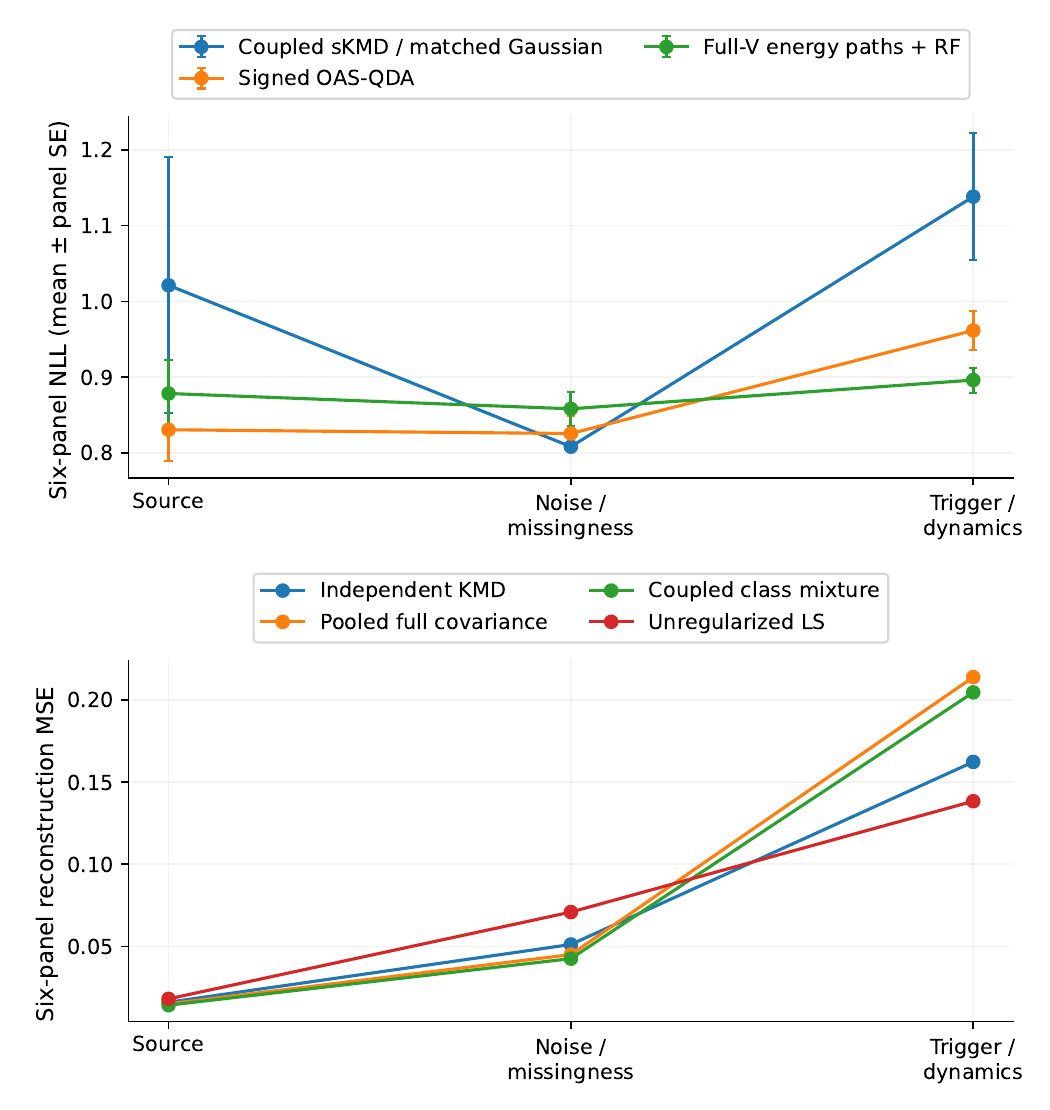}
\caption{Full-covariance validation across source, increased noise and missingness,
and trigger/dynamics-shift conditions, reporting classification and
reconstruction outcomes.}
\label{fig:ringdown_fullcov_v32}
\end{figure}

\paragraph{Source log loss and tail behavior.}
Table~\ref{tab:ringdown_fullcov_v32} and Figure~\ref{fig:ringdown_fullcov_v32}
report the full-covariance validation.
On source events, coupled NLL is $1.021$, versus $0.831$ for
OAS-QDA, although accuracies are close ($0.638$ and $0.635$). One
class-1 event in the sixth panel has clean-signal peak $9.73$ and true-class
loss $179.42$ nats; that panel's mean NLL is $1.843$. Its recovered second-mode
quadratures are approximately $(-3.89,-8.07)$, far from a small perturbation of
a trigger-aligned sine coefficient. This event contributes to all reported
metrics under the prespecified fitting procedure.

\paragraph{Classification and conditional reconstruction.}
Under increased noise and missingness, coupled NLL is $0.808$, versus
$0.858$ for the full-covariance energy-path forest. Their paired difference is
$-0.04995$ with paired $t$ interval $[-0.10318,0.00328]$, which includes
zero; the panel-bootstrap sensitivity interval $[-0.08882,-0.01616]$ excludes
it, so with six panels this classification difference is unresolved.
OAS-QDA gives $0.826$; its paired interval against sKMD includes zero,
leaving that classification difference unresolved as well. Coupled
reconstruction MSE is $0.04262$, versus $0.05120$ for independent KMD and
$0.04507$ for pooled full covariance: reductions of $16.8\%$ and $5.4\%$.
The paired MSE differences are $-0.00859$ and $-0.00245$, with paired $t$
intervals $[-0.01374,-0.00343]$ and $[-0.00441,-0.00049]$ and bootstrap
sensitivity intervals $[-0.01200,-0.00518]$ and $[-0.00369,-0.00098]$,
respectively.

Under the trigger/dynamics shift, coupled MSE is $0.2046$, versus
$0.1623$ for independent KMD and $0.1384$ for least squares. Its NLL is $1.138$,
above both OAS-QDA's $0.962$ and the full-energy-path forest's $0.896$.
The six-panel exploratory full-covariance sensitivity analysis has
noise-shift NLLs $0.785$ for coupled sKMD and $0.862$ for the energy-path forest;
its intervals use the same events as the diagonal-quality comparison.

\FloatBarrier

\subsection{Nonstationary carriers without critical transitions}
\label{sec:nonstationary_v32}

The next experiments test two nonstationary carrier mechanisms: frequency
sweeps during a decaying response, and deep amplitude fades with continuously
varying frequency. The event-level excitation class stays fixed. Neither example
contains a bifurcation or a regime change within an event. Nonstationary Gaussian
process decomposition is established prior work, including oscillatory models
with changing amplitudes \cite{wilkinson_audio_2019}. We compare signed
excitation dependence and observation uncertainty with independent or pooled
priors, using the same independently learned adaptive KMD dictionary.

\paragraph{Two prespecified signal laws.}
Every four-second record has 512 samples at 128\,Hz, with
\[x(t)=\sum_{j=1}^2 a_j e_j(t)\sin\phi_j(t).\]
The three centered Gaussian laws for the signed excitation \(a\) are those in
Section~\ref{sec:ringdown_v32}, with precision contrast \(\alpha=0.9\).
A common externally referenced trigger offset is uniform on
\([-0.5,0.5]\)\,ms and adds the corresponding initial-frequency phase offsets.
For the \emph{chirped decay},
\[
 e_j(t)=e^{-\gamma_jt},\qquad
 \phi_j(t)=2\pi(f_{0j}t+r_jt^2/2),
\]
where
\[
 f_0=(4,13),\qquad r=(1,1.25),\qquad\gamma=(0.22,0.35).
\]
Frequencies therefore sweep approximately 4--8 and 13--18\,Hz. This is an exact
linear time-varying oscillator example, not a numerical discretization: with
\(\omega_j=\phi'_j>0\), each mode satisfies
\[
 q_j''+\left(2\gamma_j-\frac{\omega'_j}{\omega_j}\right)q_j'
 +\left(\omega_j^2+\gamma_j^2
              -\gamma_j\frac{\omega'_j}{\omega_j}\right)q_j=0.
\]
For \emph{modulation and fading}, frequencies in Hz are
\[
 f_1(t)=5+0.65\sin(2\pi\,0.35t),\qquad
 f_2(t)=14+1.2\sin(2\pi\,0.23t+0.8),
\]
with analytic phase integrals starting at zero, and
\[
 e_1(t)=0.08+0.92\cos^2(\pi\,0.42t+0.2),\qquad
 e_2(t)=0.08+0.92\cos^2(\pi\,0.31t+0.9).
\]
These explicit AM--FM signals reach 8\% of peak amplitude.

\begin{figure}[tp]
\centering
\includegraphics[width=0.93\linewidth]{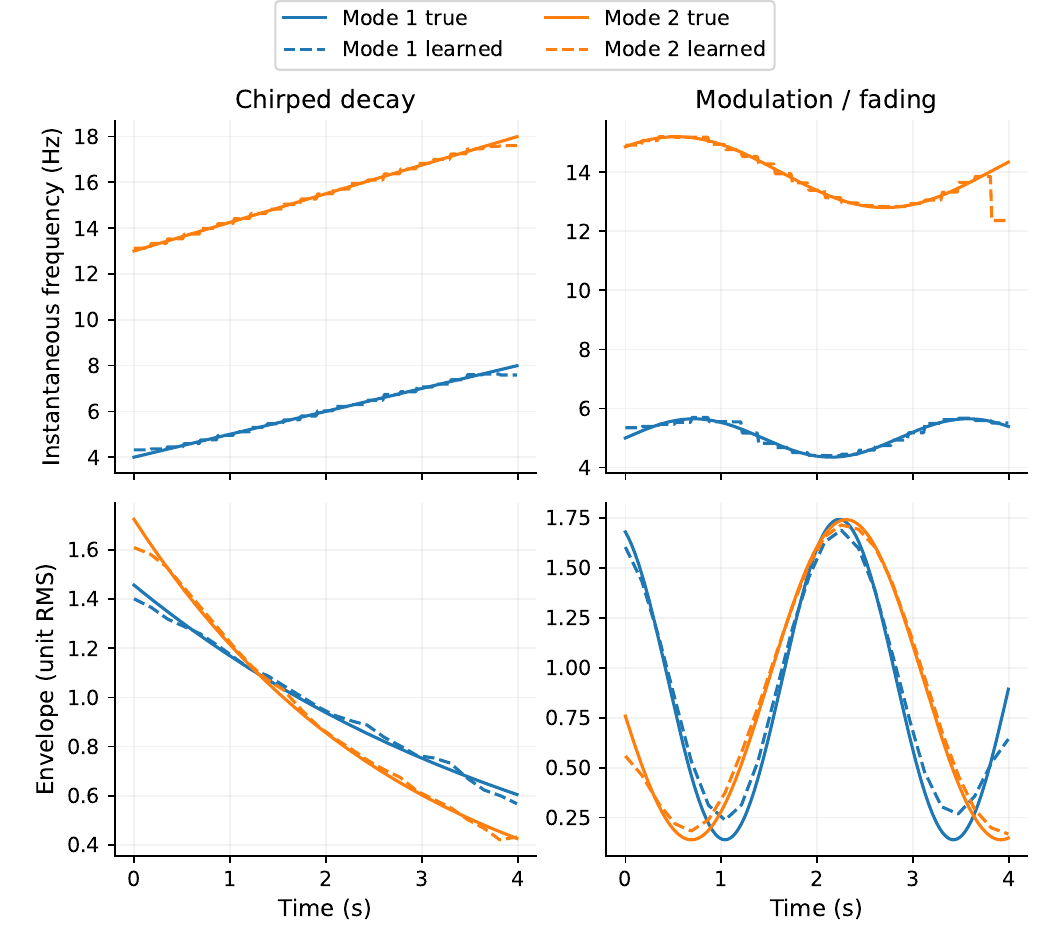}
\caption{Independent calibration in fixed panel zero. Solid curves are generating
laws, shown only for evaluation; dashed curves are learned from noisy calibration
records. Frequencies are derivatives of the learned, interpolated phase.
Endpoint error and local ambiguity near fades remain visible.}
\label{fig:nonstationary_carriers_v32}
\end{figure}

\paragraph{Independent calibration and evaluation.}
Each family has twelve independent panels. A separate bank of 30 complete
calibration events, ten per class with labels withheld and noise standard
deviation 0.25, learns the dictionary. This controlled balanced collection
counts toward the training budget. Each panel
then uses 40 fitting, 25 development, and 60 test events per class and condition,
with disjoint random seeds. No test record changes the dictionary.
The Gaussian-energy KMD routine described in
Section~\ref{sec:adaptive_experiments_v32} selects the carriers from calibration
signals in 3--20\,Hz. Twenty-four anchors, localization parameter 12, a 40\%
local search band, and 81 local frequency candidates cover the prespecified
sweeps; adjacent search bands have disjoint interiors. Calibration records with
median recovered amplitude above 0.5 supply unwrapped local phases and
normalized amplitude curves. Their pointwise medians define a fixed four-column
cosine/sine dictionary after interpolation. This uses neither generating phase
nor generating envelope. Figure~\ref{fig:nonstationary_carriers_v32} shows panel
zero.

Source noise standard deviation is uniform on \([0.3,1.3]\), with independently
retained sample fraction uniform on \([0.25,1]\). A second condition changes
these intervals to \([0.8,2]\) and \([0.15,0.4]\). A third also adds fixed carrier
offsets \((+0.20,-0.15)\)\,Hz. Every method estimates each event's noise from
64 independent pretrigger samples; the generating variance is hidden. All
hyperparameters and score temperatures use source development events only.
The six classification heads and their grids match those used in the full
observation-covariance experiment in Section~\ref{sec:ringdown_fullcov_v32}.
The signed quadratic, signed RBF, energy-interaction, and energy-path controls
receive all ten distinct entries of the four-dimensional observation covariance,
plus estimated noise and retained fraction. Signed OAS-QDA is the separate point
coefficient comparator. An OAS comparison jointly varies covariance estimation,
the zero-mean restriction, and uncertainty handling. The energy-path forest
uses 16 squared reconstructed values per mode, providing a finite sampled
representation of each energy path.

\paragraph{Classification across observation conditions.}
Table~\ref{tab:nonstationary_classification_v32} reports the prespecified classifiers.
Under changed observation quality, chirped-decay sKMD has NLL 0.76781 versus
0.77455 for OAS-QDA: the paired difference is \(-0.00673\), with paired
\(t\) 95\% interval \([-0.02100,0.00754]\) (bootstrap sensitivity interval
\([-0.01897,0.00498]\)). This comparison is inconclusive at the available precision; the interval does not establish equivalence.
The fading example is stronger: NLL is 0.77077 versus 0.80173 for OAS-QDA,
a difference of \(-0.03096\) nats, with paired \(t\) interval
\([-0.05581,-0.00612]\) (bootstrap sensitivity interval
\([-0.05302,-0.01072]\)). Accuracy is 0.6565 versus 0.6171, macro AP
0.7291 versus 0.7088, and Brier score 0.4541 versus 0.4707.
OAS-QDA has the lowest NLL among the nonidentical comparison methods in
this fading/noise condition.

The prespecified practical criteria were at least 0.02 nats improvement against
the strongest relevant nonidentical classifier, or at least 5\% reconstruction
MSE reduction against pooled full covariance under changed observation quality.
Neither criterion is met. The fading classification point estimate is
0.03096 nats, but the lower confidence limit does not establish an
improvement of at least 0.02 nats.
The chirped-decay point estimate is below that threshold. An exploratory two-sided paired
\(t\)-test sensitivity analysis for just the two family-specific OAS/noise
comparisons gives fading \(p=0.0191\), or 0.0383 after either Holm or
Bonferroni adjustment. This adjustment applies to the two family-specific
OAS/noise comparisons. All bootstrap
intervals use 20,000 paired resamples of independent panels, and the reported
intervals remain pointwise.

\begin{table}[t]
\centering\small
\caption{Test NLL across twelve independent panels per family; lower is better.
Carrier shift also includes changed noise and missingness. All prespecified
classifiers and conditions are reported.}
\label{tab:nonstationary_classification_v32}
\textbf{Chirped decay}\par\medskip
\begin{tabular}{lrrr}
\toprule
Method & Source & Noise/missingness & Carrier shift \\
\midrule
Coupled sKMD / matched Gaussian & 0.760 & 0.768 & 1.218 \\
Signed OAS-QDA & 0.760 & 0.775 & 1.221 \\
Full-V signed quadratic logistic & 0.862 & 2.599 & 7.042 \\
Full-V signed RBF-SVM & 1.049 & 1.199 & 1.212 \\
Full-V energy interaction logistic & 0.895 & 2.898 & 3.910 \\
Full-V energy paths + RF & 0.815 & 0.850 & 1.057 \\
\bottomrule
\end{tabular}

\medskip\par\textbf{Modulation and fading}\par\medskip
\begin{tabular}{lrrr}
\toprule
Method & Source & Noise/missingness & Carrier shift \\
\midrule
Coupled sKMD / matched Gaussian & 0.789 & 0.771 & 1.243 \\
Signed OAS-QDA & 0.785 & 0.802 & 1.528 \\
Full-V signed quadratic logistic & 0.896 & 2.080 & 5.440 \\
Full-V signed RBF-SVM & 1.025 & 1.198 & 1.201 \\
Full-V energy interaction logistic & 0.924 & 2.311 & 3.680 \\
Full-V energy paths + RF & 0.846 & 0.858 & 1.007 \\
\bottomrule
\end{tabular}
\end{table}

\paragraph{Reconstruction with independent, pooled, and coupled priors.}
The independent, pooled-full, and coupled estimators share the learned dictionary,
observations, and estimated noise; Table~\ref{tab:nonstationary_reconstruction_v32}
reports clean-signal error. Under changed observation quality, coupling reduces
MSE by 26.6\% and 26.0\% relative to independent KMD for chirp and fading.
Against pooled full covariance, the reductions are 2.26\% and
2.18\%, below the prespecified 5\% criterion. The absolute differences are
\(-0.000841\), with paired \(t\) interval \([-0.001467,-0.000215]\)
(bootstrap \([-0.001382,-0.000328]\)), and \(-0.000865\), with paired
\(t\) interval \([-0.001262,-0.000469]\) (bootstrap
\([-0.001213,-0.000537]\)), respectively. These intervals resolve small improvements while retaining
reductions below the practical criterion. Reconstruction uses calibrated
class weights and untempered class-conditional means, giving a calibrated
mixture estimator.

An exploratory diagnostic, specified after the primary outcomes were
available, restricts fading error to the 239 of 512 time points
where at least one generating envelope is at most 20\% of its peak. These
truth-defined regions are used for evaluation only. Under changed observation
quality the MSEs are 0.01413 (coupled), 0.01437 (pooled-full), 0.01673
(independent), and 0.02357 (least squares). The coupled--pooled difference is
\(-0.000238\), with paired \(t\) interval \([-0.000387,-0.000089]\).
The improvement within fades is 1.66\%. The generating excitation class is
constant.

\begin{table}[t]
\centering\small
\caption{Clean-signal MSE for the same learned dictionaries and the independent,
pooled-full, and coupled estimators. Incremental reductions under changed
observation quality are evaluated against the prespecified 5\% reconstruction
criterion; carrier-shift results are shown alongside.}
\label{tab:nonstationary_reconstruction_v32}
\textbf{Chirped decay}\par\medskip
\begin{tabular}{lrrr}
\toprule
Reconstruction & Source & Noise/missingness & Carrier shift \\
\midrule
Independent KMD & 0.01336 & 0.04956 & 1.14995 \\
Pooled full covariance & 0.01078 & 0.03720 & 1.23667 \\
Coupled class mixture & 0.01054 & 0.03636 & 1.22386 \\
Unregularized LS & 0.01494 & 0.06877 & 1.01315 \\
\bottomrule
\end{tabular}

\medskip\par\textbf{Modulation and fading}\par\medskip
\begin{tabular}{lrrr}
\toprule
Reconstruction & Source & Noise/missingness & Carrier shift \\
\midrule
Independent KMD & 0.01655 & 0.05240 & 0.79352 \\
Pooled full covariance & 0.01374 & 0.03964 & 0.93837 \\
Coupled class mixture & 0.01364 & 0.03878 & 0.92317 \\
Unregularized LS & 0.01824 & 0.07400 & 0.61175 \\
\bottomrule
\end{tabular}
\end{table}

\begin{figure}[tp]
\centering
\includegraphics[width=0.90\linewidth]{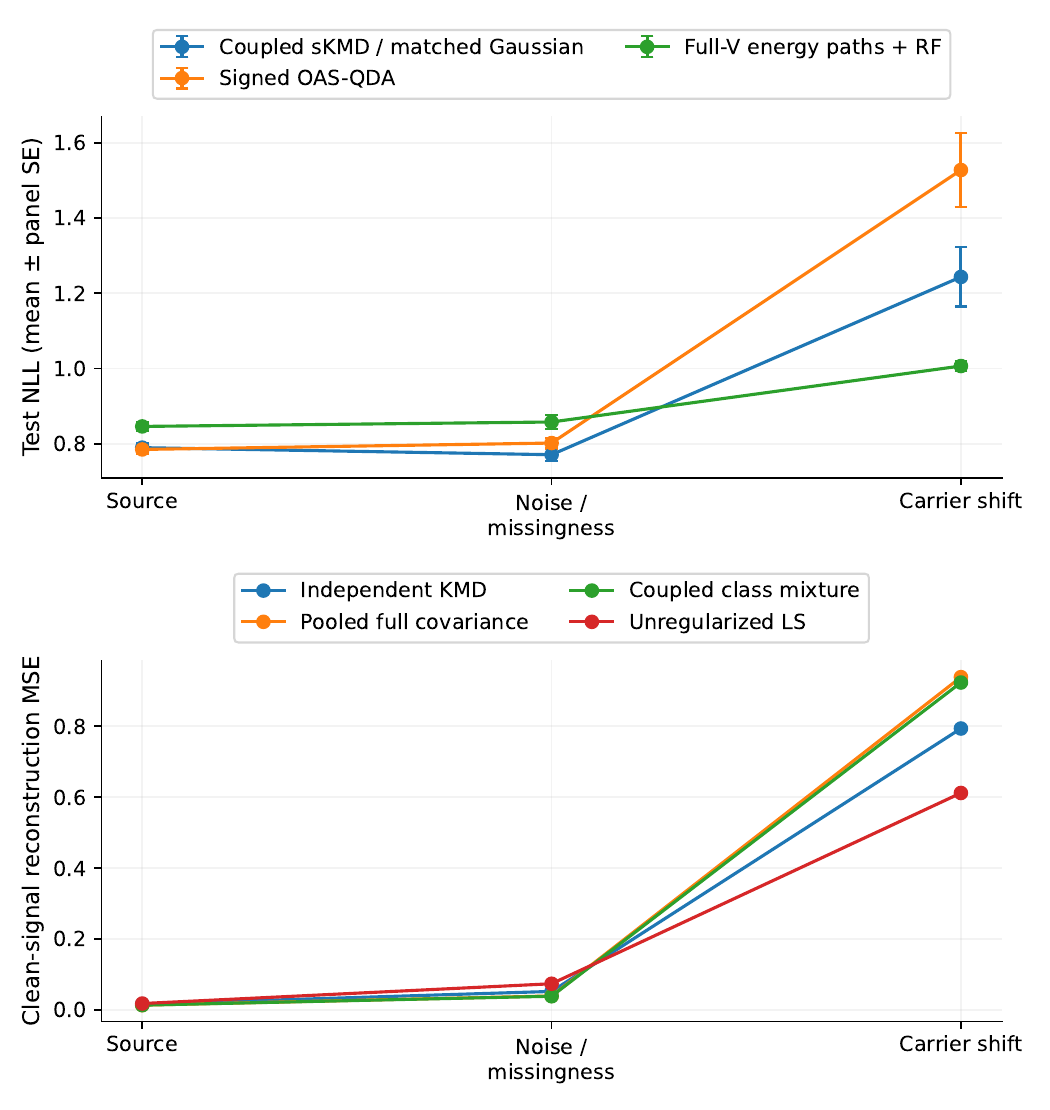}
\caption{Twelve-panel modulation/fading study across source,
observation-shift and carrier-shift conditions. Error bars are standard
errors across panels; paired intervals are reported in the text.}
\label{fig:nonstationary_fading_v32}
\end{figure}

\paragraph{Carrier-shift results.}
With the fixed frequency offsets, chirped-decay and fading coupled MSEs rise to
1.22386 and 0.92317, worse than independent KMD (1.14995 and 0.79352) and
least squares (1.01315 and 0.61175). The energy-path forest also has lower NLL
than sKMD: 1.05716 versus 1.21767 and 1.00652 versus 1.24336. In fading, the
NLL disadvantage has paired \(t\) interval \([0.06772,0.40595]\); for chirp
the bootstrap and \(t\) intervals disagree about exclusion of zero.
An evaluation-only projection onto the learned span has mean residual MSE
approximately 0.0042 and 0.0077 under changed noise alone, but 0.9229 and 0.5290
after carrier shift. Figure~\ref{fig:nonstationary_fading_v32} shows the observation-shift and
carrier-shift conditions.

\paragraph{Conditioning.}
There are no calibration peak failures or amplitude-threshold fallbacks.
All test masked Gram matrices are positive definite; the minimum retained
count is 51 and the largest condition number is 3.76. Local phase
identifiability near a fade is a separate issue from this global
four-column conditioning. The probabilities condition on the learned
dictionary and estimated noise, treating dictionary estimation, phase
uncertainty and unmodeled carrier drift as fixed.

\FloatBarrier

\subsection{Accounting for a declared family of carrier changes}
\label{sec:dictionary_mixture_v32}

The carrier-shift failures in Section~\ref{sec:nonstationary_v32} motivate
an explicit extension: retain uncertainty over a finite collection of
possible KMD dictionaries. This extension was specified after those
failures, then evaluated with new independent seeds under a frozen protocol.
Bayesian model averaging supplies the averaging operation within the
coupled KMD pipeline, allowing carrier adaptation to be compared with regime
coupling.

\subsubsection{A normalized joint model}

Let $H_j\in\mathbb R^{n\times q}$, $j=1,\ldots,J$, be dictionaries learned
or constructed independently of a new observation. Write $L$ for the latent
dictionary index, with prior $w_j>0$, $\sum_jw_j=1$, and $Y$ for the regime,
with prior $\pi_k$. The model is
\[
 L\sim w,\quad Y\sim\pi,\quad A\mid Y=k\sim\N(0,C_k),\qquad
 v_O\mid A,L=j=H_{j,O}A+\epsilon_O,\quad
 \epsilon_O\sim\N(0,R_O).
\]
The independent prior on $(L,Y)$ is a modeling assumption. All dictionaries
use the same physical coefficient convention, so that $C_k$ has a common
meaning. Changing coordinates separately in each dictionary requires
transforming its class covariance as well.

Define the information matrix and vector
\[
 J_j=H_{j,O}^{\top}R_O^{-1}H_{j,O},\qquad
 b_j=H_{j,O}^{\top}R_O^{-1}v_O,\qquad M_{jk}=C_k^{-1}+J_j.
\]
Even when the observed rows are rank deficient, $M_{jk}\succ0$. Completing
the square gives
\begin{align}
 \ell_{jk}&=-\tfrac12\log\det C_k-\tfrac12\log\det M_{jk}
               +\tfrac12 b_j^\top M_{jk}^{-1}b_j,
 \label{eq:dictionary_evidence_v32}\\
 \omega_{jk}&=\frac{w_j\pi_k e^{\ell_{jk}}}
 {\sum_{h,l}w_h\pi_l e^{\ell_{hl}}},\qquad
 \widehat s=\sum_{j,k}\omega_{jk}H_jM_{jk}^{-1}b_j.
 \label{eq:dictionary_joint_v32}
\end{align}
The regime posterior is $\sum_j\omega_{jk}$. The omitted density factor
$\exp(-v_O^\top R_O^{-1}v_O/2)/(\det(2\pi R_O))^{1/2}$ is common to
\emph{every} dictionary and regime. Omitting a dictionary-dependent residual
or reference likelihood would change these weights. Selecting a dictionary
from $v$ and then inserting it into a fixed-dictionary likelihood does not
in general yield a normalized model for $v$; the finite mixture does.
These probabilities are exact for the specified fitted model at temperature
one. Plug-in noise estimates, fitted class laws and the library restriction
remain sources of model error.

\paragraph{What the controlled comparisons isolate.}
A pooled full-covariance mixture uses the same $H_j,w_j,R_O$, but one
independently fitted coefficient covariance. A diagonal pooled mixture is
an independent-mode KMD control with exactly the same carrier adaptation.
Two MAP-dictionary controls select $j$ using either pooled evidence or
$\sum_k\pi_k e^{\ell_{jk}}$, then average over regimes in that one dictionary.
They separate averaging from the choice of a richer selection model.
A further control replaces $\omega_{jk}$ by the product of its own two
marginals while retaining every conditional mean. This directly tests the
value of dependence between the regime and dictionary indices. A second
factorized control uses the pooled dictionary posterior and the class
posterior in its MAP dictionary. Joint MAP selection of $(j,k)$ is also
reported. These controls change the inference rule.

\subsubsection{Continuous-drift experiment}

Each of the two nonstationary families receives twelve independent
panels. Thirty separate, complete, noisy calibration events supply
two learned envelopes and phases. Adding a constant frequency offset from
$\{-0.30,-0.25,\ldots,0.30\}$ Hz to each phase gives $J=169$ dictionaries,
each with four sine/cosine coefficients and uniform prior weight. The source
fit uses 120 events; development uses 75 source and 75 independently drifting
events. Class covariances select shrinkage from $\{0,0.2,0.5,1\}$ by mean
development class NLL. The pooled covariance is fitted directly to all source
coefficients after subtracting mean observation covariance, and selects its
own shrinkage by development marginal observation density. A fixed-dictionary
pooled control tunes that same criterion within its own model. All scoring
uses temperature one.

The four held-out conditions contain 180 events each: source quality;
changed noise and missingness without drift; changed quality with each
record's two offsets independently uniform on $[-0.25,0.25]$ Hz; and offsets
with independent random signs and magnitudes uniform on $[0.45,0.65]$ Hz.
Thus the third condition is continuous and generally off grid, while the
fourth is deliberately outside the library. Source sensor standard deviations
are uniform on $[0.3,1.3]$ and observation retention on $[0.25,1]$; changed
quality uses $[0.8,2]$ and $[0.15,0.4]$. Each event has an independent
64-sample noise calibration. Calibration, fitting, development and testing
seeds and outcomes are recorded in the companion protocol.
The source fitting masks have full column rank; scoring itself remains
valid for an empty or deficient observed span.

\begin{table}[htbp]
\centering\small
\caption{Finite-dictionary extension across all specified conditions. Each entry averages 12 independently fitted panels. MSE targets the generating signal. Class NLL is in nats. Fixed, MAP and mixture methods share the same source/development resources; mixture and MAP methods share 169 candidate dictionaries.}
\label{tab:dictionary_mixture_v32}
\begin{tabular}{llrrrr}
\toprule
Family & Method & Source & Noise & In library & Outside \\
\midrule
Chirp MSE & Joint mixture & 0.0129 & 0.0648 & 0.0843 & 1.4318 \\
Chirp MSE & Pooled mixture & 0.0128 & 0.0657 & 0.0856 & 1.4440 \\
Chirp MSE & Independent mixture & 0.0159 & 0.0772 & 0.0878 & 1.3831 \\
Chirp MSE & Fixed coupled & 0.0103 & 0.0393 & 0.8862 & 1.4478 \\
Chirp MSE & Coupled MAP carrier & 0.0149 & 0.0842 & 0.1109 & 1.4957 \\
Chirp MSE & Joint marginals product & 0.0129 & 0.0649 & 0.0843 & 1.4316 \\
\addlinespace
Chirp NLL & Joint mixture & 0.7720 & 0.7653 & 0.7573 & 1.2101 \\
Chirp NLL & Fixed coupled & 0.7929 & 0.7796 & 1.0640 & 1.1886 \\
Chirp NLL & Pooled MAP carrier & 0.7748 & 0.7785 & 0.7688 & 1.2091 \\
Chirp NLL & Coupled MAP carrier & 0.7749 & 0.7768 & 0.7683 & 1.2484 \\
\addlinespace
Fading MSE & Joint mixture & 0.0166 & 0.0678 & 0.1077 & 0.9820 \\
Fading MSE & Pooled mixture & 0.0168 & 0.0689 & 0.1089 & 0.9863 \\
Fading MSE & Independent mixture & 0.0195 & 0.0795 & 0.0954 & 0.9385 \\
Fading MSE & Fixed coupled & 0.0144 & 0.0442 & 0.6621 & 1.3271 \\
Fading MSE & Coupled MAP carrier & 0.0186 & 0.0898 & 0.1369 & 1.0207 \\
Fading MSE & Joint marginals product & 0.0166 & 0.0679 & 0.1077 & 0.9819 \\
\addlinespace
Fading NLL & Joint mixture & 0.7392 & 0.7986 & 0.7741 & 1.2811 \\
Fading NLL & Fixed coupled & 0.7837 & 0.8244 & 1.4557 & 1.1364 \\
Fading NLL & Pooled MAP carrier & 0.7494 & 0.8356 & 0.7884 & 1.3418 \\
Fading NLL & Coupled MAP carrier & 0.7457 & 0.8280 & 0.7912 & 1.3611 \\
\addlinespace
\bottomrule
\end{tabular}
\end{table}

\begin{figure}[tp]
\centering
\includegraphics[width=\linewidth]{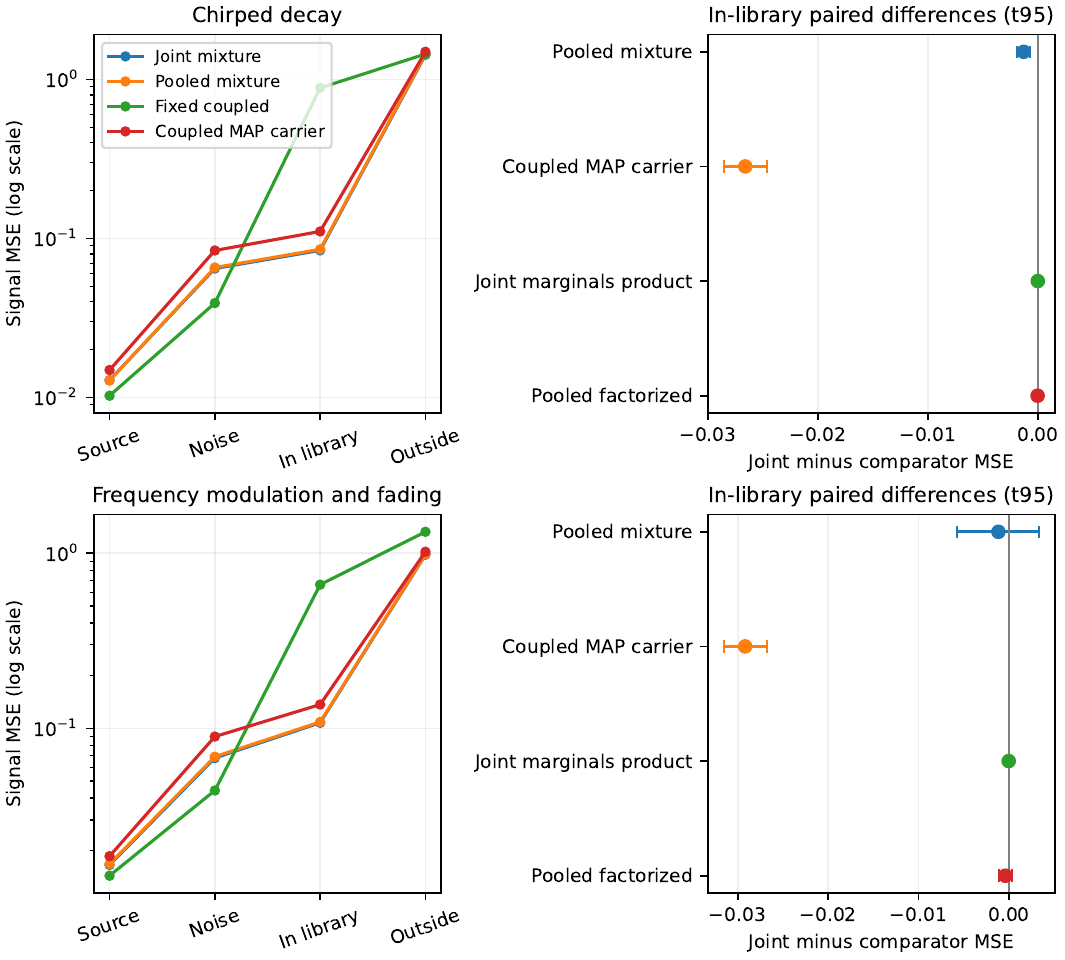}
\caption{Finite-dictionary extension on nonstationary data. Left:
reconstruction across all four prespecified conditions. Right: paired in-library
MSE differences across twelve independently fitted panels; intervals are
pointwise $t95$. The comparisons include joint averaging, a same-library
pooled model, and the independent mixture reported in
Table~\ref{tab:dictionary_mixture_v32}, which has lower MSE for fading.
The numerical model is a finite Gaussian mixture with learned KMD columns.}
\label{fig:dictionary_mixture_v32}
\end{figure}

\paragraph{Carrier averaging under in-library shifts.}
Table~\ref{tab:dictionary_mixture_v32} and Figure~\ref{fig:dictionary_mixture_v32}
report every method and condition. Within the library, joint-mixture MSE is $0.08429$ for chirped decay and
$0.10768$ for fading, compared with fixed coupled MSE $0.88618$ and $0.66205$.
The reductions are about $90.5\%$ and $83.7\%$. Relative to the coupled
MAP-dictionary control, averaging reduces MSE by $24.0\%$ and $21.3\%$.
The corresponding NLL improvements are $0.01105$ and $0.01709$ nats;
paired $t$ intervals for the improvement are $[0.00553,0.01657]$ and
$[0.00374,0.03045]$. Both are statistically resolved in the paired
four-comparison Holm analysis, but neither establishes the prespecified
$0.02$-nat practical margin. These results support averaging over plausible
carriers when observations leave more than one credible dictionary.

\paragraph{Comparison with adaptive KMD controls.}
Against the same-library full pooled prior, MSE falls by $1.52\%$
and $1.08\%$. The paired joint-minus-pooled differences are $-0.001300$
with $t95$ interval $[-0.001883,-0.000716]$ and $-0.001175$ with interval
$[-0.005713,0.003363]$. The two-family Holm-adjusted one-sided $p$ values
are $0.00047$ and $0.290$. Neither supports a reduction of at least $5\%$.
The diagonal independent mixture is stronger than the coupled mixture for
fading: MSE $0.09537$ versus $0.10768$. Multiplying the joint model's own
marginals changes MSE by only about $1.6\times10^{-5}$ and
$3.5\times10^{-5}$ in these two conditions. This experiment therefore
supports carrier averaging much more strongly than it supports the extra
dependence between the carrier and regime indices.

\paragraph{Computational cost and outside-library results.}
Allowing unnecessary carrier variation increases MSE at source quality and
under changed quality without drift, relative to the corresponding fixed
coupled model. Outside the library, the joint model gives NLL $1.2101$ and
$1.2811$, both worse than the uniform reference $\log3$, with large MSE.
The independent mixture has lower MSE in both outside-library cases.
The joint mixture's per-record calculation uses $J K$ small
$q\times q$ posterior systems rather than $K$; the library costs computation
and can increase estimation error.

The posterior offset diagnostic includes error in the learned reference
carrier.
\FloatBarrier

\subsection{Nonstationary excitation dependence}
\label{sec:switching_v32}

The preceding experiments concern independently triggered records. We next
ask whether the same interface helps when excitation regimes recur with
time-dependent persistence. This is a Gaussian hidden Markov observation
model with learned mode carriers, using the classical forward
recursion~\cite{rabiner_hmm_1989} and Gaussian emission integration.
Related probabilistic oscillatory decomposition addresses nonstationarity
\cite{song_plso_2021}. The purpose here
is to test the interaction and uncertainty interface in a nonstationary
mechanism with no critical transition, against controls sharing its temporal
information.

\paragraph{What changes with time.}
A trajectory comprises 60 externally triggered half-second blocks, sampled
at 128 Hz. Two stable physical carriers have frequencies 6 and 14 Hz.
In block $t$, a regime $Y_t\in\{0,1,2\}$ draws a new centered Gaussian pair
of sine coefficients with precision
\[
 P_0=I,\qquad P_1=I+\alpha\operatorname{diag}(1,-1),\qquad
 P_2=I+\alpha\begin{pmatrix}0&1\\1&0\end{pmatrix}.
\]
Independent cosine quadratures have standard deviation 0.12. Conditional on
the regimes, coefficients in distinct blocks are independent. This is a
sequence of aligned excitation records; it does not assert continuous
amplitude evolution across block boundaries.

The initial regime is uniform. A transition changes state with probability
$h_t$, choosing either other state equally. The source hazard is
$h_t=0.03,0.18,0.03$ in the three successive 20-block phases.
This time-inhomogeneous transition law makes the joint process nonstationary.
Its symmetry nevertheless preserves a uniform one-time regime marginal;
the nonstationarity is in temporal dependence, rather than a changing
unconditional mean or variance. Indeed, centered conditionally independent
coefficients have zero cross-block second moments. Their higher-order
dependence and the conditional modal covariance still change. No autonomous
dynamical instability or bifurcation is involved.

\paragraph{Independent collection and learning.}
Twelve independent panels each have 24 separate four-second calibration
records, eight per excitation regime with labels withheld. They have complete
observations and sensor SD 0.25. The adaptive KMD front end learns
two frequency trajectories from these records; their pooled medians define
fixed sine/cosine carriers for subsequent blocks. All 288 calibration records
return two detected peaks. Panel frequencies lie in $[6.0039,6.0046]$ and
$[13.9877,13.9979]$ Hz. Mode count two, frequency band 3--20 Hz, and the
external phase reference are supplied assumptions. The calibration
bank is part of the acquisition budget.

Each panel has 12 labeled fitting trajectories and six development
trajectories. Each test condition has 12 independent trajectories, for 144 test
trajectories per condition across panels. Labels estimate the initial regime
law and three phase-specific transition matrices, with one pseudocount per
entry. Every method receives the same clock, learned transition matrices,
and current/past blocks. Prediction is causal at the end of the current
half-second block and uses only current and past blocks.

Source sensor SD is uniform on $[0.4,1]$ and the retention probability is
uniform on $[0.5,1]$. Every block has an independent 64-sample pretrigger
noise record. Its sample SD is available to every method; the generating
noise variance is unavailable. The five prespecified conditions are:
\begin{enumerate}
 \item source observations and hazards, with $\alpha=0.9$;
 \item observation shift to SD $[0.9,3]$ and retention $[0.15,0.5]$;
 \item hazard shift to $(0.18,0.03,0.18)$, with source observation quality;
 \item the observation shift plus frequencies $(6.4,13.6)$ Hz and phase
 offsets $(0.4,1.0)$ radians, retaining the fitted source dictionary;
 \item low interaction, $\alpha=0.05$, with its own fitting/development
 trajectories and the adverse observation quality.
\end{enumerate}
The last condition uses the same separate carrier calibration bank because
the physical frequencies do not depend on $\alpha$. No shifted test label or
noise value is used for fitting or development selection.

\paragraph{An interface valid for every observation rank.}
Unrestricted Bernoulli masks can leave fewer than four samples, making
an OLS coefficient inverse undefined. The reference posterior in
Section~\ref{sec:posterior_transfer_v31} remains well defined for those
observations, so the same interface includes every mask and block. For learned observed carrier matrix $H_O$, define
\[
 J=\widehat\sigma^{-2}H_O^\top H_O,\qquad
 b=\widehat\sigma^{-2}H_O^\top y_O,\qquad
 V=(I+J)^{-1},\quad m=Vb.
\]
These are the posterior covariance and mean under reference prior $N(0,I)$.
For a fitted class covariance $C_k\succ0$, the class-dependent part of the
raw log evidence and its conditional coefficient mean are
\begin{align}
 \ell_k&=-\tfrac12\log\det C_k
          -\tfrac12\log\det(C_k^{-1}+J)
          +\tfrac12 b^\top(C_k^{-1}+J)^{-1}b,\label{eq:switching_evidence_v32}\\
 \widehat a_k&=(C_k^{-1}+J)^{-1}b.\label{eq:switching_mean_v32}
\end{align}
They remain well defined with no observations: $m=0$, $V=I$, equal raw
evidence, and zero conditional means. The final tests retain 17 rank-deficient
blocks. Source fitting blocks have full rank, checked in the saved data;
their OLS moments minus estimated noise covariance fit $C_k$, followed by
the declared shrinkage and eigenvalue floor. Rank-deficient test blocks
are handled directly by the information-form posterior.

The forward recursion is
\[
 p_t(k)\ \propto\ \exp(\ell_{t,k}/T)
               \sum_j\widehat A_t(j,k)p_{t-1}(j).
\]
Shrinkage and temperature $T$ use development trajectory NLL. At $T=1$ this
is the ordinary plug-in Gaussian HMM filter; development calibration changes
the emission weights. Reconstruction averages the untempered conditional
means in \eqref{eq:switching_mean_v32} using these probabilities. It is a
calibrated mixture estimate. The resulting probabilities condition on estimated noise and learned
carriers, with their estimation error outside the fitted model.

\paragraph{Controls and estimands.}
Signed OAS-QDA acts on reference posterior means $m$. A quadratic logistic
head additionally receives every product and square, the ten unique entries
of $V$, estimated noise and retention. A forest receives complete 16-point
squared physical-mode reconstructions and the same quality variables.
A further forest uses squared coordinates after a training-only pooled-PCA
rotation, their quadratic interactions, and the same quality variables.
All use the same causal recursion. Discriminative class probabilities are
divided by fitting class priors before use as emission scores, avoiding
counting those priors twice. These approximate discriminative score filters use calibrated classifier
scores as emissions. The energy comparison uses the specified physical
and pooled-PCA coordinates.

Reconstruction controls include diagonal and full pooled priors,
minimum-norm least squares, the coupled emission model without temporal
recursion, and a stronger control that weights the same conditional means
using point-OAS state probabilities. The last isolates regime weighting from
conditional coefficient estimation. Zero class means, covariance shrinkage,
and observation uncertainty vary jointly in the comparison with OAS.

\begin{table}[t]
\centering\small
\caption{Causal state NLL in the nonstationary dependence experiment. Each
entry averages 12 independent panels of 12 test trajectories. All
classifiers share the learned temporal information.}
\label{tab:switching_classification_v32}
\begin{tabular}{lrrrrr}
\toprule
Method & Source & Noise & Hazard & Phase & Low \\
\midrule
Coupled filter & 0.435 & 0.497 & 0.621 & 0.769 & 1.146 \\
Static coupled & 0.789 & 0.871 & 0.796 & 0.988 & 1.145 \\
Point OAS filter & 0.427 & 0.760 & 0.603 & 2.205 & 1.205 \\
Signed quadratic & 0.445 & 5.415 & 0.631 & 6.639 & 4.107 \\
Physical energies & 0.500 & 0.606 & 0.675 & 0.897 & 1.275 \\
Rotated energies & 0.780 & 0.944 & 0.904 & 1.081 & 1.344 \\
Clock only & 1.155 & 1.163 & 1.165 & 1.143 & 1.145 \\
\bottomrule
\end{tabular}

\end{table}

\begin{table}[t]
\centering\small
\caption{Clean-signal reconstruction MSE over all samples, including missing
positions. ``Point weights'' uses the same conditional coefficient means
with point-OAS regime probabilities. The table includes the phase/dictionary
and low-interaction conditions.}
\label{tab:switching_reconstruction_v32}
\begin{tabular}{lrrrrr}
\toprule
Method & Source & Noise & Hazard & Phase & Low \\
\midrule
Coupled filter & 0.0315 & 0.3776 & 0.0319 & 1.8294 & 0.3084 \\
Static coupled & 0.0315 & 0.3954 & 0.0318 & 1.8480 & 0.3084 \\
Point weights & 0.0315 & 0.3852 & 0.0319 & 1.8397 & 0.3084 \\
Full pooled KMD & 0.0316 & 0.4171 & 0.0319 & 1.8707 & 0.3084 \\
Diagonal KMD & 0.0317 & 0.4292 & 0.0320 & 1.9022 & 0.3083 \\
Minimum-norm LS & 0.0468 & 5.7407 & 0.0472 & 1.9263 & 11.0711 \\
\bottomrule
\end{tabular}

\end{table}

\begin{figure}[tp]
\centering\includegraphics[width=\linewidth]{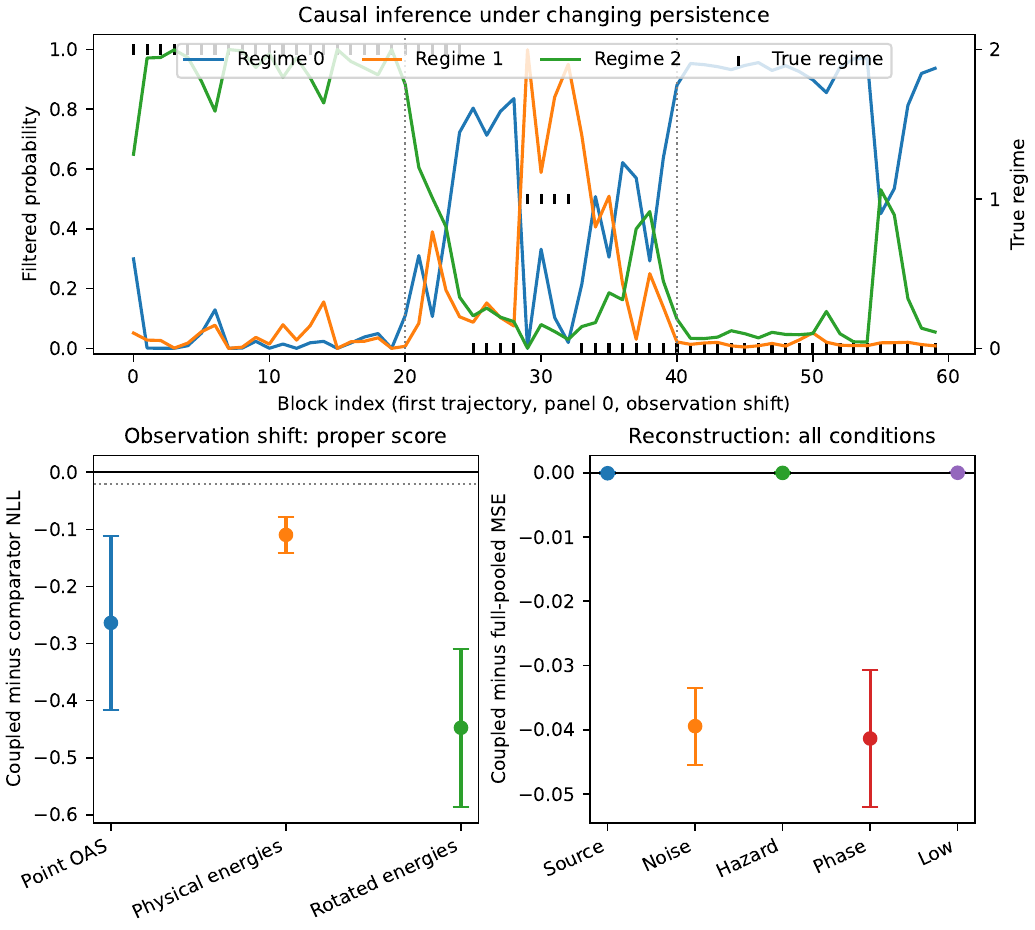}
\caption{Nonstationary regime filtering with time-varying persistence. The
first trajectory from panel 0 in the observation-shift condition is displayed. Lower panels show paired $t$ intervals
across 12 panels; the dotted score line marks $-0.02$ nats. The phase-condition
MSE difference remains favorable relative to the pooled prior while its
absolute reconstruction error is large.}
\label{fig:switching_v32}
\end{figure}

\paragraph{Observation-shift results.}
Tables~\ref{tab:switching_classification_v32}
and~\ref{tab:switching_reconstruction_v32} report every method and
condition, and Figure~\ref{fig:switching_v32} shows one trajectory and the
paired differences.
Under increased noise and missingness, coupled MSE is $0.37761$, versus
$0.41706$ for full pooled KMD: a $9.46\%$ reduction. The paired difference is
$-0.03945$, with paired $t$ interval $[-0.04546,-0.03344]$ (panel-bootstrap
sensitivity interval $[-0.04430,-0.03420]$). To assess the predeclared practical
threshold directly, we also form each panel's coupled MSE minus 0.95 times
its pooled MSE. Its mean is $-0.01860$, with paired $t$ interval
$[-0.02451,-0.01269]$ (bootstrap $[-0.02325,-0.01323]$).
Both upper endpoints are below zero, supporting at least $5\%$
improvement in the ratio of population mean MSEs under these inference
approximations. Against point-probability weighting, MSE is $0.38518$
and the incremental reduction is $1.97\%$, below that threshold;
its paired $t$ interval for the absolute difference is
$[-0.01191,-0.00324]$. The $9.46\%$ improvement uses the full pooled comparator; the
point-probability comparison isolates regime weighting.

Coupled NLL is $0.49654$, versus point-OAS $0.76031$ and physical-energy
$0.60616$. Their paired $t$ intervals for coupled-minus-control NLL are
$[-0.41609,-0.11146]$ and $[-0.14103,-0.07821]$, respectively; the mean
improvements exceed the prespecified $0.02$-nat practical threshold.
The full-quality quadratic head extrapolates poorly under this observation
shift, reaching NLL $5.415$.

\paragraph{Source, hazard, interaction, and phase conditions.}
Source reconstruction improves $0.25\%$ over the full pooled prior and
its paired intervals include zero. Under changed hazards, point OAS has
lower NLL, $0.60255$ versus $0.62075$, with coupled-minus-point paired $t$
interval $[0.00714,0.02925]$. The ranking of the two filters therefore
depends on the observation and hazard conditions.

Low interaction gives essentially identical coupled and full pooled MSE,
$0.308402$. Coupled NLL $1.146$ is close to the learned clock-only rule's
$1.145$ and exceeds the constant uniform reference $\log 3$. With
$\alpha=0.05$ no fitted head has lower NLL than the clock-only rule at the
reported precision, and the point-OAS, energy and signed quadratic heads are
worse than it; the observations add essentially no regime information
beyond the clock in this condition.

Phase/dictionary mismatch raises coupled MSE to $1.8294$; its $2.21\%$
reduction from the full pooled control is below the practical threshold.
Moreover, temporal persistence creates delay. In the first five blocks
following actual changes, observation-shift NLL is $0.937$ for the filter
versus $0.895$ for static inference; under changed hazards it is $0.985$
versus $0.806$. Whole-trajectory log-loss gains thus coexist with increased
log loss immediately after changes.

\paragraph{Uncertainty.}
NLL and MSE are averaged within whole trajectories and then within panels.
The 20,000-resample bootstrap and paired $t$ intervals use 12 independent
panels, not individual blocks. Both are pointwise, without multiplicity
adjustment and rely on panel-level resampling and paired-mean approximations.

\FloatBarrier

\section{Recorded signals: representation adequacy and observation quality}
\label{sec:recorded_studies}
Recorded ECG and bearing data evaluate the extension across choices of
mode dictionary, regime law and observation quality. The ECG studies
compare global and localized waveform spans, robust laws and supplied rhythm
features. The bearing studies compare absolute-frequency and physically
normalized order coordinates, then isolate a declared transmission-noise
problem. Reference reconstruction, native diagnosis and controlled corruption
are distinct targets throughout.
\subsection{Global-carrier ECG baseline}
\label{sec:ecg_v32}

Event alignment supplies a common time reference for the seven-coordinate
oscillatory dictionary. We evaluate this baseline on the MIT--BIH Arrhythmia
Database \cite{mitbih_moody2001}. The same cohort provides development data
for the localized representation and class law in
Section~\ref{sec:external_repair_methods_v32}.
The study uses the official version 1.0.0 waveforms and expert
annotations, with separate raw artifact recordings from the MIT--BIH Noise
Stress Test Database \cite{nstdb_moody1984}. PhysioNet provides the public
resources \cite{physionet_goldberger2000,pollard_physionet_2026}.
Dataset descriptions, downloads and attribution are at
\url{https://physionet.org/content/mitdb/1.0.0/} and
\url{https://physionet.org/content/nstdb/1.0.0/}.
The target is the literal annotation symbol N versus V (normal versus premature
ventricular contraction). This deliberately narrow annotated-beat classification
task is not the broader AAMI N/V grouping, automatic beat detection, or clinical
diagnosis. Supplied annotation times localize the beats even after corruption.

\paragraph{Patient-disjoint protocol.}
The database contains 48 records from 47 subjects, sampled at 360\,Hz. We merge
records 201 and 202 as one patient, select MLII by the header channel name
(including the reversed order in record 114), and exclude 102 and 104 because
they lack MLII. Each window has 216 samples, from 0.25\,s before to 0.35\,s after
its annotation. Eligibility checks only the target symbols, bounds, finiteness,
and availability of a preceding RR interval. A fixed, class-value-independent
ordering caps the target cohort at 200 beats per patient. Patient 232 has no
eligible target beats and is retained in the predetermined allocation without
replacement. The resulting cohort has 7,866 beats from 44 contributing patients:
6,821 N and 1,045 V. Fifteen patients have no V beats; seven have only V beats.
All contribute proper losses and reconstruction error.

Five predetermined outer folds test each patient once. Within each fold the
nominal patient budgets are nine test, five separate dictionary calibration,
eight development, and 23 fitting; empty patient 232 reduces one role's actual
count. No patient occupies two roles within a fold. Calibration uses at most
15 eligible beats per patient: 60 or 75 beats per fold. Target eligibility uses annotation labels; N/V values are hidden
from dictionary learning after eligibility is determined. All folds have both
classes in fitting and development. Full patient identifiers and role counts
are supplied with the protocol.

The source observations add independent white noise with SD uniform on
\([0.01,0.03]\)\,mV. An independent 64-sample noise-only instrument calibration
estimates the added-noise variance. This resource is supplied equally to all
methods and never uses the same beat's reference residual. Pre-existing noise
in the database waveform remains unknown. The reconstruction target is the
\emph{recorded reference}, not latent clean ECG.

\paragraph{Learned carriers, mean effects, and strong controls.}
The adaptive KMD frontend of Section~\ref{sec:adaptive_experiments_v32} extracts
three modes from calibration observations after subtraction of the calibration
mean waveform. Frequency bounds are 3--45\,Hz, localization parameter 12,
24 anchors, and 41 local candidates in a 25\% search band. Median unwrapped
phases and normalized amplitude trajectories yield three fixed sine/cosine
pairs; a constant column gives seven random coefficients. The calibration mean
is a fixed offset. A same-budget alternative uses the same calibration mean and
seven centered principal components. Both dictionaries are fixed before
test-beat evaluation. Returned KMD ridges define learned basis functions.

Class coefficient models now have distinct nonzero means. Complete source
fitting observations estimate means and covariance moments, with subtraction of
estimated coefficient-noise covariance, off-diagonal shrinkage in
\(\{0,0.2,0.5,0.8,1\}\), and eigenvalue floor \(10^{-5}\). Full-class,
diagonal-class, and shared-covariance Gaussian models separate dependence from
mean morphology. The centered Gaussian sign-loss theorem does not directly
describe these ECG laws. On retained samples \(O\), every modal head uses the
same dictionary \(H\), template \(\bar x\), and observation statistics
\[
 J=H_O^\top H_O/\widehat\sigma^2,\qquad
 b=H_O^\top(y_O-\bar x_O)/\widehat\sigma^2,\qquad
 m=(I+J)^{-1}b,\quad V=(I+J)^{-1}.
\]
Generic modal classifiers receive all 28 distinct entries of \(V\), estimated
noise, and retained fraction. The controls include point QDA, signed quadratic
logistic regression, signed RBF SVM, an energy-path forest, waveform forest,
and waveform-PCA/RBF with propagated observation covariance. The energy forest
receives 16 squared reconstructed values per mode and the squared constant
component, so it retains more than total modal energy. A normalized morphology
and RR/RBF control additionally receives the preceding RR interval and its ratio
to the median of up to five intervals ending with that interval. This explicitly
adds rhythm information from the supplied annotations; no future interval is
used. Morphology and RR features are established ECG controls
\cite{dechazal2004}; here they are implemented for the literal N/V target
described above. The prevalence-only predictor measures the effect of class imbalance.

Generic heads also train on two independently corrupted versions of each
observed fitting beat: Gaussian SD ranges \([0.1,0.3]\) and \([0.3,0.5]\)\,mV,
with retention ranges \([0.4,0.9]\) and \([0.25,0.6]\). Base and added estimated
noise variances are summed. Source Gaussian priors fit source observations;
all hyperparameter and temperature choices minimize patient-macro loss on the
same source-plus-augmented development bank. Class priors, normalizers, PCA,
and fitting weights use fitting data only, with equal total weight per patient.
Forests use 100 trees and depth choices 4, 8, or unrestricted; logistic and RBF
heads use \(C\in\{0.1,1,10\}\). Temperatures are
\(\{0.5,0.7,1,1.5,2,3\}\).

\paragraph{Observation conditions and patient-level comparisons.}
Each test beat has three separately predicted corruption repetitions in every
condition. The primary Gaussian sensor shift adds SD \([0.1,0.3]\)\,mV and
retains a fraction in \([0.4,0.9]\). The other conditions are source quality,
baseline wander (BW), muscle artifact (MA), and electrode motion (EM). Raw
artifact clips come from the final 15 minutes of the first channel of the
three named noise records; independent calibration clips come from the first
five minutes. Their imposed RMS scale is uniform on \([0.1,0.3]\)\,mV, with
the same missingness range. No artifact enters fitting or development. A scalar
noise variance for temporally correlated artifacts is a working approximation,
not their exact covariance. These tests concern held-out clips from three
records, not population-level artifact generalization.

We first average losses over each patient's beats and repetitions, then weight
patients equally. Pooled AP and AUROC are secondary, with a different weighting.
Paired 20,000-resample patient bootstrap and paired \(t\) intervals are
descriptive and conditional on the five fitted models: overlapping training
sets do not give independent model replications or refitting uncertainty.
Before testing, the primary analysis specified Holm adjustment across all
11 nonidentical classification controls and, separately, seven reconstruction
controls. Practical-margin tests use
\(\mathrm{NLL}_c-\mathrm{NLL}_j+0.02\) and
\(\mathrm{MSE}_c-0.95\mathrm{MSE}_j\).

\begin{table}[tp]
\centering\small
\caption{ECG classifiers across the specified conditions: patient-macro NLL in nats, lower is better.
BW, MA and EM are distinct real-artifact tests. Morphology + RR explicitly uses
additional rhythm information. The controls follow the implementations and
information budgets specified here.}
\label{tab:ecg_classification_v32}
\begin{tabular}{lrrrrr}
\toprule
Method & Source & Gaussian & BW & MA & EM \\
\midrule
Coupled sKMD & 0.3302 & 0.3375 & 0.3651 & 0.3748 & 0.4744 \\
Diagonal class & 0.3564 & 0.3588 & 0.3862 & 0.3921 & 0.4673 \\
Shared class & 0.3673 & 0.3985 & 0.3965 & 0.4217 & 0.4103 \\
PCA7 integrated & 0.3500 & 0.3388 & 0.3759 & 0.3804 & 0.4545 \\
Point QDA & 0.2799 & 0.2980 & 0.2870 & 0.2957 & 0.3266 \\
Signed quadratic & 0.3064 & 0.3164 & 0.3099 & 0.3261 & 0.3609 \\
Signed RBF & 0.3283 & 0.3311 & 0.3243 & 0.3399 & 0.3654 \\
Energy-path RF & 0.3821 & 0.4081 & 0.3854 & 0.4141 & 0.4575 \\
Waveform RF & 0.6496 & 0.4904 & 0.6191 & 0.5562 & 0.5296 \\
Waveform PCA/RBF & 0.2825 & 0.2804 & 0.2911 & 0.3017 & 0.3284 \\
Morphology + RR & 0.1426 & 0.1309 & 0.1408 & 0.1476 & 0.1781 \\
Prevalence & 0.5665 & 0.5665 & 0.5665 & 0.5665 & 0.5665 \\
\bottomrule
\end{tabular}
\end{table}

\begin{figure}[tp]
\centering
\includegraphics[width=0.93\linewidth]{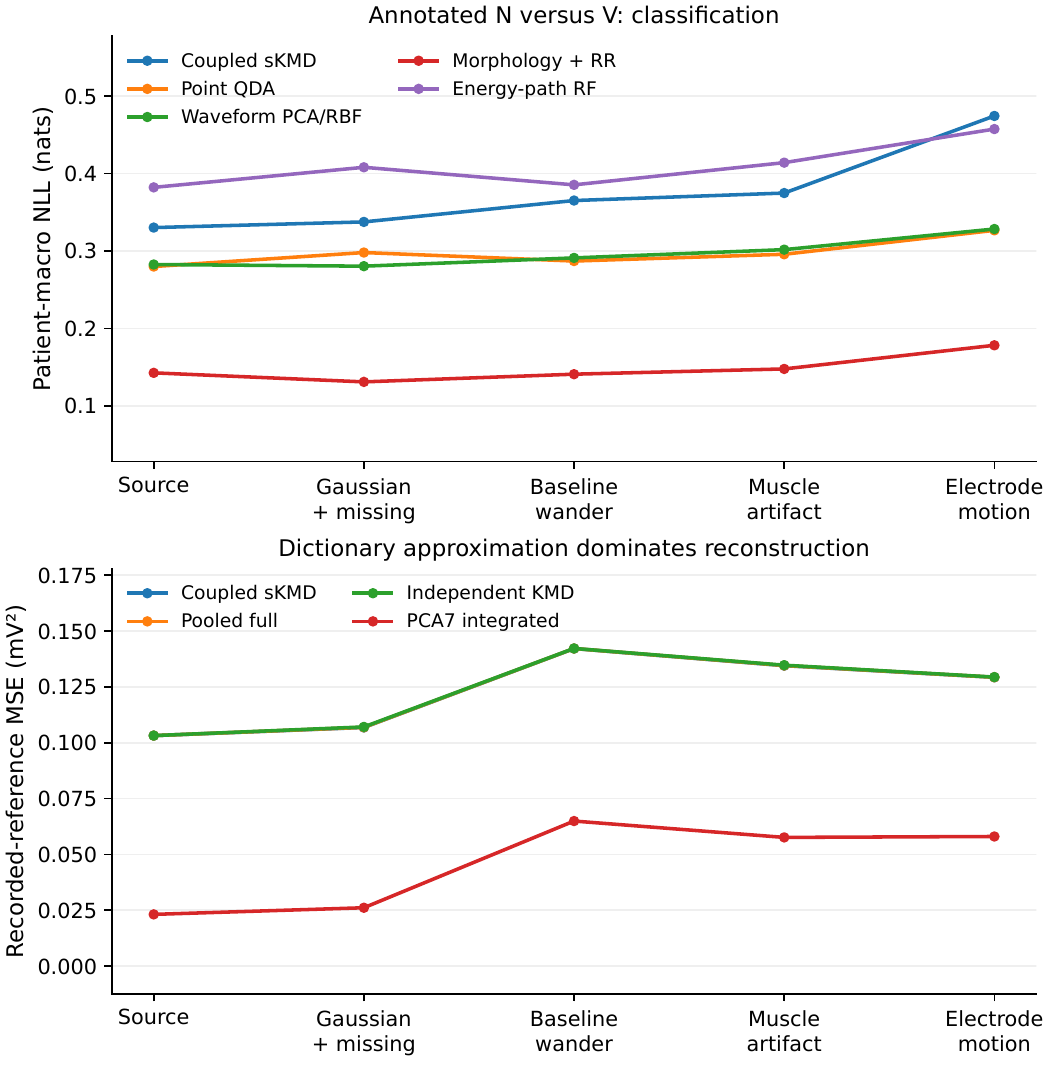}
\caption{All five prespecified ECG conditions. Lines show patient-macro means;
paired uncertainty is reported in the text and saved results. The three
KMD reconstruction curves in the lower panel nearly coincide.}
\label{fig:ecg_results_v32}
\end{figure}

\paragraph{Classification under added observation noise.}
Tables~\ref{tab:ecg_classification_v32} and~\ref{tab:ecg_quality_v32} and
Figure~\ref{fig:ecg_results_v32} report every classifier and condition.
Under the primary sensor shift, coupled NLL is 0.33754, versus 0.29797 for
point QDA, 0.28036 for waveform-PCA/RBF, and 0.13090 for morphology plus RR.
Against the energy-path forest (0.40809), the favorable difference
\(-0.07054\) has paired \(t\) interval \([-0.23940,0.09831]\) and bootstrap
interval \([-0.23795,0.08526]\). Against diagonal-class integration the mean
gain is 0.02129 nats, but the difference interval
\([-0.04473,0.00216]\) includes zero and the adjusted one-sided \(p\)-value
is 0.370. No coupled classification gain against any comparator survives the
declared Holm family; even the prevalence comparison has adjusted \(p=0.095\).
None establishes the 0.02-nat practical margin. The morphology/RR advantage
is large, but its extra rhythm inputs prevent attributing all of it to the
classifier. The waveform-only and same-modal-information point-QDA
controls also have better means. Under electrode motion, the coupled loss
rises to 0.47441, exceeding even shared-covariance integration (0.41031).

\begin{table}[tp]
\centering\small
\caption{Primary Gaussian-shift quality metrics. Brier score sums the two class
squared errors; accuracy and Brier use equal patient weight. AP and AUROC pool
all beat repetitions and are descriptive secondary metrics.}
\label{tab:ecg_quality_v32}
\begin{tabular}{lrrrr}
\toprule
Method & Brier & Accuracy & AP & AUROC \\
\midrule
Coupled sKMD & 0.1860 & 0.8853 & 0.7673 & 0.9311 \\
Diagonal class & 0.1998 & 0.8726 & 0.7341 & 0.9174 \\
Shared class & 0.2389 & 0.8524 & 0.5531 & 0.7964 \\
PCA7 integrated & 0.1892 & 0.8822 & 0.7764 & 0.9168 \\
Point QDA & 0.1668 & 0.8908 & 0.7758 & 0.9280 \\
Signed quadratic & 0.1781 & 0.8925 & 0.7838 & 0.9099 \\
Signed RBF & 0.1899 & 0.8854 & 0.7375 & 0.8926 \\
Energy-path RF & 0.2132 & 0.8780 & 0.6383 & 0.8325 \\
Waveform RF & 0.1944 & 0.8790 & 0.6848 & 0.8683 \\
Waveform PCA/RBF & 0.1667 & 0.8851 & 0.7077 & 0.9217 \\
Morphology + RR & 0.0719 & 0.9549 & 0.9663 & 0.9935 \\
Prevalence & 0.3709 & 0.7751 & 0.1175 & 0.4297 \\
\bottomrule
\end{tabular}
\end{table}

\paragraph{Reference-waveform reconstruction.}
Table~\ref{tab:ecg_reconstruction_v32} reports reconstruction error in every
condition. All KMD reconstruction controls share the same learned span and estimated
observation model. The pooled-full prior uses moments of the fitted class
mixture, including between-class mean variation; it is a shared-estimate
ablation rather than an independently tuned pooled estimator. Point-weighted
reconstruction uses the coupled conditional means with point-QDA probabilities.
All calibrated mixtures use tempered class weights with untempered conditional
means, and are not asserted to be exact untempered Bayes means.
Primary coupled reference MSE is 0.106866\,mV$^2$, versus 0.107111 for independent
KMD, 0.106934 for pooled full covariance, and 0.106855 for point-weighted
reconstruction. The first two relative reductions are only 0.229\% and
0.063\%; no comparison establishes the specified 5\% margin. The diagonal,
independent, and least-squares zero-margin comparisons are statistically
detectable after adjustment, but too small to support a practical gain.
PCA7 integration is much better at 0.026136\,mV$^2$ despite the same coefficient
dimension and calibration budget.

\begin{table}[tp]
\centering\small
\caption{Patient-macro error against the recorded ECG reference (mV$^2$).
Four-decimal rounding makes several KMD controls appear equal; complete
precision and paired intervals are saved. Existing recording noise is part
of the target, so these are not errors against latent clean ECG.}
\label{tab:ecg_reconstruction_v32}
\begin{tabular}{lrrrrr}
\toprule
Method & Source & Gaussian & BW & MA & EM \\
\midrule
Coupled sKMD & 0.1032 & 0.1069 & 0.1422 & 0.1345 & 0.1293 \\
Diagonal class & 0.1032 & 0.1070 & 0.1422 & 0.1346 & 0.1294 \\
Shared class & 0.1032 & 0.1069 & 0.1422 & 0.1346 & 0.1293 \\
PCA7 integrated & 0.0232 & 0.0261 & 0.0650 & 0.0576 & 0.0580 \\
Point-weighted mixture & 0.1032 & 0.1069 & 0.1422 & 0.1345 & 0.1293 \\
Pooled full & 0.1032 & 0.1069 & 0.1422 & 0.1346 & 0.1294 \\
Independent KMD & 0.1032 & 0.1071 & 0.1423 & 0.1348 & 0.1294 \\
Least squares & 0.1032 & 0.1079 & 0.1424 & 0.1349 & 0.1298 \\
\bottomrule
\end{tabular}
\end{table}

An exploratory projection diagnostic, conducted after scoring, measures
representation error. The best
least-squares approximation of each held-out recorded reference within its
fold's fixed affine span has patient-macro MSE 0.103206 for KMD and 0.023155 for
PCA7. The source reconstruction errors, 0.103221 and 0.023169, are already close
to those floors. Covariance integration cannot recover waveform components
outside the selected span. This diagnostic uses references only to assess
approximation error; it changes no prediction. Two calibration extractions
reported insufficient separated peaks and were retained. Across all test
masks at least 64 samples remain, the smallest Gram eigenvalue is 13.84, and
the largest condition number is 6.01: poor morphology approximation is not a
rank-deficiency effect.

MIT--BIH supplies development data for the localized robust procedure;
INCART supplies the independent evaluation cohort.

\FloatBarrier

\subsection{Localized kernels and robust regime laws}
\label{sec:external_repair_methods_v32}

We combine localized carrier packets with finite scale-mixture regime laws
to model waveform morphology and between-subject heterogeneity. MIT-BIH
patients and Paderborn bearings supply development data; procedures are fixed
before evaluation on separate physical units. The global-carrier and
absolute-frequency methods provide baseline comparisons.

\subsubsection{A waveform span built from localized KMD kernels}

Let $h_{j,c},h_{j,s}\in\R^p$, $j=1,2,3$, be the three independently
calibrated KMD carrier pairs. A constant and these six vectors give a
seven-dimensional span. This is restrictive for an ECG beat: different
parts of a beat can have different morphology even when the carrier
frequencies are appropriate. Using the development data, we
construct coefficients of smooth local wave packets,
\begin{equation}
 B=\left[\mathbf1,\,
 \{w_\ell\odot h_{j,c},w_\ell\odot h_{j,s}\}_{j=1,\ldots,3;\,
 \ell=1,\ldots,L}\right],\qquad
 w_\ell(t)=\exp\!\left[-\frac{(t-c_\ell)^2}{2b^2}\right].
 \label{eq:ecg_localized_dictionary_v32}
\end{equation}
The ECG implementation uses $p=216$, $L=7$, equally spaced centers
across the beat window, and $b=0.7$ times the center spacing; thus $q=43$.
The local amplitudes can vary across the beat. The construction remains a
finite collection of prescribed kernels, $K_j=B_jB_j^\top$, within the KMD
framework. KMD permits general waveforms. Related wave-shape
methods address nonstationary morphology directly
\cite{lin_wave_shape_2021,ruiz_wave_shape_2024,colominas_spiky_2026}.
This construction addresses waveform approximation before assessing the
added value of the stochastic mode law.

\paragraph{Stable coordinates preserve the kernel.}
The overlapping wave packets can make the columns of $B$ ill-conditioned.
Assume $B$ has full column rank, checked numerically in the implementation.
For its thin singular value decomposition $B=USV^\top$, use
\begin{equation}
 H=\sqrt p\,U,\qquad
 \Gamma=S^2/p,\qquad
 H\Gamma H^\top=BB^\top,\qquad
 x=Ha=B\bigl(\sqrt p\,VS^{-1}a\bigr).
 \label{eq:ecg_kernel_svd_v32}
\end{equation}
This transformation diagonalizes the prescribed kernel; it does not fit
principal components to evaluation waveforms. The kernel eigenvalues
$\Gamma$ must be retained when implementing the original independent
KMD reconstruction. Setting them all to one would change that baseline.
The fitted class law can instead be expressed directly in the normalized
$H$ coordinates. To report physical carrier-group energies, map the
coefficients back through $\sqrt p\,VS^{-1}$ and the original column
groups. Squaring arbitrary singular-vector coefficients would test a
different energy representation.

\paragraph{Approximation error and a matched resource control.}
For any estimate constrained to the affine span
$\bar x+\operatorname{col}(H)$, the squared distance of a reference beat
to that affine span is a lower bound on its
reconstruction error. We measure this projection error separately from
prediction risk. A data-PCA span of the same dimension, learned from the
same calibration patients, provides a strong control for whether the
particular KMD carrier structure helps. Both pipelines receive the same
recorded channels, event locations, observation masks and noise estimates.
Increasing the rank can reduce bias while raising estimation variance;
the rank is fixed before independent-cohort predictions.

\subsubsection{Finite scale mixtures with observed rhythm features}

Gaussian scale mixtures are an established robust modeling device
\cite{peel_mclachlan_tmixture_2000}. We specify the finite law used in the
experiment so that its conditioning and moment conventions are reproducible.

For class $k$, let $A\in\R^q$ be the waveform coefficients and let
$R\in\R^d$ collect the supplied past-RR features. These features use
past event times, not future intervals or diagnostic labels. We model
their joint law by the finite mixture
\begin{equation}
 \begin{pmatrix}A\\R\end{pmatrix}\Bigm| k,s
 \sim\N(\mu_k,s C_k),\qquad
 \Pr(s=s_\ell\mid k)=\omega_\ell,\qquad
 \sum_\ell\omega_\ell=\sum_\ell\omega_\ell s_\ell=1.
 \label{eq:ecg_scale_prior_v32}
\end{equation}
The fixed twelve scales start from $2^{-3},\ldots,2^8$. Their masses are
the inverse-gamma probabilities between geometric midpoint boundaries
with shape and scale $2.5$, and the scales are divided by their weighted
mean. Thus the mixture has covariance exactly $C_k$. This is a finite
Gaussian mixture with heavier tails over the observed range, rather than
an exact Student distribution. The scale law is fixed on development
data. Learning the class moments, covariance shrinkage and calibration
temperature uses only the specified fitting and development roles.

Partition $\mu_k$ and $C_k$ into waveform and rhythm blocks. Conditional
on an observed $r$ and a scale $s$, Gaussian conditioning gives
\begin{align}
 \mu_{A\mid r,k}
 &=\mu_{A,k}+C_{AR,k}C_{RR,k}^{-1}(r-\mu_{R,k}),\\
 C_{A\mid r,k,s}
 &=s\bigl(C_{AA,k}-C_{AR,k}C_{RR,k}^{-1}C_{RA,k}\bigr).
 \label{eq:ecg_rr_conditioning_v32}
\end{align}
The conditional mean does not depend on the scale, but the rhythm
likelihood does. In particular, the observed RR values update both
class weights and scale weights. Omitting this factor would not compute
the joint model in \eqref{eq:ecg_scale_prior_v32}.

Let $Y_o=M(H A+\bar x)+\varepsilon_o$, where $M$ selects the observed
samples, the template $\bar x$ is independently calibrated, and
$\varepsilon_o\sim\N(0,Q_o)$ has the supplied working covariance.
Write $H_o=MH$ and $y_o=Y_o-M\bar x$. The complete component evidence is
\begin{equation}
 e_{k\ell}=
 \pi_k\omega_\ell\,
 \N(r;\mu_{R,k},s_\ell C_{RR,k})\,
 \N\!\left(y_o;H_o\mu_{A\mid r,k},
 Q_o+H_o C_{A\mid r,k,s_\ell}H_o^\top\right).
 \label{eq:ecg_complete_scale_evidence_v32}
\end{equation}
Summing $e_{k\ell}$ over scales gives the class evidence; summing
component conditional means with normalized $e_{k\ell}$ gives the
posterior mean at calibration temperature one. The implementation uses
the equivalent information form with $J=H_o^\top Q_o^{-1}H_o$ and
$b=H_o^\top Q_o^{-1}y_o$. This form permits fewer observed samples than
coefficients and includes the zero-observation case. The positive prior
covariance supplies the missing information; no inverse of a singular
sample Gram matrix is required for prediction.

\paragraph{Calibration and the interpretation of reconstruction.}
If validation selects a temperature $T\ne1$, class probabilities are
proportional to $(\sum_\ell e_{k\ell})^{1/T}$. Within each class, the
scale weights and conditional coefficient means remain those of the
untempered model. The resulting reconstruction is a mixture of exact
class-conditional posterior means using calibrated class probabilities.
It need not equal the posterior mean under the original generative model.
The recorded-data tables use this explicitly specified estimator.

\paragraph{Matched ablations.}
The same localized kernels are supplied to independent-kernel KMD,
pooled-law reconstruction and class-dependent sKMD. The same robust scale
law and rhythm features are supplied to the matched point-estimate
classifier and to the equal-rank PCA model. These matched comparisons
isolate observation integration from the effects of the robust law,
rhythm features and representation dimension. The independent-cohort
experiment reports each comparison.

The diagonal class-covariance ablation is defined in the normalized $H$
coordinates. It removes their off-diagonal covariance entries, including
linear waveform--RR covariance, but a shared random scale still induces
statistical dependence. Because the singular vectors mix raw carrier
groups, this ablation also does not impose independence of those original
groups. The Gaussian sufficiency theorems apply to their specified
fixed-coordinate Gaussian families; this experiment uses a finite
non-Gaussian mixture.


\subsection{Independent evaluation of localized robust ECG models}
\label{sec:ecg_repair_v32}

\subsubsection{Development data and an independent cohort}

The global seven-dimensional carrier dictionary and single-Gaussian class
model provide the MIT--BIH development baselines. Those patients supply
method-development data, separate from the INCART evaluation. Two development
comparisons motivate the specification in
Section~\ref{sec:external_repair_methods_v32}. First,
localized carrier packets reduce the mean of the five development-fold
projection errors from $0.1025$ to $0.00221$\,mV$^2$ at dimension 43; the
same-rank calibration PCA floor is smaller, $0.000822$\,mV$^2$. Second, within
the expanded span, the localized Gaussian classifier has source NLL $1.4613$
in one of two inspected pilot folds. A fixed finite scale-mixture prior
reduces that fold's source NLL to $0.0583$ without RR features. These measured
development outcomes motivate both choices. Both pilots use the MIT--BIH
development cohort.

The independent evaluation corpus is the St Petersburg INCART database:
75 recordings grouped into 32 original Holter subjects, sampled at 257\,Hz
with twelve leads \cite{yakushenko_incart_2008,physionet_goldberger2000,pollard_physionet_2026}.
Before producing any INCART prediction, we select the first 180 seconds of
\emph{every} recording, choose lead II by its header name, and group records
by the official patient identifier. Header gains convert ADC values to mV;
the signals are resampled to 360\,Hz before adding experimental noise.
Models are fitted locally within INCART using the development-selected
procedure; MIT--BIH model parameters are not transferred.

The target remains literal annotation symbols N and V. The supplied times
are multilead QRS locations whose positions were not manually corrected;
they are not precise lead-II R peaks. Fixed 216-sample windows and a
class-value-independent cap of 200 eligible beats per patient yield
6,196 beats, comprising 5,548 N and 648 V, from 31 contributing patients.
Patient 8 has no eligible target beats and retains its predetermined
allocation. No patient, condition or recording prefix is replaced after
inspecting performance. These are classifications of annotated beats and
recoveries of recorded reference windows, not automatic beat detection,
clinical diagnosis or recovery of latent clean ECG.

\subsubsection{Patient roles, information and decision criteria}

Four outer folds each hold out eight nominal patients. Within each fold,
16 nominal remaining patients fit the models, four select hyperparameters, and four
supply the separate dictionary calibration bank. The latter contributes
15 eligible beats per patient, hiding N/V values after eligibility.
The same 60 recorded beats, with no additional experimental corruption,
supply the three-mode adaptive KMD carrier
estimator and the 43-dimensional PCA comparison. Seven smooth windows per
carrier pair produce the fixed 43-dimensional localized KMD span. Test
signals never adapt either dictionary. All fitting/development roles have
both classes; every eligible test patient remains in the loss averages.

All full classifiers receive the same two past-RR features: the previous
interval and its ratio to the median of up to five available intervals
including that interval. Their logarithms are standardized using only
fitting-patient weighted statistics. RR-only robust QDA and RBF controls
check how much classification can be achieved without morphology. Signed
modal heads receive every unique entry of their full coefficient covariance,
plus the estimated noise scale and retained fraction. Energy-path controls
recover the original KMD kernel-group curves before squaring, as required
by \eqref{eq:ecg_kernel_svd_v32}. Waveform and normalized-morphology controls
have their own fitting-only PCA features and observation-quality inputs.

The diagonal ablation sets off-diagonal entries to zero in the normalized
$H$/RR covariance, conditional on the scale. The shared latent scale still
induces higher-order dependence and RR-dependent scale weights. Moreover,
SVD coordinates mix the original carrier groups, so this is a
zero-cross-covariance control in the stated coordinates, not a complete
raw-mode independence ablation. The original prescribed-kernel KMD
reconstruction is a separate comparator.

The integrated models fit source observations with added white-noise SD
uniformly distributed over $[0.01,0.03]$\,mV. Point and generic heads also
train on two independently degraded copies of each observed fitting beat.
Development uses the same source and augmentation schedule for all heads.
Every noise estimate uses an independent 64-sample noise-only monitor;
no prediction receives the uncorrupted reference of its own beat. Source
recording noise remains unknown. The class shrinkage and temperature grids,
finite scale law, generic-head budgets and seeds are specified in the
accompanying protocol before evaluation on INCART.

The primary condition adds white-noise SD in $[0.1,0.3]$\,mV and retains
independently between 40\% and 90\% of samples. Source, baseline wander,
muscle artifact and electrode motion are prespecified secondary conditions.
Artifact test clips and noise-monitor clips use disjoint time portions of
the three raw Noise Stress Test Database records. They are deliberately
misspecified by the scalar white-noise likelihood. Three corruption
repetitions are predicted separately; losses are averaged within each
patient before equal-patient aggregation.

The two primary classification comparisons are against robust point QDA
with RR and integrated robust PCA43 with RR. Separate Holm corrections
cover the zero-margin and 0.02-nat practical-margin tests. The three
primary reconstruction comparisons are against an independently fitted
pooled full prior, original prescribed-kernel KMD, and the same conditional
means weighted by robust point-QDA probabilities. Their 5\% practical
criterion tests $\mathrm{MSE}_c-0.95\mathrm{MSE}_j$ directly.
Pooled priors select shrinkage using recorded development MSE; original
KMD fits a scalar kernel amplitude by source marginal likelihood and
receives a fitted pooled RR conditional mean. These are strong,
specified reconstruction controls. All-control Holm diagnostics are
also supplied. Paired bootstrap and $t$ intervals are
descriptive conditional patient analyses of overlapping cross-fits;
they do not include independent model-refitting uncertainty.

\subsubsection{Independent-cohort outcomes}

\begin{table}[t]
\centering\small
\caption{INCART patient-macro NLL (nats), all specified classifiers and
conditions. Every full classifier receives the same past-RR resource.
Lower is better; no condition selects or tunes a model.}
\label{tab:incart_nll_v32}
\begin{tabular}{lrrrrr}
\toprule
Method & Source & Gaussian & Wander & Muscle & Electrode\\
\midrule
Localized robust sKMD & 0.071 & 0.039 & 0.068 & 0.083 & 0.071\\
Robust point + RR & 0.111 & 0.096 & 0.112 & 0.106 & 0.118\\
Robust PCA43 & 0.102 & 0.063 & 0.098 & 0.103 & 0.150\\
Gaussian point + RR & 0.119 & 0.121 & 0.128 & 0.124 & 0.137\\
Signed quadratic logistic & 0.281 & 0.279 & 0.279 & 0.281 & 0.286\\
Signed RBF & 0.123 & 0.121 & 0.122 & 0.122 & 0.124\\
Original KMD energy path & 0.128 & 0.145 & 0.132 & 0.152 & 0.131\\
Waveform PCA16/RBF & 0.134 & 0.129 & 0.134 & 0.137 & 0.143\\
Normalized morphology/RBF & 0.136 & 0.117 & 0.129 & 0.128 & 0.154\\
Waveform forest & 0.215 & 0.190 & 0.204 & 0.201 & 0.195\\
RR-only robust QDA & 0.139 & 0.139 & 0.139 & 0.139 & 0.139\\
RR-only RBF & 0.306 & 0.306 & 0.306 & 0.306 & 0.306\\
Localized Gaussian & 0.200 & 0.211 & 0.316 & 0.326 & 0.322\\
Shared covariance & 0.130 & 0.104 & 0.128 & 0.130 & 0.132\\
Diagonal joint covariance & 0.111 & 0.089 & 0.112 & 0.111 & 0.117\\
Robust global KMD7 & 0.041 & 0.051 & 0.049 & 0.051 & 0.095\\
Prevalence & 0.358 & 0.358 & 0.358 & 0.358 & 0.358\\
\bottomrule
\end{tabular}
\end{table}

\begin{table}[t]
\centering\small
\caption{INCART recorded-reference MSE, in $10^{-3}$\,mV$^2$.
Projection floors are evaluation diagnostics, not feasible predictors.
The independent pooled priors use no class labels; their shrinkage is
selected on development reference recovery.}
\label{tab:incart_reconstruction_v32}
\begin{tabular}{lrrrrr}
\toprule
Method & Source & Gaussian & Wander & Muscle & Electrode\\
\midrule
Localized robust sKMD & 3.977 & 10.579 & 53.723 & 46.569 & 50.256\\
Point-weighted conditional means & 3.977 & 10.625 & 53.232 & 46.229 & 49.938\\
Independent pooled full prior & 3.972 & 9.151 & 49.565 & 44.414 & 47.202\\
Independent pooled diagonal prior & 3.980 & 13.527 & 55.279 & 48.469 & 51.529\\
Original kernel KMD & 4.012 & 89.432 & 66.383 & 61.321 & 60.933\\
Robust PCA43 & 2.663 & 8.119 & 46.596 & 41.999 & 44.423\\
Localized Gaussian & 3.980 & 11.093 & 48.518 & 43.592 & 47.057\\
Shared covariance & 3.975 & 10.266 & 52.034 & 45.403 & 47.764\\
Diagonal joint covariance & 3.980 & 13.176 & 55.301 & 48.608 & 51.374\\
Robust global KMD7 & 92.848 & 96.666 & 131.598 & 124.237 & 122.929\\
Minimum-norm least squares & 3.982 & 702.960 & 82.157 & 165.647 & 91.142\\
Localized-span floor & 3.895 & 3.895 & 3.895 & 3.895 & 3.895\\
PCA43-span floor & 2.584 & 2.584 & 2.584 & 2.584 & 2.584\\
Global KMD7-span floor & 92.834 & 92.834 & 92.834 & 92.834 & 92.834\\
\bottomrule
\end{tabular}
\end{table}

\begin{figure}[t]
\centering
\includegraphics[width=\linewidth]{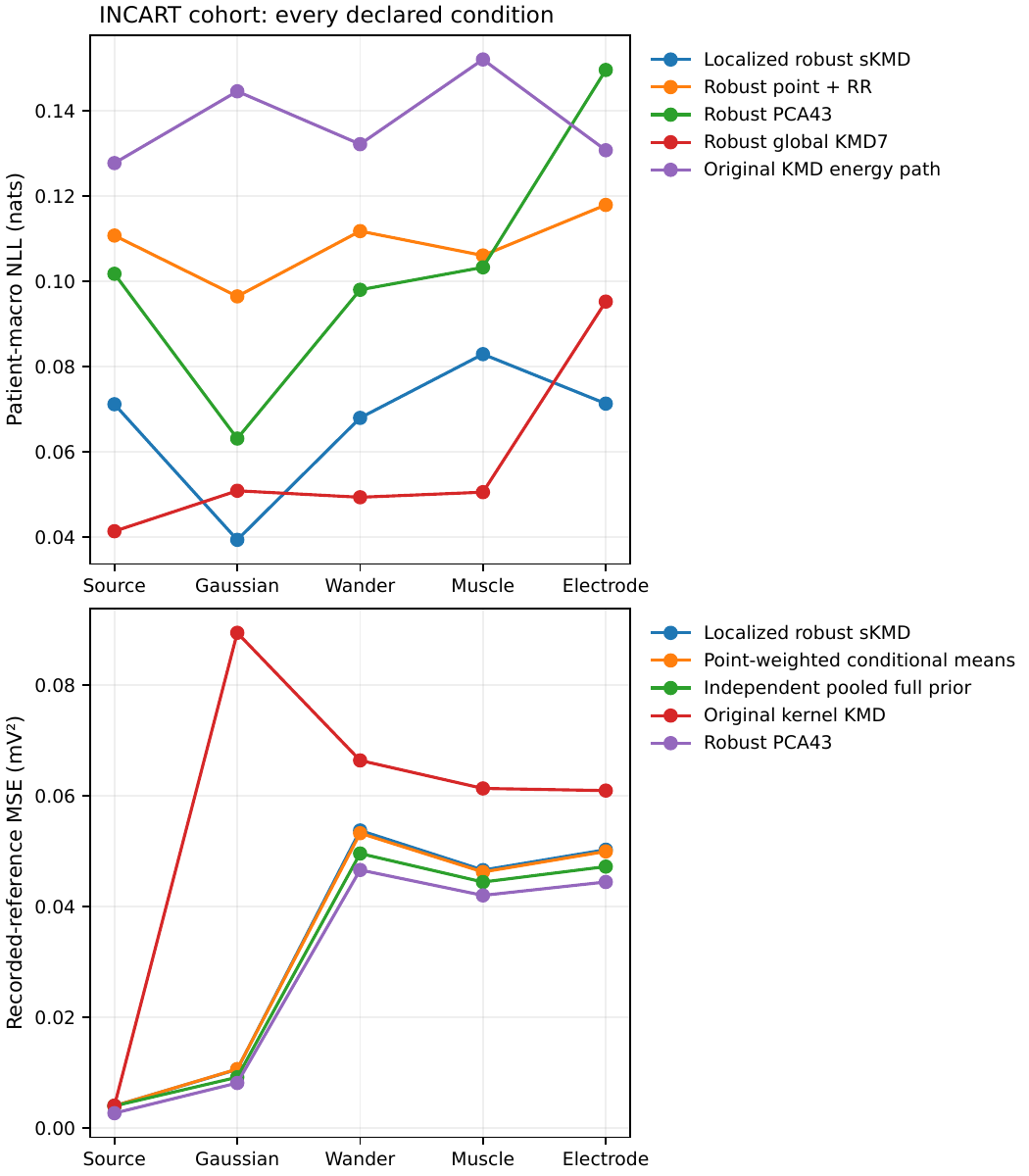}
\caption{Selected strong controls across every declared INCART condition.
The complete method set is reported in Tables~\ref{tab:incart_nll_v32}
and~\ref{tab:incart_reconstruction_v32}. Recorded-reference recovery and
class probability estimation are separate targets.}
\label{fig:incart_conditions_v32}
\end{figure}

\begin{figure}[t]
\centering
\includegraphics[width=\linewidth]{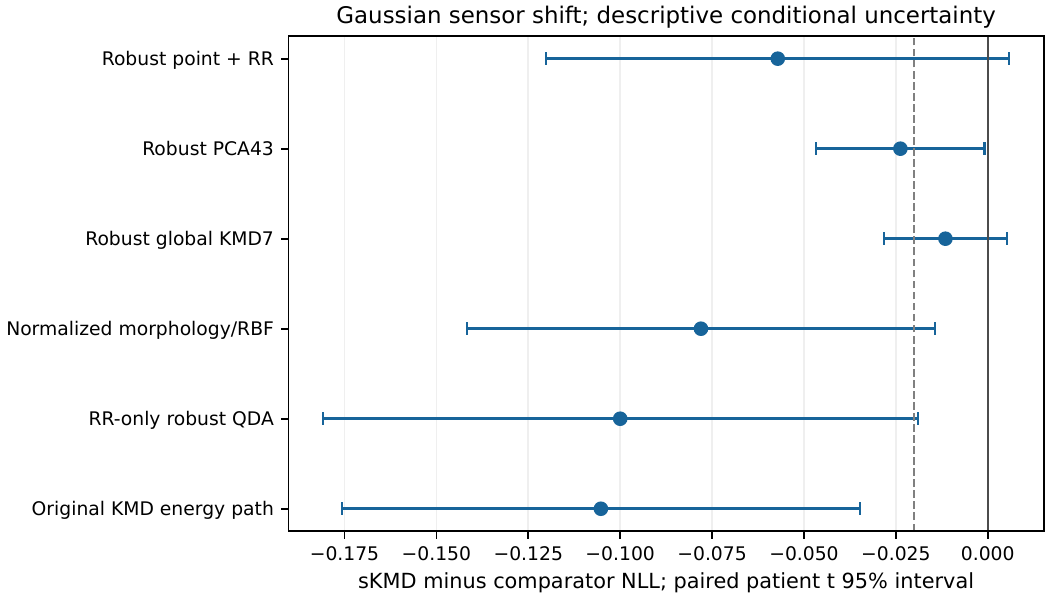}
\caption{Patient-paired NLL differences in the primary Gaussian sensor
shift. Negative values favor localized robust sKMD. The dashed line is
$-0.02$ nats. Intervals are pointwise; the text supplies the prespecified
Holm-adjusted primary and all-control tests.}
\label{fig:incart_primary_v32}
\end{figure}

Tables~\ref{tab:incart_nll_v32} and~\ref{tab:incart_reconstruction_v32}
report every method and condition, Figure~\ref{fig:incart_conditions_v32}
shows the strongest controls across conditions, and
Figure~\ref{fig:incart_primary_v32} shows the patient-paired primary
differences.
Localized robust sKMD has the lowest mean NLL among the 17 specified classifiers in
the primary Gaussian sensor shift: $0.03937$, compared with $0.09649$ for
robust point QDA and $0.06313$ for robust PCA43. The paired differences are
$-0.05712$ and $-0.02377$ nats. Both prespecified \emph{one-sided}, zero-margin
paired-$t$ comparisons have Holm-adjusted $p=0.04235$ within the two-comparison
primary family. These results are suggestive, not conclusive, evidence of a
conditional classification benefit for the localized robust model under
added noise: both tests are one-sided and pass at level $0.05$ by a small
margin, the two-sided interval for the point comparison includes zero, and
the paired $t$ test treats the 31
patients as independent although all predictions come from four fitted
models with overlapping training sets. A fold-level analysis over the four
refits resolves neither contrast. Localized robust sKMD has lower mean
primary NLL than robust point QDA in 4 of 4 folds (fold differences
$-0.134$ to $-0.020$ nats; $t$ interval over folds, 3 df,
$[-0.140,0.028]$) and than PCA43 in 3 of 4 ($-0.045$ to $+0.002$;
$[-0.061,0.015]$), so neither fold-level interval excludes zero. Leaving
out the patients of any one fold raises the larger Holm-adjusted one-sided
$p$ of the two-comparison family to between $0.059$ and $0.102$. This
dependence on the four refits is a further reason to treat the
classification evidence as suggestive.

The point-QDA difference has two-sided paired-$t$ interval
$[-0.12014,0.00589]$, which includes zero; its bootstrap sensitivity
interval is $[-0.12615,-0.01443]$. For PCA43 the corresponding intervals
are $[-0.04665,-0.00088]$ and $[-0.04793,-0.00547]$.
Thus the point comparison is sensitive to interval construction. Neither
0.02-nat practical-margin test succeeds: adjusted $p=0.23835$ against robust
point QDA and $p=0.36962$ against PCA43. Adjustment over all 16 controls does
include nonsignificant contrasts. These descriptive diagnostics are
conditional on the cross-fits and patient population described above.

The proper-loss improvement is accompanied by patient-macro Brier scores
$0.01637$, $0.04401$ and $0.02796$, and accuracies $0.99000$, $0.97225$ and
$0.98301$, for coupled, robust point and PCA43 respectively. Brier here is the
sum of squared errors over the two class probabilities. The primary model has
lower patient NLL than robust point QDA for 24 of 31 patients and than PCA43
for 23 of 31. Outcomes remain heterogeneous: robust point QDA has NLL
$1.00036$ on patient 31, while the coupled model's largest primary-condition
patient loss is $0.40157$ on patient 2. At threshold $0.5$, pooled V sensitivity
is lower for coupled than robust point QDA, $0.9141$ versus $0.9414$, while
N specificity is higher, $0.9989$ versus $0.9758$. Better NLL therefore does
not imply a better value of every diagnostic endpoint.

\paragraph{Global and localized representations.}
Adding a robust law and matched RR information also makes the global
seven-dimensional representation effective for classification. Its primary
NLL is $0.05088$, and its source NLL is \emph{better} than localized sKMD,
$0.04138$ versus $0.07119$. It also wins baseline wander and muscle artifact
in mean NLL. Localization is consequently not shown to be necessary for
classification. Patient 30 has coupled source NLL $1.02594$. The finite-mixture model
reduces aggregate loss relative
to the localized Gaussian model, but does not eliminate subject-specific
misspecification. RR-only robust QDA has primary NLL $0.13936$; waveform
information adds useful predictive information for the annotated-beat
classification target.

\paragraph{Recorded-reference reconstruction.}
Primary recorded-reference MSE is $0.010579$\,mV$^2$ for the coupled model,
$0.009151$ for the independently fitted pooled full prior, and $0.008119$ for
PCA43. Both strong controls are better: the coupled reconstruction error is
about $16\%$ larger than the pooled prior's and $30\%$ larger than PCA43's. The original prescribed-kernel KMD
mean MSE is $0.089432$, so the coupled estimate has an 88.2\% smaller point
mean, but patient variation is substantial: the zero-margin primary
reconstruction Holm diagnostic is $p=0.10061$ and the 5\% margin diagnostic
is $p=0.15436$. Point-weighted conditional means have MSE $0.010625$;
the coupled reduction is $0.43\%$, and its zero-margin adjusted
$p=0.05781$. No primary reconstruction comparison meets its Holm criterion,
so the primary reconstruction differences are not statistically established.

The localized affine-span floor is $0.003895$\,mV$^2$, compared with
$0.092834$ for global KMD7 and $0.002584$ for PCA43. Thus localization removes
much of the diagnosed approximation error, while PCA still supplies a
better reconstruction span on these data. All five calibration
peak-selection failures are included in the analysis. Raw packet-basis condition numbers
range from $5.69\times10^4$ to $1.60\times10^5$; the normalized dictionary
preserves its kernel to $2.4\times10^{-14}$. Some masked Gram matrices are
poorly conditioned, reaching $4.64\times10^8$. The information-form posterior
avoids inverting these Gram matrices. The large least-squares errors are
reported as outcomes of unregularized recovery; the primary comparisons use
the specified reconstruction controls.

\paragraph{Exploratory same-law integration comparison.}
The same-law plug-in comparison was specified after two folds had completed
and evaluated after all specified folds finished. It is exploratory and is
excluded from the confirmatory family. Keeping each coupled model's
means, covariances, scale weights, class priors and temperature fixed, evaluate
the point density at the saved reference coefficient mean and RR instead of
integrating the observation uncertainty. Primary NLL becomes $0.19181$
versus $0.03937$ for integration; the descriptive paired-$t$ interval for
the difference is $[-0.24437,-0.06051]$. This isolates the benefit of the
observation integration for that fitted law and its integration-selected
temperature. The independently fitted point control remains in the primary
comparison family; the same-law analysis is exploratory.
The informative means and RR features also distinguish this application from
the paper's centered, covariance-only information-separation example.


\FloatBarrier

\subsection{Absolute-frequency bearing baseline under operating-condition shift}
\label{sec:external-bearing-v32}

We evaluate the adaptive frontend on Paderborn bearing recordings, with
both physical bearings and an operating condition held out. This recorded-data
task extends the controlled mode-interaction comparisons to a separate
observation setting.

\subsubsection{Data, independent units, and protocol}
The official Paderborn bearing dataset is documented in~\cite{lessmeier_bearing_2016}. We use all fifteen physical bearings listed in that description's Table~10: five healthy bearings (K001--K005), five with real outer-ring damage (KA04, KA15, KA16, KA22, KA30), and five with real inner-ring damage (KI04, KI14, KI16, KI18, KI21). The complete predefined cohort avoids choosing bearings according to the present results. Mixed and artificial damage are outside this study. Recordings 1 and 2 supply four fixed, nonoverlapping half-second windows each, giving 90 original recordings and 360 windows. The data are attributed to the KAt-DataCenter, Paderborn University, and distributed under its CC BY-NC 4.0 license.

Five fixed folds each hold out one bearing per class, use another bearing per class for development, and fit on the remaining three per class. Every recording and window of a bearing has the same role within a fold. Fitting and development use two conditions at 1500 rpm: nominal torque, \texttt{N15\_M07\_F10}, and reduced torque, \texttt{N15\_M01\_F10}. The 900-rpm condition, \texttt{N09\_M07\_F10}, is never used for fitting, calibration, or hyperparameter selection. Each bearing is tested once per condition. Thus there are fifteen physical test units per condition, not 120 independent windows. The fitted training sets overlap across folds.

The protocol, candidate sets and practical threshold were frozen before feature extraction. The channel-indexed monotone \texttt{X} vectors contain 255997--278039 vibration samples and span approximately four units. Interpreting those axes as time gives mean rates of 63999.2--69509.5 samples per second. Before any classifier test metric was computed or inspected, a dated amendment to the frozen protocol changed extraction from the originally frozen nominal-index convention to interpolation of this linked axis onto a uniform 64-kHz grid within the declared windows. All methods receive the same interpolated observations. The linked axes have blank unit metadata, so this is an axis interpretation rather than an independently certified correction of physical timing. The original specification and nominal-index feature cache remain archived. An exploratory sensitivity analysis, specified after the primary outcomes were available, evaluates the nominal-index convention with the unchanged fitting and selection rules on the same units (Section~\ref{sec:external-sampling-sensitivity-v32}); it is not independent confirmation.

\subsubsection{Adaptive representations and matched controls}
Each window is filtered between 1 and 20 kHz, converted to its Hilbert envelope, and resampled to 1024 Hz. The adaptive frontend extracts two envelope modes over 10--400 Hz using the released KMD frequency-energy calculation and a fixed automatic ridge rule. All 360 windows return two modes.

The signed representation contains four central-anchor quadratures, two log median ridge frequencies, log envelope mean, and log raw RMS. The quadrature covariance is scaled by a robust envelope-noise estimate computed only from fitting-source windows. The remaining descriptors are treated as observed covariates. Gaussian models estimate class means. The coupled model subtracts mean feature-noise covariance before positive-definite projection, selects covariance shrinkage on development units, and integrates the conditional feature covariance at prediction. Its point-prior ablation uses exactly the same selected class prior and omits this integration. Both receive the common development calibration.

The uncertainty calculation is conditional on estimated ridges. It omits uncertainty from adaptive selection and the nonlinear envelope transform; the noise estimate may also contain unresolved signal. The resulting errors-in-features likelihood is consequently a working model.

Controls include signed shrinkage QDA, OAS covariance, calibrated direct sparse QDA, quadratic logistic regression, signed-feature and energy-feature forests, a diagonal uncertainty model, and spectral SVM/forest classifiers. The forests using mode features also receive covariance-quality summaries. The spectral controls use 32 raw-spectrum and 32 envelope-spectrum log-power bins plus five amplitude/shape summaries from the same half-second observation. Tunable families use four declared candidates, the same fitting and development units, and the same temperature/intercept calibration family. Candidate selection uses mean development bearing--condition NLL. The direct sparse-QDA comparator credits the published binary estimator and declares its multiclass wrapper; optimizer residuals are saved.

\subsubsection{Held-out operating-condition results}
Window log scores are averaged within bearing and condition before normalization. Table~\ref{tab:paderborn-v32} reports all fitted methods. Accuracy and Brier scores, confusion matrices, candidate scores, every window prediction, and every bearing prediction are retained in the machine-readable results.

\begin{table}[htbp]
\centering
\small
\caption{Recorded-data results on fifteen grouped physical bearings per condition. Low speed is the unseen operating condition. The last row is a policy selecting one control separately in each fold using source-development NLL; it is not the control with lowest realized test loss. A uniform three-class probability vector has NLL $\log 3=1.099$, below every learned method at the unseen speed.}
\label{tab:paderborn-v32}
\setlength{\tabcolsep}{4pt}
\begin{tabular}{lrrrr}
\toprule
& \multicolumn{3}{c}{Bearing NLL $\downarrow$} & Held-out \\
Method & Nominal & Low torque & Low speed & accuracy \\
\midrule
Coupled sKMD & 1.548 & 1.700 & 10.346 & 5/15 \\
Same-prior point & 1.392 & 1.542 & 8.465 & 5/15 \\
Diagonal uncertainty & 2.520 & 2.417 & 7.850 & 5/15 \\
Signed shrinkage QDA & 1.443 & 1.509 & 9.181 & 6/15 \\
Signed OAS & 2.251 & 2.218 & 5.709 & 3/15 \\
Direct sparse QDA & 2.151 & 1.908 & 10.449 & 6/15 \\
Energy/quality forest & 1.841 & 2.201 & 3.244 & 5/15 \\
Signed/quality forest & 2.576 & 2.784 & 10.213 & 5/15 \\
Quadratic logistic & 0.960 & 3.442 & 6.000 & 5/15 \\
Spectral SVM & 2.417 & 2.111 & 2.638 & 5/15 \\
Spectral forest & 4.863 & 5.290 & 10.145 & 6/15 \\
Source-development control & 3.598 & 3.695 & 16.064 & 7/15 \\
\bottomrule
\end{tabular}
\end{table}

The prespecified external-success criterion requires at least $0.02$ nats improvement over the source-development-selected control and a positive lower descriptive paired interval. The observed difference, control minus sKMD, is $5.718$ nats, but its stratified bearing-bootstrap 95\% interval is $[-6.192,21.834]$. The criterion is not met. The bootstrap resamples five bearings within each class, so each stratum has only $\binom{9}{5}=126$ distinct resamples ($126^3\approx2.0\times10^6$ joint resamples over the three classes, sampled by $10{,}000$ draws); as for the six-panel studies, more Monte Carlo draws do not add information. The fixed spectral SVM and energy/quality forest have substantially lower held-out loss than sKMD. The source-development selection policy chose sparse QDA in three folds and one forest in each remaining fold; its large held-out loss reflects poor transfer of this selection policy.

Integrating the conditional covariance also fails to improve the same-prior point ablation: point minus sKMD loss is $-1.880$ nats, with descriptive interval $[-5.324,0.011]$. NLL was computed directly with log-sum-exp; all log probabilities are finite and none of the aggregated probabilities underflow to zero. The worst sKMD bearing contributes $116.197$ nats, showing severe overconfidence under the speed change rather than a numerical overflow.

The three development bearings per fold provide little independent calibration information, and class-conditional feature distributions change with speed. The bootstrap resamples the five bearings in each class, conditional on the overlapping fitted models, so its interval is descriptive and does not account for training-set uncertainty. The same fifteen bearings provide development data for the order-normalized model; separate physical bearings provide independent evaluation. No clean latent vibration reference is available for this baseline, so its outcomes concern classification rather than clean-signal reconstruction or coverage.


\subsubsection{Exploratory sensitivity to the sampling axis}
\label{sec:external-sampling-sensitivity-v32}
The nominal $64$\,kHz sampling rate is given by the dataset documentation.
The archived vibration channel also points through \texttt{XIndex} to a
monotone \texttt{X} vector. Across the $90$ recordings, that vector has blank
\texttt{Name} and \texttt{Unit} metadata and \texttt{Raster=HostService}.
Its endpoint-derived rates range from $63999.219$ to $69509.507$ per presumed
second; six records differ from $64000$ by more than $1\%$.
Consequently, the amended interpolation treats this linked axis as time;
it is not an independently certified correction of the physical clock.
The originally frozen nominal-index convention is a necessary sensitivity
check against that interpretation.

We evaluated the preserved $360$ nominal-index feature windows with the
identical grouped splits, fitting functions, candidate grids, calibration
family and source-development comparator-selection rule. This exploratory check was specified after the primary outcomes were available. It reuses the same physical bearings and is not independent confirmation.
No test result changes a setting, and neither convention is selected by its
performance. Table~\ref{tab:external-sampling-sensitivity-v32} reports both
held-speed outcomes. Every method's accuracy, Brier score, other operating
conditions and selected settings are retained in the numerical package.

\begin{table}[htbp]
\centering\small
\caption{Held-speed bearing-mean NLL under the two sampling conventions.
Smaller is better. The control policy is selected on source development
bearings separately under each convention using the same selection rule.}
\label{tab:external-sampling-sensitivity-v32}
\begin{tabular}{lrr}
\toprule
Method & Linked-axis interpolation & Nominal indexing\\
\midrule
Coupled sKMD & 10.346 & 11.722\\
Same-prior point & 8.465 & 9.591\\
Diagonal uncertainty & 7.850 & 8.888\\
Signed shrinkage QDA & 9.181 & 10.511\\
Signed OAS QDA & 5.709 & 5.836\\
Calibrated direct sparse QDA & 10.449 & 13.228\\
Energy and quality forest & 3.244 & 3.064\\
Signed and quality forest & 10.213 & 7.792\\
Quadratic logistic & 6.000 & 12.607\\
Spectral SVM & 2.638 & 2.065\\
Spectral forest & 10.145 & 11.608\\
Development-selected control & 16.064 & 11.983\\
Uniform reference & 1.099 & 1.099\\
\bottomrule
\end{tabular}
\end{table}

The nominal-index control-minus-sKMD NLL difference is $0.2602$, with
bearing-bootstrap descriptive $95\%$ interval $[-15.2636,13.7195]$.
The prespecified improvement criterion is not met under either convention.
The nominal same-prior point difference is $-2.1312$
($[-5.9390,-0.0487]$), favoring the point control.
Every learned held-speed method remains worse than the uniform NLL
$\log 3$. Exact losses and rankings depend on preprocessing; the negative
external finding holds under both conventions.

\subsection{Order-normalized bearing models and observation quality}
\label{sec:paderborn_repair_v32}

The fifteen physical bearings in the absolute-frequency Paderborn experiment,
including their 900\,rpm observations, supply development data. Independent
evaluation uses a separate cohort to distinguish native discrimination from a conditional
benefit of uncertainty integration under known observation degradation.

\subsubsection{Order coordinates and development experiments}

The absolute log ridge frequencies shift by approximately $-0.49$
between 1500 and 900\,rpm, close to $\log(900/1500)=-0.511$. The files also
contain measured shaft speed. We therefore form the vibration envelope first,
map time to shaft revolutions using that channel, and retain six revolutions.
The angular footprint lasts approximately $0.24$\,s at 1500\,rpm and $0.40$\,s
at 900\,rpm; all compared methods receive that same footprint. Dense angular
interpolation followed by antialiased polyphase reduction prevents high-order
envelope content from aliasing into the selected order band. This follows established envelope/order processing
\cite{guo_envelope_order_2012,lessmeier_bearing_2016}. Nominal 64\,kHz vibration
indexing is the declared convention; the uncertainty about the linked
HostService time axis remains a limitation.

Three development pilots compare representations and calibration. First, a model using order coordinates, phase-invariant features, and calibration
using four KMD modal amplitudes and orders reduces the source-trained 900\,rpm NLL from approximately
$10.35$ to $1.086$, but the matched order-spectrum forest is better ($0.967$).
Second, fixed Fourier-order kernels and bearing-specific mixture components
avoid inconsistent ridge identities and allow heterogeneous class
populations. Their sixteen broad order bands nevertheless discard narrow
spectral structure, and a residual-power proxy does not provide calibrated
sensor-noise uncertainty. The coupled mixture remains worse than the full
spectrum forest. The plug-in power-variance rule in that pilot is explicitly
conservative: substituting observed energy for latent energy adds
$4r\sigma^4$ in expectation to the Gaussian energy-variance formula. It is
not an exact deconvolution experiment.

The third pilot retains all 64 order-spectrum bins and uses actual repeated
recordings. Let $u_1,u_2$ be the two recording-level mean power vectors. Before
the logarithm, the covariance estimator of their average is
\begin{equation}
 \widehat R_{\mathrm{record}}
 =\frac14(u_1-u_2)(u_1-u_2)^\top.
 \label{eq:paderborn_record_cov_v32}
\end{equation}
It is unbiased if the two recording means are independent and identically
distributed with finite covariance. A delta-method Jacobian transfers it to
log-power coordinates. There are only two records: the estimate has rank one,
and its Gaussian interpretation and covariance subtraction remain working
approximations. A nonnegative integration multiplier includes $\lambda=0$.
Covariance regularization and probability calibration are selected from
out-of-fold predictions on physical bearings. A Student model supplies an
established heavy-tail control \cite{peel_mclachlan_tmixture_2000}. No pilot
shows a consistent native-quality integration advantage.

\subsubsection{Independent cohort and acquisition protocol}

The independent cohort contains all fourteen bearings excluded from development with an
unambiguous healthy, outer-race, or inner-race target: K006; KA01, KA03, KA05,
KA06, KA07, KA08, KA09; and KI01, KI03, KI05, KI07, KI08, KI17. Mixed-location
KB23, KB24, and KB27 do not belong to this three-class target. The cohort
contains one healthy bearing, seven artificial outer defects, five artificial
inner defects, and one real inner defect. It therefore cannot establish
performance in a balanced population of naturally damaged or newly healthy
bearings. Reported classification macro NLL gives each class equal weight;
micro NLL gives each observed physical bearing equal weight. Both are shown.

We use the two specified recordings at each of the three specified operating
conditions. One official file, \texttt{N15\_M01\_F10\_KA08\_2.mat}, is corrupt:
its zero-filled tail begins within vibration data at approximately $2.983$\,s,
before the last required window. A second download reproduces exactly the
same archive and file hashes. Before any evaluation prediction, the file is
replaced by the next chronological complete recording, number 3, based solely
on data integrity. The provenance record includes the corrupt file; every
other window is fixed, and all fourteen bearings are included.
All 84 selected complete records pass parsing, channel, finite-value, and
required-footprint checks.

Each native decision uses the eight windows in one bearing-condition group.
The fifteen development bearings provide five physical-unit folds for candidate and
calibration selection; final parameters are fitted only to those development units.
Window-spectrum forests provide strong controls, with window log
scores averaged at the same bearing-condition target. Forest fitting uses
window spectra.

\begin{table}[htbp]
\centering\small
\caption{Independent-cohort native bearing results: macro NLL gives equal weight to the three classes. Every column averages conditions within the one-healthy/seven-outer/six-inner cohort. The uncertainty selector chooses $\lambda=0$ using development data.}
\label{tab:paderborn_repair_native_v32}
\begin{tabular}{lrrr}\toprule
Method & Macro NLL & Micro NLL & Macro accuracy\\\midrule
Replicate selector ($\lambda=0$) & 0.655 & 0.889 & 0.759\\
Same-prior point & 0.655 & 0.889 & 0.759\\
Point QDA & 0.831 & 1.084 & 0.638\\
Student & 1.040 & 1.412 & 0.622\\
Mean-spectrum forest & 0.970 & 1.031 & 0.698\\
Window-spectrum forest & 0.613 & 0.812 & 0.728\\
Window-spectrum + quality & 0.619 & 0.823 & 0.712\\
\bottomrule\end{tabular}
\end{table}

The native classifier's development-data selector chooses $\lambda=0$, so
its predictions exactly match its same-prior point ablation
(Table~\ref{tab:paderborn_repair_native_v32}).
The window-spectrum forest has lower macro NLL ($0.613$ versus $0.655$).
The development-selected model therefore uses the point prior at native
quality; these results show no native uncertainty-integration gain. On the twelve artificial defects
alone, native NLL is $0.984$ for the selector and $0.879$ for the window
forest; their accuracies are $61.1\%$ and $55.6\%$, respectively. The single
healthy and single real-inner cases are reported separately in the supplied
stratum tables, without treating either as a population sample.

\subsubsection{Linear telemetry experiment}

To test observation-quality dependence directly, the upstream six-revolution
envelope is normalized and projected onto a fixed real Fourier bank
$H\in\mathbb R^{192\times129}$. Its columns are DC and 64 sine/cosine pairs,
with order spacing $1/6$ and cutoff $10.667$. The reconstructed target is
explicitly the resulting band-limited envelope $Ha$, not the original full
vibration record. No unit-specific fault-harmonic coverage is guaranteed.
Transmission observes
\begin{equation}
 y=H_O a+\epsilon,\qquad
 \epsilon\sim N(0,\sigma^2I),
 \label{eq:paderborn_telemetry_v32}
\end{equation}
with uniformly selected sample locations $O$. Moderate quality retains 96 of
192 samples with $\sigma=0.25$; severe quality retains 48 with $\sigma=0.5$.
Every method receives the same mask, samples, and known noise variance.
These are engineered degradations after envelope preprocessing and
projection, not estimates of the native sensor noise.

A class prior mixes Gaussian components attached to training-bearing
prototypes. Each component has a DC mean and phase-invariant diagonal
covariance: the sine and cosine variances of a frequency pair are equal.
Four shrinkage choices toward the class-average variance are selected on development
physical-bearing folds. Gaussian component evidence and conditional means
under~\eqref{eq:paderborn_telemetry_v32} are exact for the fitted model. The
Gaussian priors themselves are working approximations to real envelopes.
The component laws use diagonal latent covariance; this
experiment assesses observation integration under that specification.

The classification head averages per-window marginalized log likelihoods and
then applies a development-data temperature and uniform-probability calibration.
It is a calibrated composite predictor, not the exact eight-window joint
Bayes posterior. Reconstruction combines its calibrated class weights with
component-conditional means. Three independent corruption draws are evaluated as
three separate eight-window prediction runs; losses are averaged afterward.
The draws never provide three observations of one latent window to a single
classifier. Windows and draws are not independent physical test units.

Controls include pooled Gaussian KMD, independently selected point-prior
classification, the identical-prior point ablation, and a forest given the
complete reference-posterior mean and covariance, modal powers, mask, and
noise level. Point and forest probabilities also weight the same conditional
means. Two stronger reconstruction controls use an unlabeled mixture over
training bearings, with variance shrinkage toward the all-bearing average
selected by development out-of-fold reconstruction error. The local version uses one
window; the group version has the same eight-window information as the
classifier and marginalizes a common prototype. These controls separate
class-label information from the benefit of a richer prior.

\begin{table}[htbp]
\centering\small
\caption{Independent-cohort telemetry classification, macro NLL. Moderate quality retains 96/192 samples with noise SD 0.25; severe quality retains 48 with SD 0.5. Each of three corruption draws is evaluated separately.}
\label{tab:paderborn_repair_telemetry_v32}
\begin{tabular}{lrrr}\toprule
Method & Moderate & Severe & Mean\\\midrule
Integrated sKMD & 0.797 & 0.831 & 0.814\\
Point prior & 0.774 & 1.026 & 0.900\\
Full-posterior forest & 1.043 & 1.065 & 1.054\\
\bottomrule\end{tabular}
\end{table}

\begin{table}[htbp]
\centering\small
\caption{Independent-cohort telemetry reconstruction of the declared band-limited envelope. MSE averages windows and draws within each physical bearing, then gives the 14 bearings equal weight. The unlabeled group mixture is the strongest reconstruction control.}
\label{tab:paderborn_repair_reconstruction_v32}
\begin{tabular}{lrrr}\toprule
Method & Moderate & Severe & Mean\\\midrule
Integrated sKMD & 0.04212 & 0.08403 & 0.06307\\
Point prior & 0.04293 & 0.08971 & 0.06632\\
Full-posterior forest & 0.04861 & 0.09323 & 0.07092\\
Pooled Gaussian KMD & 0.04946 & 0.09754 & 0.07350\\
Unlabeled local mixture & 0.04272 & 0.08434 & 0.06353\\
Unlabeled group mixture & 0.04215 & 0.08094 & 0.06154\\
\bottomrule\end{tabular}
\end{table}

Under severe transmission degradation, integrated classification improves
macro NLL from $1.026$ for the point-prior head to $0.831$
(Table~\ref{tab:paderborn_repair_telemetry_v32}). Over all fourteen
physical bearings, including the single healthy unit, the mean paired
micro-NLL difference (integrated minus point prior) is $-0.184$ nats, with
descriptive paired $t$ interval $[-0.272,-0.096]$; this is the primary
descriptive interval. The thirteen-bearing summary, which excludes the
single healthy unit, was not prespecified: the frozen confirmation protocol
(\path{numerics/science/docs/paderborn_repair_confirmation_v32.frozen.md},
line 9)
only notes what an interval on the thirteen damaged units would and would
not cover, and the fourteen-, thirteen- and twelve-bearing scopes were
computed side by side after scoring. As a sensitivity analysis, the
thirteen damaged bearings give $-0.180$ $[-0.275,-0.085]$. The analogous
comparison with the full-posterior forest has an interval crossing zero;
there is no statistically established superiority over every comparator.
At moderate quality, the point head has lower macro NLL ($0.774$ versus
$0.797$). The quality-averaged paired difference is also uncertain. These
results support a conditional benefit under the severe known degradation,
not universal improvement at native or moderate quality.

\begin{figure}[htbp]
 \centering
 \includegraphics[width=\linewidth]{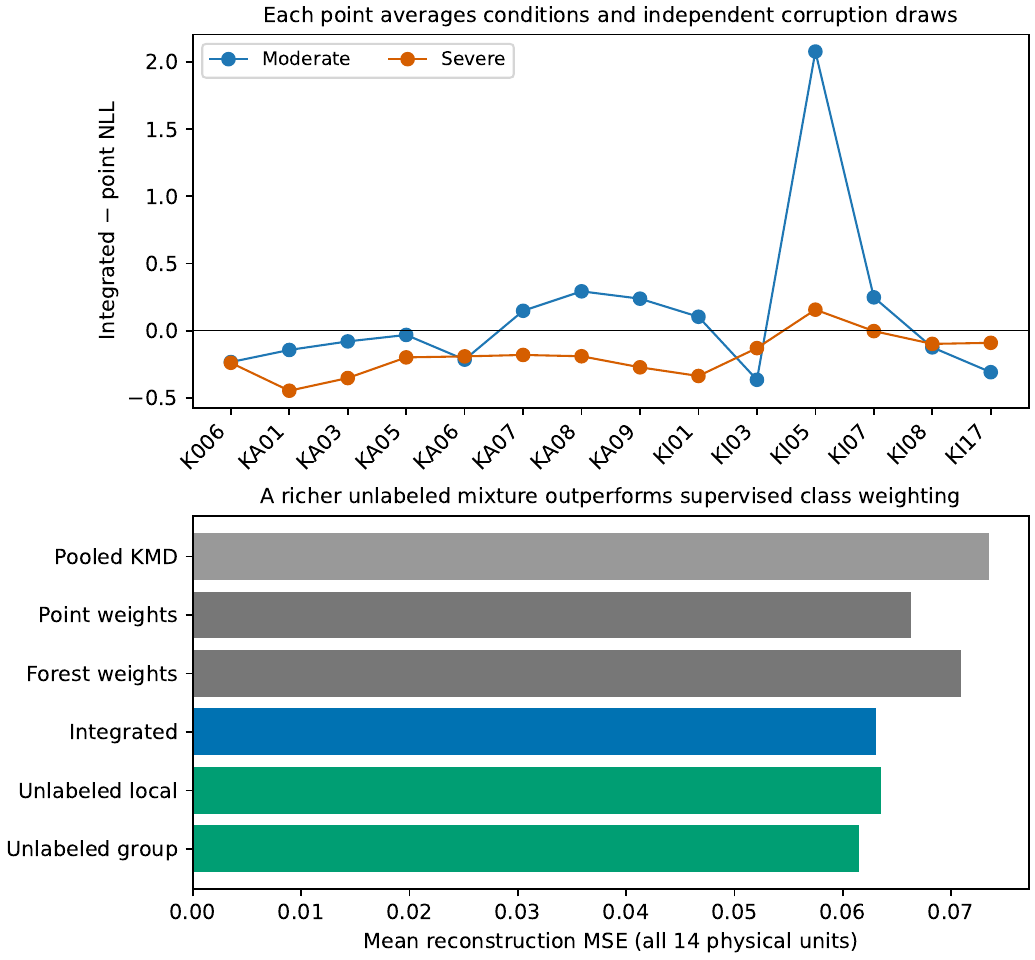}
 \caption{Independent-cohort bearing results with all physical units retained. Top:
 integrated minus point-prior NLL, separately for each known transmission
 quality; lower values favor integration. Each point averages conditions and
 separately evaluated corruption draws within one bearing. Bottom: cohort
 reconstruction MSE. The unlabeled group mixture outperforms class-weighted
 integration. The test cohort includes one healthy and one real-damaged bearing.}
 \label{fig:paderborn_repair_v32}
\end{figure}

Integrated reconstruction reduces cohort-mean MSE from $0.07350$ for pooled
Gaussian KMD to $0.06307$ (Table~\ref{tab:paderborn_repair_reconstruction_v32}). However, the unlabeled group mixture is better at
$0.06154$. Its advantage persists in the paired-unit sensitivity analysis.
The integrated head's severe-quality MSE is about $6.3\%$ below point-weighted
reconstruction, but the interval for the prespecified $5\%$ improvement
threshold crosses zero. The strongest reconstruction control has lower
MSE. Figure~\ref{fig:paderborn_repair_v32} displays the quality-dependent
classification effect and the reconstruction comparisons.

\FloatBarrier

\section{Additional controlled benchmarks}
\label{sec:controlled_benchmarks}
These benchmarks distinguish the information content of a prescribed summary
from the performance of a classifier fitted with finite data. The complete grids report oracle, fitted-classifier and reconstruction
performance separately.
\subsection{Exact energy-oracle comparisons for three Gaussian regimes}
\label{sec:energy_oracle_v31}

A comparison with a fitted energy classifier can confound representation loss with estimation error. We therefore evaluate the best possible class posterior available from each declared energy representation when the coefficient laws are known. This tests the multiclass obstruction directly. Let $C\in\mathbb R^2$ have three equally likely zero-mean Gaussian laws with precision matrices
\begin{equation}
 P_0=I,\qquad P_1=I+\alpha\begin{pmatrix}1&0\\0&-1\end{pmatrix},\qquad
 P_2=I+\alpha\begin{pmatrix}0&1\\1&0\end{pmatrix},
 \quad\alpha\in\{0.5,0.8,0.95\}.
 \label{eq:energy_oracle_laws_v31}
\end{equation}
All covariances $P_k^{-1}$ are positive definite. This experiment observes the coefficient vector directly; it isolates information in its summary and introduces no reconstruction, parameter-estimation or observation-noise error.

\paragraph{The exact posterior from two energies.}
For an invertible real basis $B$, observe $E=((BC)_1^2,(BC)_2^2)$. Put
$J_k=B^{-\top}P_kB^{-1}$, with entries $J_{k,11}=a_k$, $J_{k,12}=b_k$ and $J_{k,22}=d_k$. Summing the Gaussian density over the four sign preimages gives the class log score, up to terms common to all classes,
\begin{equation}
 \ell_k^E(e)=\tfrac12\log\det P_k
       -\tfrac12(a_ke_1+d_ke_2)
       +\log\cosh\!\left(b_k\sqrt{e_1e_2}\right).
 \label{eq:energy_oracle_score_v31}
\end{equation}
The factor $|\det B|^{-1}$ and the square-map Jacobian cancel when normalizing over classes. The formula applies away from zero coordinates, which have probability zero. We evaluate $\log\cosh t=\operatorname{logaddexp}(t,-t)-\log2$ to avoid overflow. Thus each reported energy score is its exact Bayes posterior, including the discarded-sign mixture. The class scores are evaluated analytically from the Gaussian laws.

\paragraph{Independent development and test samples.}
For every $\alpha$, eight independent panels each generate $6000$ balanced development coefficients and $15000$ balanced test coefficients from separate seeds. The full-coefficient oracle uses the known Gaussian likelihoods. The fixed-energy oracle takes $B=I$. A second energy oracle chooses one orthogonal rotation by minimum development NLL on a fixed $361$-angle grid in $[0,\pi/2)$; its test set is used once after that choice. The rotation is selected by development NLL on this grid; accuracy is evaluated separately.

A redundant control adds the third energy $E_3=((C_1+C_2)/\sqrt2)^2$ to $E_1=C_1^2,E_2=C_2^2$. It recovers $C_1C_2=E_3-(E_1+E_2)/2$, and hence every Gaussian discriminant in \eqref{eq:energy_oracle_laws_v31}. These are additional summaries of the fixed coefficient vector. Changing a KMD kernel dictionary also changes its reconstruction operator, whereas this control postprocesses fixed coordinates.

Finally, the fixed nonorthogonal bases $T_\epsilon$ of
\eqref{eq:ill_conditioned_energy_v31},
\[
 T_\epsilon=\begin{pmatrix}1&\epsilon\\1&-\epsilon\end{pmatrix},
 \qquad \epsilon\in\{1,0.3,0.1,0.03\},
\]
have condition number $1/\epsilon$. They probe the distinction between a positive information gap for each fixed invertible basis and a gap uniform over a conditioning-controlled family. Each basis is evaluated under every specified test condition.

\begingroup\small\setlength{\tabcolsep}{3.5pt}
\begin{longtable}{@{}rlrrr@{}}
\caption{Known-law Bayes comparisons, mean $\pm$ SD over eight independent panels at each $\alpha$. Accuracy is larger-is-better; natural-log NLL and multiclass Brier $\sum_k(p_k-\mathbf1_{\{Y=k\}})^2$ are smaller-is-better. Three energies are redundant postprocessing of the same coefficients.}
\label{tab:energy_oracle_v31}\\
\toprule $\alpha$ & Representation & Accuracy & NLL & Brier\\\midrule\endfirsthead
\toprule $\alpha$ & Representation & Accuracy & NLL & Brier\\\midrule\endhead
\bottomrule\endfoot
0.50 & Full coefficients & $0.4457\pm0.0025$ & $1.0327\pm0.0024$ & $0.6231\pm0.0012$ \\
0.50 & Fixed energies & $0.4181\pm0.0016$ & $1.0545\pm0.0019$ & $0.6373\pm0.0011$ \\
0.50 & Selected rotation & $0.4282\pm0.0039$ & $1.0449\pm0.0025$ & $0.6311\pm0.0015$ \\
0.50 & Three energies & $0.4457\pm0.0025$ & $1.0327\pm0.0024$ & $0.6231\pm0.0012$ \\
0.50 & $T_{1}$ & $0.4159\pm0.0013$ & $1.0557\pm0.0019$ & $0.6382\pm0.0012$ \\
0.50 & $T_{0.3}$ & $0.4256\pm0.0028$ & $1.0454\pm0.0020$ & $0.6320\pm0.0011$ \\
0.50 & $T_{0.1}$ & $0.4415\pm0.0013$ & $1.0361\pm0.0021$ & $0.6254\pm0.0011$ \\
0.50 & $T_{0.03}$ & $0.4445\pm0.0026$ & $1.0336\pm0.0023$ & $0.6237\pm0.0012$ \\
0.80 & Full coefficients & $0.5792\pm0.0040$ & $0.8580\pm0.0021$ & $0.5131\pm0.0015$ \\
0.80 & Fixed energies & $0.5505\pm0.0029$ & $0.8996\pm0.0023$ & $0.5391\pm0.0014$ \\
0.80 & Selected rotation & $0.5564\pm0.0037$ & $0.8842\pm0.0034$ & $0.5303\pm0.0021$ \\
0.80 & Three energies & $0.5792\pm0.0040$ & $0.8580\pm0.0021$ & $0.5131\pm0.0015$ \\
0.80 & $T_{1}$ & $0.5506\pm0.0024$ & $0.8996\pm0.0024$ & $0.5388\pm0.0014$ \\
0.80 & $T_{0.3}$ & $0.5195\pm0.0042$ & $0.9289\pm0.0022$ & $0.5600\pm0.0016$ \\
0.80 & $T_{0.1}$ & $0.5655\pm0.0041$ & $0.8767\pm0.0029$ & $0.5249\pm0.0019$ \\
0.80 & $T_{0.03}$ & $0.5754\pm0.0040$ & $0.8636\pm0.0022$ & $0.5166\pm0.0016$ \\
0.95 & Full coefficients & $0.7433\pm0.0021$ & $0.5871\pm0.0040$ & $0.3459\pm0.0025$ \\
0.95 & Fixed energies & $0.7272\pm0.0025$ & $0.6228\pm0.0039$ & $0.3671\pm0.0024$ \\
0.95 & Selected rotation & $0.7265\pm0.0027$ & $0.6145\pm0.0046$ & $0.3643\pm0.0029$ \\
0.95 & Three energies & $0.7433\pm0.0021$ & $0.5871\pm0.0040$ & $0.3459\pm0.0025$ \\
0.95 & $T_{1}$ & $0.7270\pm0.0030$ & $0.6228\pm0.0031$ & $0.3673\pm0.0024$ \\
0.95 & $T_{0.3}$ & $0.6212\pm0.0031$ & $0.7868\pm0.0036$ & $0.4748\pm0.0025$ \\
0.95 & $T_{0.1}$ & $0.7078\pm0.0031$ & $0.6427\pm0.0042$ & $0.3813\pm0.0027$ \\
0.95 & $T_{0.03}$ & $0.7334\pm0.0029$ & $0.6026\pm0.0037$ & $0.3558\pm0.0024$ \\

\end{longtable}\endgroup

Tables~\ref{tab:energy_oracle_v31} and~\ref{tab:energy_oracle_paired_v31} report every representation. Full information improves both proper losses over the selected orthogonal-energy representation at all three declared parameters. The mean NLL gaps are $0.01223$, $0.02619$ and $0.02746$, with paired $t$ $95\%$ intervals (eight panels, 7 df) $[0.01150,0.01297]$, $[0.02486,0.02752]$ and $[0.02586,0.02907]$; the bootstrap sensitivity intervals are $[0.01162,0.01277]$, $[0.02526,0.02729]$ and $[0.02625,0.02870]$. The corresponding Brier gaps are $0.00803$, $0.01726$ and $0.01843$. Full-coefficient accuracy exceeds selected-rotation accuracy by $0.01753$, $0.02280$ and $0.01682$ ($t$ intervals $[0.01589,0.01917]$, $[0.01991,0.02569]$ and $[0.01476,0.01887]$). Every one of the 42 accuracy and NLL $t$ intervals in Table~\ref{tab:energy_oracle_paired_v31} lies on the same side of zero as its bootstrap interval (the three-energy rows are zero to numerical precision under both), so no conclusion changes. Accuracy differences are evaluated empirically alongside the proper-score comparisons.

At $\alpha=0.95$, the NLL-selected rotation has slightly lower accuracy than the fixed-energy representation, despite its better NLL. The redundant three-energy posterior matches the full posterior to numerical precision. Together, these comparisons isolate the information loss of the specified nonredundant energy summary.

\begingroup\small\setlength{\tabcolsep}{3pt}
\begin{longtable}{@{}rlrrrr@{}}
\caption{Paired mean gaps over the eight panels. The third and fourth columns give each mean gap with its $95\%$ percentile-bootstrap interval from $20000$ resamples, reported as a sensitivity summary; the last two columns give the paired $t$ $95\%$ intervals (7 df), the reporting intervals. Accuracy gap is full minus energy; NLL gap is energy minus full, so positive values favor full information. Paired Brier intervals are supplied with the numerical results.}
\label{tab:energy_oracle_paired_v31}\\
\toprule $\alpha$ & Representation & Accuracy gap & NLL gap & Accuracy, $t$ & NLL, $t$\\\midrule\endfirsthead
\toprule $\alpha$ & Representation & Accuracy gap & NLL gap & Accuracy, $t$ & NLL, $t$\\\midrule\endhead
\bottomrule\endfoot
\input{tables/energy_oracle_v31_paired_t_rows}

\end{longtable}\endgroup

\begin{figure}[htbp]
\centering\includegraphics[width=0.97\linewidth]{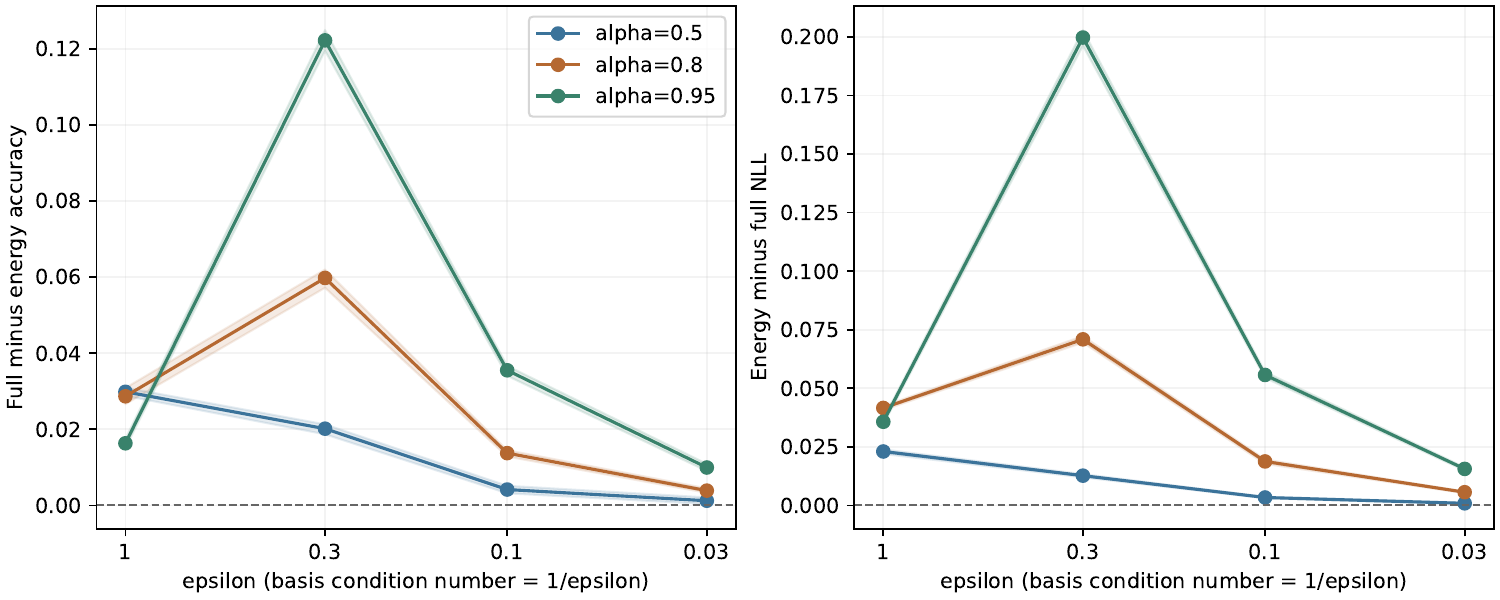}
\caption{Estimated energy-oracle gaps under increasingly ill-conditioned bases. Shading gives paired $95\%$ intervals across eight independent panels. The finite sequence need not be monotone. A restriction to orthogonal or conditioning-controlled bases is substantive.}
\label{fig:energy_oracle_v31}
\end{figure}

\paragraph{Dependence on basis conditioning.}
For $T_\epsilon$, the two energies determine both $C_1C_2$ and $C_1^2+\epsilon^2C_2^2$. The alternative coefficient preimage, apart from a simultaneous sign change, is $(\epsilon C_2,C_1/\epsilon)$. As $\epsilon$ decreases, that alternative typically lies in a Gaussian tail. The energy posterior can then approach the full posterior even though every fixed $\epsilon>0$ remains insufficient. At $\epsilon=0.03$, the observed NLL gaps are $0.00088$, $0.00563$ and $0.01558$, smaller than the selected orthogonal gaps. The dependence is not monotone over this finite set of bases: $\epsilon=0.3$ is worse than $1$ at the larger $\alpha$ values (Figure~\ref{fig:energy_oracle_v31}). These measured gaps are the reference values against which Table~\ref{tab:bound_vs_gap} compares the explicit lower bounds of Theorems~\ref{thm:energy_uniform_gap_v31} and~\ref{thm:conditioned_basis_gap}.

These comparisons establish a measurable information cost of the stated summaries under known Gaussian laws.

\subsection{Finite-sample coupled-mode benchmark}
\label{sec:interaction_experiments_v31}
This benchmark uses the algorithm in Section~\ref{sec:skmd_method} with
supplied or independently fitted carriers. It measures the finite-sample
cost of estimating the coupled law and compares it with signed controls,
energy summaries and pooled-prior reconstruction.

\subsubsection{Complete protocol and comparators}

Use the three precisions in \eqref{eq:three_class_precision_v31}, with all $\alpha\in\{0.5,0.8,0.95\}$. Two orthonormal carriers on $p=64$ samples are a cosine at three cycles per window and a sine at seven cycles. Latent coefficients have covariance $P_k^{-1}-0.04I_2$; independent nominal observation noise has standard deviation $0.2$. Equal class priors and zero means are known parts of the construction. Compare the supplied carriers with the first two right singular vectors of the \emph{unlabeled fitting windows}; this estimates an informative subspace, rather than identifying physical carriers individually; no development or test vector enters that basis fit.

For each $\alpha$, use $n=30$ and $120$ independent fitting windows per class, eight independent panels, 300 separate development windows per class, and 1,000 test windows per class. Three observation conditions are evaluated:
\begin{enumerate}[label=\arabic*,start=0,leftmargin=*]
\item complete observations with noise standard deviation $0.2$;
\item independently retain each coordinate with probability $0.5$, with noise standard deviation $0.2$;
\item use the same missingness law and draw each window's noise standard deviation uniformly on $[0.2,0.8]$.
\end{enumerate}
Masks and noise levels are known metadata and independent of class. The same complete fitting sample is reused across the three conditions within a panel; panels are independent within each condition. The 144 condition--panel evaluations share fitted models across conditions. The total labeled budget includes both the $n$ fitting observations and the development sample. The covariance eigenvalue floor is $0.02$, and off-diagonal shrinkage candidates are $0,0.1,0.3$, selected by development multiclass negative log likelihood (NLL). Set $\Gamma$ to the diagonal of the mean fitted class covariance.

Point-feature competitors receive the same reference posterior means and observation-quality metadata $(V_{11},V_{22},V_{12},\sigma_o,\text{observed fraction})$ where specified. In the two incomplete conditions, their training sample contains five augmented views per original fitting record. Added noise has variance $\sigma_o^2-0.04$, accounting for noise already present; independent masks then remove coordinates. The augmented views remain grouped by their source observation for uncertainty estimation. All methods use the same development and test units. The main controls are:
\begin{itemize}[leftmargin=*]
\item signed-point QDA and OAS covariance shrinkage, with empirical means and ridge $0.0035$;
\item logistic regression on $m_1,m_2,m_1^2,m_2^2,m_1m_2$ and quality metadata, with $C\in\{0.1,1,10\}$;
\item the same regression with $(m_1+m_2)^2$ replacing $m_1m_2$, still including signed linear terms; this is a redundant quadratic-feature control, not an energy-only classifier;
\item a 100-tree forest on energies, normalized energies, and quality metadata, with depths $3,6,12$;
\item gradient boosting on signed coefficients and quality metadata, with 100 iterations and depths $2,4,6$;
\item raw-window OAS in condition 0, and the known-law Gaussian oracle in all conditions.
\end{itemize}
Each three-candidate family is selected by development NLL. The heterogeneous forest has more information than the two squared coordinates of the theorem; its comparison measures fitted procedure performance. The separate exact oracle experiment supplies the information-loss comparison.

\paragraph{Exploratory ablations.}
Fixed sensitivity controls, specified after the primary results were available, use the same saved units with no new tuning: zero-mean point QDA using uncentered covariances and the same $0.0035$ ridge; a class-diagonal prior retaining fitted marginal mode variances and full observation integration; and a point-prior rule evaluating the \emph{same fitted coupled prior} at $m$, omitting integration over $V$. These analyses use the same independently fitted panels as the primary comparisons. A fourth control uses one pooled full coefficient covariance to reconstruct without regime mixing. It is fitted from the same training observations, without the class-specific shrinkage selection. The zero-mean control removes an unnecessary mean-estimation disadvantage in the unrestricted-mean QDA comparison. The diagonal control removes class coupling. The point-prior control isolates what is lost by treating the reconstructed mean as exact.

\subsubsection{Probability loss and reconstruction results}

Tables~\ref{tab:interaction_supplied_v31}--\ref{tab:interaction_learned_v31} report the complete grid. NLL is the primary proper score; accuracy and multiclass Brier scores are also saved. With learned modes, sKMD has lower mean NLL than the energy-and-quality forest and the quadratic logistic control in all 18 cells (point estimates). With paired $t$ intervals over the eight panels (7 df), the reporting intervals, all 18 learned-mode intervals against the forest and 17 of 18 against quadratic logistic regression exclude zero in sKMD's favor; the exception is quadratic logistic at $\alpha=0.95$, $n=30$, complete observations, with difference $-0.0194$ and interval $[-0.0449,0.0061]$. In the bootstrap sensitivity analysis (20,000 draws over eight panels, preserving the dependence among augmented views) all 18 intervals exclude zero for both controls. With supplied modes the $t$ counts are 18 against the forest and 16 against quadratic logistic, the same as the bootstrap counts.

The matched zero-mean comparison is closer: sKMD has lower mean NLL than zero-mean point QDA in 14 of 18 learned-mode cells and 13 of 18 supplied-mode cells (point-estimate counts). Paired $t$ intervals exclude zero in sKMD's favor in only 3 learned-mode and 4 supplied-mode cells (bootstrap sensitivity analysis: 7 and 6). The bootstrap intervals that excluded zero in zero-mean QDA's favor (4 learned-mode and 2 supplied-mode cells) all include zero under the $t$ interval, so no cell is resolved in that control's favor. At $\alpha=0.8$, $n=120$, complete observations, learned-mode sKMD has NLL $0.8646$ versus $0.8632$ for zero-mean point QDA (paired difference $+0.00141$, $t$ interval $[-0.00021,0.00303]$, bootstrap $[0.00033,0.00275]$). With one fixed full observation law, signed coefficients already preserve the informative projection. Differences there concern fitting, shrinkage and noise treatment, not recovered representation information.

\begin{figure}[htbp]\centering
\includegraphics[width=\linewidth]{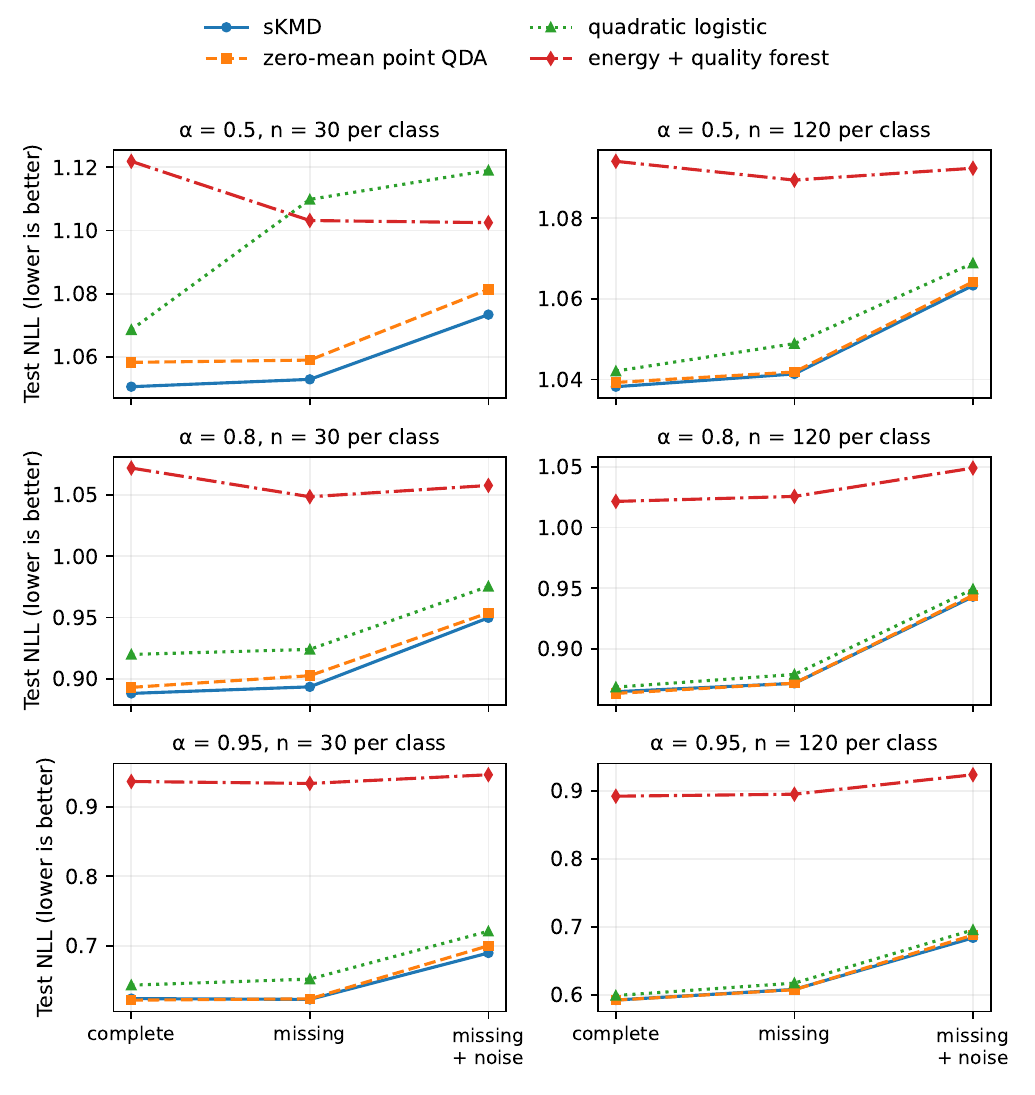}
\caption{The complete learned-mode grid. The advantage over separate energy features persists across fitting sizes and observation conditions. Zero-mean point QDA is often close, and occasionally slightly better, when the observation law is fixed. Curves join different observation conditions for readability, not a continuous parameter sweep.}
\label{fig:interaction_skmd_v31}
\end{figure}

At $\alpha=0.95$, $n=120$, missing coordinates and variable noise, learned-mode sKMD has NLL $0.6840$ (Figure~\ref{fig:interaction_skmd_v31}), compared with $0.6880$ for zero-mean point QDA, $0.6957$ for quadratic logistic regression, $0.7321$ for the point-prior rule and $0.9239$ for the energy-and-quality forest. The known-law oracle attains $0.6812$.

\begin{table}[htbp]
\centering\scriptsize\setlength{\tabcolsep}{3pt}
\caption{Test NLL with supplied carriers; lower is better. Conditions 0--2 are defined in the text. QDA/OAS estimate means; ZQ fixes zero means; Quad and Red are signed quadratic logistic controls; RF uses energies and quality; GB uses signed coefficients and quality. Every entry averages eight panels.}
\label{tab:interaction_supplied_v31}
\begin{tabular}{@{}rrr*{9}{r}@{}}\toprule
$\alpha$ & $n$ & $o$ & sKMD & QDA & OAS & ZQ & Quad & Red & RF & GB & Oracle\\\midrule
0.50 & 30 & 0 & 1.057 & 1.098 & 1.092 & 1.061 & 1.068 & 1.072 & 1.114 & 1.362 & 1.034 \\
0.50 & 30 & 1 & 1.058 & 1.100 & 1.092 & 1.061 & 1.110 & 1.108 & 1.094 & 1.276 & 1.033 \\
0.50 & 30 & 2 & 1.074 & 1.113 & 1.111 & 1.082 & 1.117 & 1.115 & 1.097 & 1.233 & 1.063 \\
0.50 & 120 & 0 & 1.038 & 1.043 & 1.042 & 1.040 & 1.042 & 1.043 & 1.076 & 1.136 & 1.033 \\
0.50 & 120 & 1 & 1.041 & 1.046 & 1.045 & 1.042 & 1.049 & 1.049 & 1.074 & 1.093 & 1.035 \\
0.50 & 120 & 2 & 1.063 & 1.067 & 1.067 & 1.064 & 1.069 & 1.069 & 1.081 & 1.102 & 1.060 \\
0.80 & 30 & 0 & 0.887 & 0.911 & 0.913 & 0.895 & 0.914 & 0.921 & 0.973 & 1.174 & 0.862 \\
0.80 & 30 & 1 & 0.894 & 0.921 & 0.908 & 0.904 & 0.919 & 0.924 & 0.970 & 1.113 & 0.871 \\
0.80 & 30 & 2 & 0.948 & 0.965 & 0.963 & 0.953 & 0.973 & 0.975 & 1.008 & 1.090 & 0.937 \\
0.80 & 120 & 0 & 0.863 & 0.866 & 0.866 & 0.863 & 0.869 & 0.869 & 0.928 & 0.952 & 0.858 \\
0.80 & 120 & 1 & 0.871 & 0.874 & 0.873 & 0.871 & 0.878 & 0.880 & 0.942 & 0.931 & 0.866 \\
0.80 & 120 & 2 & 0.942 & 0.948 & 0.948 & 0.944 & 0.949 & 0.949 & 0.992 & 0.988 & 0.940 \\
0.95 & 30 & 0 & 0.624 & 0.630 & 0.686 & 0.622 & 0.640 & 0.649 & 0.712 & 0.905 & 0.596 \\
0.95 & 30 & 1 & 0.625 & 0.636 & 0.639 & 0.624 & 0.651 & 0.660 & 0.717 & 0.888 & 0.600 \\
0.95 & 30 & 2 & 0.688 & 0.709 & 0.708 & 0.701 & 0.718 & 0.725 & 0.777 & 0.831 & 0.677 \\
0.95 & 120 & 0 & 0.593 & 0.595 & 0.606 & 0.592 & 0.599 & 0.599 & 0.669 & 0.683 & 0.587 \\
0.95 & 120 & 1 & 0.608 & 0.611 & 0.610 & 0.608 & 0.617 & 0.618 & 0.672 & 0.664 & 0.601 \\
0.95 & 120 & 2 & 0.684 & 0.692 & 0.692 & 0.689 & 0.696 & 0.696 & 0.747 & 0.736 & 0.681 \\

\bottomrule\end{tabular}
\end{table}
\begin{table}[htbp]
\centering\scriptsize\setlength{\tabcolsep}{3pt}
\caption{Test NLL with modes learned from fitting observations only. Abbreviations and replication match Table~\ref{tab:interaction_supplied_v31}. Rounding can hide small differences; saved scores retain full precision.}
\label{tab:interaction_learned_v31}
\begin{tabular}{@{}rrr*{9}{r}@{}}\toprule
$\alpha$ & $n$ & $o$ & sKMD & QDA & OAS & ZQ & Quad & Red & RF & GB & Oracle\\\midrule
0.50 & 30 & 0 & 1.051 & 1.095 & 1.090 & 1.058 & 1.069 & 1.078 & 1.122 & 1.349 & 1.034 \\
0.50 & 30 & 1 & 1.053 & 1.098 & 1.090 & 1.059 & 1.110 & 1.104 & 1.103 & 1.304 & 1.033 \\
0.50 & 30 & 2 & 1.073 & 1.113 & 1.110 & 1.081 & 1.119 & 1.117 & 1.102 & 1.235 & 1.063 \\
0.50 & 120 & 0 & 1.038 & 1.043 & 1.042 & 1.039 & 1.042 & 1.044 & 1.094 & 1.141 & 1.033 \\
0.50 & 120 & 1 & 1.041 & 1.046 & 1.045 & 1.042 & 1.049 & 1.049 & 1.089 & 1.093 & 1.035 \\
0.50 & 120 & 2 & 1.063 & 1.067 & 1.067 & 1.064 & 1.069 & 1.069 & 1.092 & 1.105 & 1.060 \\
0.80 & 30 & 0 & 0.888 & 0.910 & 0.915 & 0.893 & 0.920 & 0.930 & 1.072 & 1.169 & 0.862 \\
0.80 & 30 & 1 & 0.893 & 0.920 & 0.908 & 0.903 & 0.924 & 0.939 & 1.048 & 1.112 & 0.871 \\
0.80 & 30 & 2 & 0.950 & 0.967 & 0.965 & 0.954 & 0.975 & 0.984 & 1.058 & 1.097 & 0.937 \\
0.80 & 120 & 0 & 0.865 & 0.866 & 0.867 & 0.863 & 0.868 & 0.869 & 1.021 & 0.959 & 0.858 \\
0.80 & 120 & 1 & 0.872 & 0.874 & 0.874 & 0.872 & 0.879 & 0.880 & 1.026 & 0.929 & 0.866 \\
0.80 & 120 & 2 & 0.943 & 0.947 & 0.948 & 0.944 & 0.949 & 0.950 & 1.049 & 0.992 & 0.940 \\
0.95 & 30 & 0 & 0.624 & 0.630 & 0.686 & 0.622 & 0.643 & 0.654 & 0.937 & 0.844 & 0.596 \\
0.95 & 30 & 1 & 0.623 & 0.636 & 0.639 & 0.623 & 0.652 & 0.667 & 0.934 & 0.818 & 0.600 \\
0.95 & 30 & 2 & 0.690 & 0.708 & 0.710 & 0.700 & 0.721 & 0.728 & 0.946 & 0.823 & 0.677 \\
0.95 & 120 & 0 & 0.593 & 0.595 & 0.606 & 0.592 & 0.599 & 0.606 & 0.892 & 0.690 & 0.587 \\
0.95 & 120 & 1 & 0.608 & 0.611 & 0.610 & 0.608 & 0.617 & 0.619 & 0.895 & 0.664 & 0.601 \\
0.95 & 120 & 2 & 0.684 & 0.692 & 0.692 & 0.688 & 0.696 & 0.696 & 0.924 & 0.740 & 0.681 \\

\bottomrule\end{tabular}
\end{table}

Tables~\ref{tab:interaction_ablation_supplied_v31}--\ref{tab:interaction_ablation_learned_v31} retain the diagonal and point-prior ablations and reconstruction losses. The reference KMD uses one pooled diagonal coefficient prior; sKMD returns the regime-mixture conditional mean. With learned modes at $\alpha=0.95$, $n=120$, condition 2, mean signal MSE falls from $0.01586$ to $0.01385$, a $12.69\%$ reduction in the ratio of mean losses. Supplied modes give a $15.63\%$ reduction. In the weak-coupling complete-observation case the improvement is below $0.1\%$ at $n=30$: the coefficients are already accurately determined by the observations. The reconstruction comparison uses the stated independent-mode KMD prior. The stronger pooled full-covariance KMD control has MSE $0.01586$ in this learned-subspace case and $0.01564$ with supplied modes; sKMD reduces these mean losses by $12.69\%$ and $12.90\%$, respectively. The pooled full and diagonal learned-basis fits coincide here because the basis diagonalizes the pooled training second moment. A differently fitted dictionary remains a distinct comparison.

\begin{table}[htbp]\centering\scriptsize\setlength{\tabcolsep}{3pt}
\caption{Supplied-mode ablations and reconstruction. Diag integrates the class-diagonal prior; Point evaluates the coupled prior at the reference mean. All MSE columns are multiplied by $10^3$; Pool uses the pooled full-covariance prior; Gain is the percentage reduction in the ratio of mean MSEs.}
\label{tab:interaction_ablation_supplied_v31}
\begin{tabular}{@{}rrr*{7}{r}@{}}\toprule
$\alpha$ & $n$ & $o$ & sKMD NLL & Diag NLL & Point NLL & KMD MSE & Pool MSE & sKMD MSE & Gain\\\midrule
0.50 & 30 & 0 & 1.0567 & 1.0859 & 1.0583 & 1.208 & 1.207 & 1.207 & 0.07 \\
0.50 & 30 & 1 & 1.0584 & 1.0857 & 1.0616 & 2.454 & 2.450 & 2.450 & 0.20 \\
0.50 & 30 & 2 & 1.0742 & 1.0889 & 1.0802 & 10.965 & 10.898 & 10.885 & 0.73 \\
0.50 & 120 & 0 & 1.0384 & 1.0652 & 1.0386 & 1.206 & 1.204 & 1.203 & 0.21 \\
0.50 & 120 & 1 & 1.0409 & 1.0670 & 1.0413 & 2.433 & 2.429 & 2.426 & 0.26 \\
0.50 & 120 & 2 & 1.0634 & 1.0787 & 1.0655 & 11.047 & 10.951 & 10.924 & 1.12 \\
0.80 & 30 & 0 & 0.8875 & 0.9482 & 0.8907 & 1.219 & 1.214 & 1.210 & 0.73 \\
0.80 & 30 & 1 & 0.8944 & 0.9545 & 0.9020 & 2.484 & 2.469 & 2.449 & 1.40 \\
0.80 & 30 & 2 & 0.9478 & 0.9955 & 0.9752 & 12.973 & 12.660 & 12.292 & 5.25 \\
0.80 & 120 & 0 & 0.8632 & 0.9256 & 0.8636 & 1.230 & 1.227 & 1.221 & 0.79 \\
0.80 & 120 & 1 & 0.8709 & 0.9323 & 0.8715 & 2.498 & 2.486 & 2.458 & 1.61 \\
0.80 & 120 & 2 & 0.9423 & 0.9871 & 0.9546 & 13.144 & 12.820 & 12.462 & 5.18 \\
0.95 & 30 & 0 & 0.6237 & 0.7091 & 0.6297 & 1.251 & 1.248 & 1.227 & 1.91 \\
0.95 & 30 & 1 & 0.6250 & 0.7164 & 0.6359 & 2.584 & 2.572 & 2.474 & 4.24 \\
0.95 & 30 & 2 & 0.6883 & 0.7661 & 0.7888 & 16.225 & 15.682 & 13.586 & 16.27 \\
0.95 & 120 & 0 & 0.5926 & 0.6919 & 0.5949 & 1.233 & 1.230 & 1.207 & 2.10 \\
0.95 & 120 & 1 & 0.6082 & 0.7041 & 0.6143 & 2.552 & 2.537 & 2.438 & 4.48 \\
0.95 & 120 & 2 & 0.6839 & 0.7677 & 0.7449 & 16.147 & 15.640 & 13.623 & 15.63 \\

\bottomrule\end{tabular}\end{table}
\begin{table}[htbp]\centering\scriptsize\setlength{\tabcolsep}{3pt}
\caption{Learned-mode ablations and reconstruction, using the same conventions as Table~\ref{tab:interaction_ablation_supplied_v31}.}
\label{tab:interaction_ablation_learned_v31}
\begin{tabular}{@{}rrr*{7}{r}@{}}\toprule
$\alpha$ & $n$ & $o$ & sKMD NLL & Diag NLL & Point NLL & KMD MSE & Pool MSE & sKMD MSE & Gain\\\midrule
0.50 & 30 & 0 & 1.0506 & 1.0865 & 1.0512 & 2.089 & 2.089 & 2.089 & 0.03 \\
0.50 & 30 & 1 & 1.0529 & 1.0902 & 1.0541 & 3.357 & 3.357 & 3.356 & 0.03 \\
0.50 & 30 & 2 & 1.0734 & 1.0930 & 1.0771 & 11.656 & 11.656 & 11.645 & 0.09 \\
0.50 & 120 & 0 & 1.0382 & 1.0824 & 1.0383 & 1.432 & 1.432 & 1.431 & 0.05 \\
0.50 & 120 & 1 & 1.0414 & 1.0840 & 1.0418 & 2.667 & 2.667 & 2.664 & 0.09 \\
0.50 & 120 & 2 & 1.0633 & 1.0888 & 1.0641 & 11.163 & 11.163 & 11.138 & 0.22 \\
0.80 & 30 & 0 & 0.8881 & 1.0257 & 0.8896 & 2.136 & 2.136 & 2.131 & 0.23 \\
0.80 & 30 & 1 & 0.8935 & 1.0261 & 0.8971 & 3.433 & 3.433 & 3.410 & 0.69 \\
0.80 & 30 & 2 & 0.9497 & 1.0436 & 0.9646 & 13.525 & 13.525 & 13.171 & 2.62 \\
0.80 & 120 & 0 & 0.8646 & 1.0144 & 0.8648 & 1.436 & 1.436 & 1.430 & 0.44 \\
0.80 & 120 & 1 & 0.8715 & 1.0186 & 0.8717 & 2.698 & 2.698 & 2.670 & 1.05 \\
0.80 & 120 & 2 & 0.9430 & 1.0409 & 0.9503 & 13.014 & 13.014 & 12.663 & 2.70 \\
0.95 & 30 & 0 & 0.6237 & 0.9254 & 0.6290 & 2.248 & 2.248 & 2.227 & 0.92 \\
0.95 & 30 & 1 & 0.6226 & 0.9288 & 0.6340 & 3.618 & 3.618 & 3.517 & 2.78 \\
0.95 & 30 & 2 & 0.6895 & 0.9375 & 0.7699 & 16.724 & 16.724 & 14.627 & 12.54 \\
0.95 & 120 & 0 & 0.5926 & 0.9035 & 0.5947 & 1.448 & 1.448 & 1.425 & 1.60 \\
0.95 & 120 & 1 & 0.6082 & 0.9082 & 0.6119 & 2.758 & 2.758 & 2.658 & 3.62 \\
0.95 & 120 & 2 & 0.6840 & 0.9279 & 0.7321 & 15.860 & 15.860 & 13.848 & 12.69 \\

\bottomrule\end{tabular}\end{table}

\subsubsection{Experimental scope}

This controlled Gaussian mode-coupling problem connects the representation theorem to fitted inference. It compares KMD regime summaries and independent-mode reconstructions for both supplied and estimated dictionaries. The finite-sample theorem describes covariance estimation in known coordinates; the empirical procedure additionally estimates the basis, regularizes eigenvalues and selects tuning parameters on development data. The benchmark supplies rank two, known noise, zero means and class-independent observation metadata to all applicable methods.

\FloatBarrier

\section{Discussion}
\label{sec:discussion}
\subsection{Contribution and connections}
The sKMD construction turns a KMD reference posterior into a joint
regime-conditioned mode law. It combines a learned relation between modes
and regimes with observation uncertainty, and uses the resulting regime
posterior for classification and reconstruction. Its exact reduction to
KMD identifies the additional information supplied by the regime law.

The main contribution is the representation theory of
Part~\ref{part:theory}, which explains how observation quality changes the
information needed for inference. At fixed quality, a shared part of the
precision matrices can cancel from the decision rule
(Theorem~\ref{thm:energy_sufficiency_v31}), and a three-regime family has
no sufficient nonredundant energy basis
(Theorem~\ref{thm:three_class_obstruction_v31}). Independently varying
coordinate noise constrains the cancellation through the covariance graph
(Theorem~\ref{thm:quality_diagonal_v32}). For distinct centered covariance
laws, fully correlated noise variation requires the complete quadratic span
for a fixed linear log-odds decoder
(Theorem~\ref{thm:quality_quadratic_span_v32}). The approximation and
learning bounds connect these structural results to feature budgets,
representation error and statistical estimation. Their explicit constants
are conservative: the conditioning-controlled information bounds are
positivity statements several orders of magnitude below the numerical
estimates from exact Bayes energy posteriors (Table~\ref{tab:bound_vs_gap}).

The posterior interface itself builds on established Gaussian
conditioning, QDA, factor models, mixture inference and temporal filtering
\citep{baek_commonfactor_2010,bovy_deconvolution_2011,song_plso_2021}.
Gaussian conditioning, a matched latent-Gaussian likelihood or an identical
Gaussian HMM gives the same answer under the same parameters, so the
interface is a reformulation; its value lies in connecting these operations
to the representation results. The use of local Fourier packets,
envelope/order coordinates and principal-subspace approximation likewise
draws on established representations. Their role here is specified by the
associated KMD regime law and the matched comparisons.

\subsection{Interpreting the controlled and adaptive results}
The binary study identifies a precise connection between observation
uncertainty and calibration. Its energy posterior is uniform by symmetry,
and its noisy log odds are an exact quality-dependent rescaling of the
noiseless log odds. The prespecified gain over one globally calibrated
point rule therefore measures mainly that rule's restricted temperature
grid; quality-aware calibration removes $97.7\%$ of it. We conjecture that
the small resolved residual reflects fitting error and departure from exact
reflection symmetry in the estimated laws, but this attribution has not
been tested. This experiment supplies a mechanism example rather than
evidence that explicit integration must outperform every quality-aware
signed classifier.

The three-regime study extends the comparison to partial energy information
and continuously varying quality. A fitted estimator recovers enough signed
information to improve on the development-rotated energy oracle. The
primary gain over the selected signed classifier is $0.00246$ nats; the
retrospective centered signed control, which matches the Gaussian fit's
symmetry, gives a gain of $0.00116$ nats. These are statistically resolved
probability-score differences with small effects on geometric mean
true-class probability, and the matched comparison reverses when the noise
monitor underreports the noise variance. A conservative probability at
very high noise also need not yield useful classification accuracy.
Task-specific decision costs are needed to assign practical utility to
these score differences; the illustrative $0.02$-nat calculations are
sensitivity analyses rather than application-derived equivalence claims.

The adaptive and nonstationary studies identify useful reconstruction and
calibration settings, including heterogeneous observation quality, learned
carriers and temporal regimes. Strong signed and pooled-prior controls
identify which part of a gain comes from retained interactions, regime
conditioning or dictionary averaging. The increment over a strong signed
Gaussian head can be modest even when the contrast with separate energies
is substantial; for the adaptive oscillators, an energy-interaction logistic
classifier has lower mean NLL than sKMD, although with eight panels neither
paired contrast is resolved. In the nonstationary examples, much of the
reconstruction improvement over independent-mode KMD is also obtained by a
pooled full covariance. Carrier averaging accounts for covered changes in
the carrier family, while introducing a computational and estimation
budget; for fading, the independent-mode mixture reconstructs better than
the coupled mixture.

These comparisons also identify conditions for application design. A
trigger or phase convention must align the modal coordinates; changing the
trigger and dictionary can reverse a reconstruction gain. The combined
trigger/dynamics stress does not identify their separate causal effects. Observation-noise
integration assumes a correctly specified carrier law. The finite carrier library
does not establish robustness to arbitrary frequency, envelope or mode-count
changes. The fixed-class chirp and fading studies concern time-varying
carrier laws rather than tracking a changing excitation class. Changed
regime hazards and uncertain transition locations require
their own temporal comparisons: an improvement averaged over a trajectory
need not imply sharper localization at a switch. Supplying noise features
to a discriminative head does not itself ensure extrapolation beyond its
training-quality range; that outcome does not imply that all discriminative
estimators must behave in the same way. In the ringdown study, the
reconstruction effect is more stable across independent panels than the classification
ranking, and the large tail loss demonstrates the importance of class-law
calibration even when average accuracy is maintained.

\subsection{Representation, estimation and adaptive selection}
The information bounds are population statements under their stated class
families and decoder restrictions. Their constants may be conservative or
much smaller than finite-sample estimation error. Exact insufficiency of a
summary therefore need not produce a practically large fitted advantage.
The conditioning-controlled bound also permits its lower bound to vanish
as the allowed condition number grows; the exact coordinate calculation
does not quantify finite-precision stability of near-singular summaries.
Proper-score gaps, zero--one error
gaps and stable numerical rank remain distinct quantities.

Hausdorff-based Kernel Flows use geometric error to learn attractors and
more general invariant sets \citep{bh_kfs_p6,jahagirdar2026invariant}.
Marked metric--measure selection likewise distinguishes geometric fit
from prediction of a designated label \citep{hamzi2026category}.
Here log loss measures the cost of omitting modal contrasts relevant
to a regime label, even if a representation reconstructs trajectories
or invariant sets accurately.

The finite-library result supplies an explicit sufficient benefit condition
under independent selection, Gaussian class laws and controlled span
leakage. Its demonstrated fitted benefit concerns the specified energy
oracle and sampling design, rather than every invertible energy basis or
every adaptive microlocal extraction procedure. Shared interactions can
cancel, energies can already be sufficient, heavy noise can attenuate the
relevant interaction, and covariance estimation can cost more than the
information it recovers. The selected-dictionary perturbation bound is
conditional on its estimation errors; it does not guarantee EM convergence
or remove dictionary-learning uncertainty. Estimating a general ambient
noise covariance can also be costly.

Same-record carrier selection requires a coherent treatment of its selection
event. Conditional Gaussian integration with a selected dictionary alone
does not integrate frequency, phase or model-selection uncertainty. The
finite-grid and upstream-code checks establish numerical agreement at the
tested interfaces and resolutions. The fixed-band and global adaptive
extraction procedures remain distinct, and continuum convergence requires
analysis beyond the tested refinement. These distinctions motivate further analysis of adaptive
representations and estimated observation covariances.

\subsection{Recorded signals and physical validation}
The recorded-signal sequence shows the importance of matching the
representation and class law to the task. Robust regime laws with past-RR
features, combined with observation integration, improve the declared
INCART comparison under added noise; localization improves the
reconstruction span but was not needed for classification, since the
robust global seven-dimensional model is better in three of five
conditions. Order coordinates provide a physical normalization for the
bearing study, whose severe-telemetry condition benefits from uncertainty
integration. The complete comparisons also include native or milder
conditions favoring point inference, stronger waveform or pooled-prior
controls, and changes in ranking under operating-condition shifts.

The global-carrier ECG and absolute-frequency bearing studies show that
accurate likelihood algebra does not ensure an adequate representation or
regime law: in MIT--BIH, point QDA and waveform PCA/RBF have lower NLL than
sKMD and no comparison survives the declared Holm family, and at the unseen
bearing speed every learned method is worse than uniform prediction. Event
alignment alone does not establish physiological modes,
and a returned bearing ridge is not automatically a physical fault
mechanism. A low-dimensional span cannot recover components outside that
span; robust integration addresses a specified working noise law rather
than proving that the original recording noise follows it. Annotation
timing, correlated artifacts and the finite covariance-scale library are
further working-model components to assess in recorded signals. The bearing
sampling-axis conventions and the small number of development units are
additional sources of uncertainty whose separate causal effects were not
identified by the absolute-frequency transfer experiment.

The INCART classification gain is suggestive rather than conclusive: it
rests on one-sided tests with Holm-adjusted $p=0.04235$, conditional on four
overlapping cross-fits, and the two-sided interval for the point comparison
includes zero. Over the four fold refits, neither fold-level $t$ interval
excludes zero (point QDA $[-0.140,0.028]$, PCA43 $[-0.061,0.015]$), and
leaving out any one fold gives a family Holm $p$ between $0.059$ and
$0.102$. Neither
practical margin is met. Recorded-reference reconstruction is worse than
both strong controls: mean squared error is $0.010579$\,mV$^2$ for sKMD,
against $0.009151$ for the independently fitted pooled full prior and
$0.008119$ for PCA43. The bearing test cohort contains
one healthy and one real-damaged bearing; it does not support a
population-level claim about balanced natural damage. These cohort
compositions and the patient cross-fitting structure should guide the
design of larger physical confirmation studies.

In the recorded-data experiments, observation quality is varied through
declared corruption and missingness, with the associated observation model
supplied to the inference procedures. The recorded waveforms also contain
physiological and sensor variation, so reconstruction toward them is
recorded-reference recovery. These experiments do not yet establish a
general advantage under naturally varying, independently calibrated sensor
noise. That validation requires independent noise measurements predictive
of the scored record, externally aligned modes, and confirmation separated
by acquisition unit. MEG-SCANS offers a concrete resource with
stimulus-locked recordings from 24 participants and empty-room measurements
before and after sessions \citep{habersetzer_megscans_2025}. Empty-room
measurements provide independent noise references; their transfer to
in-task conditions remains a question for the proposed study.

\subsection{Statistical scope and research directions}
The independently fitted controlled panels quantify sampling variability
under the specified laws. The six-panel application intervals are
descriptive, and the ECG patient intervals condition on overlapping
training fits. Repeating windows, bootstrap draws or corrupted views does
not create additional independent units. Equivalence at level $0.05$
requires a specified margin and the corresponding two one-sided tests,
or a 90\% interval wholly inside that margin. A nonsignificant contrast or
numerical agreement alone does not establish it.

The comparison inventory lists the outcomes and their uncertainty units.
Its counts include overlapping outputs, so they represent reported
comparisons rather than independent discoveries. Local Holm adjustments
apply to their specified families; they do not provide a study-wide
error guarantee. Only the binary primary family, the three-regime primary
endpoint and the INCART primary classification family are confirmatory;
the retrospective calibration and sensitivity analyses and all other
application intervals are exploratory. Most prespecified practical
criteria were not met: apart from the binary experiment's illustrative
margin, the switching study's observation-shift NLL and MSE are the only
cases in which a criterion was met (Table~\ref{tab:focused_scorecard}).

Future theoretical work can sharpen the fitted-risk bounds, analyze
observation-covariance estimation and same-record selection, and extend
the Gaussian sufficiency results to additional non-Gaussian families,
including finite mixtures. Future applications can combine independently measured quality
with task-specific waveform and covariance controls, a primary decision
endpoint, and an operational tolerance set before confirmation.

\section{Conclusion}
\label{sec:conclusion}
sKMD expresses supervised regime inference and regime-conditioned
reconstruction through the posterior of a KMD mode decomposition. Its
posterior transfer is standard Gaussian conditioning; the theory in
Part~\ref{part:theory} is the main contribution. It characterizes when
fixed mode energies remain sufficient as observation quality varies, shows
that correlated noise variation forces the full quadratic span for any
fixed linear decoder, gives an exact three-regime obstruction with explicit
but conservative positivity bounds, and bounds the cost of learning mode
spans and regime covariances from finite data.

The applications in Part~\ref{part:applications} show where these
mechanisms matter in practice. In controlled experiments the integration
gains are statistically resolved but small: $0.00246$ nats in the primary
three-regime comparison and $0.00116$ nats against a symmetry-matched
retrospective control, which reverses under a biased noise monitor. In the
binary experiment most of the gain over a point rule is recovered by
quality-aware calibration. In learned and recorded signals, integration
helped under declared synthetic degradations (added ECG noise, with
suggestive confirmatory evidence, and severe simulated bearing telemetry),
whereas at native quality or under operating-condition shift simpler
controls were as good or better. Modal dependence, observation uncertainty,
carrier averaging and temporal filtering each contributed in some settings
and not in others. Together, the theory and the evidence specify what a
regime-inference method on KMD representations must retain, and when the
additional modeling is worth its cost.

\appendix
\clearpage
\phantomsection
\part*{Appendices}
\addcontentsline{toc}{part}{Appendices}

\section{Protocol and complete results for the 64-panel test}
\label{sec:controlled64supp}
The executable protocol specifies the generating laws, fitting rules and random seeds; its location and the seed formulas are recorded in Appendix~\ref{sec:reproduction}. Classes are balanced in each split. Each event is an independent draw; each panel is fitted independently. Conditions share a fitted model, and uncertainty is evaluated across independently fitted panels.

\paragraph{Carriers and observation law.}
On $48$ equally spaced times in $[0,1]$, form $e^{-t}\cos(6\pi t)$ and $e^{-0.7t}\cos(14\pi t)$, orthonormalize their columns, and choose positive first entries. Write the resulting matrix as $H$. The primary data are the modal measurements $Z=C+\varepsilon$, not an adaptively estimated carrier representation. Synthesized signals are $HC$; coefficient MSE and signal mean-square error differ only by the fixed factor $2/48$. Dense checks use the equivalent waveform covariance
\[
 R_\sigma=\sigma^2HH^\top+0.01(I-HH^\top)
\]
and a class-independent orthogonal noise realization. This controlled measurement model supplies each method with the exact noise variance.

The low-noise condition fixes $\sigma=0.05$; very high noise fixes $\sigma=3$ outside the fitting support. Student impulses multiply each Gaussian coefficient vector by $\sqrt{3/U}$, $U\sim\chi^2_5$, preserving covariance. They change the latent law, not the sensor-error law. In the null arm, an independent random sign multiplies the second coefficient in every role. Its class density is exactly the same equal mixture of the positive- and negative-correlation laws, before and after convolution with isotropic noise. The numerical null check confirms that mixture identity; the generator is verified independently against the same identity.

\paragraph{Fitting and equal information.}
Gaussian means and covariances are estimated from fitting observations; the average known noise covariance is subtracted from the within-class sample covariance. Eigenvalues are floored at $0.05$. Development NLL selects diagonal shrinkage in $\{0,0.1,0.3,0.6\}$ and temperature in $\{0.5,0.75,1,1.5,2,3,4\}$. Same-law point inference retains the selected means and covariances, omits $\sigma^2I$ from its test likelihood, and independently chooses temperature. Signed OAS-QDA fits observed coefficients and uses the same temperature grid.

The two logistic controls have $17$ features each. Signed base features are $(z_1,z_2,z_1^2,z_2^2,z_1z_2)$; energy base features are $(e_1,e_2,e_1e_2,e_1^2,e_2^2)$, $e_i=z_i^2$. Each five-vector is multiplied by $1,r,r^2$, where $r=(1+\sigma^2)^{-1}$; two additional quality coordinates are $1-r$ and $(1-r)^2$. Fitting-only standardization, $C\in\{0.01,0.1,1,10\}$ and the same temperature grid complete the common budget. Independent-mode KMD energies are invertible rescalings of $(e_1,e_2)$ when quality is supplied, so this representation does not discard additional information through normalization. These are fitted discriminative controls, not an exact energy-posterior oracle.

The robust priors are finite Gaussian covariance-scale mixtures. The initial scales $2^{-3},\ldots,2^8$ receive inverse-gamma shape/scale $5/2$ bin probabilities, with geometric bin edges, and are normalized to have weighted mean one. Finite-mixture evidence and conditional means are evaluated exactly. The finite scale law defines the fitted approximation to the continuous Student model under noise. The same-mixture point head omits observation variance but shares the fitted robust law; the independently fitted point-mixture head fits uncorrected observed moments with its own common shrinkage/temperature grid.

Table~\ref{tab:controlled64_all_nll} reports every classifier in every
condition.

\begin{table}[t]
\centering\small
\caption{Every classifier's mean NLL across $64$ panels. Heterogeneous noise is primary; all other columns are secondary. ``Selected signed'' is quality-interaction logistic in every positive panel and is separately selected in the null arm.}
\label{tab:controlled64_all_nll}
\begin{tabular}{@{}lrrrrr@{}}\toprule
Method & Heterogeneous & Low & Very high & Student & Null\\\midrule
Integrated Gaussian & 0.5826 & 0.5023 & 0.6912 & 0.6060 & 0.6937\\
Selected signed & 0.5846 & 0.5011 & 0.9609 & 0.6081 & 0.6939\\
Same-law point & 0.6511 & 0.5833 & 1.7004 & 0.6642 & 0.6954\\
Signed OAS-QDA & 0.6454 & 0.6034 & 1.3576 & 0.6557 & 0.6936\\
Signed quality interaction & 0.5846 & 0.5011 & 0.9609 & 0.6081 & 0.6936\\
Energy interaction & 0.6937 & 0.6937 & 0.7847 & 0.7236 & 0.6937\\
Integrated scale mixture & 0.5863 & 0.5122 & 0.6912 & 0.5973 & 0.6936\\
Same-mixture point & 0.6220 & 0.5567 & 0.8314 & 0.6342 & 0.6937\\
Fitted point mixture & 0.6237 & 0.5612 & 0.8824 & 0.6367 & 0.6934\\
Uniform & 0.6931 & 0.6931 & 0.6931 & 0.6931 & 0.6931\\
\bottomrule\end{tabular}

\end{table}

\paragraph{Reconstruction and scope.}
Gaussian and robust integration use temperature-calibrated class weights with untempered class-conditional means. They are therefore calibrated mixture estimators, not necessarily the posterior mean of the original untempered generative model. The two point-weighted controls reuse the same Gaussian conditional means and change only their class weights. The pooled full prior aggregates selected class covariances and between-mean variation; it is not an independently tuned pooled fit. Independent-mode KMD retains only its diagonal. The target here is simulated latent coefficients, unlike the recorded-reference targets in the external studies. Table~\ref{tab:controlled64_all_mse} reports every reconstruction in every condition.

\begin{table}[t]
\centering\small
\caption{Coefficient MSE, averaged over coordinates, events and then independent panels. The reconstruction comparisons are secondary and include all observation conditions.}
\label{tab:controlled64_all_mse}
\begin{tabular}{@{}lrrrrr@{}}\toprule
Method & Heterogeneous & Low & Very high & Student & Null\\\midrule
Integrated Gaussian & 0.29613 & 0.00249 & 0.90226 & 0.29229 & 0.29878\\
Integrated scale mixture & 0.29996 & 0.00250 & 0.90502 & 0.28300 & 0.30056\\
Point-weighted Gaussian & 0.29893 & 0.00250 & 0.92177 & 0.29533 & 0.29870\\
Selected-signed weights & 0.29645 & 0.00249 & 0.91520 & 0.29297 & 0.29875\\
Pooled full KMD & 0.29937 & 0.00250 & 0.90249 & 0.30071 & 0.29879\\
Independent-mode KMD & 0.29916 & 0.00250 & 0.90207 & 0.30049 & 0.29867\\
\bottomrule\end{tabular}

\end{table}

\paragraph{Statistical interpretation and verification.}
Two primary NLL differences use paired $t$ intervals and two-sided Holm adjustment across the two endpoints. Separate Holm families test the illustrative $0.02$-nat improvement and equivalence margins; equivalence uses TOST and $90\%$ intervals. The $0.02$-nat margin is an illustrative sensitivity threshold. The prespecified signed-versus-energy contrast and every stress/control/reconstruction comparison are secondary. Threshold calculations are interpreted for the specified NLL contrasts; MSE is reported on its own scale. Tail losses are not clipped. Panel means, rather than individual events or correlated conditions, determine uncertainty.
\FloatBarrier

\section{Three-regime protocol and complete result grid}
\label{sec:three_regime_protocol}
The executable protocol specifies the generating laws, fitting rules and
seeds; its location and the seed formulas are recorded in
Appendix~\ref{sec:reproduction}. The three-regime design develops the
representation question studied in the binary example using independently
generated test observations and model fits. Class labels are balanced and shuffled within every split.
The $64$ panels, rather than individual events or shared conditions, are
the replication units for each reported contrast.

\paragraph{Moment fit and controls.}
Known-zero-mean second moments use $n_k^{-1}\sum_i z_iz_i^\top$ minus the
class average supplied variance times $I_2$. Eigenvalues are floored at
$0.05$, and off-diagonal shrinkage is selected from $\{0,0.1,0.3,0.6\}$.
Continuous inverse temperature $\beta\in[0,100]$, including zero, minimizes
development NLL. All point and discriminative controls receive the same
continuous calibration range; the point rule therefore has no fixed upper
temperature cap. Gaussian conditional means themselves are untempered.

Let $r=(1+\sigma^2)^{-1}$ and $q=1-r$. Signed base features are
$(z_1,z_2,z_1^2,z_2^2,z_1z_2)$; energy base features are
$(e_1,e_2,e_1e_2,e_1^2,e_2^2)$ with $e_i=z_i^2$. Multiplying by
$1,r,r^2$ and appending $q,q^2$ gives $17$ features in each case.
For the quality-logit calibrator, the base vector contains the two
class-versus-zero logits of the selected untempered point Gaussian law;
the same expansion gives $8$ features. Multinomial logistic regression
uses fitting-only standardization, regularization $C\in\{0.01,0.1,1,10\}$,
and development-selected continuous temperature. All test predictions use
fixed fitted parameters. The signed selector uses development NLL among
that logit calibrator, signed quality-quadratic logistic and signed OAS-QDA.

Known-law energy likelihoods sum the Gaussian likelihood over the four
coordinate-sign choices; their common Jacobian cancels from class odds.
The rotated-energy oracle searches the $37$ angles
$\theta_j=j\pi/74$, $j=0,\ldots,36$, using development NLL. Its reported
performance is on independent test observations. Oracles receive the true
noise covariance even in the biased-monitor stress condition, whereas
fitted methods receive the stated miscalibrated monitor. They are explicit
privileged-information references in that condition.

\paragraph{Centered signed sensitivity.}
The fitted Gaussian uses central symmetry by fixing its means to zero.
A retrospective signed control also enforces that symmetry: its base is
$(z_1^2,z_2^2,z_1z_2)$, expanded by $1,r,r^2$ and $q,q^2$ to $11$ features.
It uses the same fitting standardization, regularization and continuous
calibration rules. This exploratory head was specified after the primary results and
is evaluated separately from the primary selector and endpoint.

\paragraph{Complete conditions and units.}
The primary test samples continuous source quality independently of class.
The high-noise condition fixes $\sigma=3$. The monitor-bias condition draws
the same source noise but supplies only half its true variance to every
fitted method. No parameters are retuned in either stress condition.
The single primary NLL contrast uses a paired $t$ interval and two-sided
$t$ test with $63$ degrees of freedom. Other intervals are pointwise secondary descriptions of the specified
contrasts. For NLL, $\exp(-\Delta)$ is the ratio of
geometric mean true-class probabilities; this interpretation does not apply
to squared reconstruction error. Tables~\ref{tab:three_regime_all_nll}
and~\ref{tab:three_regime_all_mse} report every method in every condition.

\begin{table}[tbp]
\centering\small
\caption{Every three-regime classifier and condition. Lower NLL is better.
The high-noise and biased-monitor columns are secondary. Oracle parameters
are known; fitted methods use the stated fitting budget.}
\label{tab:three_regime_all_nll}
\begin{tabular}{@{}rlrrr@{}}\toprule
Fit events & Method & Continuous & High noise & Biased monitor\\\midrule
600 & Integrated Gaussian & 1.03934 & 1.09599 & 1.04935\\
600 & Selected signed & 1.04273 & 1.22756 & 1.04872\\
600 & Continuous global point & 1.05992 & 1.58551 & 1.05998\\
600 & Quality-logit calibration & 1.04399 & 1.22794 & 1.05049\\
600 & Signed quality quadratic & 1.04368 & 1.21308 & 1.04820\\
600 & Energy quality polynomial & 1.06278 & 1.46583 & 1.06759\\
600 & Signed OAS-QDA & 1.05862 & 1.56495 & 1.05849\\
600 & Oracle signed & 1.03226 & 1.09570 & 1.03147\\
600 & Oracle fixed energies & 1.05206 & 1.09687 & 1.05147\\
600 & Oracle rotated energies & 1.04348 & 1.09625 & 1.04252\\
600 & Uniform & 1.09861 & 1.09861 & 1.09861\\
600 & Centered signed quadratic & 1.04059 & 1.25268 & 1.04637\\
\midrule
2400 & Integrated Gaussian & 1.03371 & 1.09578 & 1.04537\\
2400 & Selected signed & 1.03617 & 1.20310 & 1.04568\\
2400 & Continuous global point & 1.05494 & 1.61454 & 1.05535\\
2400 & Quality-logit calibration & 1.03612 & 1.18777 & 1.04630\\
2400 & Signed quality quadratic & 1.03637 & 1.20773 & 1.04477\\
2400 & Energy quality polynomial & 1.05642 & 1.44939 & 1.06526\\
2400 & Signed OAS-QDA & 1.05354 & 1.59880 & 1.05352\\
2400 & Oracle signed & 1.03185 & 1.09573 & 1.03273\\
2400 & Oracle fixed energies & 1.05196 & 1.09680 & 1.05178\\
2400 & Oracle rotated energies & 1.04290 & 1.09629 & 1.04332\\
2400 & Uniform & 1.09861 & 1.09861 & 1.09861\\
2400 & Centered signed quadratic & 1.03487 & 1.21709 & 1.04442\\
\bottomrule\end{tabular}

\end{table}
\begin{table}[tbp]
\centering\small
\caption{Every three-regime reconstruction condition, measured by coefficient
MSE. The selected-signed reconstruction shares the integrated method's
Gaussian conditional means and changes only the class weights.}
\label{tab:three_regime_all_mse}
\begin{tabular}{@{}rlrrr@{}}\toprule
Fit events & Reconstruction & Continuous & High noise & Biased monitor\\\midrule
600 & Integrated regime & 0.44660 & 1.26067 & 0.48992\\
600 & Selected-signed weights & 0.44675 & 1.27209 & 0.48918\\
600 & Pooled full KMD & 0.44867 & 1.26087 & 0.48874\\
600 & Independent KMD & 0.45448 & 1.27021 & 0.49598\\
\midrule
2400 & Integrated regime & 0.44467 & 1.25527 & 0.49124\\
2400 & Selected-signed weights & 0.44485 & 1.26189 & 0.49071\\
2400 & Pooled full KMD & 0.44704 & 1.25556 & 0.48887\\
2400 & Independent KMD & 0.45370 & 1.26505 & 0.49644\\
\bottomrule\end{tabular}

\end{table}

\FloatBarrier

\section{Computational record of the comparisons}
\label{sec:focused_inventory}
The script \path{comparison_inventory.py} of the numerical package (Appendix~\ref{sec:reproduction}) enumerates a declared set of delivered
summaries using one focal method--comparator--condition--metric tuple per entry.
Each row records its input checksum, JSON location, statistical unit and numerical value.

\begin{center}\begin{tabular}{lr}\toprule
Object & Count\\\midrule
Summary artifacts & 12\\
Artifact-condition entries & 48\\
Method-condition-metric estimates & 1655\\
Focal-comparator-condition-metric contrasts & 1445\\
Stored uncertainty/threshold records & 844\\\bottomrule
\end{tabular}\end{center}

These counts describe overlapping summaries with shared models and
observations; uncertainty is evaluated using the independent units
specified for each study. They exclude theory-only checks, early-warning
and raw-window studies outside this article, and the separately reported
two controlled experiments and retrospective calibration comparison. The sources are the oscillator, two ringdown,
two nonstationary, switching, two dictionary-mixture, MIT--BIH, INCART,
absolute-frequency bearing and combined bearing-confirmation summaries.
A stored threshold record may repeat a contrast for a different practical
margin.

Multiplicity adjustments use the comparison families specified for each
experiment. The controlled panels follow their specified protocols, and
the calibration comparison is a retrospective analysis of the binary-study
observations.
\paragraph{Additional controlled comparisons.}
The retrospective binary calibration comparison includes $14$ available
method--condition NLL estimates and $14$ integration-versus-calibrator
contrasts: five conditions for each of two calibrators and four for the
quality-bin rule, which has no extrapolation output. They reuse existing
test records. The three-regime study reports $6$ budget--condition cells,
$222$ fitted/oracle metric estimates and $60$ stored paired contrasts
($42$ NLL and $18$ reconstruction). Its centered signed sensitivity adds
$18$ metric estimates and $6$ NLL contrasts. Only one three-regime
contrast is the primary endpoint. The counts distinguish stored outputs from the comparison families used for statistical inference.

\section{Computational reproduction}
\label{sec:reproduction}

The 12 November numerical update contains the conditioning and sensor-noise
checks in \path{theory/run_theory_checks.py} and their outputs in
\path{theory/theory_results.json}.

\paragraph{Package layout.}
The LaTeX source bundle contains \path{main.tex}, \path{references.bib},
native PDF figures and table fragments. Build from the source directory with
\texttt{latexmk -pdf -interaction=nonstopmode -halt-on-error main.tex}.
The separate
numerical companion to the 8 November 2026 release contains the engine,
raw data with their licenses, saved results, \path{reproduce.py} and
\path{REPRODUCTION.json}. The generator of every figure is listed in
\path{numerics/FIGURE_SOURCES.md}. Paths below are relative to the
companion's \path{numerics/} directory. Run the commands from that
directory as described in its README.

The engine and the focused-study folders include protocols that define
generating laws, fitting rules, independent units, seeds and evaluation
conditions. Verification programs check representation inequalities,
Gaussian identities and energy likelihoods; generators and reports cover
the controlled and recorded-signal studies. Focused-study wrappers run in
empty temporary directories and compare summaries with the saved records.
Validation records distinguish checks rerun for this version from inherited
records and from studies that were not rerun. Regenerated reports do not
automatically revise this article.

\paragraph{Rerun of 8 November 2026.}
On 8 November 2026 the complete numerical companion was re-executed from
the delivered sources and raw data: all 31 registered recipe families,
including the MIT--BIH, INCART and Paderborn application studies with fresh
extraction of the bearing features from the raw recordings, the optional
\texttt{refit\_bearings} profile, and the regeneration of the omitted
prediction and model caches. Before regeneration, all 3,650 delivered files
matched their SHA-256 manifest and the raw data matched the supplied
checksums; the INCART signal files are documented 180-second prefixes and
match their download manifest. The rerun used CPython 3.12.15 with exactly
the pinned package versions on Linux aarch64 with OpenBLAS; the archived
results were produced on macOS. Every family ran to completion. Quantities
computed by sKMD agreed with the archived values: within
$2\times10^{-14}$ relative in the application summaries, except MIT--BIH
ECG ($3\times10^{-12}$) and INCART ($9\times10^{-7}$), and within
$4\times10^{-6}$ relative in the focused and development studies.
Random-forest, RBF-SVC and logistic comparators are not bit-reproducible
across platforms: 34 displayed comparator values in ten generated tables
changed, the largest by $0.013$ nats in the INCART energy-path random
forest, and across 1,445 focal contrasts no ordering of sKMD against a
comparator changed. The strict equality replays of the three frozen focused
studies failed only through such comparator differences and differences at
the $10^{-7}$ relative level or below. Of the 43 engine verification
commands, 40 passed; the three failures came from a bitwise-equality check
(discrepancy $1.8\times10^{-15}$) and a $10^{-7}$ tolerance (observed
$1.14\times10^{-7}$). Refitting the bearing development stage from the raw
recordings reproduced every frozen model selection. The tables of this
article report the archived (original-platform) values; the per-file
comparison records and the full rerun report are in
\path{RERUN_2026-11-08/} of the \path{numerics/} directory. The analyses
added for the 8 November release (the binary known-law oracle, the bound-versus-gap
values of Table~\ref{tab:bound_vs_gap}, the feature-budget qualities of
Table~\ref{tab:feature_budget_qualities}, the paired $t$ intervals for
studies with at most twelve units, the INCART fold-level analysis and the
fourteen-bearing interval) are reproducible from \path{analyses/} in
\path{numerics/}, one command each, as listed in its README. Reproduction
with fixed seeds is computational reproduction, not independent empirical
confirmation.

\subsection{Verification record by study}
\label{sec:verification_record}
The following records were moved from the body of the article. They state
what was checked for each study and, where applicable, the discrepancy
between the integrated predictions and a direct dense Gaussian (or matched
mixture or HMM) implementation, as summarized in
Section~\ref{sec:reference_reduction}.

\paragraph{Numerical interaction budget (Section~\ref{sec:quality_theory_v32}).}
The 12 qualities of Table~\ref{tab:feature_budget_qualities} are generated
by \path{quality_feature_budget.py} with seed $2026093017$; its result file
records the qualities, the eigenvalue bounds $\lambda_-$ and $\lambda_+$, the
contrast Gram eigenvalues and every bound in Figure~\ref{fig:focused_budget}.

\paragraph{Binary controlled study (Section~\ref{sec:controlled64}, Appendix~\ref{sec:controlled64supp}).}
The executable protocol is \path{controlled_study/PROTOCOL.json}; its
source-checksum manifest accompanies the code. The seed for positive-panel
fitting is $930640000+1000j$, $j=0,\ldots,63$; development adds $1$, tests
add $10,11,12,13$. The separately fitted randomized-sign arm adds $100$ to
fitting/development/primary-test seeds. All $320$ archives contain finite
arrays ($9600$ arrays checked); recorded metrics are independently
recomputed. The waveform Gaussian identity has maximum absolute log-odds
discrepancy $2.59\times10^{-13}$ and calibrated reconstruction discrepancy
$4.30\times10^{-14}$. Saved models, development losses, selected controls
and every test score are available. Summaries are regenerated with
\path{controlled_study/code/analyze_controlled_modes.py}, the declared direct
feature comparison and tables with
\path{controlled_study/code/report_controlled_modes.py}, and the native PDF
figure with \path{controlled_study/code/plot_controlled_modes.py}. Full
reruns use the supplied sources and seeds in a directory with empty result
and figure folders.

\paragraph{Retrospective calibration (Section~\ref{sec:calibration_audit}).}
The calibration sources, protocol and input hashes are provided in
\path{calibration_audit/}. All $320$ score archives recompute exactly. The
population temperature identity agrees with independently evaluated
Gaussian log odds to $2.7\times10^{-15}$, and reconstructed point odds agree
with the saved fits to $2.9\times10^{-14}$.

\paragraph{Three-regime study (Section~\ref{sec:three_regime}, Appendix~\ref{sec:three_regime_protocol}).}
The executable protocol and source-checksum manifest are supplied in
\path{three_regime_study/}. Panel $j$ and budget index $b\in\{0,1\}$ use
fitting seed $1001640000+1000j+100b$, development adds $1$, and the three
test conditions add $10,11,12$. The $384$ test archives contain $7680$
finite arrays; all saved metrics recompute exactly. Independent density and
sign-enumeration checks verify class odds, energy posteriors and conditional
means. The complete model fits, development choices and per-panel scores are
supplied, as are the centered-control protocol, fits and all $384$ sets of
its predictions. Run \path{three_regime_study/code/run_study.py}\ \texttt{run}
only in a fresh result directory with the supplied sources, followed by
\texttt{analyze}; \path{three_regime_study/code/centered_control.py}
regenerates the retrospective centered head;
\path{three_regime_study/code/report_study.py} writes complete tables and the
native PDF figure. The separate verification script checks source hashes and
saved predictions.

\paragraph{Adaptive oscillators (Section~\ref{sec:adaptive_experiments_v32}).}
The single-mode algebra of the automatic KMD subset agrees with the released
\texttt{micloc\_KMD} routine to $3.4\times10^{-16}$; the repository commit,
upstream source and diagnostic outputs accompany the code. The companion
protocol and scripts specify data generation, numerical refinement and table
production. The simulation script with \texttt{--verify} checks every saved
array for finiteness and recomputes every reported metric. All 16 test
archives pass, with metric discrepancy zero.

\paragraph{Triggered ringdowns (Sections~\ref{sec:ringdown_v32} and~\ref{sec:ringdown_fullcov_v32}).}
Conditional on $H$ and the plugged-in noise scale, direct dense checks of raw
log odds and class-conditional means agree with the coupled calculation
within $2.1\times10^{-13}$ in log odds. The companion protocol and scripts
generate all six independent panels, the tables, figure and timestep check;
the \texttt{--verify} mode checks all 648 saved arrays, recomputes all
metrics and repeats the dense Gaussian checks. Every array is finite and
every metric reproduces with zero discrepancy. The full-covariance protocol
and simulation script specify the independent validation;
\texttt{--sensitivity} runs the exploratory sensitivity analysis. Both
preserve all scores, seeds and execution hashes. Their 36 archives contain
1,044 finite arrays, with exact metric recomputation and dense
raw-likelihood discrepancy below $3.5\times10^{-13}$. The companion dynamics
runner provides full rebuild and verification, with \texttt{--verify-only}
checking the saved empirical result collections.

\paragraph{Nonstationary carriers (Section~\ref{sec:nonstationary_v32}).}
All 72 test archives and 2,088 arrays are finite; saved metrics recompute
exactly. Direct dense Gaussian likelihood contrasts agree with the
coefficient calculation to below $10^{-11}$, checked on one event per class
in every panel and condition, with matching class-conditional means. The
companion protocol specifies the two studies. Independent seeds, complete
metrics, paired intervals, conditioning and span diagnostics, and source
hashes are retained in the companion results. Simulation and artifact
scripts rebuild the studies, figures and saved-array diagnostics; a separate
sensitivity script provides the exploratory multiplicity and fade-region
checks.

\paragraph{Dictionary mixture (Section~\ref{sec:dictionary_mixture_v32}).}
All ninety-six archives accompany the numerical package. Two independent
implementations compare observation-space Gaussian densities with
\eqref{eq:dictionary_evidence_v32}, including unequal dictionary scales,
rank-zero and deficient masks. They also verify dictionary permutation,
split weights on duplicate dictionaries, congruent coefficient changes and
the single-dictionary limit. Maximum discrepancies are below $10^{-12}$ in
these algebra checks. Every saved metric is recomputed from the retained
predictions.

\paragraph{Switching (Section~\ref{sec:switching_v32}).}
The companion protocol specifies the information-form interface and
Bernoulli mask law, including rank-deficient observations. The simulation
script generates all 60 archives; its artifact script renders the tables and
figure. The source verifier finds 2,940 finite arrays and zero metric
discrepancy. An independent 221-event dense-model validation includes every
rank-deficient test block and complete-observation examples, with log-odds
error below $1.2\times10^{-12}$. Separate boundary checks cover ranks zero,
one, two and four; the no-observation filter reduces to the learned clock
prior.

\paragraph{MIT--BIH ECG (Section~\ref{sec:ecg_v32}).}
Independent dense Gaussian checks agree with the working integrated log
odds within $4.6\times10^{-7}$ and conditional means within
$5.2\times10^{-11}$. The ECG reproduction entry point generates the study and
independently recomputes saved metrics in its verification mode. The
protocol, source snapshots, official MIT--BIH file-hash checks, environment,
all 25 test archives, fitted dictionaries, per-patient losses and
full-precision comparisons are in the numerical package.

\paragraph{Localized robust ECG model (Section~\ref{sec:external_repair_methods_v32}).}
Independent ambient-space calculations evaluate the joint density of
$(Y_o,R)$, rather than reusing the reduced implementation. Thirty-two cases
cover Gaussian and twelve-scale priors, rhythm dimension zero and two, full
and deficient observation rank, no waveform observations, and temperatures
one and two. The maximum class-log-odds discrepancy is $5.2\times10^{-14}$,
and the maximum conditional-mean discrepancy is $3.1\times10^{-14}$. The
kernel and coordinate identities in \eqref{eq:ecg_kernel_svd_v32} agree to
$2.4\times10^{-14}$.

\paragraph{INCART ECG (Section~\ref{sec:ecg_repair_v32}).}
Independent verification checks 20 archives and 1,080 finite arrays,
reproduces 220 MSE values, and recomputes 2,510 summary, ranking and
multiplicity-adjusted inference values to $2.3\times10^{-16}$. Eight
ambient-space waveform/RR checks match reduced log odds to
$6.2\times10^{-11}$ and reconstructions to $6.8\times10^{-13}$. The protocol
and source snapshots specify the complete study; its reproduction entry
point rebuilds the results, and its verification mode recomputes metrics
from saved predictions. All classifier scores, coefficient estimates, full
coefficient-posterior covariances, observation-noise specifications,
patient identities, recording indices and corruption repetitions remain
available.

\paragraph{Absolute-frequency bearings (Section~\ref{sec:external-bearing-v32}).}
The 90 Paderborn files retain their original names and checksums. The
largest class-probability discrepancy between the coupled calculation and an
algebraically matched direct Gaussian calculation is $1.32\times10^{-9}$. An
independent calculation from the saved window scores reproduces all reported
scalar metrics within $2.23\times10^{-16}$. The bearing-data scripts
reproduce the specified subset and comparison. The protocol, its dated
timestamp amendment, the sampling-axis specification, source checksums,
fitted settings, optimization diagnostics, predictions and independent
arithmetic checks are supplied with the code. Verification checks the 90
recording-file hashes, grouping, finite scores and saved metrics. The saved
results record each linked-axis step distribution and metadata, and the
sensitivity script reproduces both sampling conventions.

\paragraph{Order-normalized bearings (Section~\ref{sec:paderborn_repair_v32}).}
The protocol, all candidate scores, fitted model bundle, source hashes and
raw-file provenance are supplied with the code. Independent dense and
precision-form checks verify that the telemetry model's component evidence
and conditional means equal those of a conventional noisy Gaussian mixture.
The reproduction entry points prepare the independent cohort, fit the
native and telemetry models, and regenerate the complete numerical
summaries and figures.

\paragraph{Energy oracle (Section~\ref{sec:energy_oracle_v31}).}
The protocol and generator reproduce all $24$ panels; a separate
verification mode checks the saved outputs. All $144000$ development and
$360000$ test coefficient vectors, class labels, $361$ development losses per
panel, selected angles and test log posteriors are saved. Four-sign
enumeration agrees with \eqref{eq:energy_oracle_score_v31} to maximum
log-posterior error $3.4\times10^{-12}$; redundant-energy reconstruction
agrees with the full posterior to $5.7\times10^{-14}$.

\paragraph{Coupled-mode benchmark (Section~\ref{sec:interaction_experiments_v31}).}
The generators reproduce the primary benchmark, exploratory controls and
summary tables. The saved per-window predictions, masks, coefficient targets
and reconstructions support independent verification of the reported
losses. Protocols, candidate choices, paired intervals and complete
per-method scores accompany the source. A direct Gaussian likelihood with
identical coupled parameters verifies equality with posterior transfer.

\subsection{Data, software and assistance}
\paragraph{Data and software availability.}
The numerical package contains the Python generators, pinned requirements,
experimental protocols, numerical summaries, selected frozen predictions and
inputs for the documented offline reproduction routes. The INCART data
consist of the specified three-minute waveform prefixes and complete headers
and annotations. Dataset sources, attribution, checksums and third-party
licenses are documented in \path{science/PACKAGE_DATA_LICENSES.md} and the
dataset manifests. Regenerable prediction and model caches are listed
separately and can be rebuilt with \path{reproduce.py}.

\paragraph{AI assistance.}
OpenAI Codex assisted with source auditing, computational verification and
submission-file preparation. Anthropic's Claude assisted with the
8 November 2026 revision (review-driven editing, restructuring into two
parts, and reproduction checks) under the authors' direction.
Codex assisted with the editorial and typesetting changes for the
9 November 2026 edition and the literature revision of 10 November 2026.
The article uses the archived numerical values.

\bibliographystyle{plainnat}
\bibliography{references}
\end{document}

%% file: tables/feature_budget_qualities.tex
\begin{table}[htbp]
\centering\small
\caption{The twelve observation qualities $D_l$ of the numerical interaction budget that follows Proposition~\ref{prop:stable_quality_extension}, each with weight $w_l=1/12$, and $C_0=I_3$, $C_1=\operatorname{diag}(4,1,0.5)$. Each $D_l=Q_l\operatorname{diag}(u_l)Q_l^\top$, where $Q_l$ is the orthogonal factor of the QR decomposition of a standard normal $3\times3$ matrix and $u_l$ has independent $\mathrm U(0.1,2)$ entries (NumPy \texttt{default\_rng(2026093017)}, drawn in the order $Q_1,u_1,Q_2,u_2,\dots$). Entries are rounded to five decimals, which reproduces every printed digit of the spectrum and bounds; the last three columns are the eigenvalues $u_l$ in decreasing order. Over all $\Sigma_{kl}=C_k+D_l$, $\lambda_-=\min_{k,l}\lambda_{\min}(\Sigma_{kl})=0.7689$ (attained at $k=1$, $l=2$) and $\lambda_+=\max_{k,l}\lambda_{\max}(\Sigma_{kl})=5.4602$ (at $k=1$, $l=1$), so $\lambda_+^2=29.813$. With $T_2=0.006710$ and $T_4=0.0003525$, the excess-NLL bound $(r+2)\lambda_+^2T_m/32$ of Proposition~\ref{prop:stable_quality_extension} equals $0.03126$ for $m=2$ and $0.001642$ for $m=4$.}
\label{tab:feature_budget_qualities}
\begin{tabular}{@{}r*{6}{r}@{\hspace{1.2em}}*{3}{r}@{}}
\toprule
$l$ & $(D_l)_{11}$ & $(D_l)_{22}$ & $(D_l)_{33}$ & $(D_l)_{12}$ & $(D_l)_{13}$ & $(D_l)_{23}$ & $u_{l,(1)}$ & $u_{l,(2)}$ & $u_{l,(3)}$ \\
\midrule
1 & 1.32201 & 1.43000 & 0.53633 & 0.55883 & $-$0.39534 & 0.00764 & 1.98665 & 0.98518 & 0.31651 \\
2 & 0.35523 & 0.41266 & 0.27079 & 0.01851 & $-$0.08097 & 0.00418 & 0.42334 & 0.39455 & 0.22079 \\
3 & 0.92668 & 0.88353 & 1.17183 & 0.12033 & 0.41607 & 0.13382 & 1.53253 & 0.83515 & 0.61437 \\
4 & 0.42748 & 1.19888 & 0.62136 & 0.41533 & $-$0.09907 & $-$0.03364 & 1.38651 & 0.63033 & 0.23088 \\
5 & 1.14832 & 0.75828 & 1.65831 & $-$0.41444 & 0.41813 & $-$0.06662 & 1.95537 & 1.14161 & 0.46794 \\
6 & 0.97385 & 0.96233 & 0.86144 & 0.00072 & 0.06427 & 0.02665 & 1.00636 & 0.96329 & 0.82797 \\
7 & 1.29871 & 1.34955 & 1.71119 & 0.36045 & 0.09590 & $-$0.10147 & 1.73774 & 1.68402 & 0.93769 \\
8 & 0.88169 & 1.42069 & 0.34375 & 0.21211 & 0.22986 & $-$0.26932 & 1.52473 & 0.94867 & 0.17272 \\
9 & 0.35523 & 1.11461 & 1.25006 & $-$0.01514 & $-$0.08902 & 0.07992 & 1.29519 & 1.07832 & 0.34639 \\
10 & 1.23751 & 0.51068 & 0.98561 & $-$0.06208 & $-$0.70324 & $-$0.21660 & 1.83245 & 0.66303 & 0.23832 \\
11 & 1.21818 & 1.44903 & 1.44751 & 0.01419 & $-$0.02833 & $-$0.00002 & 1.45223 & 1.44861 & 1.21389 \\
12 & 0.55303 & 0.27895 & 0.30744 & $-$0.10761 & 0.08567 & $-$0.06420 & 0.62341 & 0.29321 & 0.22280 \\
\midrule
\multicolumn{10}{@{}l@{}}{Contrast Gram eigenvalues $\lambda_j(G)$: 0.211931, 0.012581, 0.003885, 0.002473, 0.0003095, 0.0000430} \\
\multicolumn{10}{@{}l@{}}{$\lambda_-=0.76893$, $\lambda_+=5.46015$; bound $m=2$: $0.03126$; bound $m=4$: $0.001642$} \\
\bottomrule
\end{tabular}
\end{table}

%% file: tables/energy_oracle_v31_paired_t_rows.tex
0.50 & Fixed energies & $0.0276\,[0.0262,0.0293]$ & $0.0218\,[0.0204,0.0231]$ & $[0.0256,0.0296]$ & $[0.0201,0.0235]$ \\
0.50 & Selected rotation & $0.0175\,[0.0164,0.0189]$ & $0.0122\,[0.0116,0.0128]$ & $[0.0159,0.0192]$ & $[0.0115,0.0130]$ \\
0.50 & Three energies & $0.0000\,[0.0000,0.0000]$ & $0.0000\,[0.0000,0.0000]$ & $[0.0000,0.0000]$ & $[0.0000,0.0000]$ \\
0.50 & $T_{1}$ & $0.0298\,[0.0288,0.0309]$ & $0.0230\,[0.0220,0.0239]$ & $[0.0285,0.0312]$ & $[0.0217,0.0243]$ \\
0.50 & $T_{0.3}$ & $0.0202\,[0.0187,0.0213]$ & $0.0127\,[0.0120,0.0134]$ & $[0.0184,0.0219]$ & $[0.0118,0.0136]$ \\
0.50 & $T_{0.1}$ & $0.0042\,[0.0032,0.0052]$ & $0.0034\,[0.0031,0.0037]$ & $[0.0029,0.0055]$ & $[0.0030,0.0038]$ \\
0.50 & $T_{0.03}$ & $0.0012\,[0.0004,0.0021]$ & $0.0009\,[0.0006,0.0012]$ & $[0.0001,0.0023]$ & $[0.0005,0.0013]$ \\
0.80 & Fixed energies & $0.0288\,[0.0271,0.0306]$ & $0.0417\,[0.0411,0.0422]$ & $[0.0265,0.0310]$ & $[0.0409,0.0424]$ \\
0.80 & Selected rotation & $0.0228\,[0.0205,0.0250]$ & $0.0262\,[0.0253,0.0273]$ & $[0.0199,0.0257]$ & $[0.0249,0.0275]$ \\
0.80 & Three energies & $0.0000\,[0.0000,0.0000]$ & $0.0000\,[0.0000,0.0000]$ & $[0.0000,0.0000]$ & $[0.0000,0.0000]$ \\
0.80 & $T_{1}$ & $0.0287\,[0.0266,0.0308]$ & $0.0416\,[0.0407,0.0424]$ & $[0.0259,0.0314]$ & $[0.0405,0.0427]$ \\
0.80 & $T_{0.3}$ & $0.0598\,[0.0572,0.0620]$ & $0.0709\,[0.0695,0.0723]$ & $[0.0567,0.0629]$ & $[0.0691,0.0727]$ \\
0.80 & $T_{0.1}$ & $0.0137\,[0.0128,0.0145]$ & $0.0188\,[0.0177,0.0196]$ & $[0.0126,0.0148]$ & $[0.0175,0.0200]$ \\
0.80 & $T_{0.03}$ & $0.0039\,[0.0033,0.0045]$ & $0.0056\,[0.0052,0.0060]$ & $[0.0031,0.0046]$ & $[0.0051,0.0061]$ \\
0.95 & Fixed energies & $0.0162\,[0.0153,0.0171]$ & $0.0358\,[0.0345,0.0369]$ & $[0.0150,0.0173]$ & $[0.0342,0.0373]$ \\
0.95 & Selected rotation & $0.0168\,[0.0151,0.0183]$ & $0.0275\,[0.0262,0.0287]$ & $[0.0148,0.0189]$ & $[0.0259,0.0291]$ \\
0.95 & Three energies & $0.0000\,[0.0000,0.0000]$ & $0.0000\,[0.0000,0.0000]$ & $[0.0000,0.0000]$ & $[0.0000,0.0000]$ \\
0.95 & $T_{1}$ & $0.0163\,[0.0151,0.0175]$ & $0.0357\,[0.0346,0.0369]$ & $[0.0148,0.0178]$ & $[0.0342,0.0372]$ \\
0.95 & $T_{0.3}$ & $0.1222\,[0.1195,0.1249]$ & $0.1997\,[0.1962,0.2029]$ & $[0.1187,0.1257]$ & $[0.1953,0.2041]$ \\
0.95 & $T_{0.1}$ & $0.0355\,[0.0340,0.0367]$ & $0.0557\,[0.0544,0.0570]$ & $[0.0338,0.0372]$ & $[0.0540,0.0574]$ \\
0.95 & $T_{0.03}$ & $0.0100\,[0.0090,0.0109]$ & $0.0156\,[0.0149,0.0162]$ & $[0.0087,0.0112]$ & $[0.0147,0.0165]$ \\